\documentclass{article}
\usepackage{booktabs}
\usepackage[backend=bibtex, sorting=none]{biblatex}
\usepackage{pdflscape}
\usepackage[a4paper, margin=1in]{geometry}
\usepackage{xcolor}
\usepackage[disable]{todonotes}
\usepackage{authblk}
\usepackage{float}
\usepackage[hidelinks]{hyperref}
\usepackage{bbding}
\usepackage{booktabs}
\usepackage{multirow}
\usepackage[table]{xcolor}
\usepackage{makecell}
\usepackage{subcaption}
\usepackage{comment}
\usepackage[T1]{fontenc} 
\usepackage{longtable} 
\usepackage{threeparttable} 
\usepackage{listings} 
\usepackage{booktabs}
\usepackage{amsmath}
\usepackage{amssymb}
\usepackage{float}
\usepackage{xcolor}
\usepackage{listings}
\floatstyle{plain}
\restylefloat{listing}
\definecolor{lightgray}{gray}{0.95}
\usepackage{graphicx} 

\floatname{listing}{Listing}
\floatstyle{plain}
\newfloat{listing}{thp}{lop}

\definecolor{lightgray}{gray}{0.95}

\usepackage[table]{xcolor}
\newcommand{\reused}[1]{\cellcolor{cyan!12}#1}
\usepackage{array}
\usepackage{tabularx}
\usepackage{ragged2e}
\usepackage{listings}
\usepackage{subcaption}

\title{\textbf{Benchmarking EngGPT2-16B-A3B against Comparable Italian \& International Open-source LLMs}}

\author{
Andrea Sassella$^{2}$\footnote{These authors contributed equally to this work.},
Andrea Chizzola$^{1*}$,
Tommaso Bianchi$^{1*}$,
Luca Alessandrelli$^{1*}$,
Mark J. Carman$^{2}$ \\
\small $^{1}$AIRIC, Politecnico di Milano \\
\small $^{2}$DEIB, Politecnico di Milano \\
\small \texttt{andrea.sassella@polimi.it, andrea.chizzola@polimi.it, tommaso.bianchi@polimi.it, luca.alessandrelli@polimi.it, mark.carman@polimi.it}
}
\begin{document}

\date{}
\maketitle

\begin{abstract}
This report benchmarks the performance of ENGINEERING Ingegneria Informatica S.p.A.'s EngGPT2MoE-16B-A3B Large Language Model (LLM), a 16 billion parameter Mixture of Experts (MoE) model with 3 billion active parameters. Performance is investigated across a wide variety of representative LLM benchmarks\footnote{Complete code used for this project is available at \url{https://github.com/AIRIC-Polimi/EngGPT-evaluation}}, including both international and Italian datasets, and is compared against comparably-sized open-source MoE and dense models.

In comparison with popular Italian models, namely FastwebMIIA-7B, Minerva-7B-instruct-v1.0, Velvet-14B, and LLaMAntino-3-ANITA-8B, the new EngGPT2MoE-16B-A3B model performs as well or better on the international benchmarks: ARC-Challenge, GSM8K, AIME2024, AIME2025, MMLU, and HumanEval. It achieves the best performance for the longest context setting (32k) of the RULER benchmark. On the Italian benchmark dataset ITALIC, the model performs as well or better than the other models except for Velvet-14B, which outperforms it.

Compared with popular Mixture-of-Experts models of comparable size, the new model reports higher values than DeepSeek-MoE-16B-Chat on all considered benchmarks. It has higher values than Moonlight-16B-A3B-Instruct on HumanEval, MMLU, AIME-2024, AIME-2025, GSM8K, and the 32k RULER setting, but lower on the function calling benchmark BFCL and some ARC-Challenge and ITALIC settings. Finally it has lower values than GPT-OSS-20B on most benchmarks, including HumanEval, MMLU, AIME 2024, AIME 2025, GSM8K, ARC, BFCL, and the RULER 32k context setting.
When compared with popular dense models, EngGPT2MoE-16B-A3B reports higher values on AIME-2024 and AIME-2025 than Llama-3.1-8B-Instruct, Gemma-3-12b-it, and Ministral-3-8B-Instruct-2512-BF16, but lower values on ITALIC, BFCL, and RULER with a 32k context.

When performance is aggregated in a dataset-wise fashion across all benchmark metrics (in terms of either score or rank), EngGPT2MoE-16B-A3B shows higher performance than the Italian models under evaluation - namely FastwebMIIA-7B, Minerva-7B-instruct-v1.0, Velvet-14B, and LLaMAntino-3-ANITA-8B - while achieving lower results than some of the most performant international models, in particular GPT-5 nano and Qwen3-8B. Taken together, our findings find the new model to be a step forward for native Italian Large Language Models.

\end{abstract}

\section{Introduction}
\label{section:introduction}
Over the last few years, Large Language Modelling (LLM) technology~\cite{brown2020language} based on generative pretrained Transformer models~\cite{vaswani2017attention} have demonstrated remarkable capabilities across all types of Natural Language Processing tasks from machine translation \cite{ganesh2025advances}-\cite{feng2025tear} to question answering \cite{yue2025survey}, text generation \cite{liang2024controllable}, etc. 

LLM technology has evolved over time into two branches, large proprietary models, that can only be accessed via paid programatic interfaces (APIs) and smaller open-source models that can be deployed locally with the advantage of increased control and reduced risk of data loss \cite{yao2024survey}-\cite{lu2024small}. The amount of training tokens used to pre-train these models has steadily increased over time and along with it the performance of the models, particularly the smaller open-source ones \cite{hoffmann2022training}.

Moreover, the development of instruction tuning techniques to improve the model's ability to follow instructions \cite{wei2021finetuned}, chain of thought prompting to improve the model's ability to generate and leverage intermediate results~\cite{wei2022chain}, and reinforcement learning for improving the reasoning capabilities of the models~\cite{ouyang2022training}, have all resulted in performance improvements across difficult benchmarks \cite{chung2024scaling}. Finally, specialized training has enabled the models to be used for tool calling~\cite{schick2023toolformer} and skill acquisition \cite{wang2023voyager}. 

On the architecture side, there have been a steady series of improvements to increase the \cite{ding2024longrope} the performance of the models, in terms of throughput \cite{ainslie2023gqa}, answer quality, and ability to work with large inputs \cite{ding2024longrope}. Sliding window attention \cite{fu2025sliding} with key-value caches have allowed the models to work with very long context lengths, greatly reducing the limitations of the early models to process only relatively short documents. In order to further increase inference speed, Mixture of Experts (MoE) models~\cite{2017olln,fedus2022switch} allow for a subset of the parameters of the model to be used during forward passes, thereby speeding up model inference, while also reducing training time~\cite{lepikhin2020gshard}. 

EngGPT2-16B-A3B\footnote{Huggingface link: https://huggingface.co/engineering-group/EngGPT2-16B-A3B} \cite{ciarfaglia2026enggpt2sovereignefficientopen} is a 16 billion parameter Mixture of Experts (MoE) model with 3 billion active parameters, developed and trained by ENGINEERING Ingegneria Informatica S.p.A. for the Italian context. In this paper, we evaluate the performance of the model across a wide variety of representative LLM benchmark datasets, including general international benchmarks which are primarily in English as well as more specific benchmarks for Italian. Performance is compared with comparable sized state-of-the-art open-source models (both MoE and dense) from leading international research teams as well as competitive LLMs that have been developed specifically for the Italian market.

\begin{figure}[t]
    \centering
    \includegraphics[width=0.9\linewidth]{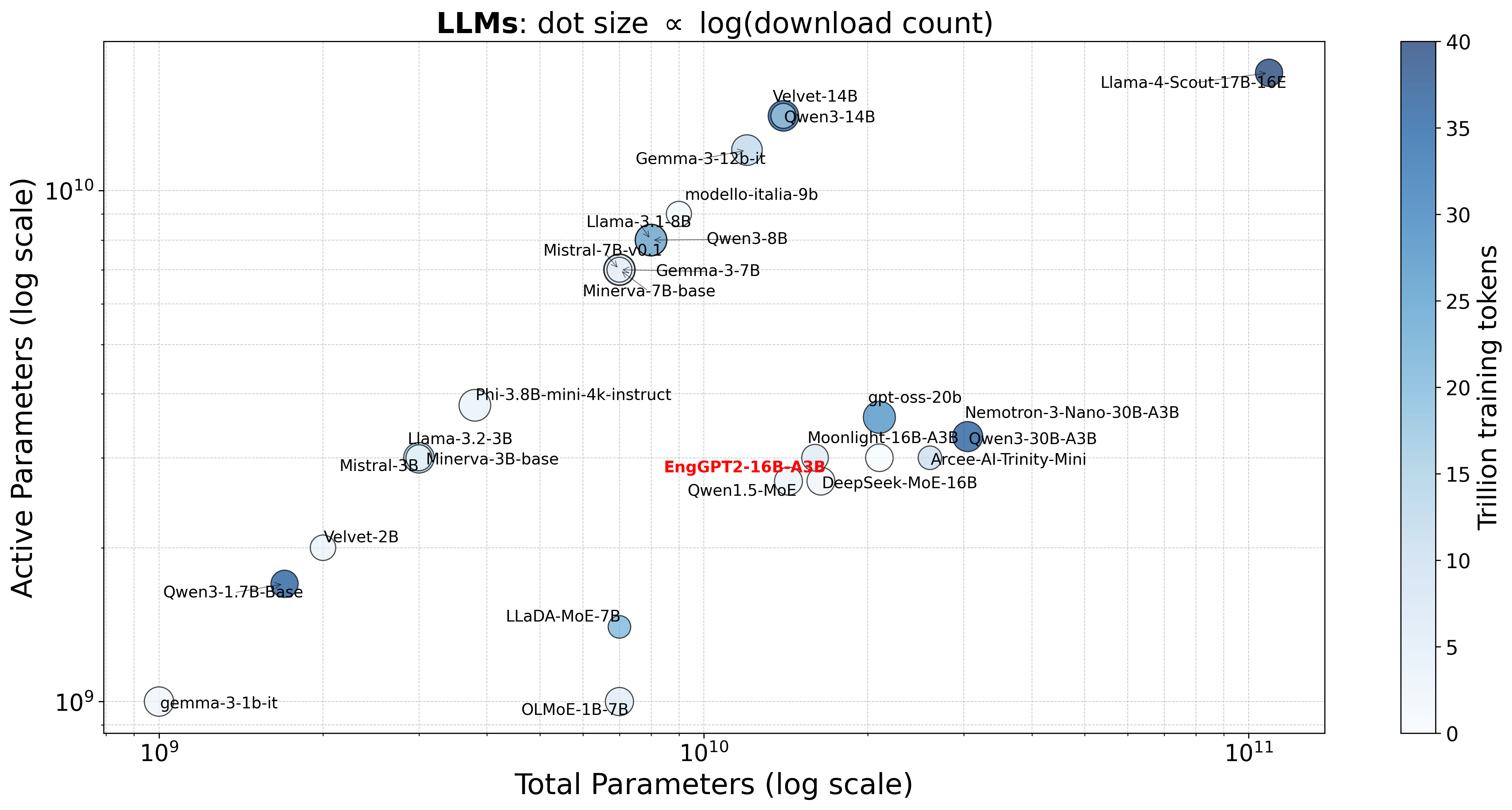} 
    \caption{Comparable open-source models in April 2026. The horizontal axis shows the total parameter count, and the vertical axis is the active parameter count, both on a logarithmic scale. The size of each circle shows the popularity of the model as measured by the number of downloads from the Huggingface repository, while the colour indicates the size of the training corpus used to pretrain the model (with darker indicating more tokens). We see the set of dense models (where active parameters equals the total parameter count) along a diagonal on the top left of the graph, and a cluster of Mixture of Expert (MoE) models in the centre of the graph.}
    \label{fig:models_overview}
\end{figure}

\section{Evaluation Methodology}

We now describe the methodology adopted for selecting appropriately dimensioned comparison models and benchmarks for evaluating them. 

\begin{table}[th]
\centering
\footnotesize
\begin{tabular}{l r c c c c c}
\toprule
 & \textbf{Date} & \textbf{\#} & \textbf{\# Active} & \textbf{Context} & \textbf{}& \textbf{Tool} \\
\textbf{Model} & \textbf{Released} & \textbf{Parameters} & \textbf{Parameters} & \textbf{Length} & \textbf{Thinking} & \textbf{Calling} 
\\
\midrule
EngGPT2-MoE-16B & Mar 30, 2025 & 16b & 3b & 32k & Yes & Yes  \\
FastwebMIIA-7B & May 29, 2025 & 7.4b & 7.4b & 16k & No & No \\
Minerva-7B-instruct-v1.0 & Sep 18, 2024 & 7b & 7b & 4k & No & No \\
Velvet-14B & Jan 31, 2025  & 14b & 14b & 128k & No & Yes \\
LLaMAntino-3-ANITA-8B & May 10, 2024 & 8b & 8b & 8k & No & Yes  \\
\hline
Qwen3-4B & Apr 28, 2025 & 4b & 4b & 32k & Yes & Yes \\
Qwen3-8B & Apr 28, 2025 & 8b & 8b & 32k & Yes & Yes \\
Llama-3.2-3B-Instruct & Sep 18, 2024 & 3b & 3b & 128k & No & Yes\\
Llama-3.1-8B-Instruct & Jul 15, 2024 & 8b & 8b & 128k & No & Yes \\
\textcolor{black}{gemma-3-4b-it} & Mar 12, 2025 & 4b & 4b & 128k & No & Yes \\
\textcolor{black}{gemma-3-12b-it} & Mar 12, 2025 & 12b & 12b & 128k & No & Yes \\
Ministral-3-8B & Jan 13, 2026 & 8b & 8b & 256k & No & Yes  \\
GPT-5 nano & Mar 12, 2025 & N.A. & N.A. & 400k & Yes & Yes  \\
\hline
deepseek-moe-16b-chat & Jan 9, 2024 & 16.4b & 2.7b & 4k & No & No\\
gpt-oss-20b & Aug 5, 2025 & 21b & 3.6b & 128k & Yes & Yes \\
\textcolor{black}{Moonlight-16B-A3B-Instruct} & Feb 24, 2025 & 12b & 3b & 128k & No & No \\
\bottomrule
\end{tabular}
\caption{Models chosen for comparison with their relevant characteristics.}
\label{tab:chosen_models}
\end{table}

\subsection{Selection of Models for Comparison}

In order to select an appropriate set of models for comparison, we first visualise popular models in Figure \ref{fig:models_overview} in terms of their (i) total parameter count, (ii) active parameter count (parameters used during a forward inference pass), (iii) popularity (as estimated by the number of downloads from the HuggingFace repository), and (iv) amount of training data (number of tokens) used to pre-train the model. A set of dense models (where all parameters are active) are seen along a diagonal on the top left of the graph, while a cluster of Mixture of Expert (MoE) models (containing also EngGPT2-16B-A3B) appears in the centre of the graph. Our aim is to select the most representative and popular models from both clusters. 

For the analysis, we will compare the EngGPT2MoE-16B-A3 model with the models listed in Table~\ref{tab:chosen_models}, specifically: 
\begin{itemize}
\item \textbf{Dense Italian models} that have been developed specifically for the Italian context, namely \textit{FastwebMIIA-7B}\footnote{\url{https://huggingface.co/Fastweb/FastwebMIIA-7B}} and \textit{Minerva-7B-instruct-v1.0} \cite{orlando-etal-2024-minerva}, both leveraging seven billion parameters, \textit{Velvet-14B} \footnote{\url{https://huggingface.co/Almawave/Velvet-14B}} with 14 billion parameters, and \textit{LLaMAntino-3-ANITA-8B-Inst-DPO-ITA} \footnote{From here on \textit{LLaMAntino-3-ANITA-8B-Inst-DPO-ITA} and \textit{LLaMAntino-3-ANITA-8B} will be used interchangeably} \cite{polignano2026advanced}, a model with 8 billion parameters that has been fine-tuned from a LlaMA3 model on Italian datasets; 
\item \textbf{Dense International models} that are extremely popular, including two models from Alibaba, \textit{Qwen3-4B} and \textit{Qwen3-8B} \cite{yang2025qwen3} with four and eight billion parameters respectively, two models from Meta, \textit{Llama-3.2-3B-Instruct} and the older \textit{Llama-3.1-8B-Instruct} \cite{grattafiori2024llama} with 3 and 8 billion parameters respectively, a model from mistral \textit{Ministral-3-8B-Base-2512}\footnote{From here on \textit{Ministral-3-8B-Base-2512} and \textit{Ministral-3-8B} will be used interchangeably} \cite{liu2026ministral3} and finally two models from Google, \textit{gemma-3-4b-it} and \textit{gemma-3-12b-it} \cite{gemmateam2025gemma3technicalreport} with 4 and 12 billion parameters respectively. We additionally include a proprietary model with undisclosed weights, \textit{GPT-5 nano}, to establish a baseline using commercially available APIs and to better understand the gap between open-weight and closed models.
\item \textbf{Mixture of Expert (MoE) models} of comparable size: \textit{deepseek-moe-16b-chat} \cite{dai2024deepseekmoe} from Deepseek with 16 billion parameters of which 2.7 billion are active, \textit{gpt-oss-20b} \cite{agarwal2025gpt} from OpenAI with 20 billion parameters and 3.6 billion active, and \textit{Moonlight-16B-A3B-Instruct} \cite{liu2025muon}, with the most similar amount of total and active parameters with respect to EngGPT2-MoE-16B.
\end{itemize}

We note that the chosen models in Table~\ref{tab:chosen_models} differ not only in terms of number of total and active parameters, but also in their context length (affecting the maximum size of conversations they can deal with), whether or not they have been trained as "thinking models" \cite{chen2025towards,xu2025toward} to output their reasoning in a structured format when responding, and whether they have been trained for tool/function calling \cite{schick2023toolformer,patil2023gorilla,
song2023restgpt}. These architectural and training differences are highly consequential and directly dictate performance across specific benchmarks; for instance, native tool-calling capabilities are decisive for success on the BFCL benchmark, while structured chain-of-thought training is critical for achieving high accuracy on complex, multi-step problem-solving tasks such as AIME, MMLU, and GSM8K. 

\begin{table}[th]
\centering
\tiny
\begin{tabular}{lcccc}
\toprule
Dataset & Llama3 & Qwen3 & gpt\_oss & DeepSeek MoE \\
\midrule
\reused{AIME \cite{aime2024aimei}-\cite{aime2025aimei}} &  & \Checkmark & \Checkmark &  \\
AlignBench v1.1 \cite{liu2024alignbench} &  & \Checkmark &  &  \\
API-Bank \cite{li2023api} & \Checkmark &  &  &  \\
\reused{ARC-Challenge \cite{clark2018think}} & \Checkmark &  &  & \Checkmark \\
Arena-Hard \cite{li2024crowdsourced} &  & \Checkmark &  &  \\
AutoLogi \cite{zhu2025autologi} &  & \Checkmark &  &  \\
\reused{BFCL \cite{patil2025berkeley}} & \Checkmark & \Checkmark &  &  \\
C-Eval \cite{huang2023c} &  & \Checkmark &  & \Checkmark \\
CHID \cite{zheng2019chid} &  &  &  & \Checkmark \\
CLUEWSC \cite{xu-etal-2020-clue} &  &  &  & \Checkmark \\
CMMLU \cite{li2024cmmlu} &  &  &  & \Checkmark \\
CodeElo \cite{quan2025codeelo} &  & \Checkmark & \Checkmark &  \\
Creative Writing V3 &  & \Checkmark &  &  \\
DROP \cite{dua2019drop} &  &  &  & \Checkmark \\
Gorilla API-Bench \cite{patil2024gorilla} & \Checkmark &  &  &  \\
GPQA \cite{rein2024gpqa} & \Checkmark &  & \Checkmark &  \\
GPQA-Diamond \cite{rein2024gpqa} &  & \Checkmark &  &  \\
\reused{GSM8K \cite{cobbe2021training}} & \Checkmark &  &  & \Checkmark \\
HealthBench \cite{arora2025healthbench} &  &  & \Checkmark &  \\
HellaSwag \cite{zellers2019hellaswag} &  &  &  & \Checkmark \\
HLE \cite{phan2025humanity} &  &  & \Checkmark &  \\
\reused{HumanEval} \cite{chen2021evaluating} & \Checkmark &  &  &  \\
IFEval \cite{zhou2023instruction} & \Checkmark & \Checkmark &  &  \\
INCLUDE \cite{romanou2024include} &  & \Checkmark &  &  \\
InfiniteBench \cite{zhang2024bench} & \Checkmark &  &  &  \\
LiveBench \cite{white2024livebench} &  & \Checkmark &  &  \\
LiveCodeBench \cite{jain2024livecodebench} &  & \Checkmark &  &  \\
MATH \cite{hendrycks2021measuring} & \Checkmark &  &  & \Checkmark \\
MATH-500 \cite{lightman2023let} &  & \Checkmark &  &  \\
MBPP EvalPlus \cite{liu2023your} & \Checkmark &  &  &  \\
MGSM \cite{shi2022language} & \Checkmark &  &  &  \\
MlogiQA \cite{zhang2025p} &  & \Checkmark &  &  \\
MMMLU \cite{hendrycks2020measuring} &  & \Checkmark & \Checkmark &  \\
\reused{MMLU \cite{hendrycks2020measuring}} & \Checkmark &  & \Checkmark & \Checkmark \\
MMLU-Pro \cite{wang2024mmlu} & \Checkmark &  &  &  \\
\reused{MMLU-Redux} \cite{gema2025we} &  & \Checkmark &  &  \\
Multi-IF \cite{he2024multi} &  & \Checkmark &  &  \\
MultiPL-E \cite{cassano2023multipl} & \Checkmark &  &  &  \\
Needle-in-a-Haystack \cite{kamradt2023llmtest_needleinahaystack} & \Checkmark &  &  &  \\
Nexus \cite{srinivasan2023nexusraven} & \Checkmark &  &  &  \\
Pile \cite{gao2020pile} &  &  &  & \Checkmark \\
PolyMath \cite{wang2025polymath} &  & \Checkmark &  &  \\
\reused{RULER} \cite{hsieh2024ruler} & & \Checkmark & & \\
TruthfulQA \cite{lin2022truthfulqa} &  &  &  & \Checkmark \\
WinoGrande \cite{sakaguchi2021winogrande} &  &  &  & \Checkmark \\
WritingBench \cite{wu2025writingbench} &  & \Checkmark &  &  \\
ZebraLogic \cite{lin2025zebralogic} &  & \Checkmark &  &  \\
ZeroSCROLLS \cite{shaham2023zeroscrolls} & \Checkmark &  &  &  \\
$\tau$-Bench Retail \cite{yao2024tau} &  &  & \Checkmark &  \\
\bottomrule
\end{tabular}
\caption{Datasets used for benchmarking in the papers introducing four popular state-of-the-art post-trained LLMs: \textit{Llama3}, \textit{Qwen3}, \textit{GPT-oss}, and \textit{DeepSeekMoE}. We highlight the ones that were used in this work to perform the benchmarking.}
\label{tab:all_datasets}
\end{table}

\subsection{Selection of Benchmarks}

In order to identify a comprehensive set of general benchmarks that can be used to evaluate the chosen models across a representative set of tasks, we start from the articles describing recent and popular LLMs: \textit{Llama3} \cite{grattafiori2024llama}, \textit{Qwen3} \cite{yang2025qwen3}, \textit{GPT-Oss} \cite{agarwal2025gpt}, and \textit{DeepSeekMoE} \cite{dai2024deepseekmoe}, and investigate the set of datasets that were used to evaluate them. Table~\ref{tab:all_datasets} lists all the datasets used to evaluate the capabilities of these state-of-the-art \textit{post-trained} LLMs. We note that there is significant variation in the set of benchmarks used in the evaluation across the different models, an issue that we aim to address in this work by adopting a standardized, representative and reproducible set of benchmarks.  

\begin{figure}[t]
    \centering
    \includegraphics[width=0.8\linewidth]{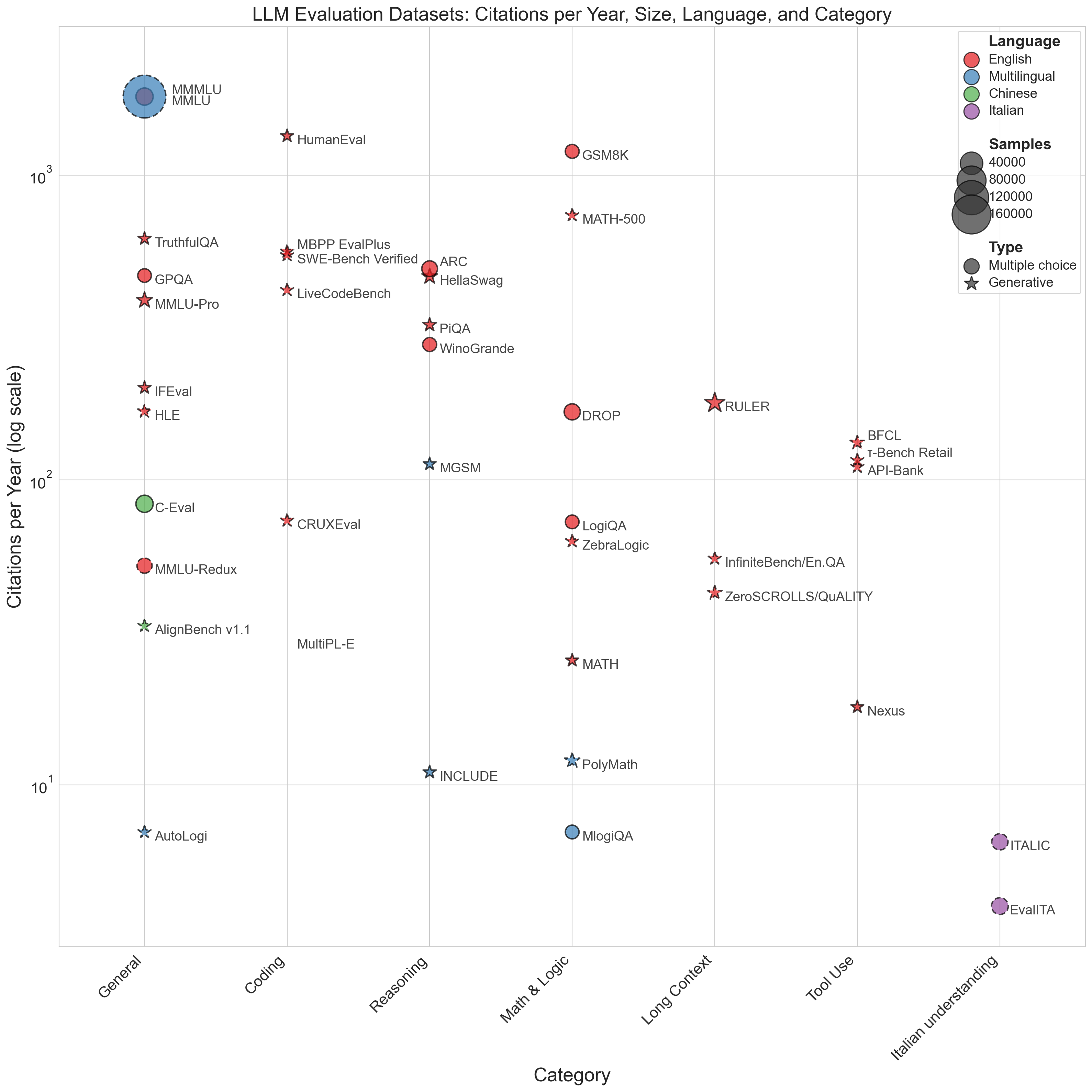} 
    \caption{Overview of popular potential datasets, including their popularity (measured by yearly downloads), category, size (in terms of the number of samples), language, and type (multiple choice vs generative).}
    \label{fig:datasets_overview}
\end{figure}

We investigate the properties of the benchmark datasets in Figure~\ref{fig:datasets_overview}, where we categorise them in terms of their category (general knowledge vs coding, etc.), their language (English, Multilingual or Italian), size (total number of samples), popularity (downloads from Huggingface) and whether they are generative tasks or multiple choice tasks. We split the dataset into 7 categories, and for each of them we identify the selected benchmarks, which are summarized in Table~\ref{tab:selected_benchmarks}. Based on the visualisation, we then decide to start our preliminary evaluation using the most popular of the datasets within their category: \textit{MMLU} \cite{hendrycks2020measuring,gema2025we} for general tasks, \textit{HumanEval} \cite{chen2021evaluating} for coding, \textit{ARC-Challenge} \cite{clark2018think} for reasoning, \textit{GSM8K} \cite{cobbe2021training} and \textit{AIME 2024 \& 2025} \cite{aime2024aimei,aime2025aimei} for math reasoning, \textit{RULER} \cite{hsieh2024ruler} for Long context, and \textit{BFCL} \cite{patil2025berkeley} for Tool Use.

Since we aim at evaluating the model also on the Italian language, we consider including at least one native Italian dataset. Work by NLP researchers in Italy has resulted in the creation of two standard set of benchmark datasets in Italian, called \textit{Evalita-LLM} \cite{magnini2025evalita} (extending a previous ITA-bench benchmark \cite{moroni2024towards}) and \textit{ITALIC} \cite{seveso2025italic}. Both \textit{Evalita-LLM} and \textit{ITALIC} contain a large amount of native Italian text, distinguishing it from other Italian datasets which tended to be created through a process of machine translation. For the sake of our evaluation, we consider the \textit{ITALIC} benchmark, as it is more aligned with current benchmarks adopted to assess LLMs capabilities.

\subsection{Implementation Details}

To conduct a rigorous and reproducible evaluation across the selected benchmarks, we utilized a combination of standardized frameworks and custom integrations tailored to the specific requirements of each task. The majority of the evaluations were executed using EleutherAI’s \texttt{lm-evaluation-harness} Python package \cite{eval-harness}. Within this framework, all associated tasks were configured using the \texttt{generate\_until} output type. This setting instructs the model to perform open-ended text generation until a predefined stopping criterion (such as a specific token or maximum length limit) is met, thereby capturing complete reasoning trajectories rather than merely calculating log-likelihoods for predefined multiple-choice tokens. To automatically evaluate these free-form responses, the harness employs regular expression (regex) functions to parse and isolate the final answer from the surrounding generated text. For our evaluations, we utilized both \textit{strict} and \textit{flexible} extraction methodologies. Strict extraction enforces rigid formatting constraints (e.g., requiring the answer to appear at the very end of the text or within specific delimiters), penalizing models that fail to follow precise formatting instructions. Conversely, flexible extraction uses broader regex patterns to identify the correct answer anywhere within the output. Employing both methods is crucial, as it allows us to effectively decouple a model's underlying reasoning and factual capabilities from its strict formatting compliance.

To maximize inference throughput and computational efficiency, the evaluated models were predominantly served utilizing the vLLM backend \cite{kwon2025vllm}. There were, however, two necessary deviations from this default serving configuration. First, the evaluation of the \textit{GPT-5-nano} model was conducted via API calls using the \texttt{azure-openai-chat-completions} model option of the harness. During this process, we encountered instances where specific task prompts inadvertently triggered Azure's built-in safety and content filters, resulting in API exceptions that halted the execution. To ensure an uninterrupted evaluation pipeline, we modified the underlying \texttt{lm-evaluation-harness} source code to include an exception-handling mechanism; this allowed the system to catch safety-related errors, bypass the blocked queries, and automatically proceed to the subsequent test sample. Second, for the evaluation of \textit{Ministral-3-8B-Instruct-2512-BF16}, we were required to use the \texttt{hf-mistral3} model option of the harness, which uses Huggingface's \texttt{transformers} library instead of vLLM as a backend. This fallback was necessitated by compatibility constraints between the model's underlying architecture and the Hugging Face version currently supported by the \texttt{lm-evaluation-harness}. Consequently, this compatibility workaround resulted in significantly slower inference and overall evaluation times for this specific model compared to the other ones.

For benchmarks outside the scope of the primary evaluation \textit{harness}, we employed alternative software solutions. The BFCL evaluation was carried out using the \texttt{evalscope} Python package \cite{evalscope_2024}, since tool calling is currently not compatible with the architecture of EleutherAI’s \texttt{lm-evaluation-harness}. Furthermore, since the \textit{ITALIC} benchmark is not natively supported by these standard evaluation libraries, we conducted its evaluation by directly importing the original task repository into our custom codebase, in addition to developing custom tasks for integrating it with the \texttt{lm-evaluation-harness} framework.

For reproducibility purposes, comprehensive technical specifications, including the exact software versions and environment utilized during all experiments, are documented in Appendix \ref{appendix:technical_specifications}.

\begin{table}[t]
\centering
\footnotesize
\begin{tabular}{ll}
\hline
\textbf{Task category} & \textbf{Selected dataset} \\
\hline
General tasks         & MMLU \\
Coding                & HumanEval \\
Reasoning             & ARC-Challenge \\
Math reasoning        & GSM8K,  AIME-2024, AIME-2025 \\
Long context          & RULER \\
Tool use              & BFCL \\
Language understanding & ITALIC \\
\hline
\end{tabular}
\caption{Datasets selected for the evaluation, chosen as the most popular benchmark within each task category.}
\label{tab:selected_benchmarks}
\end{table}

\section{Benchmarking Datasets}
This section presents the benchmarking datasets selected for our evaluation. For each benchmark, we briefly describe its main characteristics, we provide an illustrative example to clarify the type of task it contains and we describe the selected evaluation setting. For additional technical details on how each evaluation task was configured (including custom modifications) starting from the selected benchmarking datasets, see Appendix \ref{appendix:config}.

\begin{table}[th]
\centering
\scriptsize
\begin{tabular}{|p{0.6\linewidth} | p{0.3\linewidth}|}
\hline
\textbf{Example Questions} & \textbf{Desired Answers} \\ 
\hline
\hline
\textbf{ARC-Challenge:} & \\
George wants to warm his hands quickly by rubbing them. Which skin surface will produce the most heat? &  A. dry palms\\
A. dry palms & \\
B. wet palms  & \\
C. palms covered with oil & \\
D. palms covered with lotion & \\ 
\hline
\hline
\textbf{GSM8K:} &  \\ 
Natalia sold clips to 48 of her friends in April, and then she sold half as many clips in May. How many clips did Natalia sell altogether in April and May? & Natalia sold 48/2 = \textless\textless48/2=24\textgreater\textgreater24 clips in May.
Natalia sold 48+24 = \textless\textless48+24=72\textgreater\textgreater72 clips altogether in April and May.
\#\#\#\# 72 \\ 
\hline
\hline
\textbf{AIME-2024:} &  \\ 
Let $x,y$ and $z$ be positive real numbers that satisfy the following system of equations:
\[\log_2\left({\frac{x}{yz}}\right) = {\frac{1}{2}}\]
\[\log_2\left({\frac{y}{xz}}\right) = {\frac{1}{3}}\]
\[\log_2\left({\frac{z}{xy}}\right) = {\frac{1}{4}}\]
Then the value of $\left|\log_2(x^4y^3z^2)\right|$ is $\frac{m}{n}$ where $m$ and $n$ are relatively prime positive integers. Find $m+n$. 
& 
Denote $\log_2(x) = a$, $\log_2(y) = b$, and $\log_2(z) = c$.
Then, we have:
$a-b-c = \frac{1}{2}$,
$-a+b-c = \frac{1}{3}$,
$-a-b+c = \frac{1}{4}$.
Now, we can solve to get $a = \frac{-7}{24}, b = \frac{-9}{24}, c = \frac{-5}{12}$.
Plugging these values in, we obtain $|4a + 3b + 2c| = \frac{25}{8} = 0.3125$. \\ 
\hline
\hline
\textbf{MMLU:} & \\ 
Find the degree for the given field extension $Q(\sqrt{2}, \sqrt{3}, \sqrt{18})$ over $Q$. & B\\
A. 0 & \\
B. 4 & \\
C. 2 & \\
D. 6 & \\
Which of these principles is not an element of the responsibility to protect? & C\\
A. The responsibility to prevent & \\
B. The responsibility to react & \\
C. The responsibility to remain sovereign & \\
D. The responsibility to rebuild & \\
\hline
\end{tabular}
\caption{Illustrative examples of questions and correct answers from the \textit{ARC-Challenge}, \textit{GSM8K}, \textit{AIME-2024}, and \textit{MMLU} benchmarks.}
\label{tab:arc-gsm8k-aime2024-MMLU-examples}
\end{table}

\subsection{Reasoning-Intensive Benchmark - ARC-Challenge}
ARC-Challenge (AI2 Reasoning Challenge) \cite{clark2018think} is a benchmark composed of grade-school-level multiple-choice science questions designed to evaluate commonsense and scientific reasoning abilities (see Table \ref{tab:arc-gsm8k-aime2024-MMLU-examples}). The dataset focuses on questions that are difficult for simple retrieval-based and word co-occurrence methods, making it a more challenging subset of the broader ARC benchmark. In particular, we consider the Challenge split, which is composed of 2.6k examples (1,119 train, 299 validation, and 1,172 test).\\
To assess the impact of intermediate reasoning on model performance, we evaluate three distinct configurations of this task: 
\begin{itemize}
    \item \texttt{arc\_challenge\_chat}, which evaluates direct answering \textit{without enabling reasoning};
    \item \texttt{arc\_challenge\_cot\_custom}, which we introduced as an extension of the previous one to \textit{explicitly prompt the model to reason} (see Appendix \ref{appendix:arc_config} for implementation details);
    \item \texttt{arc\_challenge\_cot\_it\_custom}, which is an \textit{Italian machine-translated version} of the previous one (see Appendix \ref{appendix:arc_config} for implementation details).
\end{itemize}
This dual approach allows for a comprehensive comparison, ensuring that models natively supporting reasoning paradigms can be fairly evaluated alongside those that do not. For all the configurations, the evaluation is formulated as an open-ended generation task utilizing the \texttt{generate\_until} output type, wherein the entirety of the model's generated text is considered. Model performance is quantified using the mean exact match score. Finally, since the reasoning-enabled \texttt{arc\_challenge\_cot\_custom} variant produces verbose outputs, we use both flexible and strict extraction functions to accurately parse the final answer from the preceding reasoning traces. The same applies to the Italian version. The exact code implementation for these extraction methodologies is detailed in Appendix \ref{appendix:arc_config}.

\subsection{Math - GSM8K}

GSM8K (Grade School Math 8K) \cite{cobbe2021training} is a benchmark composed of grade-school level mathematical word problems designed to evaluate numerical and multi-step reasoning abilities. The dataset focuses on problems expressed in natural language, where solving the task typically requires performing a sequence of arithmetic operations rather than retrieving a direct fact. The dataset is composed by 7.5k examples (see Table \ref{tab:arc-gsm8k-aime2024-MMLU-examples} for an example).\\
For our evaluation, we specifically selected the \texttt{gsm8k\_llama} variant of the task, as it incorporates superior parsing rules for extracting mathematical answers from complex model outputs. The task is structured using the \texttt{generate\_until} output type, meaning that the entirety of the model's generated text is considered during evaluation. Performance is measured using the mean exact match score. To reliably isolate the final numerical answer from the generated reasoning traces, we applied both flexible and strict extraction methodologies using their default configurations.

\subsection{Hard Math - AIME-2024 \& AIME-2025}

AIME-2024 \cite{aime2024aimei} and AIME-2025 \cite{aime2025aimei} are benchmarks derived from the 2024 and 2025 editions of the American Invitational Mathematics Examination (AIME), a competition aimed at evaluating advanced problem-solving skills in pre-college mathematics. Compared with datasets such as GSM8K, AIME problems are significantly more challenging (see Table~\ref{tab:arc-gsm8k-aime2024-MMLU-examples}), often requiring advanced reasoning, algebraic manipulation, combinatorics, number theory, or geometry. This benchmark is designed to assess high-level mathematical reasoning on competition-style questions, where success depends not only on obtaining the correct final answer but also on correctly identifying the sequence of intermediate logical and computational steps needed to solve the problem.\\
Consistent with our other evaluations, this task utilizes the \texttt{generate\_until} output type, meaning the entirety of the model's generated text is considered to capture the complete reasoning trajectory. For the standard evaluation runs, overall performance is measured using the mean exact match score. In addition to the standard evaluations, we performed custom runs to compute the pass@k and avg@k metrics, specifically for k=1 and k=8. These custom runs are identical to the standard configurations in all respects except for the evaluation metrics applied; the exact code used to compute these custom metrics is provided in Appendix \ref{appendix:aime_config}. Finally, for models evaluated via API calls (e.g., OpenAI's {GPT-5-nano}), we modified the task prompt to explicitly instruct the model to enclose its final answer using the \texttt{\textbackslash boxed\{\}} syntax, ensuring reliable extraction. The exact prompt template utilized for these API-based evaluations is detailed in Appendix \ref{appendix:aime_config}.

\subsection{General Knowledge Benchmarks - MMLU \& MMLU Redux}

MMLU (Massive Multitask Language Understanding) \cite{hendrycks2020measuring} is a benchmark composed of multiple-choice questions drawn from diverse branches of knowledge, designed to evaluate broad world knowledge and problem-solving abilities (see example in Table \ref{tab:arc-gsm8k-aime2024-MMLU-examples}). The benchmark spans subjects in the humanities, social sciences, hard sciences, and other domains, covering 57 tasks. The dataset includes 99.8k examples in the auxiliary training split, 285 examples in the dev split, 1.5k examples in the validation split, and 14k examples in the test split.
For our evaluation, we selected two distinct variants of this benchmark. The first is \textit{MMLU\_cot\_llama}, a chain-of-thought version of MMLU designed to elicit step-by-step reasoning. To effectively handle the outputs of this variant, we integrated enhanced parsing rules and created a variant of the benchmark named \textit{MMLU\_cot\_llama\_custom}, the details of which are described in Appendix \ref{appendix:mmlu_config}. The second variant is \textit{mmlu\_redux} \cite{gema2025we}, a carefully annotated and corrected version of the original dataset that provides a more accurate and reliable benchmark for evaluating language model performance. For both configurations, the task is structured using the \texttt{generate\_until} output type, meaning that the entirety of the model's generated text is considered during evaluation. Overall, performance across both variants is quantified using the mean exact match score. Additionally, because the \textit{MMLU\_cot\_llama} variant involves verbose reasoning traces, we applied both flexible and strict extraction methodologies to accurately isolate the final multiple-choice answer from the generated text.

\begin{table}[t]
\centering
\scriptsize
\setlength{\tabcolsep}{4pt}
\renewcommand{\arraystretch}{1.1}
\begin{tabularx}{\linewidth}{|X|p{0.18\linewidth}|p{0.3\linewidth}|}
\toprule 
\textbf{Prompt} & \textbf{Canonical Solution} & \textbf{Test} \\ 
\midrule 
\begin{minipage}[t]{\linewidth}
def greatest\_common\_divisor(a: int, b: int) -> int: \\
"""Return a greatest common divisor \\
of two integers a and b \\
{>}{>}{>} greatest\_common\_divisor(3, 5) \\
1 \\
{>}{>}{>} greatest\_common\_divisor(25, 15) \\
5 \\
""" \\
\end{minipage}
&
\begin{minipage}[t]{\linewidth}
while b: \\
\ \ \ a, b = b, a \% b \\
return a
\end{minipage}
&
\begin{minipage}[t]{\linewidth}
METADATA = \{ \\
\ \ \ 'author': 'jt', \\\ \ \ 'dataset': 'test' \\
\} \\
def check(candidate): \\
\ \ \ assert candidate(3, 7) == 1 \\
\ \ \ assert candidate(10, 15) == 5 \\
\ \ \ assert candidate(49, 14) == 7 \\
\ \ \ assert candidate(144, 60) == 12 \\
\end{minipage}
\\ \bottomrule 
\end{tabularx}
\caption{Illustrative example from the HumanEval benchmark style.}
\label{tab:humaneval-example}
\end{table}

\subsection{Coding - HumanEval}
HumanEval is a benchmark composed of handwritten Python programming problems designed to evaluate code generation and functional reasoning abilities \cite{chen2021evaluating}. The dataset focuses on tasks where a model must generate a correct function implementation from a natural-language specification and function signature, and correctness is assessed through unit tests. Each example includes a prompt, a canonical solution, a test suite, and an entry point. The dataset is composed of 164 examples.\\
For our evaluation, we utilized the default variant of the task, enhanced with custom parsing rules. Specifically, we observed that several models generated functionally correct code that inadvertently included unsupported Unicode characters, which caused Python syntax errors during execution. To address this, we implemented an additional sanitization step to remove these characters prior to running the unit tests. The generation phase was configured with the \texttt{generate\_until} output type, meaning that the entirety of the model's generated text is considered. Finally, to rigorously evaluate code generation performance, we measured the pass@k and avg@k metrics for $k=1$ and $k=8$.

\subsection{Tool Calling - BFCL}
\label{sec:datasets-bfcl}

\textit{BFCL} (Berkeley Function Calling Leaderboard) \cite{patil2025berkeley} is a benchmark dataset for evaluating large language models on function/tool calling. It pairs user requests with function specifications and measures whether a model selects the right tool, formats the call correctly, fills arguments accurately, abstains when no tool is appropriate, and, in newer versions, handles multi-turn and multi-step agentic interactions. \textit{BFCL} covers multiple tool-use settings such as single, multiple, and parallel calls, and spans several programming-language/tool formats.
In particular, we consider the version \verb|v3| of the benchmark, which includes four main groups of tasks that we scored separately:
\begin{itemize}
    \item \textbf{Non-live}: contains expert-curated samples spanning the Python, Java and JavaScript programming languages in which each one describes a single function that needs to be called correctly, a multiple category in which the model is given more than one tool and needs to call the correct one, a parallel category in which the model needs to call more than one tool, and a parallel multiple category which is the most complex and require selection of $m$ correct tools out of $N$ provided.
    \item \textbf{Live}: contains user-contributed samples, spanning the same categories as the Non-live group.
    \item \textbf{Hallucination Measurement}: contains both expert-curated and user-contributed samples where the model is given non relevant tools and should correctly recognize that it does not need to use them to answer. This group measures how well the model behaves in the presence of irrelevant data.
    \item \textbf{Multi-Turn}: contains samples in which multiple tool-based interactions are needed to complete the task, such as listing the files in a folder and then reading the correct one. This group contains a base category, a missing functions category in which some tools are not available in the first turn of conversation but are revealed later, a missing parameters category in which some key information is held out from the original user message and needs to be requested during the conversation, and a long context category in which large volumes of extraneous data (e.g., long files, long directory listings) are added to measure how well the model can extract important information out of noisy data.
\end{itemize}

\begin{table}[t]
\centering
\footnotesize
\begin{tabular}{|*{4}{c|}}
\hline
\multicolumn{4}{|c|}{\makecell{\textbf{Overall Score} \\ Unweighted Average \\ (4423)}} \\ \hline
\makecell{\textbf{Non-Live (Single-Turn)} \\ Unweighted Average \\ (1150)} & 
\makecell{\textbf{Live (Single-Turn)} \\ Weighted Average \\ (1351)} & 
\makecell{\textbf{Multi-Turn} \\ Unweighted Average \\ (800)} & 
\makecell{\textbf{Hallucination Measurement} \\ Unweighted Average \\ (1122)} \\ \hline
\end{tabular}
\caption{Structure of the BFCL dataset used for the evaluations in terms of task groups. The number in parentheses indicate the number of samples for each task and task group, while the averaging rules indicate how scores were grouped together for each task group and then globally. Further details on the specific tasks for each category can be found in Table \ref{tab:bfcl-dataset-details} in the Appendix.}
\label{tab:bfcl-dataset}
\end{table}

Table \ref{tab:bfcl-dataset} summarizes the structure of the BFCL dataset, including the number of samples for each task group and the aggregation logic used to report the overall scores. Further details about each specific task can be found in Appendix \ref{appendix:bfcl-dataset-details}.

The metric used to evaluate model responses depends on the task: for the live and non-live tasks, the generated tool calls are parsed into their Abstract Syntax Tree (AST), which is in turn used for determining whether the functions and their respective parameters and types are correct; for multi-turn tasks, at each turn both the state (e.g., a virtual filesystem) and the response of the models are checked against a ground truth, and both checks must succeed for all turns for the entry to be marked as a success; for hallucination measurement, we simply check that the model outputs no tool calls at all (which may cause non-responsive models to inadvertently perform well on this task).

\begin{table}[th]
\centering
\footnotesize
\begin{threeparttable}
\begin{tabular}{|l c c c|}
\toprule
\textbf{Model} & \textbf{FC mode} & \textbf{Tool call parser} & \textbf{Reasoning parser} 
\\
\midrule
EngGPT2-MoE-16B & \Checkmark & hermes & qwen3  \\
FastwebMIIA-7B & & & \\
Minerva-7B-instruct-v1.0 & & & \\
Velvet-14B & & unknown\tnote{1} & \\
LLaMAntino-3-ANITA-8B & & &  \\
\hline
Qwen3-4B & \Checkmark & hermes & qwen, deepseek\_r1 \\
Qwen3-8B & \Checkmark & hermes & qwen, deepseek\_r1 \\
Llama-3.2-3B-Instruct & \Checkmark & llama3\_json, pythonic & \\
Llama-3.1-8B-Instruct & \Checkmark & llama3\_json & \\
\textcolor{black}{gemma-3-4b-it} & \Checkmark & pythonic\tnote{2} & \\
\textcolor{black}{gemma-3-12b-it} & \Checkmark & pythonic\tnote{2} & \\
Ministral-3-8B-Instruct-2512-BF16 & \Checkmark & mistral &  \\
GPT-5 nano & \Checkmark & N/A\tnote{3} & N/A\tnote{3}  \\
\hline
deepseek-moe-16b-chat & & & \\
gpt-oss-20b & \Checkmark & openai & auto-detected by vLLM\tnote{4} \\
\textcolor{black}{Moonlight-16B-A3B-Instruct} & & & \\
\bottomrule
\end{tabular}
\begin{tablenotes}\footnotesize
    \item[1] Velvet models have reportedly been trained for tool calling, but we were unable to find details regarding the parsing style used. Tests with plausible parsers (mistral, granite) proved ineffective.
    \item[2] During evaluation the pythonic tool call parser for gemma was unable to properly parse model output.
    \item[3] The GPT5 model family was accessed through the Azure OpenAI Cloud service, so we did not have explicit control over tool and reasoning parsing, though it was in effect server-side.
    \item[4] The gpt-oss family of models uses Harmony tokenizer response format, which is natively auto-detected and parsed by the vLLM engine without specific configuration.
\end{tablenotes}
\end{threeparttable}
\caption{vLLM tool call parsers used for each model in the BFCL evaluation.}
\label{tab:bfcl_tool_parsers}
\end{table}

This dataset required ad hoc treatment due to the complexity and many nuances of function calling. In particular, since function calling in general is not supported by EleutherAI’s LM Evaluation Harness suite due to architectural limitations, we adopted the Evalscope evaluation library \cite{evalscope_2024}, which in turn wraps the original implementation of the BFCL evaluation benchmark. In addition, we served each model using the vLLM  \cite{kwon2023efficient} inference server, which enabled us to make use of their built-in parsers for tool calling for the models that support it. More details on which parser was used for each model can be found in Table \ref{tab:bfcl_tool_parsers}.

It is worth noting that the evaluation logic supports both injecting tool descriptions in the model input as raw strings and then parsing them from the raw textual output (so-called non-FC or PROMPT mode), as well as using dedicated fields in the Request and Response objects of the OpenAI standard API format and leaving serialization and deserialization to and from the model to the inference server (so-called native function calling or FC mode). For models that support both, we tested them and reported the best results; for the other ones we used PROMPT mode.

\begin{table}[th]
\centering
\scriptsize
\begin{tabular}{|p{0.6\linewidth}|p{0.21\linewidth}|p{0.08\linewidth}|}
\hline
\textbf{Messages} & \textbf{Options} & \textbf{Expected Answer} \\ \hline
La frase "Secondo recenti sondaggi il candidato repubblicano gode di scarsissima popolarità: la possibilità di una sua vittoria si allontana sempre di più" contiene un verbo: & [
``riflessivo diretto'', ``impersonale'', ``passivo'', ``intransitivo pronominale''
] & D
\\
\hline
\end{tabular}
\caption{Illustrative example from the ITALIC benchmark.}
\label{tab:italic-example}
\end{table}

\subsection{Italian Benchmark - ITALIC}

ITALIC is a large-scale benchmark introduced to assess the ability of large language models to understand Italian language, culture, and commonsense knowledge. The dataset comprises 10{,}000 multiple-choice questions drawn from authentic public examination materials and covers 12 domains across two broad dimensions: \textit{Culture and Commonsense} and \textit{Language Capability}. By focusing on culturally grounded and linguistically rich Italian content, ITALIC provides a more representative evaluation setting for Italian than benchmarks obtained through simple translation from English\cite{seveso2025italic}. 

While ITALIC is not natively supported by the \texttt{lm\_evaluation\_harness} library, we employed a dual-evaluation strategy to ensure a robust and fair assessment of the models. The original ITALIC evaluation suite structures prompts as a multi-turn conversation, injecting few-shot examples and the target question as a history of prior chat messages. However, to decouple intrinsic factual knowledge from chat-formatting adherence, we evaluated the benchmark using both the original repository's chat-based format (manually integrated into our codebase) and custom task configurations designed for native compatibility with the \texttt{lm\_evaluation\_harness}.

For the original chat-formatted evaluation, we maintained the two distinct prompting strategies available in the original implementation: a "fast" variant, which strictly instructs the model to output only the correct letter, and a "chain-of-thought" (\emph{CoT}) variant, which prompts the model for a brief reasoning trace. Both utilize a 5-shot conversational setting. While we adopted the original implementation, we introduced specific enhancements to ensure reliable extraction. For the \emph{CoT} variant, we implemented more robust parsing rules to mitigate false negatives caused by functionally correct answers being misclassified. Furthermore, we refined the few-shot formatting to improve structural clarity, explicitly formulating the example answers as 'Risposta: A' rather than a standalone 'A' (see Appendix \ref{appendix:italic_config} for exact changes).

In our custom \texttt{lm\_evaluation\_harness} integration, questions are provided directly to the models, stripping away the conversational overhead. All evaluations within this harness were conducted in a 5-shot setting. To systematically isolate the effect of the prompt structure, we implemented this integration in two phases. First, we evaluated the models using the exact same prompt wording as the original chat-based version, simply adapted for direct input; we designated these variants as \texttt{italic\_fast} and \texttt{italic\_cot} (see Appendix \ref{appendix:italic_config}). Second, we explored the impact of more highly structured prompting by adapting templates from established benchmarks. This resulted in a standard non-\emph{CoT} variant (\texttt{italic\_fast\_custom}) and a \emph{CoT} variant (\texttt{italic\_cot\_custom}), whose prompt templates were directly translated from the \texttt{arc\_challenge\_chat} and \texttt{mmlu\_cot\_llama} tasks, respectively (see Appendix \ref{appendix:italic_config}).

Adapting the \emph{CoT} variants (\texttt{italic\_cot} and \texttt{italic\_cot\_custom}) for the direct-prompting harness required addressing a limitation in the source dataset: the standard few-shot examples provide only the final answer without underlying reasoning traces. Consequently, we could not rely on the standard \texttt{num\_fewshot} parameter for dynamic prompt construction. Instead, we utilized Gemini 3.1 Pro to synthetically generate accurate, high-quality explanations for the five baseline few-shot examples. These generated reasoning traces were statically embedded into the prompt templates of both \emph{CoT} configurations to properly condition the models for deliberative generation.

As with previous tasks, the entirety of the model's generated text is considered, and overall performance is quantified using accuracy as the primary metric. Analyzing the results across both evaluation formats (chat-based versus direct) and prompting modes (fast versus \emph{CoT}) provides a highly nuanced perspective on model capabilities. This multi-faceted approach allows us to accurately determine whether a model's performance on Italian-centric tasks genuinely benefits from deliberative processing, or if it is fundamentally constrained by an underlying lack of localized knowledge or an inability to navigate complex conversational templates.

\begin{table}[th]
\centering
\scriptsize
\begin{tabular}{|p{0.75\linewidth}|p{0.15\linewidth}|}
\hline
\textbf{Prompt} & \textbf{Desired Output} \\ \hline
Memorize and track the chain(s) of variable assignment hidden in the following text.  & ["VYI", "SJZ",\\  
The grass is green. The sky is blue. The sun is yellow. Here we go. There and back again. &   "OFG", "LHF", \\ 
The grass is green. The sky is blue. The sun is yellow. Here we go. There and back again. &  "EVV"]\\
$\dots$ &   \\
VAR VYI = 71476 The grass is green. The sky is blue. The sun is  $\dots$ & \\
The grass is green. The sky is blue. The sun is yellow. Here we go. There and back again. &  \\
$\dots$ &   \\
VAR SJZ = VAR VYI The grass is green. The sky is blue. The sun is  $\dots$  & \\
The grass is green. The sky is blue. The sun is yellow. Here we go. There and back again. & \\
$\dots$ & \\
VAR OFG = VAR SJZ The grass is green. The sky is blue. The sun is  $\dots$  & \\
The grass is green. The sky is blue. The sun is yellow. Here we go. There and back again. & \\
$\dots$ & \\
VAR LHF = VAR OFG The grass is green. The sky is blue. The sun is  $\dots$  & \\
The grass is green. The sky is blue. The sun is yellow. Here we go. There and back again. & \\
$\dots$ & \\
VAR EVV = VAR LHF  The grass is green. The sky is blue. The sun is  $\dots$  & \\
The grass is green. The sky is blue. The sun is yellow. Here we go. There and back again. & \\
$\dots$ & \\
Find all the variables that are assigned the value 71476 in the text above. & \\
A: & \\ 
\hline
\end{tabular}
\caption{Illustrative example from the RULER benchmark.}
\label{tab:ruler-example}
\end{table}

\subsection{Long context benchmark - RULER}

RULER is a synthetic benchmark proposed for assessing the long-context capabilities of large language models (LLMs). It builds upon the vanilla \textit{Needle-in-a-Haystack} \cite{nelson2024needle} test by incorporating multiple variations that differ in the types and quantities of needles inserted into the context. As a result, RULER provides a more fine-grained evaluation of a model's ability to identify and utilize relevant information within extended input contexts.

For our evaluation, we utilized the default variant of the benchmark. To systematically assess performance across varying context sizes, we tested the models using maximum sequence lengths of 4,096, 16,384, and 32,768 tokens. For models whose context windows are smaller than the target sequence length, the input prompt is truncated to their maximum supported capacity. It should be noted that, despite this truncation, a model may still successfully retrieve the target information if the inserted 'needle' happens to fall within the retained portion of the context. Consistent with our methodology for other tasks, the generation phase was configured using the \texttt{generate\_until} output type, ensuring that the entirety of the model's generated text is considered. Finally, the overall performance is measured by computing the aggregated metric independently for each specific sequence length.

\section{Results and Discussion}

In this Section, we illustrate and discuss the main results of the evaluation. For complete examples of model responses for each of the evaluated benchmarks, see Appendix \ref{appendix:examples}. Additionally, Table~\ref{tab:throughputs} in the Appendix contains detailed reports on the time required to run each benchmark for each model in our evaluation environment, as well as an estimation of the model throughput under ideal conditions. Statistical tests were conducted to compare results across models and evaluate their significance, with the corresponding outcomes reported in Appendix \ref{app:pairwise-significance-heatmaps}.

\begin{table}[th]
\centering
\footnotesize
\begin{tabular}{lrrr}
\toprule
& \multicolumn{3}{c}{\textbf{arc\_challenge}} \\
\cmidrule(lr){2-4}
 & \textbf{chat} & \textbf{cot\_custom} & \textbf{cot\_it\_custom} \\
\textbf{Model} & \textbf{exact\_match} & \textbf{flexible\_extract} & \textbf{flexible\_extract} \\
\midrule
EngGPT2-16B-A3B & 71.0\% $\pm$1.3 & 87.2\% $\pm$1.0 & 82.3\% $\pm$1.1 \\
FastwebMIIA-7B & 45.5\% $\pm$1.5 & 56.2\% $\pm$1.4 & 58.7\% $\pm$1.4 \\
Minerva-7B-instruct-v1.0 & 26.2\% $\pm$1.3 & 25.4\% $\pm$1.3 & 26.1\% $\pm$1.3 \\
Velvet-14B & 65.9\% $\pm$1.4 & 74.1\% $\pm$1.3 & 70.8\% $\pm$1.3 \\
LLaMAntino-3-ANITA-8B & 45.7\% $\pm$1.5 & 75.4\% $\pm$1.3 & 75.5\% $\pm$1.3 \\
\hline
Qwen3-4B & 88.0\% $\pm$0.9 & 93.0\% $\pm$0.7 & 89.8\% $\pm$0.9 \\
Qwen3-8B & 90.5\% $\pm$0.9 & 95.1\% $\pm$0.6 & \textbf{93.5}\% $\pm$0.7 \\
Llama-3.2-3B-Instruct & 75.6\% $\pm$1.3 & 73.7\% $\pm$1.3 & 60.0\% $\pm$1.4 \\
Llama-3.1-8B-Instruct & 81.7\% $\pm$1.1 & 80.5\% $\pm$1.2 & 72.0\% $\pm$1.3 \\
gemma-3-4b-it & 75.7\% $\pm$1.2 & 77.8\% $\pm$1.2 & 72.5\% $\pm$1.3 \\
gemma-3-12b-it & 90.0\% $\pm$0.9 & 92.3\% $\pm$0.8 & 87.5\% $\pm$1.0 \\
Ministral-3-8B & 89.6\% $\pm$0.9 & 94.2\% $\pm$0.7 & 68.0\% $\pm$0.8 \\
GPT-5 nano & \textbf{95.1}\% $\pm$0.6 & 94.4\% $\pm$0.7 & 93.0\% $\pm$0.7 \\
\hline
deepseek-moe-16b-chat & 39.8\% $\pm$1.4 & 54.9\% $\pm$1.4 & 46.4\% $\pm$1.5 \\
gpt-oss-20b & 88.1\% $\pm$0.9 & \textbf{95.4}\% $\pm$0.6 & 92.7\% $\pm$0.8 \\
Moonlight-16B-A3B-Instruct & 80.9\% $\pm$1.1 & 78.9\% $\pm$1.2 & 62.4\% $\pm$1.4 \\
\bottomrule
\end{tabular}
\caption{Results for the ARC Challenge (reasoning-intensive) benchmark under different evaluation settings: reasoning disabled, chain-of-thought reasoning, and chain-of-thought reasoning in Italian. Uncertainty in the Accuracy values (after the $\pm$ symbol) is given in terms of their standard error.}
\label{tab:ARC}
\end{table}
 
\subsection{Reasoning-Intensive Benchmark - ARC Challenge}

The results of the ARC Challenge are summarized in Table \ref{tab:ARC}. We use three different settings, the \textit{arc\_challenge\_chat} task where reasoning is not enabled and the exact match is used for evaluation, the \textit{arc\_challenge\_cot\_custom} chain-of-thought task and its Italian version \textit{arc\_challenge\_cot\_it\_custom}, both evaluated under flexible extraction. Examples of model responses under the different settings can be found in Appendix \ref{appendix:arc_challenge_examples}, which showcase the ability of different models to handle them correctly.
Under these settings, EngGPT2-16B-A3B achieves 71.0\% exact match on the \textit{chat} variant, and 87.2\% and 82.3\% under flexible extraction on the \textit{cot\_custom} and \textit{cot\_it\_custom} variants, respectively. The model outperforms all Italian-oriented systems, but remains below stronger multilingual models such as Qwen3-8B and Qwen3-4B.

In the \textit{arc\_challenge\_chat} setting, we observe a high variability in the results, due to the exact match logic for answer extraction. Among the italian models, EngGPT2-16B-A3B and Velvet-14B show comparable performance under a statistical t-test performed using Bonferroni correction \textcolor{black}{(see Figure \ref{fig:arc-challenge-chat-exact-match})}. Despite being the best performing among the italian group, the most accurate results are achieved by international models, with GPT-5 nano obtaining the best results. 
Several models exhibit a significant gap between the standard setting and CoT-based variants. For example, LLaMAntino-3-ANITA-8B-Inst-DPO-ITA improves from 45.7\% on \textit{arc\_challenge\_chat} to 75.4\% on \textit{arc\_challenge\_cot\_custom}, indicating sensitivity to the evaluation protocol and to the prompting - which is different in the two evaluation setting -, as well as effectiveness of Chain-of-Thought reasoning in unlocking latent knowledge in some of the models. Similarly,  EngGPT2-16B-A3B benefits from the CoT protocol, confirming its best performance among italian models on the \textit{cot\_custom} and \textit{cot\_it\_custom} variants with statistical significance. Again, there still exist a statistically significant difference in favour of the best international models like GPT-5 nano, Qwen3-8B and gpt-oss-20b, which is shown \textcolor{black}{in Figure \ref{fig:arc-challenge-cot-it-fixed-flexible-extract}}.



\begin{table}[t]
\centering
\footnotesize
\begin{tabular}{l r}
\toprule
 & \textbf{GSM8K} \\
\textbf{Model} & \textcolor{black}{\textbf{strict\_match}} \\
\midrule
EngGPT2-16B-A3B & 86.3\% $\pm$0.9 \\
FastwebMIIA-7B & 53.1\% $\pm$1.4 \\
Minerva-7B-instruct-v1.0 & 9.0\% $\pm$0.8 \\
Velvet-14B & 61.0\% $\pm$1.3 \\
LLaMAntino-3-ANITA-8B-Inst-DPO-ITA & 69.0\% $\pm$1.3 \\
\hline
Qwen3-4B & 86.8\% $\pm$0.9 \\
Qwen3-8B & 94.3\% $\pm$0.6 \\
Llama-3.2-3B-Instruct & 77.9\% $\pm$1.1 \\
Llama-3.1-8B-Instruct & 85.7\% $\pm$1.0 \\
gemma-3-4b-it & 81.6\% $\pm$1.1 \\
gemma-3-12b-it & 90.7\% $\pm$0.8 \\
Ministral-3-8B-Instruct-2512-BF16 & 75.1\% $\pm$1.2 \\
GPT-5 nano & 94.4\% $\pm$0.6 \\
\hline
deepseek-moe-16b-chat & 51.5\% $\pm$1.4 \\
gpt-oss-20b & 92.5\% $\pm$0.7 \\
Moonlight-16B-A3B-Instruct & 73.3\% $\pm$1.2 \\
\bottomrule
\end{tabular}
\caption{Results on the GSM8K (arithmetic problem solving) benchmark dataset.}
\label{tab:gsm8k}
\end{table}

\subsection{Arithmetic \& Mathematical Benchmarks - GSM8K \& AIME-2024/25}

Tables \ref{tab:gsm8k}-\ref{tab_aime2024-results}-\ref{tab_aime2025-results} present the results on \textit{GSM8K} and on the more challenging \textit{AIME24} and \textit{AIME25} benchmarks, respectively. All the evaluated models show a marked difference between arithmetic problem solving (GSM8K) and competition-level mathematical reasoning (AIME24-25). While several models obtain competitive scores on \textit{GSM8K}, FastwebMIIA-7B, Minerva-7B-instruct-v1.0, Velvet-14B, LLaMAntino-3-ANITA-8B, Llama-3.1-8B-Instruct and deepseek-moe-16b-chat collapse to 0.0\% on both \textit{AIME} benchmarks, indicating that strong performance on arithmetic problems does not necessarily translate into success on multi-step reasoning-heavy olympiad-style mathematics. The inspection of the failure cases for these zero-scoring models (examples of which can be found in Appendix \ref{appendix:aime24_examples} and \ref{appendix:aime25_examples}) reveals that this collapse is a genuine reflection of mathematical shortcomings rather than merely an artifact of strict evaluation parsing. While we observed instances of extraction failures where a model provided an answer without the required syntax (i.e., placed inside \verb|\boxed{}|), a manual review of these unparsed outputs confirmed that the underlying mathematical conclusions were fundamentally incorrect in the vast majority of cases. In this context, EngGPT2-16B-A3B stands out among Italian-oriented models as the only system to achieve non-zero results on both \textit{AIME24} and \textit{AIME25}, while also maintaining a strong \textit{GSM8K} score of 86.3\%. Notably, this family of mathematical and reasoning tasks emerges as the strongest evaluation domain for EngGPT2-16B-A3B; within this specific context, the model not only leads the Italian-focused baselines but also occasionally approaches the performance levels of the best international open models.

\begin{table}[th]
\centering
\footnotesize
\begin{tabular}{l r r r r}
\toprule
 & \multicolumn{4}{c}{\textbf{aime24\_pass\_avg\_8}} \\
\cmidrule(lr){2-5}
\textbf{Model} & \textbf{avg@1,e\_m} & \textbf{avg@8,e\_m} & \textbf{pass@1,e\_m} & \textbf{pass@8,e\_m} \\
\hline
EngGPT2-16B-A3B & 53.3\% $\pm$9.3 & 42.5\% $\pm$6.8 & 53.3\% $\pm$9.3 & 73.3\% $\pm$8.2 \\
FastwebMIIA-7B & 0.0\% $\pm$0.0 & 0.0\% $\pm$0.0 & 0.0\% $\pm$0.0 & 0.0\% $\pm$0.0 \\
Minerva-7B-instruct-v1.0 & 0.0\% $\pm$0.0 & 0.0\% $\pm$0.0 & 0.0\% $\pm$0.0 & 0.0\% $\pm$0.0 \\
Velvet-14B & 0.0\% $\pm$0.0 & 0.0\% $\pm$0.0 & 0.0\% $\pm$0.0 & 0.0\% $\pm$0.0 \\
LLaMAntino-3-ANITA-8B & 0.0\% $\pm$0.0 & 0.0\% $\pm$0.0 & 0.0\% $\pm$0.0 & 0.0\% $\pm$0.0 \\
\hline
Qwen3-4B & 60.0\% $\pm$9.1 & 66.2\% $\pm$7.3 & 60.0\% $\pm$9.1 & 80.0\% $\pm$7.4 \\
Qwen3-8B & 70.0\% $\pm$8.5 & 71.2\% $\pm$7.2 & 70.0\% $\pm$8.5 & \textbf{86.7}\% $\pm$6.3 \\
Llama-3.2-3B-Instruct & 3.3\% $\pm$3.3 & 3.3\% $\pm$3.3 & 3.3\% $\pm$3.3 & 3.3\% $\pm$3.3 \\
Llama-3.1-8B-Instruct & 0.0\% $\pm$0.0 & 0.0\% $\pm$0.0 & 0.0\% $\pm$0.0 & 0.0\% $\pm$0.0 \\
gemma-3-4b-it & 13.3\% $\pm$6.3 & 11.7\% $\pm$5.6 & 13.3\% $\pm$6.3 & 13.3\% $\pm$6.3 \\
gemma-3-12b-it & 16.7\% $\pm$6.9 & 18.3\% $\pm$6.8 & 16.7\% $\pm$6.9 & 23.3\% $\pm$7.8 \\
Ministral-3-8B & 23.3\% $\pm$7.8 & 23.3\% $\pm$7.8 & 23.3\% $\pm$7.8 & 23.3\% $\pm$7.8 \\
GPT-5 nano & \textbf{73.3\%} $\pm$8.2 & \textbf{75.0\%} $\pm$6.7 & \textbf{73.3\%} $\pm$8.2 & \textbf{86.7\%} $\pm$6.3 \\
\hline
deepseek-moe-16b-chat & 0.0\% $\pm$0.0 & 0.0\% $\pm$0.0 & 0.0\% $\pm$0.0 & 0.0\% $\pm$0.0 \\
gpt-oss-20b & 70.0\% $\pm$8.5 & 67.5\% $\pm$7.2 & 70.0\% $\pm$8.5 & 83.3\% $\pm$6.9 \\
Moonlight-16B-A3B-Instruct & 3.3\% $\pm$3.3 & 3.3\% $\pm$3.3 & 3.3\% $\pm$3.3 & 3.3\% $\pm$3.3 \\
\bottomrule
\end{tabular}
\caption{Results on the AIME-2024 (competition-level mathematical reasoning) benchmark dataset.}
\label{tab_aime2024-results}
\end{table}

\begin{table}[th]
\centering
\footnotesize
\begin{tabular}{l r r r r}
\toprule
 & \multicolumn{4}{c}{\textbf{aime25\_pass\_avg\_8}} \\ 
\cmidrule(lr){2-5}
\textbf{Model} & \textbf{avg@1,e\_m} & \textbf{avg@8,e\_m} & \textbf{pass@1,e\_m} & \textbf{pass@8,e\_m} \\ 
\hline
EngGPT2-16B-A3B & 30.0\% $\pm$8.5 & 31.7\% $\pm$7.4 & 30.0\% $\pm$8.5 & 50.0\% $\pm$9.3 \\
FastwebMIIA-7B & 0.0\% $\pm$0.0 & 0.0\% $\pm$0.0 & 0.0\% $\pm$0.0 & 0.0\% $\pm$0.0 \\
Minerva-7B-instruct-v1.0 & 0.0\% $\pm$0.0 & 0.0\% $\pm$0.0 & 0.0\% $\pm$0.0 & 0.0\% $\pm$0.0 \\
Velvet-14B & 0.0\% $\pm$0.0 & 0.0\% $\pm$0.0 & 0.0\% $\pm$0.0 & 0.0\% $\pm$0.0 \\
LLaMAntino-3-ANITA-8B & 0.0\% $\pm$0.0 & 0.0\% $\pm$0.0 & 0.0\% $\pm$0.0 & 0.0\% $\pm$0.0 \\
\hline
Qwen3-4B & 60.0\% $\pm$9.1 & 58.3\% $\pm$8.3 & 60.0\% $\pm$9.1 & 70.0\% $\pm$8.5 \\
Qwen3-8B & 63.3\% $\pm$8.9 & 61.2\% $\pm$7.7 & 63.3\% $\pm$8.9 & 76.7\% $\pm$7.9 \\
Llama-3.2-3B-Instruct & 0.0\% $\pm$0.0 & 0.0\% $\pm$0.0 & 0.0\% $\pm$0.0 & 0.0\% $\pm$0.0 \\
Llama-3.1-8B-Instruct & 0.0\% $\pm$0.0 & 0.0\% $\pm$0.0 & 0.0\% $\pm$0.0 & 0.0\% $\pm$0.0 \\
gemma-3-4b-it & 13.3\% $\pm$6.3 & 13.3\% $\pm$6.3 & 13.3\% $\pm$6.3 & 13.3\% $\pm$6.3 \\
gemma-3-12b-it & 13.3\% $\pm$6.3 & 17.5\% $\pm$6.8 & 13.3\% $\pm$6.3 & 20.0\% $\pm$7.4 \\
Ministral-3-8B & 26.7\% $\pm$8.2 & 26.7\% $\pm$8.2 & 26.7\% $\pm$8.2 & 26.7\% $\pm$8.2 \\
GPT-5 nano & 73.3\% $\pm$8.2 & \textbf{72.9\%} $\pm$6.7 & 73.3\% $\pm$8.2 & 86.7\% $\pm$6.3 \\
\hline
deepseek-moe-16b-chat & 0.0\% $\pm$0.0 & 0.0\% $\pm$0.0 & 0.0\% $\pm$0.0 & 0.0\% $\pm$0.0 \\
gpt-oss-20b & \textbf{76.7\%} $\pm$7.8 & 66.7\% $\pm$6.1 & \textbf{76.7\%} $\pm$7.9 & \textbf{93.3\%} $\pm$4.6 \\
Moonlight-16B-A3B-Instruct & 6.7\% $\pm$4.6 & 9.6\% $\pm$5.3 & 6.7\% $\pm$4.6 & 10.0\% $\pm$5.6 \\
\bottomrule
\end{tabular}
\caption{Results on the AIME-2025 (competition-level mathematical reasoning) benchmark dataset.}
\label{tab_aime2025-results}
\end{table}

The detailed \textit{AIME} results (see Tables \ref{tab_aime2024-results}-\ref{tab_aime2025-results}) further show that EngGPT2-16B-A3B benefits substantially from multi-sample evaluation. On \textit{AIME-2024}, its score increases from 40.0\% in single-sample exact\_match to 73.3\% in pass@8; on \textit{AIME-2025}, it improves from 33.3\% to 50.0\%. These performance improvements indicate that the model is often able to recover the correct solution when given multiple attempts, suggesting the presence of non-trivial mathematical reasoning ability even when single-sample decoding is imperfect. When performing a statistical t-test on both AIME benchmarks, the differences between the performance of EngGPT2-MoE-16B and that of the best-performing models are not statistically significant, likely due in part to the small number of samples in the datasets. This places EngGPT2-MoE-16B within a cluster of models showing non-negligible mathematical reasoning ability. Interestingly, under the pass@8 metric, all Italian models but EngGPT2-MoE-16B show statistically significant differences in performance with respect to the English models. The only model that performs significantly better than EngGPT2-MoE-16B under the statistical test is gpt-oss-20b for \texttt{pass@8} on \textit{AIME-2025} (see Table \ref{fig:aime25-pass-avg-8-pass-8-em}).

Additional examples of model responses for the \textit{GSM8K} benchmark can be found in Appendix \ref{appendix:gsm8k_examples}.


\begin{table}[th]
\centering
\footnotesize
\begin{tabular}{l r r r r r}
\toprule
& \textbf{mmlu\_cot} & \multicolumn{4}{c}{\textbf{mmlu\_redux generative}} \\
\cmidrule(lr){3-6}
 & \makecell{\textbf{llama} \\ \textbf{custom}} & 
\makecell{\textbf{humanities}} &
\makecell{\textbf{other} } &
\makecell{\textbf{social} \\ \textbf{sciences} } &
\makecell{\textbf{stem} } \\
\cmidrule(lr){3-6}
\textbf{Model} & \textbf{flexible} & \multicolumn{4}{c}{\textbf{exact match}} \\
\midrule
EngGPT2-16B-A3B & 71.1\% $\pm$0.7 & 72.5\% $\pm$4.4 & 74.7\% $\pm$4.4 & 75.4\% $\pm$4.3 & 72.4\% $\pm$4.5 \\
FastwebMIIA-7B & 47.8\% $\pm$0.8 & 46.4\% $\pm$5.1 & 45.4\% $\pm$5.2 & 46.9\% $\pm$5.0 & 38.6\% $\pm$4.9 \\
Minerva-7B-instruct-v1.0 & 27.9\% $\pm$0.7 & 30.6\% $\pm$4.8 & 31.2\% $\pm$5.0 & 30.3\% $\pm$4.7 & 26.0\% $\pm$4.5 \\
Velvet-14B & 52.6\% $\pm$0.7 & 60.4\% $\pm$4.8 & 59.7\% $\pm$5.1 & 62.9\% $\pm$4.8 & 46.7\% $\pm$4.9 \\
LLaMAntino-3 & 21.2\% $\pm$0.6 & 71.9\% $\pm$4.4 & 73.6\% $\pm$4.5 & 74.3\% $\pm$4.4 & 56.9\% $\pm$4.9 \\
\hline
Qwen3-4B & 79.2\% $\pm$0.6 & 81.0\% $\pm$3.8 & 81.4\% $\pm$4.0 & 83.6\% $\pm$3.7 & 88.8\% $\pm$3.0 \\
Qwen3-8B & 83.0\% $\pm$0.5 & 84.3\% $\pm$3.5 & 83.0\% $\pm$3.6 & 85.6\% $\pm$3.5 & 80.8\% $\pm$4.8 \\
Llama-3.2-3B-Instruct & 65.9\% $\pm$0.7 & 59.5\% $\pm$4.9 & 61.9\% $\pm$5.0 & 60.0\% $\pm$4.9 & 44.7\% $\pm$4.9 \\
Llama-3.1-8B-Instruct & 74.0\% $\pm$0.6 & 72.2\% $\pm$4.3 & 73.5\% $\pm$4.5 & 74.3\% $\pm$4.3 & 56.9\% $\pm$4.9 \\
gemma-3-4b-it & 63.2\% $\pm$0.7 & 64.3\% $\pm$4.6 & 65.2\% $\pm$4.8 & 69.0\% $\pm$4.6 & 50.6\% $\pm$5.0 \\
gemma-3-12b-it & 76.9\% $\pm$0.6 & 77.4\% $\pm$3.9 & 78.0\% $\pm$4.1 & 83.2\% $\pm$3.7 & 66.4\% $\pm$4.6 \\
Ministral-3-8B & 79.7\% $\pm$0.6 & 78.8\% $\pm$3.9 & 80.5\% $\pm$3.9 & 84.0\% $\pm$3.6 & 69.9\% $\pm$4.4 \\
GPT-5 nano & \textbf{87.2\%} $\pm$0.5 & \textbf{87.8\%} $\pm$3.2 & \textbf{89.1\%} $\pm$3.1 & \textbf{88.6\%} $\pm$3.1 & \textbf{92.7\%} $\pm$2.5 \\
\hline
deepseek-moe-16b-chat & 45.3\% $\pm$0.7 & 40.2\% $\pm$5.0 & 39.7\% $\pm$5.2 & 43.9\% $\pm$5.0 & 30.8\% $\pm$4.7 \\
gpt-oss-20b & 83.7\% $\pm$0.5 & 85.8\% $\pm$3.5 & 87.1\% $\pm$3.4 & 87.6\% $\pm$3.2 & 93.5\% $\pm$2.3 \\
Moonlight-16B-A3B-Instruct & 68.9\% $\pm$0.7 & 71.5\% $\pm$4.3 & 72.3\% $\pm$4.6 & 75.7\% $\pm$4.3 & 57.6\% $\pm$4.8 \\
\bottomrule
\end{tabular}
\caption{Comparison of evaluation settings and metrics across tasks, highlighting mmlu\_cot (llama custom) assessed with {\color{black} flexible match}, and four mmlu\_redux generative domain variants: \textit{humanities}, \textit{other}, \textit{social sciences}, and \textit{stem}, each evaluated using exact match.}
\label{tab:mmlu}
\end{table}

\subsection{General Knowledge Benchmarks - MMLU \& MMLU Redux}

In Table~\ref{tab:mmlu}, we summarize the results on several \textit{MMLU} variants, covering both a chain-of-thought evaluation protocol (\textit{MMLU\_cot\_llama\_custom}, strict match) and four generative subsets of \textit{MMLU-Redux} spanning humanities, social sciences, STEM and other domains. The \textit{MMLU} benchmark family is designed to assess academic and professional knowledge breadth across a wide array of subjects, ranging from mathematics and natural sciences to law, history, and medicine.


Among Italian-oriented models, EngGPT2-MoE-16B achieves the highest scores across nearly all subset, consistently outperforming FastwebMIIA-7B and Minerva-7B-instruct-v1.0 . Performance gaps relative to Velvet-14B are also notable, particularly on STEM, where EngGPT2-MoE-16B leads by approximately 25 percentage points. LLaMAntino-3 presents an interesting anomaly: it achieves competitive generative scores on humanities (71.9\%), other (73.6\%), and social sciences (74.3\%), broadly comparable to EngGPT2-MoE-16B, yet its \textit{MMLU\_cot\_llama\_custom} score is markedly low (21.2\%), suggesting that this model generates well-formed responses in open-ended settings but fails to comply with the formatting requirements imposed by chain-of-thought evaluation (See Appendix \ref{appendix:mmlu_examples} for detailed examples). This result may be due to the fact that Llamantino-3 did not undergo neither a specific CoT training nor a reasoning fine-tuning.
On \textit{mmlu\_cot\_llama\_custom}, EngGPT2-MoE-16B performs comparably to the models with the smallest number of active parameters such as Llama-3.2-3B-Instruct, gemma-3-4b-it, and Moonlight-16B-A3B-Instruct, while remaining slightly below the bigger Llama-3.1-8B-Instruct and more clearly behind gemma-3-12b-it, Qwen3-4B, Qwen3-8B, and gpt-oss-20b.

Across the \textit{MMLU-Redux} generative subsets, EngGPT2-MoE-16B remains consistently strong, and compares favorably with several dense and MoE models. In particular, it stays close to Llama-3.1-8B-Instruct and Moonlight-16B-A3B-Instruct, while clearly outperforming gemma-3-4b-it and deepseek-moe-16b-chat. However, stronger references such as the dense models gemma-3-12b-it, Qwen3-4B, and especially the MoE gpt-oss-20b  remain ahead overall.

\begin{table}[th]
\centering
\footnotesize
\begin{tabular}{lrr}
\toprule
 & \multicolumn{2}{c}{\textbf{humaneval\_8\_instruct\_cot}} \\
\cmidrule(lr){2-3}
\textbf{Model} & \textbf{pass@1} & \textbf{pass@8} \\
\midrule
EngGPT2-16B-A3B & 45.2\% $\pm$2.0 & 92.1\% $\pm$2.1 \\
FastwebMIIA-7B & 18.5\% $\pm$1.8 & 56.7\% $\pm$3.9 \\
Minerva-7B-instruct-v1.0 & 5.0\% $\pm$1.1 & 16.5\% $\pm$2.9 \\
Velvet-14B & 12.7\% $\pm$1.8 & 36.0\% $\pm$3.8 \\
LLaMAntino-3-ANITA-8B-Inst-DPO-ITA & 26.0\% $\pm$2.2 & 57.9\% $\pm$3.9 \\
\hline
Qwen3-4B & 93.3\% $\pm$1.3 & \textbf{99.4}\% $\pm$0.6 \\
Qwen3-8B & 80.1\% $\pm$1.7 & 97.6\% $\pm$1.2 \\
Llama-3.2-3B-Instruct & 33.9\% $\pm$2.9 & 59.2\% $\pm$3.8 \\
Llama-3.1-8B-Instruct & 28.7\% $\pm$2.7 & 54.9\% $\pm$3.9 \\
gemma-3-4b-it & 64.1\% $\pm$2.9 & 81.1\% $\pm$3.1 \\
gemma-3-12b-it & 76.2\% $\pm$2.9 & 87.8\% $\pm$2.6 \\
Ministral-3-8B-Instruct-2512-BF16 & 25.1\% $\pm$1.5 & 80.5\% $\pm$3.1 \\
GPT-5 nano & 84.8\% $\pm$1.6 & 98.8\% $\pm$0.9 \\
\hline
deepseek-moe-16b-chat & 17.8\% $\pm$1.7 & 54.9\% $\pm$3.9 \\
gpt-oss-20b & \textbf{94.7}\% $\pm$1.0 & \textbf{99.4}\% \textcolor{black}{$\pm$0.6} \\
Moonlight-16B-A3B-Instruct & 61.1\% $\pm$2.6 & 87.2\% $\pm$2.6 \\
\bottomrule
\end{tabular}
\caption{HumanEval results under the \textit{humaneval\_8\_instruct\_cot} evaluation setting.}
\label{tab:humaneval}
\end{table}

\subsection{Coding - HumanEval} 

Table~\ref{tab:humaneval} reports pass@1 and pass@8 results on \textit{humaneval\_8\_instruct\_cot}, a benchmark for functional code generation under a chain-of-thought instruction-following setting. Since pass@k measures the probability that at least one among $k$ sampled solutions passes all unit tests, pass@1 reflects single-sample reliability, whereas pass@8 better captures the problem-solving capabilities under repeated sampling.

EngGPT2-MoE-16B achieves the highest score on both metrics among italian models. It also outperforms international models such as Llama-3.1-8B-Instruct, Llama-3.2-3B-Instruct, and deepseek-moe-16b-chat. While the strongest overall systems remain gpt-oss-20b, Qwen3-4B, GPT-5 nano, and Qwen3-8B, EngGPT2-MoE-16B stands out as strong coding model.

A notable pattern is the large improvement seen for some models when comparing the pass@1 and the pass@8 metrics, which is particularly pronounced for EngGPT2-MoE-16B. Its increase from 45.2\% to 92.0\% is one of the largest among the evaluated models, indicating that it often finds correct solutions when allowed multiple attempts. This suggests that EngGPT2-MoE-16B has substantial underlying code reasoning capacity, even if its first-sample reliability still lags behind the very strongest reference models. Examples of model responses for the best performing models can be found in Appendix \ref{appendix:humaneval_examples}.

\begin{table}[!ht]
\centering
\footnotesize
\begin{threeparttable}
\setlength{\tabcolsep}{4pt}
\begin{tabular}{lrrrr|r}
\toprule
\textbf{Model} &
\makecell{\textbf{single\_turn} \\ \textbf{non\_live}} &
\makecell{\textbf{single\_turn} \\ \textbf{live}} &
\makecell{\textbf{multi\_turn}} &
\makecell{\textbf{hallucination}} &
\makecell{\textbf{overall}\tnote{1}} \\
\midrule
EngGPT2-16B-A3B\tnote{F} & 73.8\% $\pm$1.4 & 65.3\% $\pm$1.3 & 1.0\% $\pm$1.8 & 56.1\% $\pm$0.4 & 49.1\% $\pm$0.7 \\
FastwebMIIA-7B\tnote{P} & 8.7\% $\pm$0.9 & 13.2\% $\pm$0.9 & 0.0\% $\pm$1.6 & 76.0\% $\pm$0.0 & 24.5\% $\pm$0.5 \\
Minerva-7B-instruct-v1.0\tnote{P} & 0.0\% $\pm$0.0 & 0.0\% $\pm$0.0 & 0.0\% $\pm$0.1 & 99.9\% $\pm$0.0 & 25.0\% $\pm$0.0 \\
Velvet-14B\tnote{P} & 5.8\% $\pm$0.6 & 2.8\% $\pm$0.4 & 0.0\% $\pm$1.0 & 94.1\% $\pm$0.0 & 25.7\% $\pm$0.3 \\
LLaMAntino-3-ANITA-8B\tnote{P} & 39.6\% $\pm$1.6 & 35.5\% $\pm$1.3 & 0.1\% $\pm$1.4 & 78.1\% $\pm$0.1 & 38.3\% $\pm$0.6 \\
\midrule
Qwen3-4B\tnote{F} & 83.2\% $\pm$1.1 & \textbf{81.8\%} $\pm$1.0 & 29.8\% $\pm$1.4 & 79.1\% $\pm$1.6 & \textbf{68.5\%} $\pm$0.7 \\
Qwen3-8B\tnote{F} & \textbf{84.2\%} $\pm$1.1 & \textbf{81.1\%} $\pm$1.1 & 30.8\% $\pm$1.5 & 77.8\% $\pm$1.6 & \textbf{68.5\%} $\pm$0.7 \\
Llama-3.2-3B-Instruct\tnote{P} & \textbf{84.5\%} $\pm$1.2 & 63.1\% $\pm$1.3 & 0.4\% $\pm$1.7 & 33.0\% $\pm$0.2 & 45.3\% $\pm$0.6 \\
Llama-3.1-8B-Instruct\tnote{P} & \textbf{85.2\%} $\pm$1.1 & 70.4\% $\pm$1.2 & 7.4\% $\pm$1.7 & 54.0\% $\pm$0.9 & 54.2\% $\pm$0.6 \\
gemma-3-4b-it\tnote{P} & 64.3\% $\pm$1.5 & 60.4\% $\pm$1.3 & 0.5\% $\pm$1.8 & 47.6\% $\pm$0.2 & 43.2\% $\pm$0.7 \\
gemma-3-12b-it\tnote{P} & 81.8\% $\pm$1.3 & 74.5\% $\pm$1.2 & 6.1\% $\pm$1.6 & 70.4\% $\pm$0.8 & 58.2\% $\pm$0.6 \\
Ministral-3-8B\tnote{F} & 79.1\% $\pm$1.2 & 73.7\% $\pm$1.2 & 15.0\% $\pm$1.4 & 80.1\% $\pm$1.2 & 62.0\% $\pm$0.6 \\
GPT-5 nano\tnote{F} & 65.4\% $\pm$1.5 & 66.8\% $\pm$1.3 & \textbf{37.0\%} $\pm$1.1 & \textbf{88.5\%} $\pm$1.7 & 64.4\% $\pm$0.7 \\
GPT-5 nano\tnote{P} & 76.7\% $\pm$1.4 & 63.6\% $\pm$1.3 & 0.8\% $\pm$1.8 & 45.4\% $\pm$0.3 & 46.6\% $\pm$0.7 \\
\midrule
deepseek-moe-16b-chat\tnote{P} & 10.2\% $\pm$1.0 & 4.6\% $\pm$0.6 & 0.0\% $\pm$1.7 & 75.2\% $\pm$0.0 & 22.5\% $\pm$0.5 \\
gpt-oss-20b\tnote{F} & 33.8\%\tnote{*} $\pm$0.9 & 73.6\%\tnote{*} $\pm$1.1 & 26.5\% $\pm$1.2 & 84.5\% $\pm$1.6 & 54.6\% $\pm$0.6 \\
Moonlight-16B-A3B-Instruct\tnote{P} & 2.2\% $\pm$0.5 & 5.6\% $\pm$0.6 & 0.0\% $\pm$0.4 & 97.5\% $\pm$0.0 & 26.3\% $\pm$0.2 \\
\bottomrule
\end{tabular}%
\begin{tablenotes}
    \footnotesize
    \item[1] The Overall score was computed as the unweighted average of the four categories, following the style of BFCL v3, with the only difference of treating hallucination measurements as a standalone category as in BFCL v4.
    \item[F] Evaluated in native Function Calling mode (see Section \ref{sec:datasets-bfcl} for details).
    \item[P] Evaluated in Prompt mode (see Section \ref{sec:datasets-bfcl} for details).
    \item[*] The Harmony response format used by the gpt-oss-20b model has a known problem (see the open issue at \href{https://github.com/openai/harmony/issues/68}{https://github.com/openai/harmony/issues/68}) with parallel tool calls, which resulted in a score of zero in all parallel and parallel multiple tasks, thus lowering the average accuracy.
\end{tablenotes}
\end{threeparttable}
\caption{BFCL tool calling results. Comparison of model performance across single-turn, multi-turn, hallucination, and overall evaluation settings.}
\label{tab:model_eval_comparison}
\end{table}

\subsection{Tool Calling - BFCL} 

Table~\ref{tab:model_eval_comparison} summarizes performance across five evaluation dimensions of the BFCL benchmark: \textit{single-turn non-live}, \textit{single-turn live}, \textit{multi-turn}, \textit{hallucinations}, and the aggregated \textit{Overall} score.

The highest scores on the \textit{Overall} metric are obtained by Qwen3-4B and Qwen3-8B, which record the highest scores also on the live category, and among the best ones in the other categories.
From this perspective, the most important contrast is between EngGPT2-16B-A3B and the top overall models, especially Qwen3-8B and Qwen3-4B. These two models lead the ranking because they combine strong results in both single-turn settings with clearly superior \textit{multi-turn} performance, which ultimately drives their higher \textit{Overall} scores. A second group, including GPT-5 nano, Ministral-3-8B-Instruct-2512-BF16, gpt-oss-20b, Llama-3.1-8B-Instruct, and gemma-3-12b-it, appears somewhat less dominant overall but still more balanced across evaluation dimensions than the other models in the table. Relative to these systems, the key issue for EngGPT2-16B-A3B is not simply absolute performance, but the fact that its profile is much less even across settings.

Indeed, one of the clearest patterns in the table is the difficulty of the \textit{multi-turn} setting, and EngGPT2-16B-A3B is one of the most informative cases in this respect. Its multi-turn score is only 1.00\%, placing it far below the strongest models and much closer to systems whose performance on this dimension collapses almost entirely. This suggests that, for EngGPT2-16B-A3B, the main limitation is not captured by single-turn evaluation alone, but rather emerges during sustained interaction and in highly complex scenarios, involving multiple tool calls across multiple conversation turns. Detailed examples can be found in Appendix \ref{appendix:bfcl_examples}, and show that while able to produce consistent tool calls across multiple conversational turns, the model often slips over some details that make the overall scenario fail.

The hallucination-related results show another interesting pattern: some of the models that really struggle even in simple scenarios - such as Minerva-7B-instruct-v1.0 and Moonlight-16B-A3B-Instruct - reach almost-perfect scores. This is due to the fact that the task with the most samples in this group is the irrelevance task, where the objective is to abstain from using any tool even if some are provided; this in turn causes models that are almost never able to call tools to perform very well, which should be treated as an outlier result. Among the other models, the best performing one is GPT-5 nano, closely followed by other strong models such as gpt-oss-20b, Ministral-3-8B-Instruct-2512-BF16 and both Qwen3 models. The EngGPT2-16B-A3B scores in the lower half of the ranking, clearly showing some ability to distinguish cases in which the provided tools are just noise, but not at the level of state of the art models.

For models that supported native Function Calling mode we evaluated both that and Prompt mode, and reported the best results. Surprisingly, for some models (most notably the Llama-3.1-8B-Instruct and Llama-3.2-3B-Instruct models) not using the structured tool call parser they were trained with but falling back to a more naïve serialization of functions specifications directly in the prompt resulted in higher accuracy across the board. The only exception to this is the GPT-5 nano model, for which Prompt mode was clearly superior for simple tasks (live and non-live categories), while native Function Calling worked a lot better for more complex tasks in the multi-turn category. Inspecting raw results, this seems to be stemming from the fact that - being a very capable model - in simple math-related tasks it is able to get to correct answers without the use of the provided tools, which is considered an invalid result by the BFCL evaluation framework (see Table \ref{tab:bfcl-examples-simple-python-gpt-5} in Appendix \ref{appendix:bfcl_examples} for some examples).

\begin{table}[t]
\centering
\footnotesize
\begin{tabular}{lrrrr}
\toprule
 & \multicolumn{2}{c}{\textbf{italic}} & \multicolumn{1}{c}{\textbf{italic\_fast}} & \multicolumn{1}{c}{\textbf{italic\_cot\_custom}} \\
\cmidrule(lr){2-3} \cmidrule(lr){4-4} \cmidrule(lr){5-5}
\textbf{Model} & \textbf{fast} & \textbf{COT} & \textbf{exact\_match} & \textbf{flexible\_extract} \\
\midrule
EngGPT2-16B-A3B & 59.3\% $\pm$0.5 & 55.9\% $\pm$0.5 & 60.2\% $\pm$0.5 & 68.3\% $\pm$0.5 \\
FastwebMIIA-7B & 59.3\% $\pm$0.5 & 0.3\% $\pm$0.1 & 61.7\% $\pm$0.5 & 53.7\% $\pm$0.5 \\
Minerva-7B-instruct-v1.0 & 49.9\% $\pm$0.5 & 45.0\% $\pm$0.5 & 34.7\% $\pm$0.5 & 32.9\% $\pm$0.5 \\
Velvet-14B & 68.1\% $\pm$0.5 & 64.6\% $\pm$0.5 & 67.2\% $\pm$0.5 & 66.6\% $\pm$0.5 \\
LLaMAntino-3-ANITA-8B-Inst-DPO-ITA & 44.8\% $\pm$0.5 & 67.8\% $\pm$0.5 & 69.9\% $\pm$0.4 & 67.3\% $\pm$0.5 \\
\hline
Qwen3-4B & 65.4\% $\pm$0.5 & 67.6\% $\pm$0.5 & 66.7\% $\pm$0.5 & 71.4\% $\pm$0.4 \\
Qwen3-8B & 70.8\% $\pm$0.5 & 70.9\% $\pm$0.5 & 73.0\% $\pm$0.4 & 77.4\% $\pm$0.4 \\
Llama-3.2-3B-Instruct & 58.4\% $\pm$0.5 & 57.4\% $\pm$0.5 & 57.4\% $\pm$0.5 & 49.6\% $\pm$0.5 \\
Llama-3.1-8B-Instruct & 70.7\% $\pm$0.5 & 69.9\% $\pm$0.5 & 67.5\% $\pm$0.5 & 65.3\% $\pm$0.5 \\
gemma-3-4b-it & 62.5\% $\pm$0.5 & 62.8\% $\pm$0.5 & 62.8\% $\pm$0.5 & 65.6\% $\pm$0.5 \\
gemma-3-12b-it & 76.5\% $\pm$0.5 & 76.5\% $\pm$0.5 & 76.5\% $\pm$0.4 & 79.0\% $\pm$0.4 \\
Ministral-3-8B-Instruct-2512-BF16 & \textbf{87.4\%} $\pm$0.4 & 81.0\% $\pm$0.4 & 80.2\% $\pm$0.4 & 81.3\% $\pm$0.4 \\
GPT-5 nano & 86.8\% $\pm$0.3 & \textbf{86.6\%} $\pm$0.3 & \textbf{86.1\%} $\pm0.4$ & \textbf{86.8\%} $\pm0.3$ \\
\hline
deepseek-moe-16b-chat & 50.9\% $\pm$0.5 & 50.0\% $\pm$0.5 & 48.5\% $\pm$0.5 & 36.8\% $\pm$0.5 \\
gpt-oss-20b & 26.1\% $\pm$0.4 & 65.3\% $\pm$0.5 & 67.2\% $\pm$0.5 & 76.9\% $\pm$0.4 \\
Moonlight-16B-A3B-Instruct & 61.7\% $\pm$0.5 & 60.9\% $\pm$0.5 & 57.0\% $\pm$0.5 & 53.0\% $\pm$0.5 \\
\bottomrule
\end{tabular}
\caption{Combined model performance evaluated on ITALIC, italic\_fast, and italic\_cot\_custom.}
\label{tab:italic_combined}
\end{table}

\subsection{Italian benchmark - ITALIC}
\label{ssec:italic-results}

Table~\ref{tab:italic_combined} presents the performance on the \textit{ITALIC} benchmark. To provide a concise and highly informative comparison between the original chat-based evaluation and the direct-prompting \texttt{lm\_evaluation\allowbreak\_harness} integration, we report a carefully selected subset of the total evaluations. For the direct non-\emph{CoT} settings, we retained \emph{italic\_fast}: its results were virtually identical to the alternative \textit{italic\_fast\allowbreak\_custom} configuration, but it carries the advantage of utilizing the exact wording from the original prompt. For the \emph{CoT} settings, we retained \emph{italic\_cot\_custom}, as it yielded the most stable results across all architectures; the structural improvements it provided over the more fragile \emph{italic\_cot} variant were too significant to ignore.

Overall, the strongest result is obtained by GPT-5 nano, which reaches 86.8\% in the original \emph{fast} setting and remains essentially unchanged under \emph{CoT} (86.6\%), indicating both high accuracy and strong robustness to conversational prompting styles. Among the open models, Ministral-3-8B-Instruct-2512-BF16 is the best-performing system across all reported configurations, peaking at 87.4\% in the original \emph{fast} setting and reaching 81.3\% under the structured \emph{italic\_cot\_custom} setup. These two models consistently achieve the highest scores, surpassing all Italian-oriented models.

When analysing the original chat-based evaluation format (\textit{italic\_fast} and \textit{COT}), chain-of-thought prompting does not seem to yield systematic gains. For most models, the difference between the two settings is negligible or slightly negative. A qualitative inspection of the generated outputs reveals the underlying cause for this behaviour: models exhibiting similar results across both configurations consistently failed to follow the CoT directive. Despite explicit instructions to generate a brief reasoning trace prior to answering, these models bypassed the deliberative step entirely, directly outputting the final answer. Consequently, their responses in the \emph{CoT} setting were structurally identical to those in the fast setting. In contrast, FastwebMIIA-7B exhibits a dramatic collapse under the original \emph{CoT} ($59.3\%\rightarrow0.3\%$). The model's outputs indicate that this degradation stems from a severe formatting failure—frequently halting after the "ASSISTANT:" role indicator—rather than a genuine loss of task competence (see Appendix \ref{ssec:italic_examples} for some examples).

To decouple these chat-formatting issues from true culturally grounded knowledge, we examine the direct-prompt configurations. Across the non-\emph{CoT} variants (comparing original \emph{fast} with \emph{italic\_fast}), model performance generally remains consistent, confirming baseline factual recall. Introducing the structured \emph{italic\_cot\_custom} setting enabled the models to successfully generate intermediate reasoning steps, a stark contrast to their behaviour in the original chat format. Despite generating reasoning, some models still struggled with the final output formatting, as evidenced by the lower \emph{flexible\_extract} scores; for example, deepseek-moe-16b-chat decreased from 50.0\% to 36.8\%, and Minerva-7B-instruct-v1.0 from 45.0\% to 32.9\%.

Interestingly, even when properly adhering to the CoT directive in this structured configuration, a clear dichotomy emerges. Certain models (such as Llama-3.1-8B-Instruct and Velvet-14B) show no meaningful improvement—or even slight degradation—when forced to reason, suggesting a preference for direct fact retrieval. Conversely, models that received extensive post-training focused specifically on reasoning capabilities (such as the Qwen3 family, gpt-oss-20b, and EngGPT2-16B-A3B) exhibited the most significant gains from the \emph{italic\_cot\_custom} prompt. For instance, gpt-oss-20b improved from a baseline \emph{italic\_fast} of 67.2\% to 76.9\%, and EngGPT2-16B-A3B jumped from 60.2\% to 68.3\%. These gains indicate that instruction-tuned models with robust reasoning priors rely heavily on deliberative, highly structured prompting to unlock their latent capabilities on Italian-language tasks.

On \textit{ITALIC}, EngGPT2-16B-A3B remains behind the top-performing international dense models. However, within the group of Italian-oriented models, it demonstrates a distinct advantage in reasoning: while it trails LLaMAntino-3, FastwebMIIA-7B, and Velvet-14B in direct retrieval (\emph{italic\_fast}), it achieves the highest performance among regional models when evaluated with structured reasoning (\emph{italic\_cot\_custom} at 68.3\%). Among MoE architectures, EngGPT2-16B-A3B achieves competitive performance, outperforming deepseek-moe-16b-chat (36.8\%) and Moonlight-16B-A3B-Instruct (53.0\%), but trailing gpt-oss-20b (76.9\%), further highlighting its capacity for deliberative problem-solving in Italian contexts.

\begin{table}[t]
\centering
\footnotesize
\begin{tabular}{lr|rrr}
\toprule
 & \textbf{max} & \multicolumn{3}{c}{\textbf{RULER}} \\
\textbf{Model} & \textbf{context} & \textbf{4k context} & \textbf{16k context} & \textbf{32k context}  \\
\midrule
EngGPT2-16B-A3B & 32k & 79.5\% $\pm$0.5 & 57.9\% $\pm$0.6 & 42.6\% $\pm$0.6  \\
FastwebMIIA-7B & 16k & 74.9\% $\pm$0.5 & 58.7\% $\pm$0.6 & 33.9\% $\pm$0.6 \\
Minerva-7B-instruct-v1,0 & 4k & 55.0\% $\pm$0.6 & 16.1\% $\pm$0.5 & 10.1\% $\pm$0.4  \\
Velvet-14B & 128k & 85.5\% $\pm$0.4 & 34.4\% $\pm$0.6 & 0.0\% $\pm$0.0  \\
LLaMAntino-3-ANITA-8B-Inst-DPO-ITA & 8k & 91.6\% $\pm$0.3 & 48.2\% $\pm$0.6 & 28.5\% $\pm$0.6 \\
\hline
Qwen3-4B & 32k & 92.7\% $\pm$0.3 & 91.0\% $\pm$0.4 & 89.0\% $\pm$0.4  \\
Qwen3-8B & 32k & 94.8\% $\pm$0.3 & 92.0\% $\pm$0.3 & 91.0\% $\pm$0.4 \\
Llama-3.2-3B-Instruct & 128k & 92.5\% $\pm$0.3 & 83.1\% $\pm$0.5 & 77.8\% $\pm$0.5  \\
Llama-3.1-8B-Instruct & 128k & \textbf{95.0\%}$\pm$0.3 & 93.3\% $\pm$0.3 & 87.3\% $\pm$0.4 \\
gemma-3-4b-it & 128k & 90.9\% $\pm$0.4 & 68.5\% $\pm$0.6 & 62.3\% $\pm$0.6 \\
gemma-3-12b-it & 128k & 94.7\% $\pm$0.3 & 87.5\% $\pm$0.4 & 80.4\% $\pm$0.5 \\
Ministral-3-8B-Instruct-2512-BF16  & 256k & 94.6\% $\pm$0.3 & \textbf{93.4\%} $\pm$0.3 & \textbf{91.7\%}$\pm$0.3 \\
GPT-5 nano & 400k & 87.2\% $\pm$0.4 & 91.7\% $\pm$0.3 & 88.9\% $\pm$0.4  \\
\hline
deepseek-moe-16b-chat  & 4k & 77.5\% $\pm$0.5 & 25.2\% $\pm$0.5 & 16.9\% $\pm$0.5 \\
gpt-oss-20b  & 128k & 79.6\% $\pm$0.5 & 77.2\% $\pm$0.5 & 77.1\% $\pm$0.5 \\
Moonlight-16B-A3B-Instruct  & 128k & 92.2\% $\pm$0.3 & 55.7\% $\pm$0.6 & 32.7\% $\pm$0.6 \\
\bottomrule
\end{tabular}
\caption{RULER results under different context lengths (4096, 16384, 32768). Performance for some models on the longer context lengths is hampered by the fact that they have been developed with smaller maximum context lengths, shown in the first column. }
\label{tab:ruler}
\end{table}

\subsection{Long context benchmark - RULER}

Table~\ref{tab:ruler} reports performance on \textit{RULER} as a function of input context length, and reveals substantial heterogeneity in models' ability to maintain performance as the sequence grows. In general, most systems achieve their strongest results at the shortest context setting (4k tokens), while longer contexts introduce varying degrees of degradation. However, the magnitude of this decline differs markedly across models, suggesting that advertised context capacity and effective long-context robustness should be treated as distinct properties.

Among the strongest models, Llama-3.1-8B-Instruct, Qwen3-8B, Qwen3-4B and Ministral-3-8B-Instruct stand out for combining high accuracy with relatively limited degradation across all three settings. Llama-3.1-8B-Instruct achieves $95.0\%$ at 4k tokens and still retains $87.3\%$ at 32k tokens, Qwen3-8B remains highly stable, decreasing only from $94.8\%$ to $91.0\%$, and Ministral-3-8B-Instruct exhibiting the best performance at the longest context length considered. These results indicate that the Qwen3 family, in particular, exhibits strong effective utilization of extended context windows. gpt-oss-20b is the most stable model in relative terms, with performance remaining essentially flat from $79.6\%$ at 4k tokens to $77.1\%$ at 32k tokens.

A second group of models shows good short-context accuracy but more pronounced deterioration as the context increases. This pattern is visible for Llama-3.2-3B-Instruct ($92.5\%\rightarrow77.8\%$), gemma-3-12b-it ($94.7\%\rightarrow80.4\%$), and gemma-3-4b-it ($90.9\%\rightarrow62.3\%$). Although these systems remain functional at longer lengths, the drop is substantial enough to suggest that their nominal support for large contexts does not fully translate into robust retrieval or reasoning over long inputs.

By contrast, several models exhibit severe performance collapse beyond the shortest setting. Velvet-14B declines from $85.5\%$ at 4k tokens to $34.4\%$ at 16k tokens, and reaches $0.0\%$ at 32k tokens despite a reported context length of 128k. Similarly, Moonlight-16B-A3B-Instruct drops from $92.2\%$ to $32.7\%$, while LLaMAntino-3-ANITA-8B-Inst-DPO-ITA falls from $91.6\%$ to $28.5\%$. The same general pattern is observed for EngGPT2-16B-A3B, FastwebMIIA-7B, and deepseek-moe-16b-chat, all of which degrade sharply as context length increases. These results suggest limited robustness in long-context settings and may reflect weaknesses in positional generalization, retrieval across long spans, or instruction-following under high memory load.

For long-context settings, the international models appear better trained to handle extended input sequences. Among the Italian models, although EngGPT2-16B-A3B does not achieve the strongest performance on the shortest contexts, it maintains the highest accuracy at the greatest sequence length.

\begin{figure}[th]
    \centering
    \includegraphics[width=0.75\linewidth]{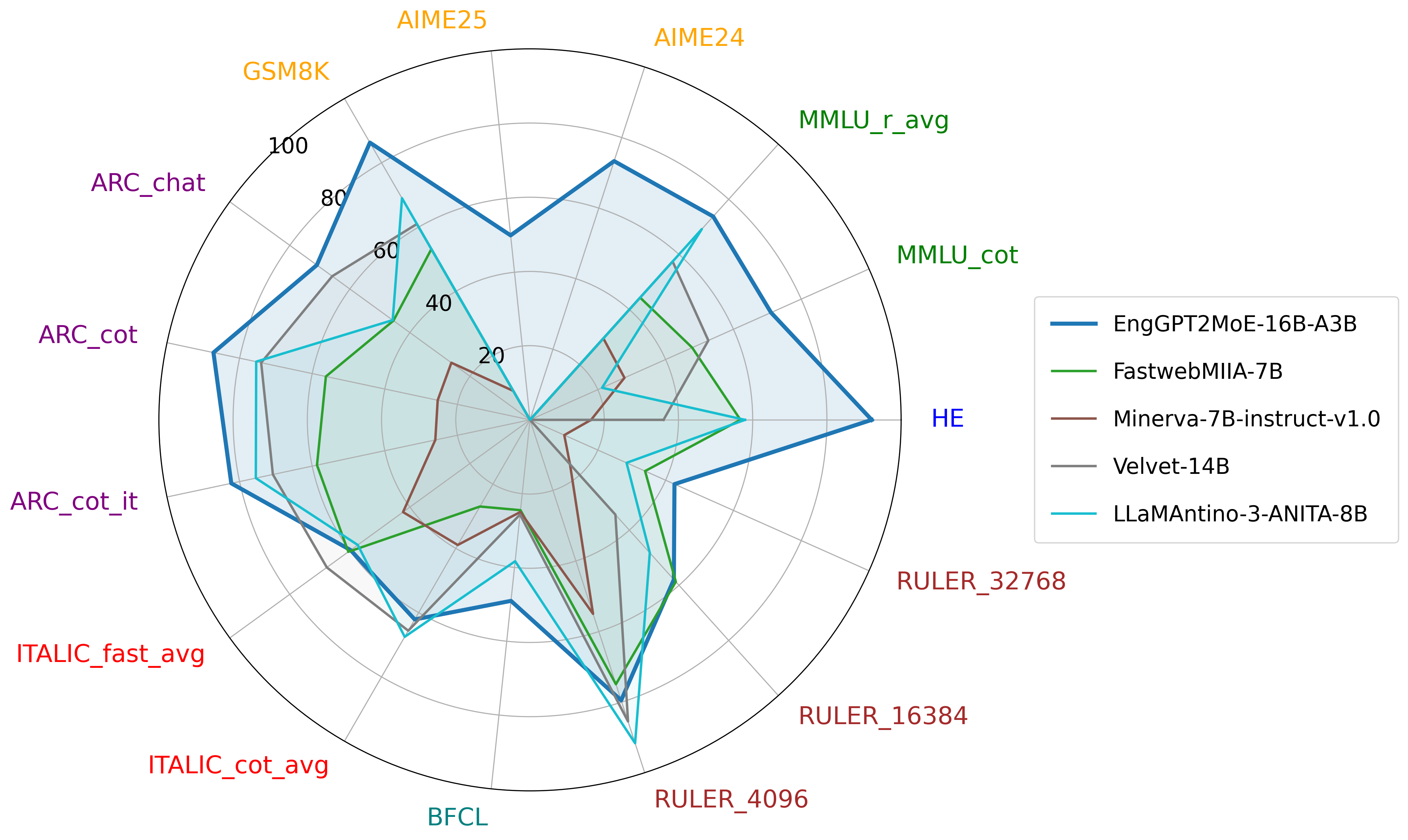}
    \caption{Comparison with Italian models across the considered benchmarks.}
    \label{fig:spider_ita}
\end{figure}

\begin{figure}[th]
    \centering        
    \includegraphics[width=0.75\linewidth]{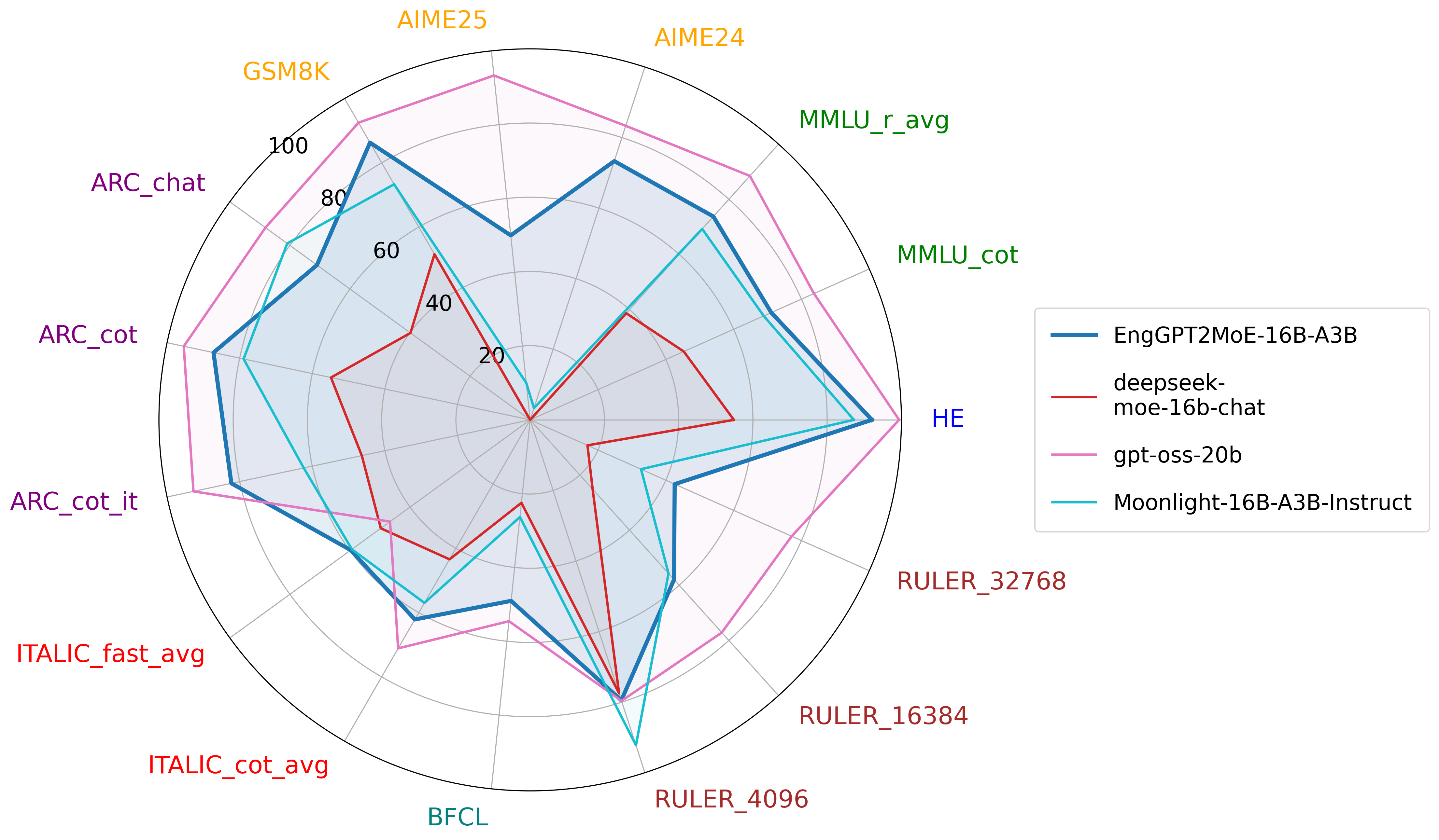}
    \caption{Comparison with Mixture-of-Experts models across the considered benchmarks.}
    \label{fig:spider_moe}
\end{figure}

\begin{figure}[th]
    \centering
    \includegraphics[width=0.75\linewidth]{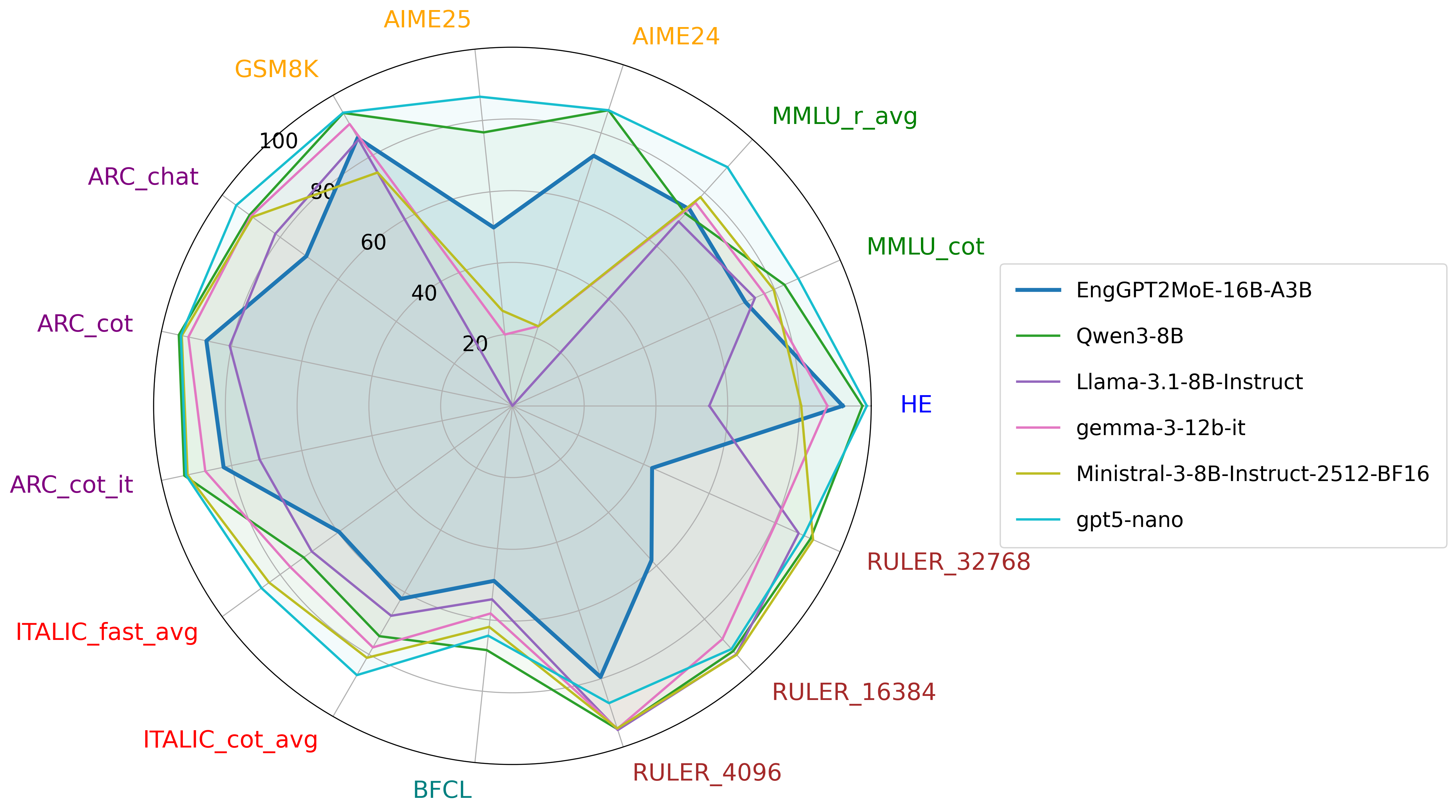}
    \caption{Comparison with the larger dense models across the considered benchmarks.}
    \label{fig:spider_large_dense}
\end{figure}
    
\begin{figure}[th]
    \centering       
    \includegraphics[width=0.75\linewidth]{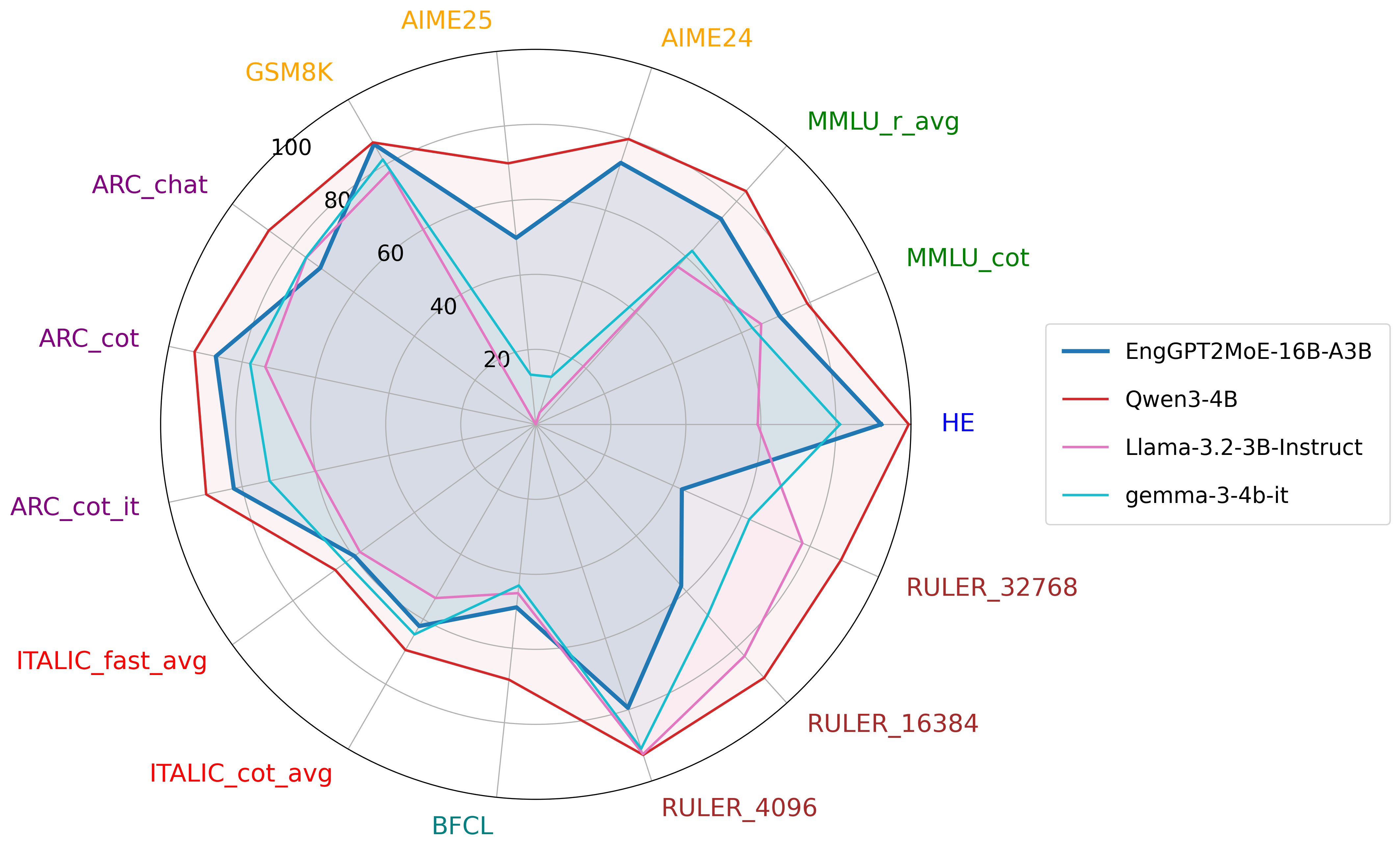}
    \caption{Comparison with the smaller dense models across the considered benchmarks.}
    \label{fig:spider_small_dense}
\end{figure}

\subsection{Visualising Performance across Benchmarks}

Figures \ref{fig:spider_ita} to \ref{fig:spider_small_dense} provide a visual comparison of the performance across the selected benchmarks for various subsets of the models under consideration. To highlight differences in performance, we focus on a set of representative indicators. In particular, the results on \textit{MMLU-Redux} are aggregated into a single average score. Similarly, for a clearer view, we also aggregate the results of the ITALIC dataset computing the mean of the fast and cot settings. The AIME24 and AIME25 scores shown below are the \textit{pass@8}, in order to reflect the potential of the model in solving complex mathematical problems. The different models are divided into four subgroups, to better understand the overall capabilities of similar models: italian-based models, MoE models, larger and smaller dense models.

In Figure \ref{fig:spider_ita}, we can observe that EngGPT2-MoE-16B outperforms the competitor Italian models on most of the evaluated benchmarks. It stands out particularly for its advanced reasoning capabilities, coding and tool calling. Figure \ref{fig:spider_moe} shows the comparison with Mixture-of-Experts models. EngGPT2-MoE-16B is trailing behind gpt-oss-20B on most benchmarks. On the other hand, it shows comparable performance with respect to Moonlight-16B-A3B, while clearly outperforming Deepseek-MoE-16B. Interestingly, when compared with Moonlight-16B-A3B-Instruct, which has the most similar parameters distribution (16 billion total, 3 billion active), EngGPT2-MoE-16B appears to achieve comparable-to-better performance, especially leading in mathematical reasoning and tool use, suggesting that it makes more effective use of its parameters. 

Concerning larger dense models (with more than 5B parameters), Figure \ref{fig:spider_large_dense} shows that Qwen3-8B and GPT5-nano achieve the most balanced and strongest overall performance across the evaluated benchmarks. In contrast, EngGPT2-MoE-16B exhibits a less uniform performance profile. Its results are comparatively solid on challenging mathematical reasoning tasks, particularly the AIME benchmarks, and remain competitive on other reasoning and mathematical benchmarks (ARC, GSM8K). However, the model lags behind the strongest baselines on long-context benchmarks, especially at extended context lengths, and also shows weaker results on Italian and tool-call-related evaluations. Overall, this pattern suggests that EngGPT2-MoE-16B is relatively effective on high-difficulty reasoning tasks, but still requires improvements in italian language, function-calling, and long-context robustness.

Finally, Figure \ref{fig:spider_small_dense} compares EngGPT2-MoE-16B against smaller dense models with a more similar number of active parameters. Qwen3-4B clearly achieves the strongest overall performance, showing the most consistent results across all benchmarks. EngGPT2-MoE-16B, however, remains competitive with the other baselines and often delivers comparable or better performance on several reasoning-oriented tasks. Its main weakness remain long-context evaluation, where performance degrades more visibly, especially on the RULER benchmarks at longer context lengths.

\begin{table}[th]
\centering
\footnotesize
\begin{tabular}{l|rrrrrrrr|r}
\toprule
\textbf{Model} & \textbf{AIME} & \textbf{ARC} &	\textbf{GSM8K} &	\textbf{HE} & \textbf{MMLU} & \textbf{IT} & \textbf{RU} & \textbf{BFCL} & \textbf{Average} \\
\midrule
EngGPT2-16B-A3B & 5.0 & 8.3 & 7.0 & 6.5 & 7.0 & 10.8 & 11.3 & 8.0 & 8.0 $\pm$0.8 \\
FastwebMIIA-7B & 13.3 & 14.0 & 14.5 & 12.5 & 13.8 & 12.3 & 12.0 & 15.0 & 13.4 $\pm$0.4 \\
Minerva-7B & 13.3 & 16.0 & 16.0 & 16.0 & 15.8 & 15.3 & 15.7 & 14.0 & 15.2 $\pm$0.4 \\
Velvet-14B & 13.3 & 11.7 & 13.0 & 15.0 & 12.2 & 7.8 & 13.7 & 13.0 & 12.4 $\pm$0.8 \\
LLaMAntino-3-8B & 13.3 & 10.7 & 12.0 & 11.0 & 10.4 & 8.5 & 11.3 & 11.0 & 11.0 $\pm$0.5 \\ \hline
Qwen3-4B & 3.3 & 5.3 & 5.5 & 1.8 & 4.2 & 7.3 & 4.3 & 2.0 & 4.2 $\pm$0.7 \\
Qwen3-8B & 2.5 & \textbf{1.7} & 2.0 & 4.0 & 3.2 & 4.0 & 2.3 & \textbf{1.0} & 2.6 $\pm$0.4 \\
Llama-3.2-3B & 11.3 & 12.0 & 10.0 & 9.5 & 12.2 & 12.8 & 6.7 & 9.0 & 10.4 $\pm$0.7 \\
Llama-3.1-8B & 13.3 & 8.3 & 7.0 & 11.8 & 8.6 & 6.8 & 2.7 & 7.0 & 8.2 $\pm$1.1 \\
gemma-3-4b-it & 8.0 & 9.3 & 8.8 & 7.0 & 11.0 & 9.5 & 9.0 & 10.0 & 9.1 $\pm$0.4 \\
gemma-3-12b-it & 6.9 & 5.0 & 4.5 & 5.5 & 6.2 & 3.0 & 5.0 & 5.0 & 5.1 $\pm$0.4 \\
Ministral-3-8B & 6.1 & 4.0 & 7.0 & 10.5 & 4.8 & 1.8 & \textbf{2.0} & 4.0 & 5.0 $\pm$1.0 \\
GPT-5 nano & \textbf{1.8} & 2.0 & \textbf{1.0} & 3.0 & \textbf{1.2} & \textbf{1.3} & 6.0 & 3.0 & \textbf{2.4} $\pm$0.6 \\ \hline
deepseek-moe-16b & 13.3 & 15.0 & 14.5 & 13.8 & 14.8 & 14.3 & 14.3 & 16.0 & 14.5 $\pm$0.3 \\
gpt-oss-20b & 2.5 & 3.0 & 3.0 & \textbf{1.3} & 1.8 & 9.3 & 9.3 & 6.0 & 4.5 $\pm$1.2 \\
Moonlight-16B-A3B & 9.3 & 9.7 & 10.3 & 7.0 & 8.8 & 11.8 & 10.3 & 12.0 & 9.9 $\pm$0.6 \\
\bottomrule
\end{tabular}
\caption{Average ranking for each benchmark dataset, computed by averaging the ranks of the model over the metrics for the benchmark. The overall average rank across all benchmarks is shown on the right. Lower rank indicates better performance. (Some model names have been abbreviated as have the benchmark names (HE = HumanEval, IT = ITALIC, RU = Ruler).}
\label{tab:average_ranking}
\end{table}

\begin{figure}[ht]
    \centering
    \includegraphics[width=0.8\textwidth]{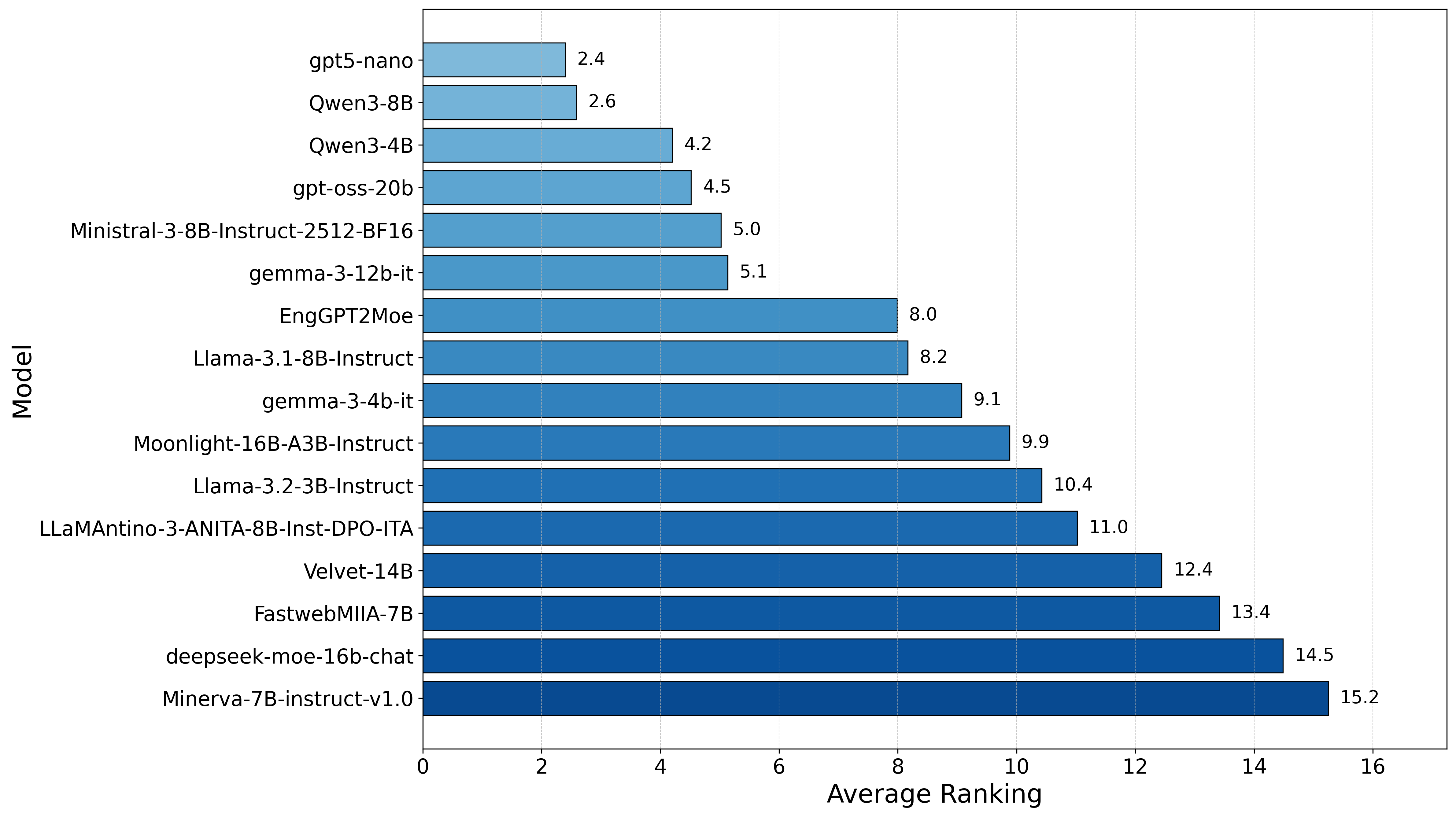}
    \caption{Average ranking of models across the various benchmark datasets. The ranking for each benchmark contributes equally to the overall ranking. For each benchmark, the rank of the model on each metric is first averaged, and then the average rankings for each benchmark are themselves aggregated.}
    \label{fig:avg-ranking}
\end{figure}

\begin{table}[t]
\centering
\footnotesize
\begin{tabular}{l|rrrrrrrr|r}
\toprule
\textbf{Model} & \textbf{AIME} & \textbf{ARC} &	\textbf{GSM8K} &	\textbf{HE} & \textbf{MMLU} & \textbf{IT} & \textbf{RU} & \textbf{BFCL} & \textbf{Average} \\
\midrule
EngGPT2-16B-A3B & 49.2\% & 80.2\% & 86.4\% & 68.6\% & 73.2\% & 60.9\% & 60.0\% & 49.1\% & 65.9\% $\pm$3.6\% \\
FastwebMIIA-7B & 0.0\% & 53.5\% & 53.1\% & 37.6\% & 45.0\% & 43.7\% & 55.9\% & 24.5\% & 39.0\% $\pm$2.1\% \\
Minerva-7B & 0.0\% & 25.9\% & 10.5\% & 10.7\% & 29.2\% & 40.6\% & 27.1\% & 25.0\% & 21.0\% $\pm$1.8\% \\
Velvet-14B & 0.0\% & 70.3\% & 61.0\% & 24.4\% & 56.5\% & 66.6\% & 40.0\% & 25.7\% & 43.0\% $\pm$2.0\% \\
LLaMAntino-3-8B & 0.0\% & 65.6\% & 69.0\% & 42.0\% & 59.6\% & 62.5\% & 56.1\% & 38.3\% & 49.0\% $\pm$2.0\% \\ \hline
Qwen3-4B & 70.8\% & 90.3\% & 87.5\% & 96.3\% & 82.8\% & 67.8\% & 90.9\% & \textbf{68.5\%} & 82.0\% $\pm$3.2\% \\
Qwen3-8B & 74.2\% & 93.0\% & 94.3\% & 88.8\% & 83.3\% & 73.0\% & 92.6\% & \textbf{68.5\%} & 83.0\% $\pm$3.1\% \\
Llama-3.2-3B & 1.7\% & 69.8\% & 78.1\% & 46.5\% & 58.4\% & 55.7\% & 84.5\% & 45.3\% & 55.0\% $\pm$2.3\% \\
Llama-3.1-8B & 0.0\% & 78.1\% & 86.1\% & 41.8\% & 70.2\% & 68.3\% & 91.9\% & 54.2\% & 61.0\% $\pm$2.0\% \\
gemma-3-4b-it & 12.5\% & 75.3\% & 81.5\% & 72.6\% & 62.5\% & 63.4\% & 73.9\% & 43.2\% & 61.0\% $\pm$2.9\% \\
gemma-3-12b-it & 20.0\% & 89.9\% & 90.7\% & 82.0\% & 76.4\% & 77.1\% & 87.5\% & 58.2\% & 73.0\% $\pm$3.1\% \\
Ministral-3-8B & 25.0\% & 92.1\% & 83.7\% & 52.8\% & 78.6\% & 82.5\% & \textbf{93.2\%} & 62.0\% & 71.0\% $\pm$3.3\% \\
GPT-5 nano & \textbf{81.7\%} & \textbf{94.2\%} & \textbf{94.6\%} & 91.8\% & \textbf{89.1\%} & \textbf{86.6\%} & 89.2\% & 64.4\% & \textbf{86.0\%} $\pm$2.8\% \\ \hline
deepseek-moe-16b & 0.0\% & 47.0\% & 53.4\% & 36.4\% & 40.0\% & 46.6\% & 39.9\% & 22.5\% & 36.0\% $\pm$2.1\% \\
gpt-oss-20b & 79.2\% & 92.1\% & 92.9\% & \textbf{97.0\%} & 87.5\% & 58.9\% & 78.0\% & 54.6\% & 80.0\% $\pm$2.8\% \\
Moonlight-16B-A3B & 5.8\% & 74.1\% & 77.4\% & 74.2\% & 69.2\% & 58.2\% & 60.2\% & 26.3\% & 56.0\% $\pm$2.4\% \\
\bottomrule
\end{tabular}
\caption{Aggregated Accuracy values for each benchmark, computed by averaging accuracy values across metrics for the benchmark. The average across benchmarks is  overall average across all categories; higher is better. Model names have been abbreviated to fit the table (full names can be found in Table \ref{tab:chosen_models}). Some benchmark names have been abbreviated to fit the table (HE = HumanEval, IT = ITALIC, RU = Ruler).}
\label{tab:average_accuracy}
\end{table}

\begin{figure}[!ht]
    \centering
    \includegraphics[width=0.8\textwidth]{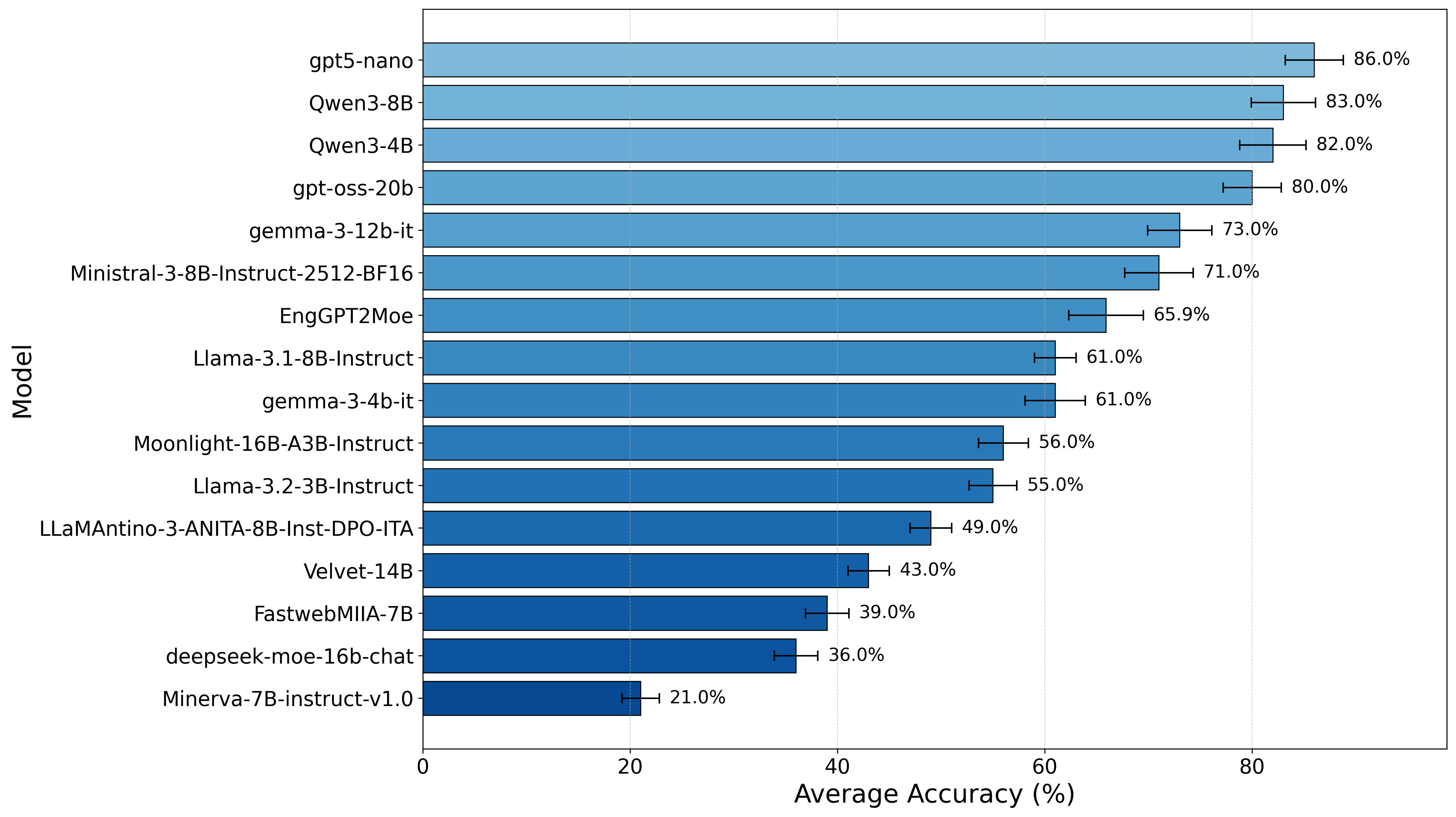}
    \caption{Average of the accuracy score of models across the various benchmarks datasets. Each benchmark contributes equally to the overall average. For each benchmark, the Accuracy score of the model on each metric is first averaged, and then the averages for each benchmark are  themselves aggregated. Error bars show standard errors values, calculated by aggregating variances across the datasets from the standard error estimates for each metric value.}
    \label{fig:median-stderr}
\end{figure}

\subsection{Aggregating Results across Benchmark datasets}

In order to provide a high level comparison across the models, we report aggregate measures of performance across benchmarks for each model in Tables \ref{tab:average_ranking} and \ref{tab:average_accuracy}. We caution the reader when interpreting aggregated results, since the ordering of systems depends heavily on the choice of datasets included in the survey. 

For both tables, we first compute an average score for each benchmark over the individual metrics collected, and then average those results at the dataset level into a single aggregate score. This procedure weights each benchmark equally in terms of importance, regardless of the number of metrics taken. For most benchmarks we considered all reported metrics, with the exception of AIME, where we only considered pass@1 and pass@8 metrics, for both 2024 and 2025 versions. Table \ref{tab:average_ranking} contains a ranking-based aggregate score: for each metric, the performance of each model in terms of accuracy is translated into a relative ranking with respect to other models, breaking ties taking the mid point. Table \ref{tab:average_accuracy} contains an accuracy-based aggregate score: each metric accuracy score is used directly, and the standard error is computed considering the underlying standard error of each metric.

Figures \ref{fig:avg-ranking}-\ref{fig:median-stderr} show the considered models ordered by ranking and average performance, respectively. The ranking-based aggregation clearly highlights several clusters of performance: GPT-5 nano and Qwen3-8B lead all other models, followed by a second group of similarly-performing models including Qwen3-4B, gemma-3-12b-it, Ministral-3-8B and gpt-oss-20b. Then, we find EngGPT2-16B-A3B and Llama-3.1-8B-Instruct, which are almost paired despite having a clearly distinct performance profile, the former leading in reasoning-heavy and coding tasks, the latter leading by a large margin in long-context understanding and less prominently in tool calling.

The more direct accuracy-based aggregation portrays a similar picture, although with minor numerical differences related to the different scale and the different importance given to the actual accuracy gap in each specific benchmark. Four models reach an average accuracy score of 80\% or above: GPT-5 nano, both Qwen3 models and gpt-oss-20b. It is worth noting that, using both aggregation methods, the EngGPT2-16B-A3B largely leads all other Italian models, with a gap of more than 15 percentage points when aggregating the accuracy score.

Overall, the results portray EngGPT2MoE-16B-A3 as one of the most capable models across many of the various considered tasks. Compared with the other Italian-focused systems, it appears to be more robust and versatile. In particular, it is generally more reliable on reasoning-intensive evaluation, clearly stronger on mathematical problem solving, often superior on broad knowledge benchmarks, and solid in code generation. This makes it the only italian model with competitive general-purpose capabilities. At the same time, the model does not yet reach the level of the strongest international dense models, which remain the reference point for overall performance. In particular, Qwen, GPT-5 nano, gpt-oss, and in some cases Ministral or the larger Gemma variant display a more mature overall profile, combining stronger first-try reliability, better compliance with complex evaluation protocols, stronger long-context behavior, and more effective multi-turn tool use. 


\section{Conclusions}

In this work, we evaluated EngGPT2MoE-16B-A3 across a broad set of benchmarks covering reasoning, mathematics, coding, tool use, general knowledge, Italian-language understanding, and long-context capabilities, comparing it with a set of recent state-of-the-art models. The results show that the model achieves a strong and balanced performance profile, consistently ranking above all the other tested Italian-oriented models. At the same time, the evaluation highlights that EngGPT2MoE-16B-A3 remains below the strongest multilingual open models on the most challenging capabilities, especially in long context understanding, complex tool interaction, and high-precision coding tasks.

Overall, these findings position EngGPT2MoE-16B-A3 as one of the most complete Italian-focused models, combining local language specialization with credible general-purpose performance. 




\section{Acknowledgements}


The research was carried out in the context of the European Union–funded “AVANT” project, with Engineering Ingegneria Informatica S.p.A. sponsorship, participating as an industrial research partner.

\printbibliography


\newpage

\appendix


\section{Timings and Model Throughput}
\label{appendix:throughput}

Tables \ref{tab:timings_aime}, \ref{tab:timings_arc}, \ref{tab:timings_gsm8k_humaneval_ruler}, and \ref{tab:timings_mmlu} contain details regarding the wall-clock time taken to run each evaluation benchmark for each model, in the evaluation environment described in Appendix \ref{appendix:technical_specifications}, as well as the number of samples for each benchmarking dataset. It is worth noting that the per-sample time can vary quite dramatically between both different tasks and different models, mainly as a consequence of the number of generated tokens; for instance, tasks requiring long reasoning traces such as AIME24 and AIME25 tipically consume two to three orders of magnitude more tokens than knowledge-based tasks such as MMLU, and models heavily trained to produce long and detailed reasoning traces (e.g., Qwen models) can easily generate hundreds of tokens more even for simple tasks than models that did no undergo specific reasoning training (e.g., Llama3 models).

\begin{table}[h]
\caption{Evaluation Times (in seconds) for the AIME 24 \& AIME 25 tasks}
\label{tab:timings_aime}
\centering
\scriptsize
\begin{threeparttable}
\begin{tabular}{l r r r r}
\toprule
\textbf{Model \textbackslash\ Task (n samples)} & \makecell[r]{\textbf{aime24} \\ \textbf{single\_pass} \\ (30)} & \makecell[r]{\textbf{aime24} \\ \textbf{pass\_avg\_8} \\ (30)} & \makecell[r]{\textbf{aime25} \\ \textbf{single\_pass} \\ (30)} & \makecell[r]{\textbf{aime25} \\ \textbf{pass\_avg\_8} \\ (30)} \\
\midrule
EngGPT2-16B-A3B & 2379 & 16496 & 2634 & 16903 \\
FastwebMIIA-7B & 107 & 192 & 111 & 241 \\
Minerva-7B-instruct-v1.0 & 77 & 128 & 82 & 150 \\
Velvet-14B & 3633 & 12279 & 3877 & 10090 \\
LLaMAntino-3-ANITA-8B-Inst-DPO-ITA & 151 & 198 & 153 & 199 \\
Qwen3-4B & 718 & 6592 & 1461 & 8522 \\
Qwen3-8B & 1312 & 8008 & 1919 & 13213 \\
Llama-3.2-3B-Instruct & 925 & 3809 & 881 & 4232 \\
Llama-3.1-8B-Instruct & 1895 & 6339 & 1988 & 8206 \\
gemma-3-4b-it & 635 & 969 & 426 & 497 \\
gemma-3-12b-it & 2735 & 10038 & 971 & 4526 \\
deepseek-moe-16b-chat & 136 & 301 & 289 & 350 \\
gpt-oss-20b & 723 & 1686 & 744 & 1733\\
Moonlight-16B-A3B-Instruct & 227 & 376 & 187 & 231 \\
Ministral-3-8B-Instruct-2512-BF16\tnote{*} & 14432 & 110570 & 16507 & 135549 \\
\bottomrule
\end{tabular}
\begin{tablenotes}\footnotesize
    \item[*] Evaluation times computed serving the model via Huggingface (not vLLM).
\end{tablenotes}
\end{threeparttable}
\end{table}

\begin{table}[h]
\caption{Evaluation Times (in seconds) for the ARC tasks}
\label{tab:timings_arc}
\centering
\scriptsize
\begin{threeparttable}
\begin{tabular}{l r r r r}
\toprule
\textbf{Model \textbackslash\ Task (n samples)} & \makecell[r]{\textbf{arc\_challenge} \\ \textbf{chat} \\ (1172)} & \makecell[r]{\textbf{arc\_challenge} \\ \textbf{cot\_custom} \\ (1172)} & \makecell[r]{\textbf{arc\_challenge} \\ \textbf{cot\_it\_custom} \\ (1169)} \\
\midrule
EngGPT2-16B-A3B & 81 & 605 & 1034 \\
FastwebMIIA-7B & 61 & 248 & 206 \\
Minerva-7B-instruct-v1.0 & 61 & 188 & 250 \\
Velvet-14B & 84 & 280 & 354 \\
LLaMAntino-3-ANITA-8B-Inst-DPO-ITA & 135 & 601 & 268 \\
Qwen3-4B & 162 & 340 & 404 \\
Qwen3-8B & 69 & 441 & 611 \\
Llama-3.2-3B-Instruct & 125 & 130 & 137 \\
Llama-3.1-8B-Instruct & 216 & 286 & 343 \\
gemma-3-4b-it & 209 & 192 & 236 \\
gemma-3-12b-it & 462 & 425 & 598 \\
deepseek-moe-16b-chat & 427 & 251 & 858 \\
gpt-oss-20b & 69 & 197 & 183 \\
Moonlight-16B-A3B-Instruct & 386 & 202 & 179 \\
Ministral-3-8B-Instruct-2512-BF16\tnote{*} & 324 & 12026 & 18766 \\
\bottomrule
\end{tabular}
\begin{tablenotes}\footnotesize
    \item[*] Evaluation times computed serving the model via Huggingface (not vLLM).
\end{tablenotes}
\end{threeparttable}
\end{table}

\begin{table}[h]
\caption{Evaluation Times (in seconds) for the  GSM8K, HumanEval, and RULER tasks}
\label{tab:timings_gsm8k_humaneval_ruler}
\centering
\scriptsize
\begin{threeparttable}
\begin{tabular}{l r r r r}
\toprule
\textbf{Model \textbackslash\ Task (n samples)} & \makecell[r]{\textbf{gsm8k} \\ \textbf{llama} \\ (1319)} & \makecell[r]{\textbf{humaneval\_8} \\ \textbf{instruct\_cot} \\ (164)} & \makecell[r]{\textbf{ruler} \\ (19500)} \\
\midrule
EngGPT2-16B-A3B & 5729 & 16434 & 27890 \\
FastwebMIIA-7B & 168 & 3045 & 18082 \\
Minerva-7B-instruct-v1.0 & 138 & 3036 & 5429 \\
Velvet-14B & 164 & 3067 & 65225 \\
LLaMAntino-3-ANITA-8B-Inst-DPO-ITA & 269 & 3173 & 9003 \\
Qwen3-4B & 202 & 5504 & 25965 \\
Qwen3-8B & 1040 & 7027 & 33237 \\
Llama-3.2-3B-Instruct & 98 & 2811 & 17493 \\
Llama-3.1-8B-Instruct & 131 & 3037 & 30983 \\
gemma-3-4b-it & 189 & 3091 & 15138 \\
gemma-3-12b-it & 274 & 3076 & 50193 \\
deepseek-moe-16b-chat & 211 & 3303 & 6659 \\
gpt-oss-20b & 147 & 3061 & 33640 \\
Moonlight-16B-A3B-Instruct & 211 & 3108 & 7850 \\
Ministral-3-8B-Instruct-2512-BF16\tnote{*} & 10203 & 88368 & 196186 \\
\bottomrule
\end{tabular}
\begin{tablenotes}\footnotesize
    \item[*] Evaluation times computed serving the model via Huggingface (not vLLM).
\end{tablenotes}
\end{threeparttable}
\end{table}

\begin{table}[h]
\caption{Evaluation Times (in seconds) for the MMLU tasks}
\label{tab:timings_mmlu}
\centering
\scriptsize
\begin{threeparttable}
\begin{tabular}{l r r r r r}
\toprule
& \textbf{mmlu\_cot} & \multicolumn{4}{c}{\textbf{mmlu\_redux\_generative}} \\
\cmidrule(lr){3-6}
 & \makecell[r]{\textbf{llama} \\ \textbf{custom} } & \textbf{humanities} &  \textbf{other} & \makecell[r]{ \textbf{social} \\ \textbf{sciences} } & \textbf{stem}  \\
\textbf{Model \textbackslash\ Task} (n samples) & (14042) & (1208) & (1162) & (1147) & (1813) \\
\midrule
EngGPT2-16B-A3B & 30517 & 879 & 692 & 718 & 3039 \\
FastwebMIIA-7B & 1709 & 83 & 123 & 79 & 102 \\
Minerva-7B-instruct-v1.0 & 942 & 88 & 82 & 84 & 92 \\
Velvet-14B & 1449 & 112 & 95 & 93 & 117 \\
LLaMAntino-3-ANITA-8B & 6642 & 187 & 187 & 175 & 240 \\
Qwen3-4B & 36693 & 329 & 299 & 327 & 1108 \\
Qwen3-8B & 32258 & 441 & 427 & 323 & 143 \\
Llama-3.2-3B-Instruct & 587 & 88 & 68 & 64 & 79 \\
Llama-3.1-8B-Instruct & 943 & 101 & 73 & 88 & 91 \\
gemma-3-4b-it & 649 & 123 & 104 & 100 & 115 \\
gemma-3-12b-it & 2101 & 149 & 129 & 136 & 142 \\
deepseek-moe-16b-chat & 2703 & 203 & 146 & 142 & 232 \\
gpt-oss-20b & 894 & 140 & 245 & 123 & 216 \\
Moonlight-16B-A3B-Instruct & 1138 & 100 & 108 & 106 & 876 \\
Ministral-3-8B-Instruct-2512-BF16\tnote{*} & 159499 & 160 & 152 & 143 & 212 \\
\bottomrule
\end{tabular}
\begin{tablenotes}\footnotesize
    \item[*] Evaluation times computed serving the model via Huggingface (not vLLM).
\end{tablenotes}
\end{threeparttable}
\end{table}

\begin{table}[h]
\centering
\scriptsize
\begin{threeparttable}
\caption{Model throughputs computed via vLLM.}
\begin{tabular}{l r r}
\toprule
\textbf{Model} & \textbf{Tokens per second} & \textbf{Requests per second}
\\
\midrule
EngGPT2-MoE-16B & 1023 & 8.00  \\
FastwebMIIA-7B & 1265 & 9.89 \\
Minerva-7B-instruct-v1.0 & 1277 & 9.98 \\
Velvet-14B & 572 & 4.48 \\
LLaMAntino-3-ANITA-8B & 1267 & 9.90  \\
\hline
Qwen3-4B & 1848 & 14.44 \\
Qwen3-8B & 1228 & 9.59 \\
Llama-3.2-3B-Instruct & 2435 & 19.03 \\
Llama-3.1-8B-Instruct & 1259 & 9.84 \\
gemma-3-4b-it & 2058 & 16.08 \\
gemma-3-12b-it & 442 & 3.46 \\
Ministral-3-8B-Instruct-2512-BF16 & 1159 & 9.06  \\
\hline
deepseek-moe-16b-chat & 531 & 4.16 \\
gpt-oss-20b & 2057 & 16.07 \\
Moonlight-16B-A3B-Instruct & 1531 & 11.96 \\
\bottomrule
\end{tabular}
\label{tab:throughputs}
\end{threeparttable}
\end{table}

To provide a more standardized perspective on the raw performance of each model (although specific performance testing remains outside the scope of this paper), Table \ref{tab:throughputs} reports the models' throughput measured as wall-clock speed over a full batch run of requests, using the \textit{vLLM} serving framework. To isolate core model inference speed from tokenization overhead, we constructed a benchmark of 1,000 synthetic requests per model. Each prompt consisted of exactly 1,024 token IDs uniformly sampled from the model's vocabulary.

Models were instantiated via the \texttt{LLM} class, utilizing a constrained context window (\texttt{max\_model\_len=4096}) to optimize KV-cache memory and prevent out-of-memory errors during batch execution. To ensure uniform, maximum-load evaluation conditions, we configured generation parameters using the \texttt{SamplingParams} class with \texttt{max\_tokens=128} and \texttt{ignore\_eos=True}. This forced every model to output exactly 128 tokens per prompt. Throughput metrics were derived by measuring the end-to-end execution time of \texttt{llm.generate()}, which processes the entire batch of requests.

\clearpage


\section{Heatmaps for Statistical t-test with Bonferroni correction}
\label{app:pairwise-significance-heatmaps}

This section reports the full set of pairwise significance heatmaps of every dataset. For each metric and for each benchmark, pairs of models were compared using a two-sided Welch test using the reported means and standard errors. Since many pairwise tests are performed for each metric, $p$-values were adjusted using a Bonferroni correction within each column. Each heatmap displays the Bonferroni-adjusted pairwise $p$-values for one benchmark+metric combination. Models are ordered by the mean score for that metric, from strongest to weakest. Green cells indicate statistically significant differences at $\alpha=0.05$, yellow cells lie near the decision threshold, and red cells indicate that the comparison is not significant after correction. Because the figures show adjusted $p$-values rather than signed effect sizes, the color encodes confidence in a difference, while the axis ordering mimics the performance ranking.

\newcommand{\pairwiseplot}[4]{%
\begin{figure}[h]
    \centering
    \includegraphics[width=0.45\linewidth]{\detokenize{images_test/#1}}
    \caption{#2. #3}
    \label{#4}
\end{figure}
}

\newcommand{\pairwiseplotsub}[3]{%
\begin{subfigure}[t]{0.45\linewidth}
    \centering
    \includegraphics[width=\linewidth]{\detokenize{images_test/#1}}
    \caption{#2}
    \label{#3}
\end{subfigure}
}

\subsection{ARC-Challenge}
ARC-Challenge is shown for three evaluation settings. These plots make it easy to see whether the same groups of models remain statistically distinguishable when the prompting or extraction setting is changed.

\begin{figure}[h]
    \centering
    \pairwiseplotsub
        {arc_challenge_chat___exact_match.png}
        {\texttt{chat | exact\_match}}
        {fig:arc-challenge-chat-exact-match}
    \hfill
    \pairwiseplotsub
        {arc_challange_cot_fixed___flexible_extract.png}
        {\texttt{cot\_fixed | flexible\_extract}}
        {fig:arc-challenge-cot-fixed-flexible-extract}

    \vspace{0.5em}

    \begin{subfigure}[t]{0.45\linewidth}
        \centering
        \includegraphics[width=\linewidth]{\detokenize{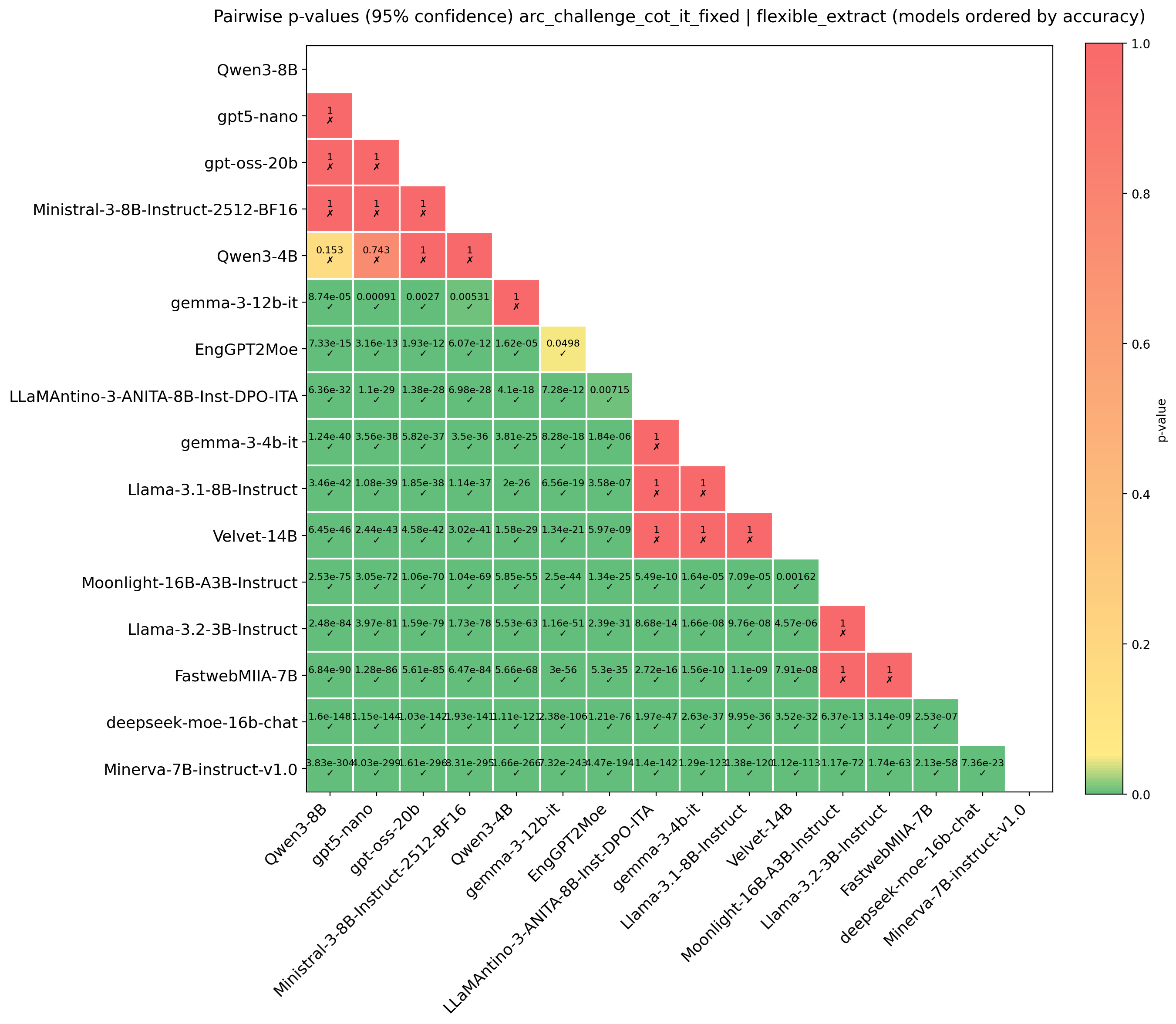}}
        \caption{\texttt{cot\_it\_fixed | flexible\_extract}}
        \label{fig:arc-challenge-cot-it-fixed-flexible-extract}
    \end{subfigure}

    \caption{Bonferroni-adjusted pairwise significance heatmaps for ARC-Challenge under three settings: \texttt{chat | exact\_match}, \texttt{cot\_fixed | flexible\_extract}, and \texttt{cot\_it\_fixed | flexible\_extract}.}
    \label{fig:arc-challenge-pairwise}
\end{figure}

\clearpage

\subsection{GSM8K}
For GSM8K, the plot below summarizes the pairwise evidence under the strict-match metric. This figure heps us identify whether models achieve statistically indistinguishable performance at the chosen threshold after correction.

\pairwiseplot{gsm8k_llama___strict_match.png}
{GSM8K, \texttt{llama | strict\_match}}
{Bonferroni-adjusted pairwise significance heatmap for GSM8K under the \texttt{strict\_match} metric.}
{fig:gsm8k-llama-strict-match}

\clearpage

\subsection{AIME-2024 \& AIME-2025}
We report the comparison of the considered models on AIME-2024 and AIME-2025 in Figures \ref{fig:aime24-pass-avg-8-pairwise} and \ref{fig:aime25-pass-avg-8-pairwise}, respectively. The comparison is performed across all four metrics.

\begin{figure}[h]
    \centering
    \pairwiseplotsub
        {aime24_pass_avg_8___avg_1_e_m.png}
        {\texttt{pass\_avg\_8 | avg\_1\_e\_m}}
        {fig:aime24-pass-avg-8-avg-1-em}
    \hfill
    \pairwiseplotsub
        {aime24_pass_avg_8___avg_8_e_m.png}
        {\texttt{pass\_avg\_8 | avg\_8\_e\_m}}
        {fig:aime24-pass-avg-8-avg-8-em}

    \vspace{0.5em}

    \pairwiseplotsub
        {aime24_pass_avg_8___pass_1_e_m.png}
        {\texttt{pass\_avg\_8 | pass\_1\_e\_m}}
        {fig:aime24-pass-avg-8-pass-1-em}
    \hfill
    \pairwiseplotsub
        {aime24_pass_avg_8___pass_8_e_m.png}
        {\texttt{pass\_avg\_8 | pass\_8\_e\_m}}
        {fig:aime24-pass-avg-8-pass-8-em}

    \caption{Bonferroni-adjusted pairwise significance heatmaps for AIME24 under the \texttt{pass\_avg\_8} setting, for the metrics \texttt{avg\_1\_e\_m}, \texttt{avg\_8\_e\_m}, \texttt{pass\_1\_e\_m}, and \texttt{pass\_8\_e\_m}.}
    \label{fig:aime24-pass-avg-8-pairwise}
\end{figure}

\begin{figure}[th]
    \centering
    \pairwiseplotsub
        {aime25_pass_avg_8___avg_1_e_m.png}
        {\texttt{pass\_avg\_8 | avg\_1\_e\_m}}
        {fig:aime25-pass-avg-8-avg-1-em}
    \hfill
    \pairwiseplotsub
        {aime25_pass_avg_8___avg_8_e_m.png}
        {\texttt{pass\_avg\_8 | avg\_8\_e\_m}}
        {fig:aime25-pass-avg-8-avg-8-em}

    \vspace{0.5em}

    \pairwiseplotsub
        {aime25_pass_avg_8___pass_1_e_m.png}
        {\texttt{pass\_avg\_8 | pass\_1\_e\_m}}
        {fig:aime25-pass-avg-8-pass-1-em}
    \hfill
    \pairwiseplotsub
        {aime25_pass_avg_8___pass_8_e_m.png}
        {\texttt{pass\_avg\_8 | pass\_8\_e\_m}}
        {fig:aime25-pass-avg-8-pass-8-em}

    \caption{Bonferroni-adjusted pairwise significance heatmaps for AIME25 under the \texttt{pass\_avg\_8} setting, for the metrics \texttt{avg\_1\_e\_m}, \texttt{avg\_8\_e\_m}, \texttt{pass\_1\_e\_m}, and \texttt{pass\_8\_e\_m}.}
    \label{fig:aime25-pass-avg-8-pairwise}
\end{figure}

\clearpage

\subsection{MMLU \& MMLU-Redux}
Figure \ref{fig:mmlu-cot-fixed-llama-flexible-extract} reports the pairwise significance for the MMLU dataset.
MMLU-Redux is decomposed into four domain groups as shown in Figure \ref{fig:mmlu-redux-generative-exact-match-pairwise}. 

\pairwiseplot{mmlu_cot_fixed_llama___flexible_extract.png}
{}
{Bonferroni-adjusted pairwise significance heatmap for MMLU under the \texttt{cot\_fixed\_llama} / \texttt{flexible\_extract} setting.}
{fig:mmlu-cot-fixed-llama-flexible-extract}

\begin{figure}[h]
    \centering
    \pairwiseplotsub
        {mmlu_redux_humanities_generative___exact_match.png}
        {Humanities, \texttt{generative | exact\_match}}
        {fig:mmlu-redux-humanities-generative-exact-match}
    \hfill
    \pairwiseplotsub
        {mmlu_redux_other_generative___exact_match.png}
        {Other, \texttt{generative | exact\_match}}
        {fig:mmlu-redux-other-generative-exact-match}

    \vspace{0.5em}

    \pairwiseplotsub
        {mmlu_redux_social_sciences_generative___exact_match.png}
        {Social Sciences, \texttt{generative | exact\_match}}
        {fig:mmlu-redux-social-sciences-generative-exact-match}
    \hfill
    \pairwiseplotsub
        {mmlu_redux_stem_generative___exact_match.png}
        {STEM, \texttt{generative | exact\_match}}
        {fig:mmlu-redux-stem-generative-exact-match}

    \caption{Bonferroni-adjusted pairwise significance heatmaps for the MMLU-Redux subsets Humanities, Other, Social Sciences, and STEM under the \texttt{generative | exact\_match} setting.}
    \label{fig:mmlu-redux-generative-exact-match-pairwise}
\end{figure}

\clearpage

\subsection{HumanEval}
HumanEval is shown in Figure \ref{fig:humaneval-8-instruct-cot-pairwise} for two \texttt{create\_test} metrics. 

\begin{figure}[h]
    \centering
    \pairwiseplotsub
        {humaneval_8_instruct_cot___pass_1_create_test.png}
        {\texttt{8\_instruct\_cot | pass\_1\_create\_test}}
        {fig:humaneval-8-instruct-cot-pass-1-create-test}
    \hfill
    \pairwiseplotsub
        {humaneval_8_instruct_cot___pass_8_create_test.png}
        {\texttt{8\_instruct\_cot | pass\_8\_create\_test}}
        {fig:humaneval-8-instruct-cot-pass-8-create-test}

    \caption{Bonferroni-adjusted pairwise significance heatmaps for HumanEval under the \texttt{8\_instruct\_cot} setting, for the metrics \texttt{pass\_1\_create\_test} and \texttt{pass\_8\_create\_test}.}
    \label{fig:humaneval-8-instruct-cot-pairwise}
\end{figure}

\subsection{BFCL}
\begin{figure}[h]
    \centering
    \includegraphics[width=0.45\linewidth]{\detokenize{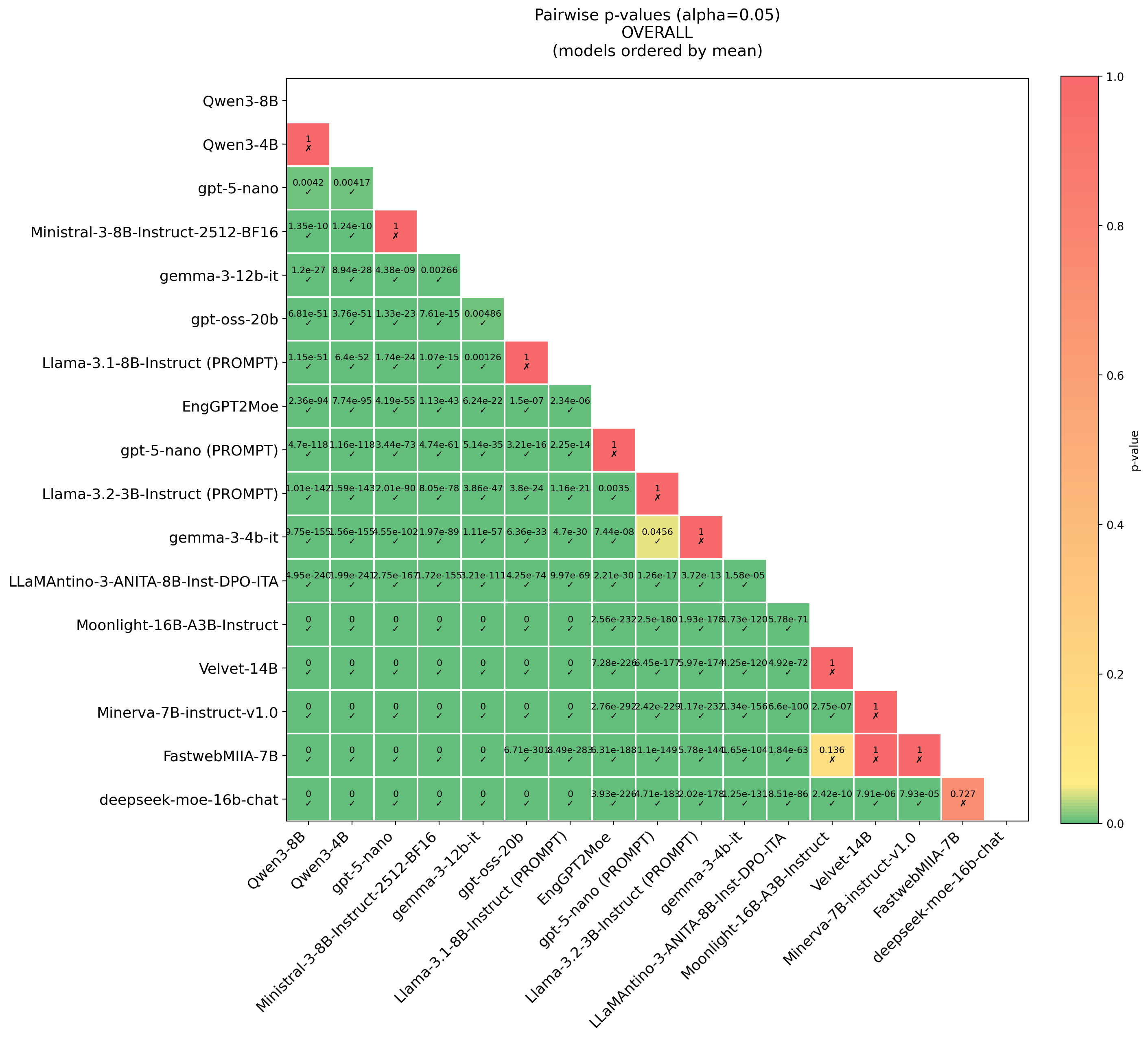}}
    \caption{Bonferroni-adjusted pairwise significance heatmap for the BFCL \texttt{Overall} score.}
    \label{fig:ttest-bfcl}
\end{figure}

Figure \ref{fig:ttest-bfcl} shows the results of the statistical test on BFCL dataset.

\clearpage

\subsection{ITALIC}
ITALIC is reported in Figure \ref{fig:italic-pairwise} for the four evaluation modes. 

\begin{figure}[h]
    \centering
    \pairwiseplotsub
        {ITALIC___COT.png}
        {\texttt{COT}}
        {fig:italic-cot}
    \hfill
    \pairwiseplotsub
        {ITALIC___fast.png}
        {\texttt{fast}}
        {fig:italic-fast}

    \vspace{0.5em}

    \pairwiseplotsub
        {italic_fast___exact_match.png}
        {\texttt{fast | exact\_match}}
        {fig:italic-fast-exact-match}
    \hfill
    \pairwiseplotsub
        {italic_mmlu_cot___flexible_extract.png}
        {\texttt{MMLU COT | flexible\_extract}}
        {fig:italic-mmlu-cot-flexible-extract}

    \caption{Bonferroni-adjusted pairwise significance heatmaps for ITALIC under the settings \texttt{COT}, \texttt{fast}, \texttt{fast | exact\_match}, and \texttt{MMLU COT | flexible\_extract}.}
    \label{fig:italic-pairwise}
\end{figure}

\clearpage

\subsection{RULER}
\begin{figure}[h]
    \centering
    \pairwiseplotsub
        {ruler___4096.png}
        {context length 4096}
        {fig:ruler-4096}
    \hfill
    \pairwiseplotsub
        {ruler___16384.png}
        {context length 16384}
        {fig:ruler-16384}

    \vspace{0.5em}

    \begin{subfigure}[t]{0.45\linewidth}
        \centering
        \includegraphics[width=\linewidth]{\detokenize{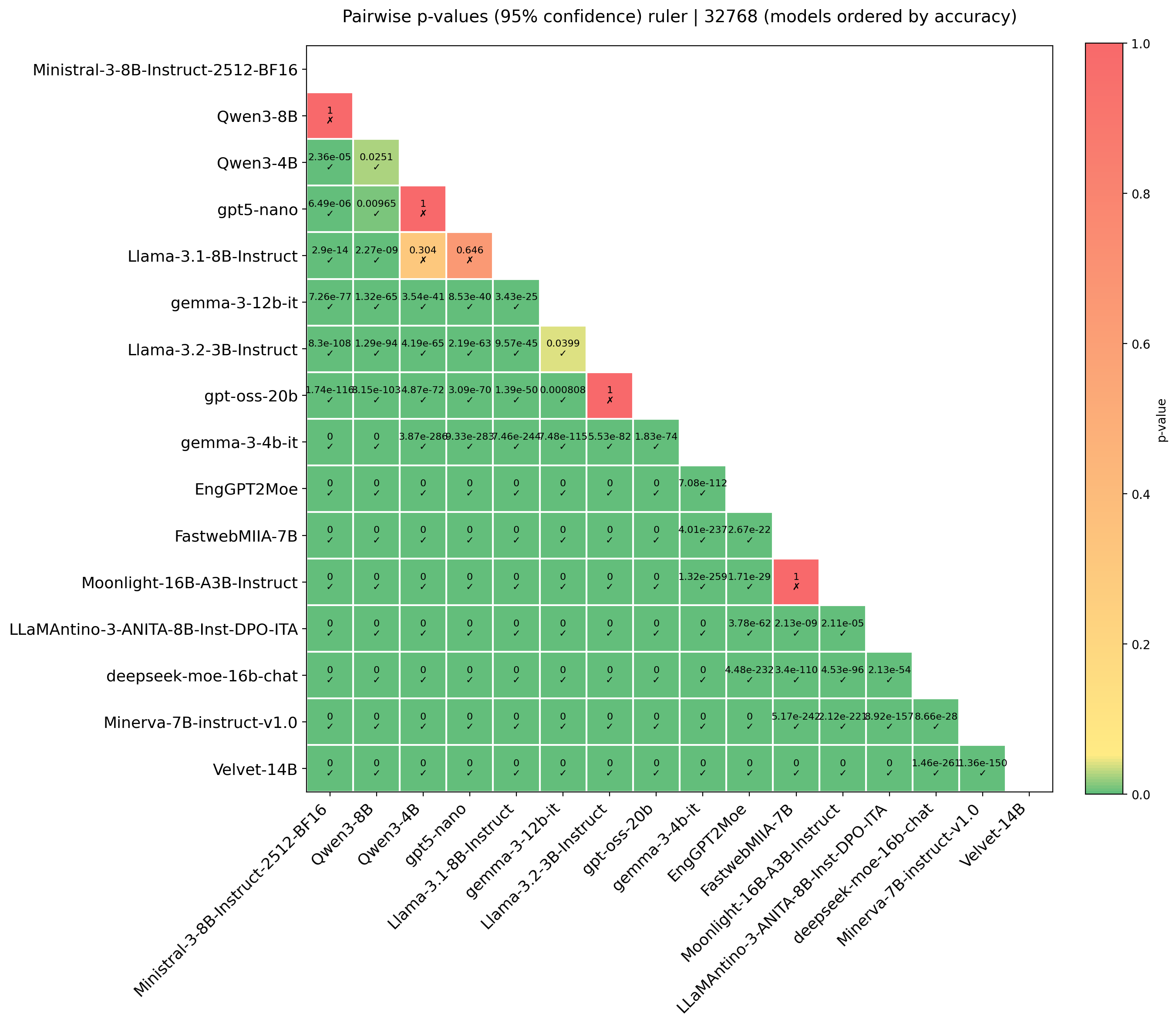}}
        \caption{context length 32768}
        \label{fig:ruler-32768}
    \end{subfigure}

    \caption{Bonferroni-adjusted pairwise significance heatmaps for RULER at context lengths 4096, 16384, and 32768.}
    \label{fig:ruler-pairwise}
\end{figure}

The RULER benchmark is shown in Figure \ref{fig:ruler-pairwise} at three different context lengths. These plots are useful for assessing that longer context length makes model differences become more pronounced.

\clearpage


\section{BFCL dataset detailed structure}
\label{appendix:bfcl-dataset-details}

Table \ref{tab:bfcl-dataset-details} contains the detailed description of each single task in the BFCL dataset, including number of samples, subdivision in task groups, and aggregation functions used to combine together scores obtained in each task towards group scores and then a single overall score. \texttt{Unweighted Average} means averaging each task score (accuracy) directly - meaning each tasks is considered equally, regardless of the number of samples it contains -, while \texttt{Weighted Average} means averaging each task score using the number of samples in the task as weights. All details were taken from the original BFCL implementation.

\begin{table}[ht]
\centering
\scriptsize
\renewcommand{\arraystretch}{1.6}
\begin{tabular}{|*{6}{c|}}
\hline
\multicolumn{6}{|c|}{\makecell{\textbf{Non-Live (Single-Turn)} \\ Unweighted Average (1150)}} \\ \hline
\multicolumn{3}{|c|}{\makecell{\textbf{Simple} \\ Unweighted Average}} & 
\multirow{2}{*}{\makecell{\textbf{Python} \\ \textbf{Multiple} \\ (200)}} & 
\multirow{2}{*}{\makecell{\textbf{Python} \\ \textbf{Parallel} \\ (200)}} & 
\multirow{2}{*}{\makecell{\textbf{Python} \\ \textbf{Parallel} \\ \textbf{Multiple} \\ (200)}} \\ \cline{1-3}
\makecell{\textbf{Python} \\ (400)} & 
\makecell{\textbf{Java} \\ (100)} & 
\makecell{\textbf{JavaScript} \\ (50)} & & & \\ \hline
\end{tabular}

\vspace{5pt}

\begin{tabular}{|*{4}{c|}}
\hline
\multicolumn{4}{|c|}{\makecell{\textbf{Live (Single-Turn)} \\ Weighted Average (1351)}} \\ \hline
\makecell{\textbf{Python} \\ \textbf{Simple} \\ (258)} & 
\makecell{\textbf{Python} \\ \textbf{Multiple} \\ (1053)} & 
\makecell{\textbf{Python} \\ \textbf{Parallel} \\ (16)} & 
\makecell{\textbf{Python} \\ \textbf{Parallel} \\ \textbf{Multiple} \\ (24)} \\ \hline
\end{tabular}

\vspace{5pt}

\begin{tabular}{|*{4}{c|}}
\hline
\multicolumn{4}{|c|}{\makecell{\textbf{Multi-Turn} \\ Unweighted Average \\ (800)}} \\ \hline
\multirow{2}{*}{\makecell{\textbf{Multi Turn} \\ \textbf{Base} \\ (200)}} & 
\multicolumn{3}{c|}{\makecell{\textbf{Augmented Cases} \\ Independent Categories \\ (600)}} \\ \cline{2-4}
 & 
\makecell{\textbf{Missing} \\ \textbf{Function} \\ (200)} & 
\makecell{\textbf{Missing} \\ \textbf{Parameter} \\ (200)} & 
\makecell{\textbf{Long Context} \\ (200)} \\ \hline
\end{tabular}

\vspace{5pt}

\begin{tabular}{|*{2}{c|}}
\hline
\multicolumn{2}{|c|}{\makecell{ \\ \textbf{Hallucination Measurement} \\ Unweighted Average \\ (1122)}} \\ \hline
\makecell{\textbf{Non-Live} \\ \textbf{Irrelevance} \\ (240)} & 
\makecell{\textbf{Live} \\ \textbf{Irrelevance} \\ (882)} \\ \hline
\end{tabular}
\caption{Detailed structure of the BFCL dataset used for the evaluations, in terms of task composition within each task group. The number in parentheses indicate the number of samples for each task and task group, while the averaging rules indicate how scores were grouped together for each task group.}
\label{tab:bfcl-dataset-details}
\end{table}

\clearpage


\section{Custom Task and Model Configurations for lm\_eval}
\label{appendix:config}

The following sections provide detailed descriptions of the modifications and extensions applied to the \textit{lm\_eval} harness tasks during our evaluation. We outline the base configuration template used to define the generation parameters passed to the evaluated models. Any tasks not explicitly discussed in these sections were executed using their default configurations as provided by the evaluation framework, without further modification.

\begin{listing}[th]
    \centering
    \includegraphics[
        width=\linewidth,
        trim={1.2cm 0 25cm 0},
        clip
    ]{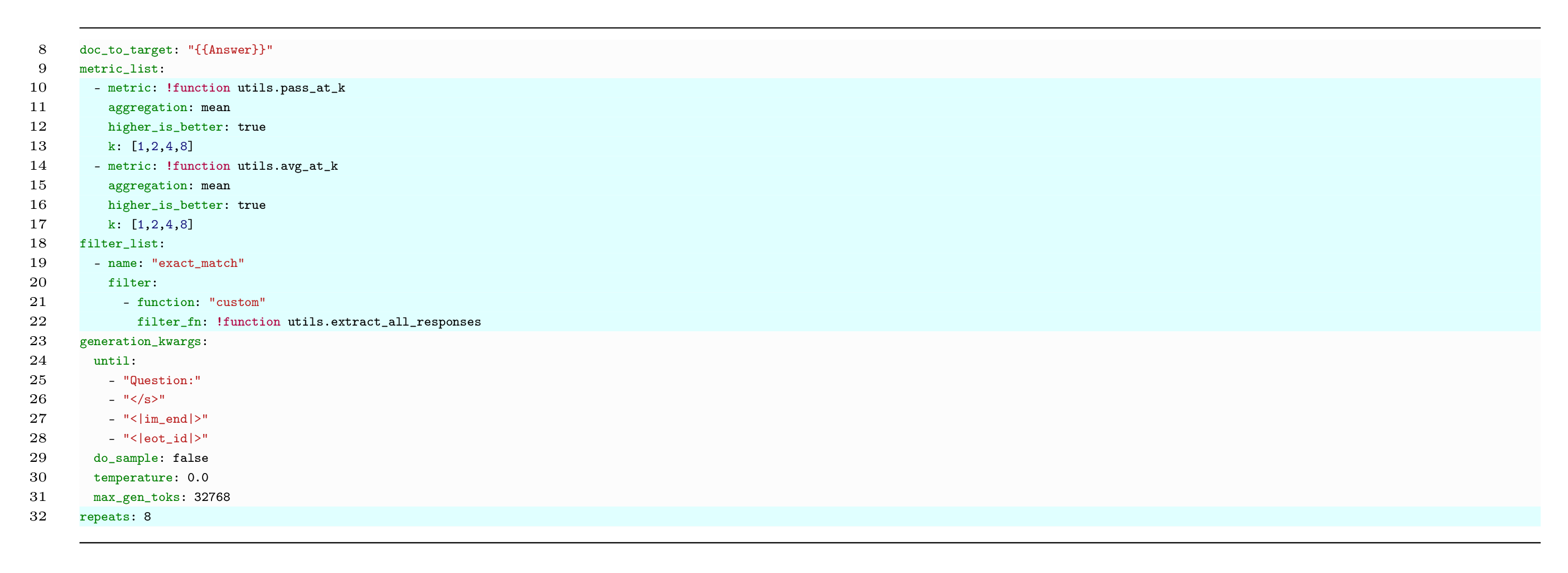}
    \caption{Configuration for aime24\_pass\_avg\_8}
    \label{lst:config_aime24}
\end{listing}

\begin{listing}[th]
    \centering
    \includegraphics[
        width=\linewidth,
         trim={1.2cm 0 25cm 0},
        clip
    ]{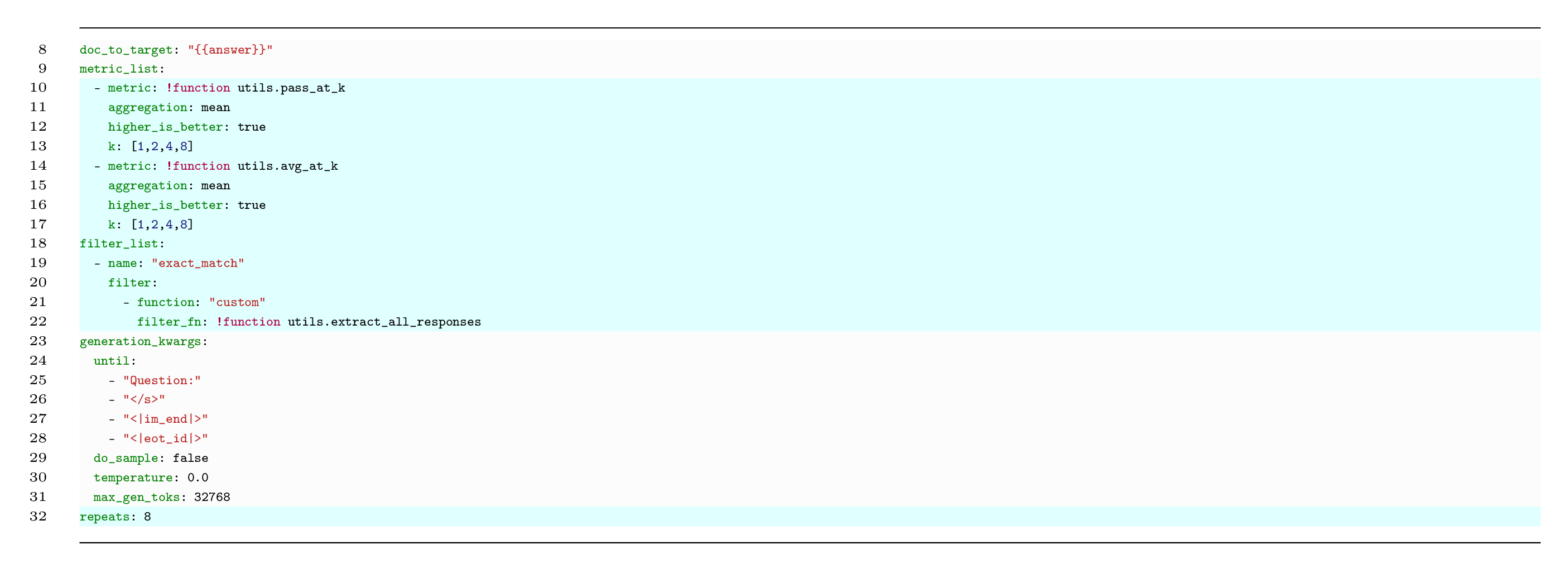}
    \caption{Configuration for aime25\_pass\_avg\_8}
    \label{lst:config_aime25}
\end{listing}

\begin{listing}[th]
    \centering
    \includegraphics[
        width=\linewidth,
         trim={1.2cm 0 25cm 0},
        clip
    ]{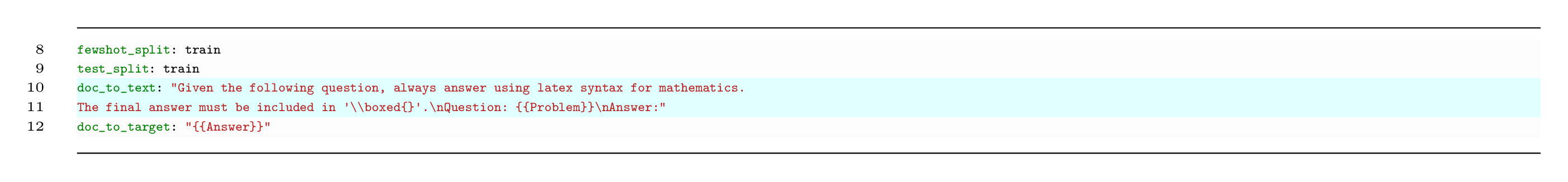}
    \caption{Configuration for aime24\_openai}
    \label{lst:config_aime24_openai}
\end{listing}

\begin{listing}[th]
    \centering
    \includegraphics[
        width=\linewidth,
         trim={1.2cm 0 25cm 0},
        clip
    ]{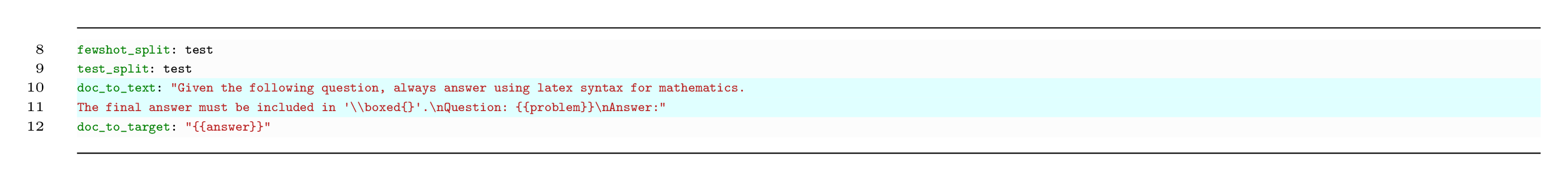}
    \caption{Configuration for aime25\_openai}
    \label{lst:config_aime25_openai}
\end{listing}

\subsection{AIME-2024 and AIME-2025}
\label{appendix:aime_config}
The AIME-2024 and AIME-2025 tasks where complemented with additional metrics, as shown in Listing \ref{lst:config_aime24} and \ref{lst:config_aime25} respectively, including pass@k and avg@k with k taking the values of 1, 2, 4, and 8. For evaluations performed through API calls (i.e., for OpenAI's GPT-5-nano model), the prompt was changed (Listing \ref{lst:config_aime24_openai} and \ref{lst:config_aime25_openai}) to clarify the output format as the final number had to be expressed with \verb|\boxed{}| syntax.

\subsection{ARC-Challenge}
\label{appendix:arc_config}
The code in Listing \ref{lst:config_arc_challenge_chat_openai} shows the modified regex rule to correctly parse the output in case of evaluations performed through API calls (i.e., for OpenAI's GPT-5-nano model). 

A similar change is shown in Listings \ref{lst:config_arc_challenge_cot} and \ref{lst:config_arc_challenge_cot_it}. The latter was applied to the arc\_challenge task to improve the filtering functions.

\begin{listing}[th]
    \centering
    \includegraphics[
        width=\linewidth,
         trim={1.2cm 0 25cm 0},
        clip
    ]{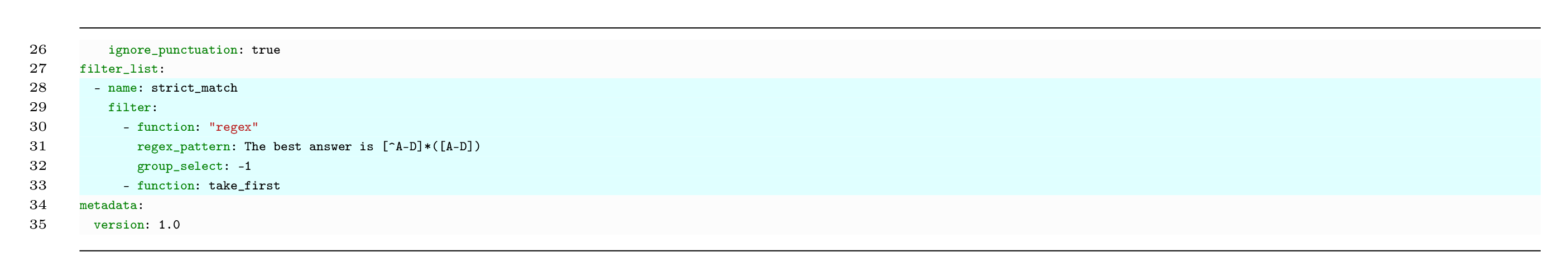}
    \caption{Configuration for arc\_challenge\_chat\_openai}
    \label{lst:config_arc_challenge_chat_openai}
\end{listing}

\begin{listing}[th]
    \centering
    \includegraphics[
        width=\linewidth,
         trim={1.2cm 0 25cm 0},
        clip
    ]{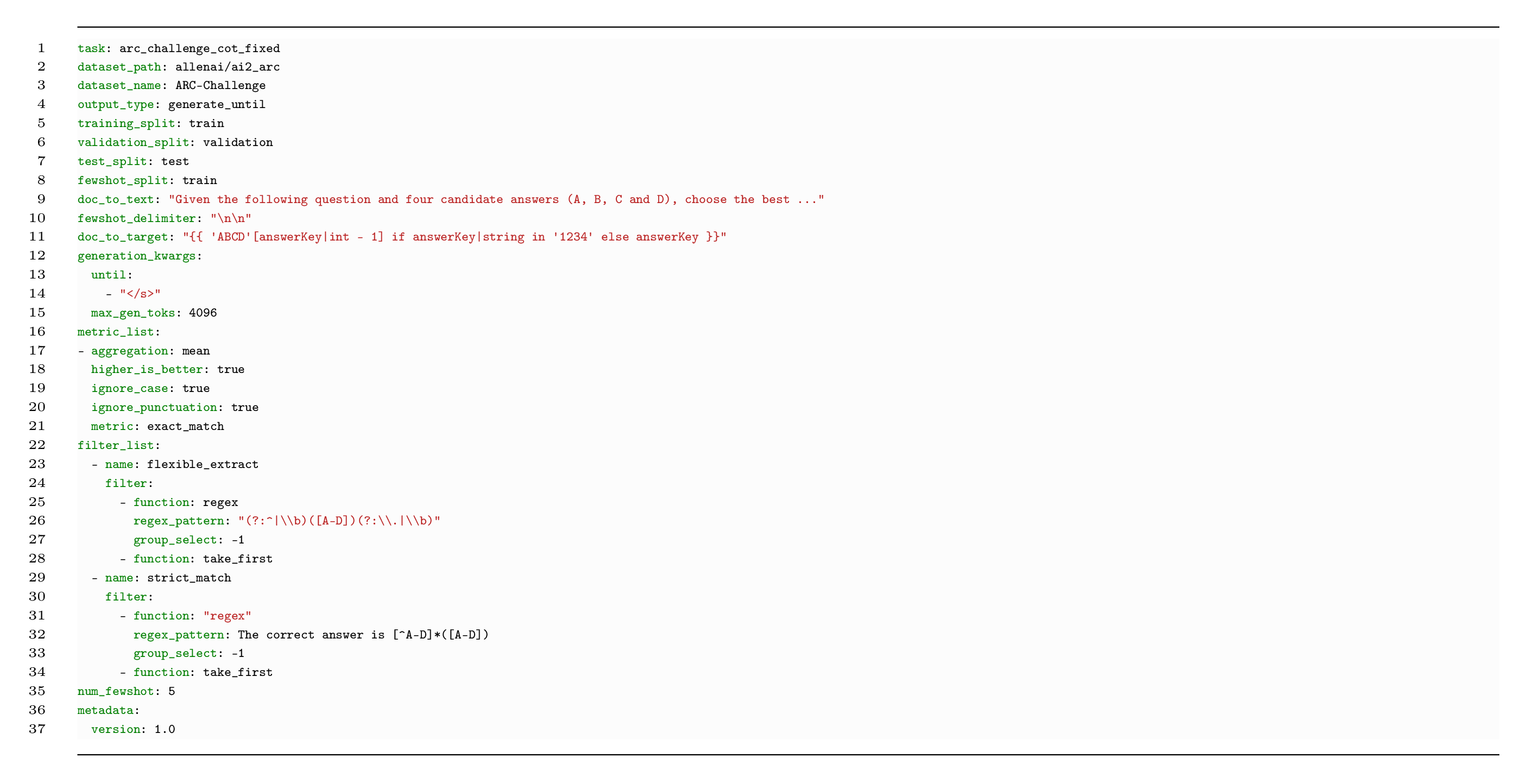}
    \caption{Configuration for arc\_challenge\_cot\_custom}
    \label{lst:config_arc_challenge_cot}
\end{listing}

\begin{listing}[th]
    \centering
    \includegraphics[
        width=\linewidth,
         trim={1.2cm 0 25cm 0},
        clip
    ]{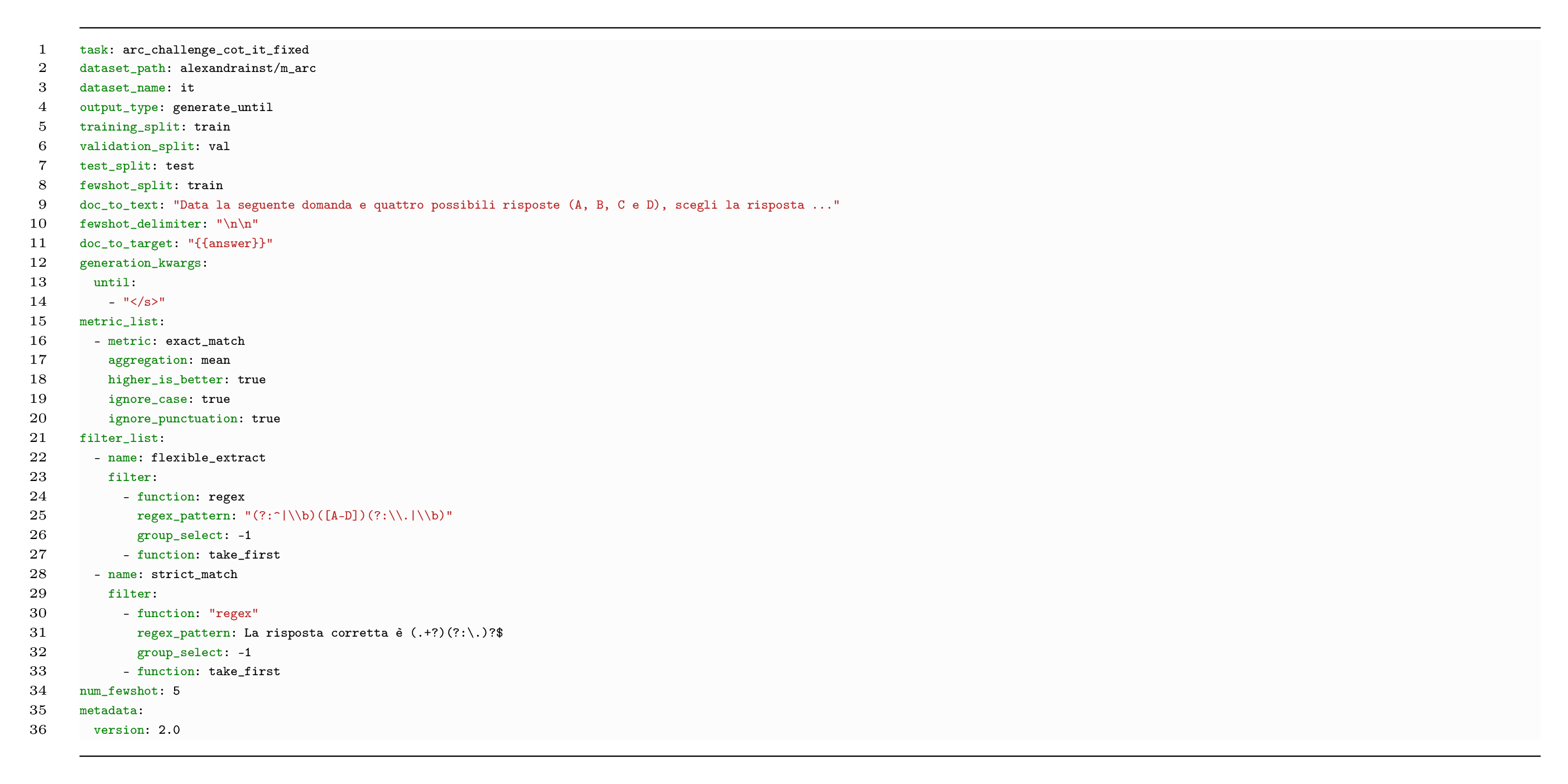}
    \caption{Configuration for arc\_challenge\_cot\_it\_custom}
    \label{lst:config_arc_challenge_cot_it}
\end{listing}

\subsection{MMLU}
\label{appendix:mmlu_config}
The code in Listing \ref{lst:config_mmlu_cot_llama} shows the new regex rule added to compute accuracy also under flexible extraction.

\begin{listing}[th]
    \centering
    \includegraphics[
        width=\linewidth,
         trim={1.2cm 0 25cm 0},
        clip
    ]{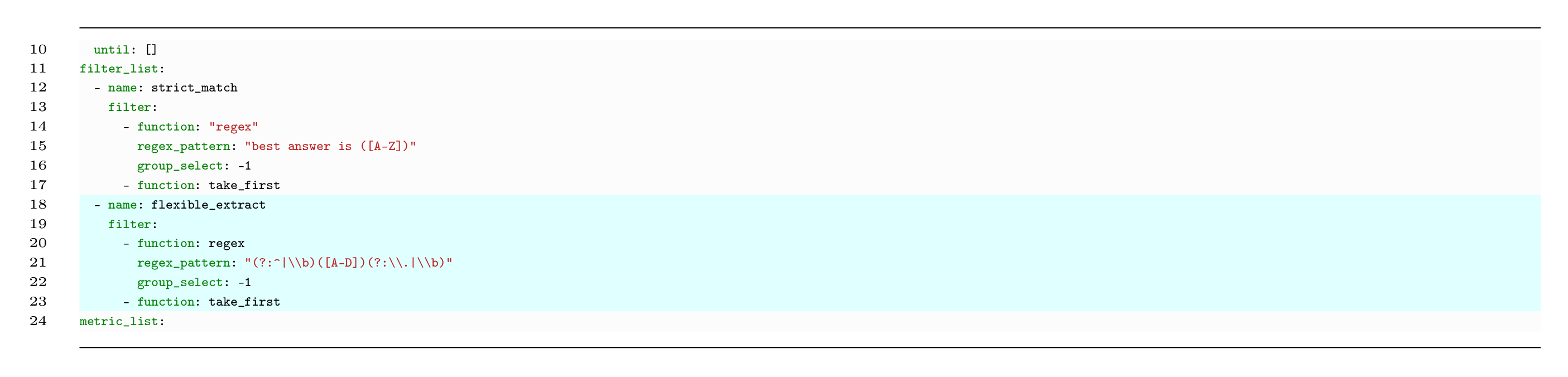}
    \caption{Configuration for mmlu\_cot\_llama\_custom}
    \label{lst:config_mmlu_cot_llama}
\end{listing}

\subsection{ITALIC}
\label{appendix:italic_config}
Since ITALIC is not natively supported by the \texttt{lm\_evaluation\_harness} library, we first performed the original chat-formatted evaluation, maintaining the two distinct prompting strategies available in the original implementation (a "fast" variant and a "chain-of-thought" (\emph{CoT}) variant). While we adopted the original implementation, we introduced some enhancements to the \texttt{extract\_answer} and \texttt{extract\_answer\_fast} for better answer extraction (both functions are located in the \texttt{ITALIC/run\_eval.py} script). Specifically, we implemented more robust parsing rules to mitigate false negatives caused by functionally correct answers being misclassified (see Listing \ref{lst:code_extract_aswer_fast} and \ref{lst:code_extract_aswer}). Furthermore, we refined the few-shot formatting in the \texttt{configure\_payload} function, explicitly formulating the example answers as 'Risposta: A' rather than a standalone 'A' (see Listing \ref{lst:code_fewshot}).

The ITALIC task was then made compatible with the \texttt{lm\_evaluation\_harness} library by creating custom configuration files. First, mirroring the original setup, both a standard non-\emph{CoT} variant named \texttt{italic\_fast} (Listing \ref{lst:config_italic_fast}) and a \emph{CoT} variant named \texttt{italic\_cot} (Listing \ref{lst:config_italic_cot}) were implemented. Second, we used more structured prompting by adapting templates from established benchmarks. This resulted in a non-\emph{CoT} variant (\texttt{italic\_fast\_custom}) and a \emph{CoT} variant (\texttt{italic\_cot\_custom}). To ensure methodological consistency, the prompt templates for these configurations were directly translated from the established \texttt{arc\_challenge\_chat} and \texttt{mmlu\_cot\_llama} tasks, respectively. We report only the code for the \texttt{italic\_fast} and \texttt{italic\_cot} tasks as the \texttt{italic\_fast\_custom} and \texttt{italic\_cot\_custom} tasks follow the same configurations except for the \texttt{doc\_to\_text} and \texttt{gen\_prefix} parameters.
\begin{listing}[ht]
    \centering
    \includegraphics[
        width=\linewidth,
         trim={1.2cm 0 25cm 0},
        clip
    ]{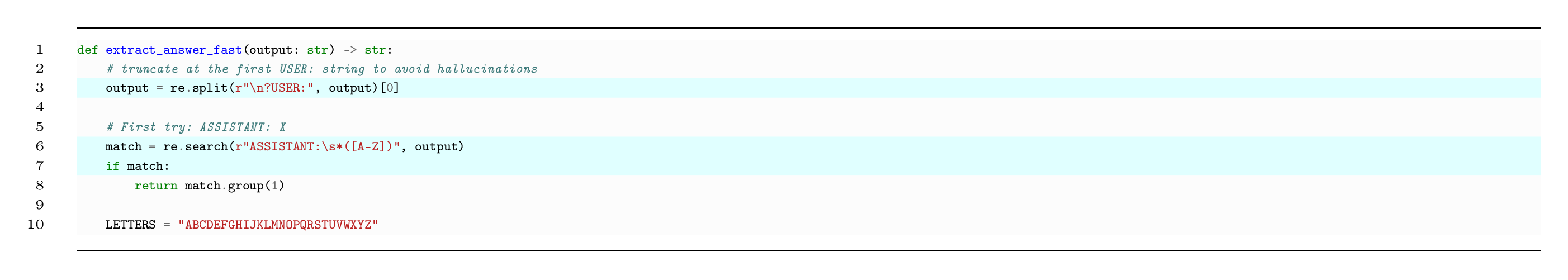}
    \caption{Changes to the \texttt{extract\_answer\_fast} function}
    \label{lst:code_extract_aswer_fast}
\end{listing}

\begin{listing}[ht]
    \centering
    \includegraphics[
        width=\linewidth,
         trim={1.2cm 0 25cm 0},
        clip
    ]{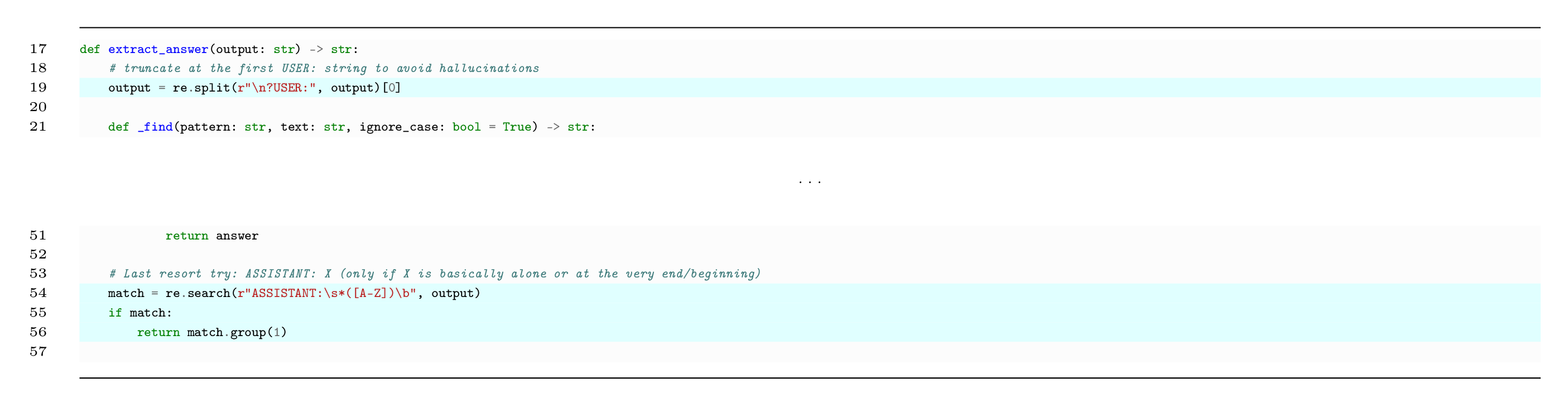}
    \caption{Changes to the \texttt{extract\_answer} function}
    \label{lst:code_extract_aswer}
\end{listing}

\begin{listing}[ht]
    \centering
    \includegraphics[
        width=\linewidth,
         trim={1.2cm 0 25cm 0},
        clip
    ]{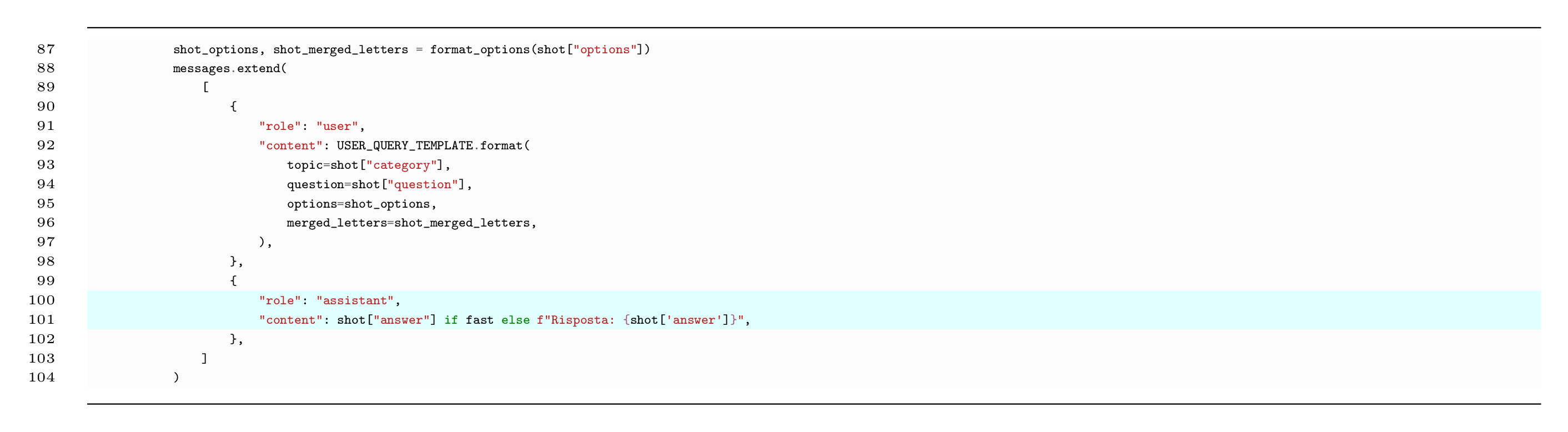}
    \caption{Changes to the \texttt{configure\_payload} function}
    \label{lst:code_fewshot}
\end{listing}

\begin{listing}[ht]
    \centering
    \includegraphics[
        width=\linewidth,
         trim={1.2cm 0 25cm 0},
        clip
    ]{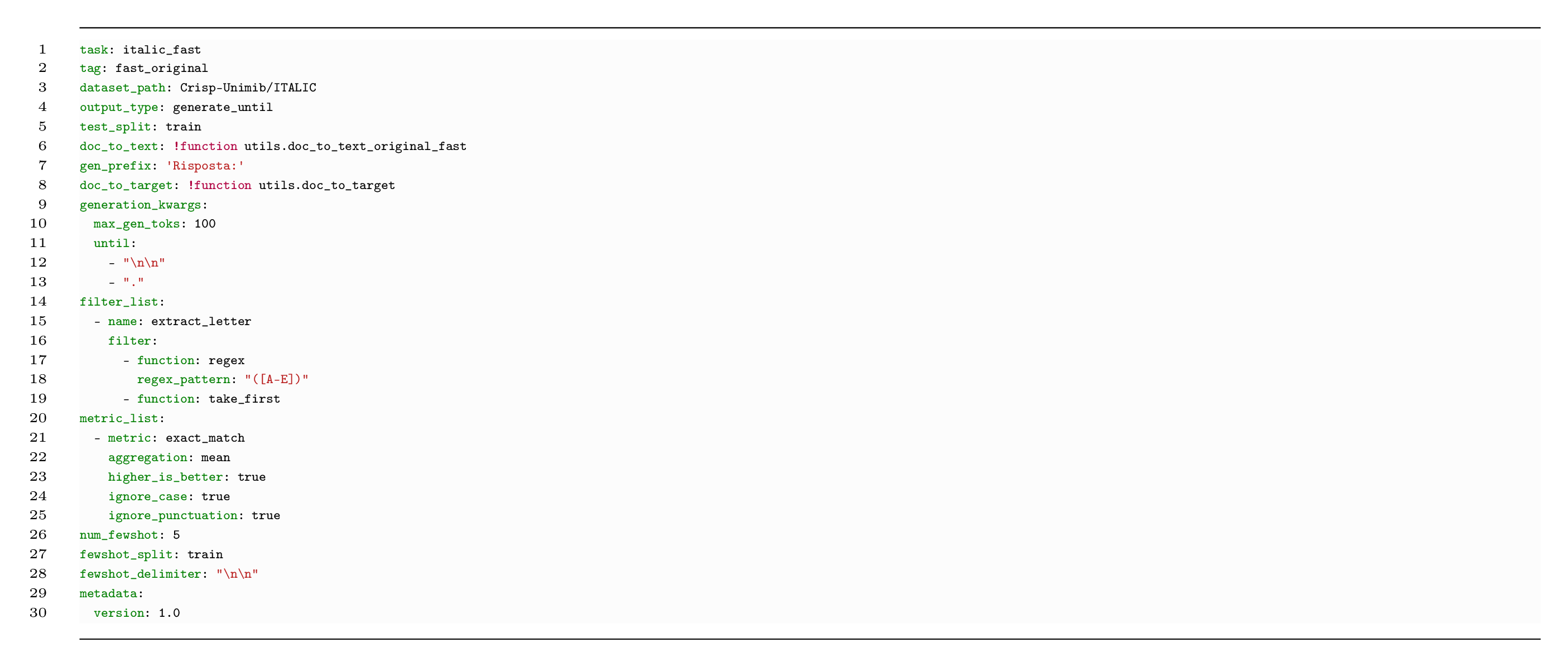}
    \caption{Configuration for italic\_fast}
    \label{lst:config_italic_fast}
\end{listing}

\begin{listing}[ht]
    \centering
    \includegraphics[
        width=\linewidth,
         trim={1.2cm 0 25cm 0},
        clip
    ]{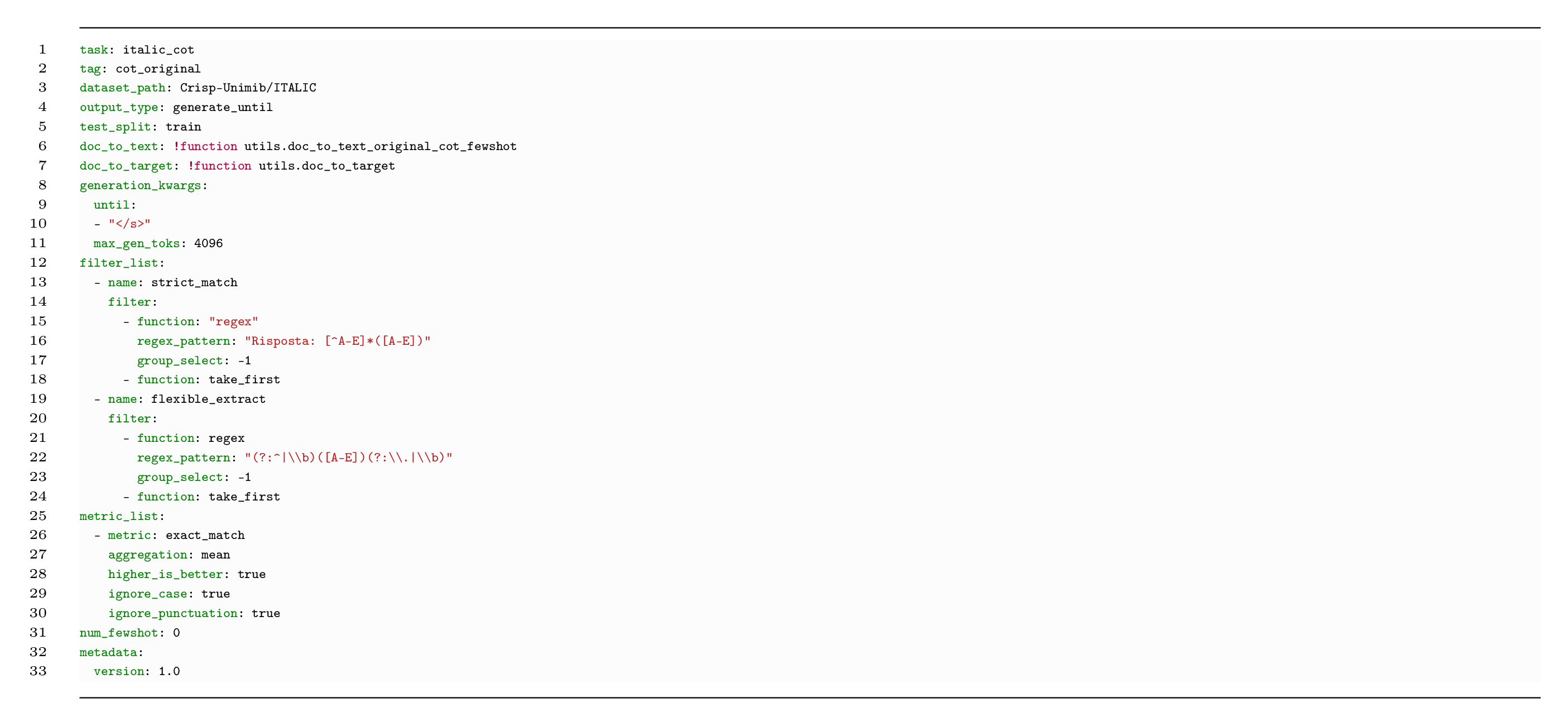}
    \caption{Configuration for italic\_cot}
    \label{lst:config_italic_cot}
\end{listing}

\subsection{Model Configurations}
\label{appendix:eval_params}

Listing \ref{lst:base_template} illustrates the base configuration template used to define the execution parameters for the evaluations. While this structure remained largely consistent across our experiments, specific arguments were dynamically adapted depending on the architecture and capabilities of the model under evaluation.
The \texttt{model} parameter designates the inference backend used to serve the model. As previously discussed, \texttt{vllm} was our default choice for high-throughput evaluation. However, this parameter was modified to \texttt{hf-mistral3} for the Ministral-3-8B-Instruct-2512-BF16 evaluation and to \texttt{azure-openai-chat-completions} for the API-based evaluation of gpt-5-nano.
Within the \texttt{model\_args} definition, the reasoning-specific parameters—namely \texttt{enable\_thinking} and \texttt{think\_end\_token}—were conditionally included. These were passed strictly to reasoning-capable models designed to output explicit intermediate thought traces (e.g., delimited by \texttt{\textless\text{think}\textgreater} tags) before providing a final answer.
Additionally, the \texttt{include\_path} directive was utilized exclusively when evaluating custom versions of the tasks, pointing to our local task implementations. Finally, to prevent excessively long generation and manage computational overhead, we optionally appended the \texttt{max\_gen\_toks} parameter within the generation keyword arguments (\texttt{gen\_kwargs}) to strictly limit the maximum number of tokens a model could produce per sample.

\begin{listing}[ht]
    \centering
    \includegraphics[
        width=\linewidth,
         trim={1.2cm 0 25cm 0},
        clip
    ]{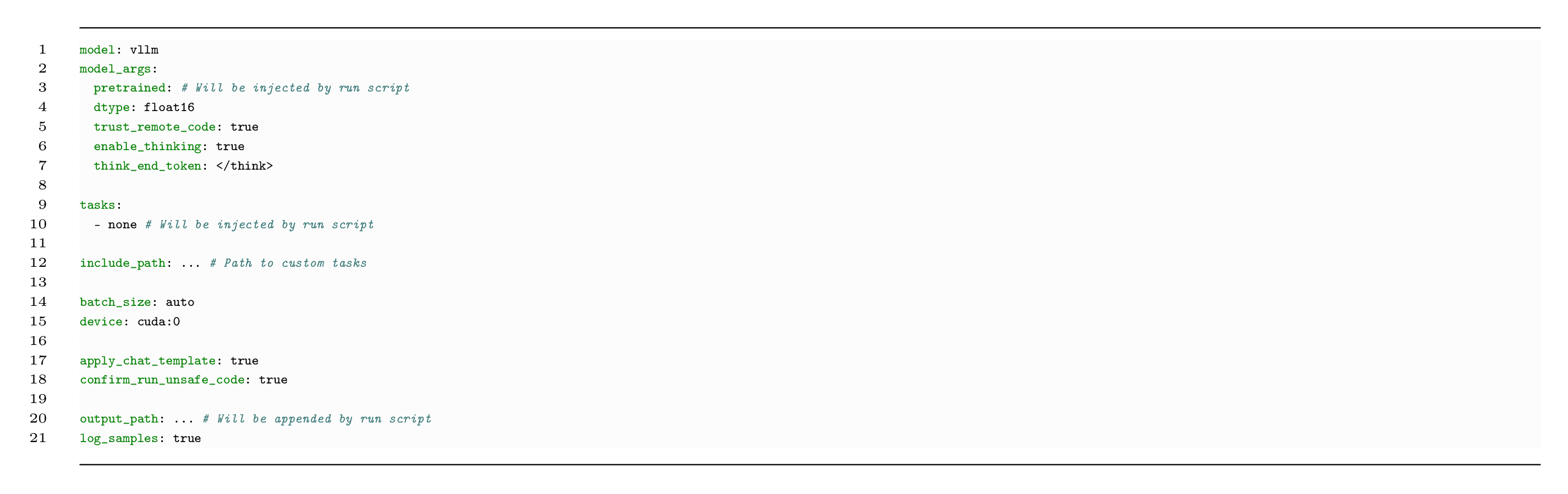}
    \caption{Base configuration template utilized for the \textit{lm\_eval} harness.}
    \label{lst:base_template}
\end{listing}

\clearpage


\section{Evaluation Environment Technical Specifications}
\label{appendix:technical_specifications}

Tables \ref{tab:hardware_specs} and \ref{tab:software_specs} contains, respectively, the specifications for the hardware used to run all of the reported evaluations, and for the software used for each evaluation suite, for full reproducibility.

\begin{table}[ht]
\caption{Hardware specifications used for the evaluation.}
\label{tab:hardware_specs}
\centering
\footnotesize
\begin{tabular}{l l}
\toprule
\textbf{Component / Software} & \textbf{Specification} \\
\midrule
GPU Model & NVIDIA A100-SXM4 \\
GPU Memory (VRAM) & 40 GB \\
NVIDIA Driver Version & 590.48.01 \\
CUDA Version & 13.1 \\
Max Power Limit (TDP) & 400W \\
\bottomrule
\end{tabular}
\end{table}

\begin{table}[ht]
\caption{Software libraries used for the evaluation and their respective versions, grouped by the different evaluation suites that were used.}
\label{tab:software_specs}
\centering
\scriptsize
\begin{tabular}[t]{l l}
\toprule
\multicolumn{2}{c}{LM Evaluation Harness} \\
\midrule
python & 3.12.11 \\
lm\_eval & 0.4.12 \\
torch & 2.9.1 \\
transformers & 4.57.6 \\
accelerate & 1.12.0 \\
openai & 2.20.0 \\
PyJWT & 2.11.0 \\
requests & 2.32.5 \\
aiohttp & 3.13.3 \\
tenacity & 9.1.4 \\
tiktoken & 0.12.0 \\ 
\bottomrule
\end{tabular}
\hspace{6pt}
\begin{tabular}[t]{l l}
\toprule
\multicolumn{2}{c}{BFCL} \\
\midrule
python & 3.12.11 \\
bfcl-eval & 2026.2.9 \\
evalscope & 1.4.2 \\
requests & 2.32.5 \\
vllm & 0.16.0 \\
\bottomrule
\end{tabular}
\hspace{6pt}
\begin{tabular}[t]{l l}
\toprule
\multicolumn{2}{c}{HF Evaluation} \\
\midrule
python & 3.12.11 \\
lm\_eval & 0.4.12 \\
torch & 2.10.0 \\
transformers & 5.3.0 \\
accelerate & 1.12.0 \\
openai & 2.16.0 \\
PyJWT & 2.10.1 \\
wonderwords & 3.0.1 \\
nltk & 3.9.2 \\
\bottomrule
\end{tabular}
\end{table}

\clearpage


\section{Example responses}
\label{appendix:examples}
In this Section, we present a selection of illustrative samples drawn from most of the evaluated benchmarks. Each example includes the original input and target response from the dataset, actual responses from some of the models under evaluation, and an indication of correctness or score. These examples should be interpreted as qualitative evidence to help understand the different characteristics of both evaluation datasets and evaluated models, rather than as definitive indication on how a specific model responds to all samples and prompts.

\subsection{Example responses for AIME-2024}
\label{appendix:aime24_examples}

This subsection reports examples from the AIME24 benchmark, specifically for the \texttt{aime24\_pass\_avg\_8} task. Table \ref{tab:aime24-data-sample} presents a selected input instance (ID 2), while Table \ref{tab:aime24-responses} summarizes the corresponding model outputs. In particular, we provide pass@1 responses for a meaningful subset of models, which is composed of the best performing models.

\begin{table}[ht]
\centering
\scriptsize
\begin{tabular}{|p{0.25\linewidth}|p{0.6\linewidth}|c|}
\hline
\textbf{Problem} & \textbf{Solution} & \textbf{Answer} \\ \hline
Jen enters a lottery by picking $4$ distinct numbers from $\{1,2,3,\cdots,9,10\}$. $4$ numbers are randomly chosen from $S$. She wins a prize if at least two of her numbers were $2$ of the randomly chosen numbers, and wins the grand prize if all four of her numbers were the randomly chosen numbers. The probability of her winning the grand prize given that she won a prize is $\tfrac{m}{n}$ where $m$ and $n$ are relatively prime positive integers. Find $m+n$.
& 
This is a conditional probability problem. Bayes' Theorem states that: $[P(A|B)=\dfrac{P(B|A)\cdot P(A)}{P(B)}$]. In other words, the probability of $A$ given $B$ is equal to the probability of $B$ given $A$ times the probability of $A$ divided by the probability of $B$. In our case, $A$ represents the probability of winning the grand prize, and $B$ represents the probability of winning a prize. Clearly, $P(B|A)=1$, since by winning the grand prize you automatically win a prize. Thus, we want to find $\dfrac{P(A)}{P(B)}$. Let us calculate the probability of winning a prize. We do this through casework: how many of Jen's drawn numbers match the lottery's drawn numbers To win a prize, Jen must draw at least $2$ numbers identical to the lottery. Thus, our cases are drawing $2$, $3$, or $4$ numbers identical. Let us first calculate the number of ways to draw exactly $2$ identical numbers to the lottery. Let Jen choose the numbers $a$, $b$, $c$, and $d$; we have $\dbinom42$ ways to choose which $2$ of these $4$ numbers are identical to the lottery. We have now determined $2$ of the $4$ numbers drawn in the lottery; since the other $2$ numbers Jen chose can not be chosen by the lottery, the lottery now has $10-2-2=6$ numbers to choose the last $2$ numbers from. Thus, this case is $\dbinom62$, so this case yields $\dbinom42\dbinom62=6\cdot15=90$ possibilities. Next, let us calculate the number of ways to draw exactly $3$ identical numbers to the lottery. Again, let Jen choose $a$, $b$, $c$, and $d$. This time, we have $\dbinom43$ ways to choose the identical numbers and again $6$ numbers left for the lottery to choose from; however, since $3$ of the lottery's numbers have already been determined, the lottery only needs to choose $1$ more number, so this is $\dbinom61$. This case yields $\dbinom43\dbinom61=4\cdot6=24$. Finally, let us calculate the number of ways to all $4$ numbers matching. There is actually just one way for this to happen. In total, we have $90+24+1=115$ ways to win a prize. The lottery has $\dbinom{10}4=210$ possible combinations to draw, so the probability of winning a prize is $\dfrac{115}{210}$. There is actually no need to simplify it or even evaluate $\dbinom{10}4$ or actually even know that it has to be $\dbinom{10}4$; it suffices to call it $a$ or some other variable, as it will cancel out later. However, let us just go through with this. The probability of winning a prize is $\dfrac{115}{210}$. Note that the probability of winning a grand prize is just matching all $4$ numbers, which we already calculated to have $1$ possibility and thus have probability $\dfrac1{210}$. Thus, our answer is $\dfrac{\frac1{210}}{\frac{115}{210}}=\dfrac1{115}$. Therefore, our answer is $1+115=\boxed{116}$.
& 116 \\ 
\hline
\end{tabular}
\caption{Representative example (Sample ID 2) from the AIME24 benchmark. The Problem column reports the original question, the Solution column provides the reference step-by-step derivation, and the Answer column contains the final ground-truth value used for evaluation.}
\label{tab:aime24-data-sample}
\end{table}

\begin{scriptsize}
\begin{longtable}[t]{|p{0.09\linewidth}|p{0.75\linewidth}|p{0.11\linewidth}|}
\caption{Representative model responses on the AIME24 benchmark (sample ID 2). For each model, the Answer column reports the pass@1 response, while the Result column presents the corresponding pass@1, pass@8, and avg@8 metrics for the given data sample. NOTE: the response generated by FastwebMIIA-7B is truncated due to repetitive token generation. Minor formatting adjustments have been applied to model responses for clarity and consistency of presentation.}
\label{tab:aime24-responses} \\
\hline
\textbf{Model} & \textbf{Answer} & \textbf{Result} \\ \hline
\hline
EngGPT2-16B-A3B &
To solve the problem, we are given a scenario where:
\begin{itemize}
    \item **Jen** selects **4 distinct numbers** from the set $ S = {1, 2, 3, \dots, 10} $.
    \item **4 numbers** are randomly selected from the same set $S$.
    \item Jen **wins a prize** if **at least two** of her selected numbers are among the 4 randomly chosen.
    \item She **wins the grand prize** if **all four** of her selected numbers are among the 4 randomly chosen.
\end{itemize}
We are asked to compute the **conditional probability** that Jen **won the grand prize**, **given that she won a prize**. That is:
$$P(\text{Grand Prize} \mid \text{Prize}) = \frac{P(\text{Grand Prize})}{P(\text{Prize})}$$
Step 1: Total number of ways to choose 4 numbers from 10 \newline
The total number of possible 4-number combinations that can be drawn from the 10-number set is: $$\binom{10}{4} = 210$$
Step 2: Count the number of favorable outcomes for the **grand prize** \newline
Jen{\textbackslash}u2019s 4 numbers are a specific set of 4 distinct numbers. To win the **grand prize**, all 4 of her numbers must be among the 4 randomly selected numbers. There is **only one** such combination (the set of her 4 numbers), so: $$\text{Number of grand prize outcomes} = 1$$\
Step 3: Count the number of favorable outcomes for the **prize** \newline
Jen wins a **prize** if **at least two** of her 4 numbers are among the 4 selected. So, we need to count the number of 4-number combinations that include **at least 2** of her 4 numbers. To compute this, we use the **complement**:
\begin{itemize}
    \item **Number of combinations with 0 of her numbers** (i.e., all 4 numbers are from the other 6 numbers in $S$):$$\binom{6}{4} = 15$$
    \item **Number of combinations with 1 of her numbers** (i.e., 1 from her 4, and 3 from the other 6):$$\binom{4}{1} \cdot \binom{6}{3} = 4 \cdot 20 = 80$$
\end{itemize}
So, the number of combinations with **fewer than 2** of her numbers is: $$15 + 80 = 95$$
Therefore, the number of combinations with **at least 2** of her numbers is:$$210 - 95 = 115$$
Step 4: Compute the conditional probability \newline
We now compute the conditional probability that the **grand prize** was won, **given** that a prize was won: $P(\text{Grand Prize} \mid \text{Prize}) = \frac{1}{115}$
This is already in lowest terms, since 115 = $ 5 \times 23 $, and 1 has no common factors with 115.\newline
Final Step: Find $ m + n $\newline
Given that the probability is $ \frac{m}{n} = \frac{1}{115} $, and $ m = 1 $, $ n = 115 $, we have: $$m + n = 1 + 115 = \boxed{116}$$
&
pass@1: 100\% \newline
pass@8: 100\% \newline
avg@8: 75\% \newline
\\ 
\hline
Velvet-14B &
    There are $10$ choices for the first number, $9$ choices for the second number, $8$ choices for the third number, and $7$ choices for the fourth number, for a total of $10 \cdot 9 \cdot 8 \cdot 7 = 5040$ ways to choose the numbers. There are $10$ ways to choose the numbers if all four are $2$, and $9 \cdot 8 \cdot 7 = 504$ ways to choose the numbers if at least two are $2$. Thus, the probability of winning the grand prize given that she won a prize is $\tfrac{504}{5040} = \tfrac{1}{10}$. The answer is $1+10=\boxed{11}$
&
pass@1: 0\% \newline
pass@8: 0\% \newline
avg@8: 0\% \newline
\\ \hline
FastwebMIIA-7B &
Let us outline the process to tackle this challenge:
\begin{enumerate}
    \item Start by understanding the problem: Jen has to pick 4 distinct numbers from a set of 10 numbers. She wins a prize if at least two of her numbers match the randomly chosen numbers, and the grand prize if all four of her numbers match the randomly chosen numbers.
    \item Calculate the total number of ways to choose 4 numbers from 10, which is $\binom{10}{4}$.
    \item Calculate the number of ways to choose 4 numbers such that none of them match the randomly chosen numbers, which is $\binom{6}{4}$.
    \item Calculate the number of ways to choose 4 numbers such that exactly two of them match the randomly chosen numbers, which is $\binom{4}{2} \cdot \binom{6}{2}$.
    \item Calculate the number of ways to choose 4 numbers such that exactly three of them match the randomly chosen numbers, which is $\binom{2}{3} \cdot \binom{6}{1}$.
    \item Calculate the number of ways to choose 4 numbers such that all four of them match the randomly chosen numbers, which is $\binom{1}{1} \cdot \binom{6}{4}$.
    \item Add up all these cases to get the total number of ways to win a prize, which is $\binom{10}{4} - \binom{6}{4} + \binom{4}{2} \cdot \binom{6}{2} - \binom{2}{3} \cdot \binom{6}{1} + \binom{1}{1} \cdot \binom{6}{4}$.
    \item Calculate the total number of ways to choose 4 numbers from 10, which is $\binom{10}{4}$.
    \item Calculate the number of ways to choose 4 numbers such that at least two of them match the randomly chosen numbers, which is $\binom{10}{4} - \binom{6}{4} + \binom{4}{2} \cdot \binom{6}{2} - \binom{2}{3} \cdot \binom{6}{1} + \binom{1}{1} \cdot \binom{6}{4}$.
    \item The probability of winning the grand prize given that she won a prize is the number of ways to choose 4 numbers such that all four of them match the randomly chosen numbers divided by the total number of ways to choose 4 numbers such that at least two of them match the randomly chosen numbers.
    \item Simplify this to $\frac{\binom{1}{1} \cdot \binom{6}{4}}{\binom{10}{4} - \binom{6}{?}}$
\end{enumerate}
To conclude the reasoning, let's move on with the detailed explanation of the solution. \newline
There are $\binom{10}{4}$ ways to choose 4 numbers from 10. There are $\binom{6}{4}$ ways to choose 4 numbers such that none of them match the randomly chosen numbers. There are $\binom{4}{2} \cdot \binom{6}{2}$ ways to choose 4 numbers such that exactly two of them match the randomly chosen numbers. There are $\binom{2}{3} \cdot \binom{6}{1}$ ways to choose 4 numbers such that exactly three of them match the randomly chosen numbers. There are $\binom{1}{1} \cdot \binom{6}{4}$ ways to choose 4 numbers such that all four of them match the randomly chosen numbers. So the total number of ways to win a prize is $\binom{10}{4} - \binom{6}{4} + \binom{4}{2} \cdot \binom{6}{2} - \binom{2}{3} \cdot \binom{6}{1} + \binom{1}{1} \cdot \binom{6}{4}$. The total number of ways to choose 4 numbers from 10 is $\binom{10}{4}$. The number of ways to choose 4 numbers such that at least two of them match the randomly chosen numbers is $\binom{10}{4} - \binom{6}{4} + \binom{4}{2} \cdot \binom{6}{2} - \binom{2}{3} \cdot \binom{6}{1} + \binom{1}{1} \cdot \binom{6}{4}$. The probability of winning the grand prize given that she won a prize is $\frac{\binom{1}{1} \cdot \binom{6}{4}}{\binom{10}{4} - \binom{6}{4} + \binom{4}{2} \cdot \binom{6}{2} - \binom{2}{3} \cdot \binom{6}{1} + \binom{1}{1} \cdot \binom{6}{4}}$. This simplifies to $\frac{\binom{6}{4}}{\binom{10}{4} - \binom{6}{4} + \binom{4}{2} \cdot \binom{6}{2} - \binom{2}{3} \cdot \binom{6}{1} + \binom{1}{1} \cdot \binom{6}{4}}$. This further simplifies to $\frac{\binom{6}{4}}{\binom{10}{4} - \binom{6}{4} + \binom{4}{2} \cdot \binom{6}{2} - \binom{2}{3} \cdot \binom{6}{1} + \binom{1}{1} \cdot \binom{6}{4}}$. This further simplifies to $\frac{\binom{6}{4}}{\binom{10}{4} - \binom{6}{4} + \binom{4}{2} \cdot \binom{6}{2} - \binom{2}{3} \cdot \binom{6}{1} + \binom{1}{1} \cdot \binom{6}{4}}$.
$\cdots$ 
\newline
&
pass@1: 0\% \newline
pass@8: 0\% \newline
avg@8: 0\% \newline
\\ \hline
Qwen3-4B &
To solve the problem, we are given that Jen picks 4 distinct numbers from the set $ S = {1, 2, 3, \ldots, 10} $, and then 4 numbers are randomly chosen from $ S $. She wins a prize if **at least two** of her numbers match the 4 chosen numbers, and she wins the **grand prize** if **all four** of her numbers match the 4 chosen numbers.\newline
We are asked to find the probability that she **wins the grand prize** given that she **won a prize**, and express this probability as a reduced fraction $ \frac{m}{n} $, then compute $ m + n $.\newline
Step 1: Total Number of Possible Outcomes\newline
The total number of ways to choose 4 numbers from the set $ S $ of 10 elements is: $$\binom{10}{4} = 210$$\newline
Step 2: Probability of Winning the Grand Prize\newline
This is the probability that all 4 of her numbers match the 4 chosen numbers. Since there is only **one** such favorable outcome (her exact set), the probability is:$$P(\text{Grand Prize}) = \frac{1}{210}$$\newline
Step 3: Probability of Winning a Prize (At Least Two Matches)\newline
We compute the probability that **at least two** of her numbers match the 4 chosen numbers. This is the complement of the event that **fewer than two** numbers match, i.e., **0 or 1** match.\newline
Let{\textbackslash}u2019s compute the number of favorable outcomes for each case:
\begin{itemize}
    \item **0 matches**: Choose 4 numbers from the 6 that are **not** in her set.  $$ \binom{6}{4} = 15 $$
    \item **1 match**: Choose 1 number from her 4, and 3 from the 6 not in her set. $$ \binom{4}{1} \cdot \binom{6}{3} = 4 \cdot 20 = 80 $$
\end{itemize}
So, the number of outcomes with **at least two matches** is: $$210 - 15 - 80 = 115$$ Thus, the probability of winning a prize is: $$P(\text{At least 2 matches}) = \frac{115}{210} = \frac{23}{42}$$\newline
Step 4: Conditional Probability\newline
We are asked to find the probability that she **won the grand prize** given that she **won a prize**. That is:$$P(\text{Grand Prize} \mid \text{At least 2 matches}) = \frac{P(\text{Grand Prize})}{P(\text{At least 2 matches})}$$
Substituting the values:$$P = \frac{1/210}{115/210} = \frac{1}{115}$$\newline
Step 5: Final Answer\newline
The probability is $ \frac{1}{115} $, so $ m = 1 $, $ n = 115 $, and: $$m + n = 1 + 115 = \boxed{116}$$
& 
pass@1: 100\% \newline
pass@8: 100\% \newline
avg@8: 100\% \newline
\\ \hline
Qwen3-8B &
To solve the problem, we are given a lottery scenario where:
\begin{itemize}
    \item Jen picks **4 distinct numbers** from the set $ S = {1, 2, 3, \ldots, 10} $.
    \item The lottery randomly selects **4 numbers** from the same set $ S $
    \item Jen **wins a prize** if **at least two** of her numbers match the lottery numbers.
    \item She **wins the grand prize** if **all four** of her numbers match the lottery numbers.
\end{itemize}
We are asked to compute the **conditional probability** that Jen wins the **grand prize**, **given** that she has won a prize. That is:$$P(\text{Grand Prize} \mid \text{Prize}) = \frac{P(\text{Grand Prize} \cap \text{Prize})}{P(\text{Prize})}$$
Since winning the grand prize implies that she has also won a prize (because all four numbers match), the intersection is just the event of winning the grand prize. Therefore:$$P(\text{Grand Prize} \mid \text{Prize}) = \frac{P(\text{Grand Prize})}{P(\text{Prize})}$$
Step 1: Total Number of Possible Lottery Combinations\newline
The total number of ways to choose 4 numbers from 10 is:$$\binom{10}{4} = 210$$
Step 2: Probability of Winning the Grand Prize\newline
Jen wins the grand prize if the lottery draws **exactly** the 4 numbers she picked. Since there is only **one** such combination (her chosen numbers), the probability is:$$P(\text{Grand Prize}) = \frac{1}{210}$$
Step 3: Probability of Winning a Prize (At Least Two Matches)\newline
To compute this, we consider the number of lottery combinations that have **at least two** of Jen's numbers. This is the complement of the cases where **fewer than two** of her numbers match.\newline
Let{\textbackslash}u2019s compute the number of combinations with:
\begin{itemize}
    \item **0 matches**: Choose 4 numbers from the 6 not in her set: $$ \binom{6}{4} = 15 $$
    \item **1 match**: Choose 1 number from her 4, and 3 from the 6 not in her set: $$ \binom{4}{1} \cdot \binom{6}{3} = 4 \cdot 20 = 80 $$
\end{itemize}
So, the number of combinations with **fewer than two matches** is:$$15 + 80 = 95$$
Therefore, the number of combinations with **at least two matches** is:$$210 - 95 = 115$$
Thus, the probability of winning a prize is:$$P(\text{Prize}) = \frac{115}{210} = \frac{23}{42}$$
Step 4: Compute the Conditional Probability\newline \newline Now, we compute the conditional probability: $$P(\text{Grand Prize} \mid \text{Prize}) = \frac{P(\text{Grand Prize})}{P(\text{Prize})} = \frac{\frac{1}{210}}{\frac{23}{42}} = $$
$$ \frac{1}{210} \cdot \frac{42}{23} = \frac{42}{210 \cdot 23} = \frac{1}{5 \cdot 23} = \frac{1}{115}$$
Step 5: Final Answer\newline
The probability is $ \frac{1}{115} $, where $ m = 1 $ and $ n = 115 $. Since they are relatively prime, the final answer is:$$m + n = 1 + 115 = \boxed{116}$$
& 
pass@1: 100\% \newline
pass@8: 100\% \newline
avg@8: 100\% \newline
\\ \hline
deepseek-moe-16b-chat &
First, let's find the probability of Jen winning a prize. There are a total of $\binom{10}{4}=210$ possible ways to choose 4 numbers from the set $S$. If Jen wins a prize, she must have chosen 2 of the 4 randomly chosen numbers. So, the probability of winning a prize is $\frac{210}{210}=\frac{1}{1}$.\newline
Now, let's find the probability of Jen winning the grand prize given that she won a prize. There are $\binom{10}{4}=210$ ways to choose 4 numbers from the set $S$ if Jen wins the grand prize. Since she won a prize, she must have chosen 2 of the 4 randomly chosen numbers. So, the probability of winning the grand prize given that she won a prize is $\frac{210}{210}=\frac{1}{1}$.\newline
Therefore, the probability of Jen winning the grand prize given that she won a prize is $\frac{1}{1}=\frac{1}{1}$.\newline
So, $m+n=1+1=\boxed{2}$.\newline
The answer is: 2 \newline
& 
pass@1: 0\% \newline
pass@8: 0\% \newline
avg@8: 0\% \newline
\\ \hline
gpt-oss-20b &
<|channel|>analysis<|message|>We need probability that all four numbers match given that at least two match. So conditional probability. We have sample space: all possible 4-number sets chosen from 10 numbers. Jen picks a set of 4 distinct numbers. The lottery draws 4 numbers uniformly from 10. Equivalent to random draw of 4 numbers; we can treat Jen's set fixed. We want probability that intersection size $>=2$. Among those, probability that intersection size $=4$. Compute: total number of possible draws: $C(10,4)=210$. Number of draws that match exactly $k$ of Jen's numbers: choose $k$ from her $4$, and choose $4-k$ from remaining $6$. So count $= C(4,k)*C(6,4-k)$. We need $k>=2$. So favorable for prize: $k=2,3,4$. Grand prize: $k=4$. Compute counts:
$$k=2: C(4,2)*C(6,2)=6*15=90.$$
$$k=3: C(4,3)*C(6,1)=4*6=24.$$
$$k=4: C(4,4)*C(6,0)=1*1=1.$$
Total prize draws: $90+24+1=115$.\newline
Grand prize draws: $1$.\newline
Probability grand prize given prize $= 1/115$.\newline
Simplify fraction: $\frac{1}{115}$ already in lowest terms. $m=1$, $n=115$, $sum=116$.\newline
But check: Are we missing that Jen picks $4$ distinct numbers; but lottery picks $4$ numbers from $S$. Yes. But maybe the lottery picks $4$ numbers with replacement? No, \"4 numbers are randomly chosen from S\" likely without replacement. So correct. Thus answer $116$.\newline
But double-check: Are we assuming that Jen's numbers are distinct? Yes. So intersection counts as above. Thus answer $116$.\newline
<|end|><|start|>assistant<|channel|>final<|message|>For a fixed choice of Jen{\textbackslash}u2019s four distinct numbers, the lottery draws a 4{\textbackslash}2011element subset of the set  $$[S={1,2,\dots ,10}]$$ uniformly at random. The total number of possible draws is  $$[\binom{10}{4}=210 ].$$\newline
Count the draws that give a prize\newline
Let $(k)$ be the number of Jen{\textbackslash}u2019s numbers that appear in the lottery draw. For a given $(k)$ the number of draws is $$[\binom{4}{k}\binom{6}{4-k}],$$ because we choose $(k)$ of Jen{\textbackslash}u2019s 4 numbers and $(4-k)$ of the remaining 6 numbers. \newline
We need $$(k\ge 2).$$
$$\begin{aligned}
k=2 &: \binom{4}{2}\binom{6}{2} = 6 \cdot 15 = 90, \\
k=3 &: \binom{4}{3}\binom{6}{1} = 4 \cdot 6 = 24, \\
k=4 &: \binom{4}{4}\binom{6}{0} = 1 \cdot 1 = 1.
\end{aligned}$$
Total draws that give at least a prize:$$[90+24+1=115].$$
Count the draws that give the grand prize
\newline The grand prize occurs only when all four of Jen{\textbackslash}u2019s numbers are drawn: $$[k=4 \quad\Longrightarrow\quad 1\text{ draw}].$$
Conditional probability 
\[
P(\text{grand prize}\mid\text{prize})
= 
\frac{\text{\# grand{\textbackslash}u2011prize draws}}{\text{\# prize draws}}
=
\frac{1}{115}.
\]

The fraction $(\frac{1}{115})$ is already in lowest terms, so 
\[
m=1, n=115
\]

\[
m+n = 1+115 = 116]
\]
\[
\boxed{116}]
\]
&
pass@1: 100\% \newline
pass@8: 100\% \newline
avg@8: 100\% \newline
\\ \hline
Moonlight-16B-A3B-Instruct &
To solve the problem, we need to determine the probability that Jen wins the grand prize given that she wins a prize. We will break this down into several steps.\newline
Step 1: Calculate the total number of ways to choose 4 numbers from 10.\newline
The total number of ways to choose 4 numbers from a set of 10 is given by the binomial coefficient:$$[\binom{10}{4} = \frac{10!}{4!(10-4)!} = \frac{10!}{4!6!} = 210]$$
Step 2: Calculate the number of ways to choose 4 numbers such that none of them are among the 4 numbers Jen picked.\newline
If Jen does not win a prize, none of her 4 numbers are among the 4 numbers chosen. The number of ways to choose 4 numbers from the remaining 6 numbers is:$$[\binom{6}{4} = \frac{6!}{4!(6-4)!} = \frac{6!}{4!2!} = 15]$$
Step 3: Calculate the number of ways to choose 4 numbers such that at least two of them are among the 4 numbers Jen picked.\newline
To find this, we subtract the number of ways to choose 4 numbers such that none of them are among the 4 numbers Jen picked from the total number of ways to choose 4 numbers:$$[\binom{10}{4} - \binom{6}{4} = 210 - 15 = 195]$$
Step 4: Calculate the number of ways to choose 4 numbers such that all four of them are among the 4 numbers Jen picked.\newline
This is the number of ways to choose 4 numbers from the 4 numbers Jen picked:$$[\binom{4}{4} = 1]$$
Step 5: Calculate the probability that Jen wins the grand prize given that she wins a prize.\newline
The probability that Jen wins the grand prize given that she wins a prize is the ratio of the number of ways to choose 4 numbers such that all four are among the 4 numbers Jen picked to the number of ways to choose 4 numbers such that at least two are among the 4 numbers Jen picked:$$[\frac{\binom{4}{4}}{\binom{10}{4} - \binom{6}{4}} = \frac{1}{195}]$$
Step 6: Simplify the fraction and find $( m + n )$.\newline
The fraction $(\frac{1}{195})$ is already in simplest form, where $( m = 1 )$ and $( n = 195 )$. Therefore, $( m + n = 1 + 195 = 196)$.\newline
Thus, the final answer is:$$[\boxed{196}]$$
& 
pass@1: 0\% \newline 
pass@8: 0\% \newline
avg@8: 0\% \newline
\\ \hline
gpt5-nano &
Let $X$ = number of matches between Jen's 4 numbers and the lottery's 4 numbers.
For fixed Jen set, $$(X\sim \text{Hypergeometric}(N=10, K=4, n=4)),$$ so $$[P(X=k)=\frac{\binom{4}{k}\binom{6}{4-k}}{\binom{10}{4}},\quad k=0,1,2,3,4].$$ Thus $$[P(X=0)=\frac{15}{210},\quad P(X=1)=\frac{80}{210},\quad$$ \newline $$P(X=2)=\frac{90}{210},\quad P(X=3)=\frac{24}{210},\quad P(X=4)=\frac{1}{210}].$$ Win a prize $$(\iff X\ge 2.$$ Hence $$[P(\text{prize})=P(X\ge 2)=\frac{90+24+1}{210}=\frac{115}{210}=\frac{23}{42}].$$ Grand prize $$(\iff X=4).$$ Therefore $$[P(\text{grand prize} \mid \text{prize})=\frac{P(X=4)}{P(X\ge 2)}=\frac{1/210}{115/210}=\frac{1}{115}].$$ Thus $$(m=1, n=115)$$ $$(gcd( (1,115)=1)),$$ and $$[m+n=1+115= \boxed{116}].$$
&
pass@1: 100\% \newline
pass@8: 100\% \newline
avg@8: 100\% \newline
\\ \hline
\end{longtable}
\end{scriptsize}


\subsection{Example responses for AIME-2025}
\label{appendix:aime25_examples}
\renewcommand{\arraystretch}{1.3}
\setlength{\extrarowheight}{2pt}

This section reports representative examples from the AIME25 benchmark, specifically for the \texttt{aime25\_pass\_avg\_8} task. Table \ref{tab:aime25-data-sample} presents a selected input instance (ID 23), while Table \ref{tab:aime25-responses} summarizes the corresponding model outputs. In particular, we provide pass@\_ responses for a meaningful subset of models.

{
\renewcommand{\arraystretch}{1.3}
\setlength{\extrarowheight}{2pt}
\begin{table}[th]
\centering
\footnotesize
\begin{tabular}{|p{0.6\linewidth}|c|}
\hline
\textbf{Problem} & \textbf{Answer} \\ \hline
There are $n$ values of $x$ in the interval $0 < x < 2\pi$ where
\[
f(x) = \sin\!\bigl(7\pi \cdot \sin(5x)\bigr) = 0.
\]
For $t$ of these $n$ values of $x$, the graph of $y = f(x)$ is tangent to the $x$-axis. Find $n + t$.
& 149 \\ \hline
\end{tabular}
\caption{Representative example (Sample ID 23) from the AIME25 benchmark. The Problem column reports the original question, while the Answer column contains the final ground-truth value used for evaluation.}
\label{tab:aime25-data-sample}
\end{table}
}

\begin{scriptsize}
\renewcommand{\arraystretch}{1.3}
\setlength{\extrarowheight}{2pt}
\begin{longtable}[th]{|p{0.1\linewidth}|p{0.75\linewidth}|p{0.11\linewidth}|}
\caption{Representative model responses on the AIME25 benchmark (sample ID 23). For each model, the Answer column reports the pass@1 response, while the Result column presents the corresponding pass@1, pass@8, and avg@8 metrics for the given data sample. We report pass@1 outputs for each model. NOTE: the responses generated by LLaMAntino-3-ANITA-8B and Ministral-3-8B-Instruct-2512-BF16 are truncated due to repetitive token generation. Minor formatting adjustments have been applied to model responses for clarity and consistency of presentation.}
\label{tab:aime25-responses} \\
\hline
\textbf{Model} & \textbf{Answer} & \textbf{Result} \\ \hline
EngGPT2-16B-A3B &
To solve the problem, we are given the function: $$f(x) = \sin(7\pi \cdot \sin(5x)), \quad \text{for } 0 < x < 2\pi$$ We are to find the number $ n $ of values of $ x $ in this interval where $ f(x) = 0 $, and the number $ t $ of those values where the graph of $ f(x) $ is **tangent** to the $ x $-axis (i.e., where $ f(x) = 0 $ and $ f'(x) = 0 $), and then compute $ n + t $.\newline
Step 1: Count the number of zeros of $ f(x) $\newline
The function $ f(x) = \sin(7\pi \cdot \sin(5x)) $ is zero when the argument of the sine function is an integer multiple of $ \pi $:
$$7\pi \cdot \sin(5x) = k\pi \quad \Rightarrow \quad \sin(5x) = \frac{k}{7}$$
So, for each integer $ k $, we solve $ \sin(5x) = \frac{k}{7} $ in the interval $ (0, 2\pi) $. The sine function $ \sin(\theta) = c $ has:
\begin{itemize}
    \item **Two solutions** in $ (0, 2\pi) $ if $ c \in (-1, 1) $
    \item **One solution** if $ c = \pm 1 $
    \item **No solution** if $ |c| > 1 $
\end{itemize}
So, for each $ k \in \mathbb{Z} $, we check if $ \left| \frac{k}{7} \right| \leq 1 \Rightarrow |k| \leq 7 $. Thus, $ k \in {-7, -6, -5, -4, -3, -2, -1, 0, 1, 2, 3, 4, 5, 6, 7} $, a total of **15 values**. Now, for each $ k $, we count the number of $ x \in (0, 2\pi) $ such that $ \sin(5x) = \frac{k}{7} $:
\begin{itemize}
    \item **$ k = \pm 7 $:** $ \sin(5x) = \pm 1 $, which has **1 solution per period** of the sine function. Since $ \sin(5x) $ has period $ \frac{2\pi}{5} $, and $ x \in (0, 2\pi) $ spans 5 full periods, we get **5 solutions** for each $ k = 7 $ and $ k = -7 $.
    \item **$ k = \pm 1 $:** $ \sin(5x) = \pm 1/7 $, which is in $ (-1, 1) $, so **2 solutions per period**. With 5 periods, that's **10 solutions** for each $ k = 1 $ and $ k = -1 $.
    \item **For $ |k| = 2, 3, 4, 5, 6 $:** $ \sin(5x) = \pm \frac{k}{7} $, still in $ (-1, 1) $, so **10 solutions** for each of these 5 values of $ k $ (excluding $ \pm 1 $)
    \item **$ k = 0 $:** $ \sin(5x) = 0 $, which has **2 solutions per period**, 5 periods {\textbackslash}u21d2 **9 solutions** in $ (0, 2\pi) $
\end{itemize}

& 
pass@1: 0\% \newline
pass@8: 100\% \newline
avg@8: 62.5\% \newline
\\ 

EngGPT2-16B-A3B (continued) &

Now, summing all zeros:
\begin{itemize}
    \item $ k = -7 $: 5
    \item $ k = -6 $ to $ -1 $ and $ 1 $ to $ 6 $: 11 values $\times 10 = 110$
    \item $ k = 0 $: 9
    \item $ k = \pm 7 $: 10
\end{itemize}
$$n = 5 + 110 + 9 + 10 = \boxed{134}$$
Step 2: Count the number of points where the graph is tangent to the $ x $-axis**\newline
A graph is tangent to the $ x $-axis at a point $ x $ if: $ f(x) = 0 $, and $ f'(x) = 0 $.
We compute the derivative: $$f'(x) = \frac{d}{dx} \sin(7\pi \cdot \sin(5x)) = \cos(7\pi \cdot \sin(5x)) \cdot 35\pi \cdot \cos(5x)$$
So, the derivative is zero when: $\cos(5x) = 0 $, or $ \cos(7\pi \cdot \sin(5x)) = 0 $
We already know the 10 points where $ \cos(5x) = 0 $, and at those points, $ f(x) = 0 $, so they are **tangent points**. That gives us **10** such points. Now, we look for the other 100 points where $ f(x) = 0 $ and $ f'(x) = 0 $, i.e., where: $$\sin(5x) = \frac{m + 1/2}{7} \quad \text{for some integer } m$$
We determine how many such $ x \in (0, 2\pi) $ satisfy this equation, and at which $ \cos(5x) \ne 0 $ (so the tangency is not already counted in the 10 points). The values of $ m $ such that $ \left| \frac{m + 1/2}{7} \right| \le 1 $ are:$$m \in {-7, -6, -5, -4, -3, -2, -1, 0, 1, 2, 3, 4, 5, 6} \Rightarrow 14 \text{ values}$$ Excluding $ m = -7 $ and $ m = 6 $ (which correspond to $ \sin(5x) = \pm 1/7 $, the 10 points already counted), we are left with **12 values** of $ m $, each of which gives **10** solutions to $ \sin(5x) = c $, and at which $ f'(x) = 0 $, since $ \cos(7\pi \cdot \sin(5x)) = 0 $. So, the number of such points is: $$12 \times 10 = 120$$ Adding the 10 from the earlier set: $$t = 10 + 120 = \boxed{130}$$\newline
Final Step: Compute $ n + t $** $$n + t = 134 + 130 = \boxed{264}$$
& 
\\ \hline
Velvet-14B &
Since $f(x)=\sin(7\pi\cdot\sin(5x))=\sin(35\pi x)=\sin(5(7\pi x))=\sin(5k\pi+35\pi x)=0$, we have $5k\pi+35\pi x=0$, which implies $x=\frac{5k}{35}=\frac{k}{7}$. Since $0<x<2\pi$, we have $0<\frac{k}{7}<2\pi$, which implies $1\leqslant k\leqslant 14$. Therefore, there are $14$ values of $x$ in the interval $0<x<2\pi$ where $f(x)=0$, and the graph of $y=f(x)$ is tangent to the $x$-axis at $t=14$ points. Thus, $n+t=14+14=\boxed{28}$.
& 
pass@1: 0\% \newline
pass@8: 0\% \newline
avg@8: 0\% \newline
\\ \hline
LLaMAntino-3-ANITA-8B &
Since the sine function is zero only at the integer multiples of $\pi$, the seven possible values of $5x$ are $0, 2\pi/5, 4\pi/5, \pi, 3\pi/5, 6\pi/5, 8\pi/5$, and the corresponding values of $x$ are $\frac{0}{5}, \frac{2\pi}{5}, \frac{4\pi}{5}, \pi,\frac{6\pi}{5}, \frac{8\pi}{5}, 2\pi$, and we discard the last one since it is not in the interval, so the seven values of $x$ are $\frac{2\pi}{5}, \frac{4\pi}{5}, \pi,\frac{6\pi}{5}, \frac{8\pi/5-2\pi}{5}, \frac{2\pi-2\pi/5}{5}, \frac{4\pi-4\pi/5}{5}$, and the last three are $\frac{\pi/5}{1}, \frac{3\pi/5}{1}, \frac{3\pi/5}{1}$, and we discard the last one since it is the same, so the final 3 are $\frac{\pi/5}{1}, \frac{3\pi/5}{1}, \frac{\pi/5}{1}$ of a correct one that was discarded in mistake, it is two not three, the three are $\frac{\pi}{5}, \frac{3\pi}{5}, 2 /$ of a one that is there with a different value of a solution, it is one, the one is $2 /$ of a one that is there with a different of of a of a a of a a a of a of a a of a of a no of a of a of a of a a of a of a of a of a of a not, it is one that is there, it is one, the one is there, it is one, the one that is there is one, the one is one, the one is one, the one is one, the one is one, the one, the one, the one, the one, the one, the one, the one, the one, the one, the one, the one, the one, the one, the one, the one, the one, the one, the one, the one, the one, the one, the one, the one, the one, the one, the one, the one, the one, the one, the one, the one, the one, the one, the one, the one, the one, the one, the one, the one, the one, the one, the one, the one, the one, the one, the one, the one, the one, the one, the one, the one, the one, the one, the one, the one, the one, the one, the one, the one, the one, the one, the one, the one, the one, the one, the one, the one, the one, the one, the one, the one, the one, the one, the one, the one, the one, the one, the one, the one, the one, the one, the one, the one, the one, the one, the one, the one, the one, the one, the one, the one, the one, the one, the one, the one, the one, the one, the one, the one, the one, the one, the one, the one, the one, the one, the one, the one, the one, the one, the one, the one, the one, the one, the one, the one, the one, the one, the one, the one, the one, the one, the one, the one, the one, the one, the one, the one, the one, the one, the one, the one, the one, the one, the one, the one, the one, the one, the one, the one, the one, the one, the one, the one, the seven that are there, are seven, the seven are there, the seven, the seven, the two are not found by counting the seven as eight, the two are found by the seven not being eight, the two are there, the two are found, the two are found there, the two are there, the two, the two, the two, the two, the two, the two, the two, the two, the two, the two, the two, the two, the two, the two, the two, the two, the two, the two, the one is found how the two is found, it is found by the one of a not of a of a of a of a of a of a of a of a of a of a of a of a of a of a of a of a of a of a of a of a of a of a of a of a of a of a of a of a of a of a of a of a of a of a of a of a of a of a of a of a of a of a $\cdots$ \newline
& 
pass@1: 0\% \newline
pass@8: 0\% \newline
avg@8: 0\% \newline
\\ \hline
Qwen3-4B &
To solve the problem, we are given the function:
$$f(x) = \sin(7\pi \cdot \sin(5x))$$
We are to find the number of values of $ x $ in the interval $ 0 < x < 2\pi $ such that $ f(x) = 0 $, and then determine how many of those values correspond to points where the graph of $ f(x) $ is **tangent** to the x-axis. Finally, we are to compute $ n + t $, where $ n $ is the total number of zeros of $ f(x) $, and $ t $ is the number of those zeros where the graph is tangent to the x-axis.\newline
Step 1: Solve $ f(x) = 0 $\newline
We start by solving the equation:
$$\sin(7\pi \cdot \sin(5x)) = 0$$
This implies: $7\pi \cdot \sin(5x) = k\pi \quad \text{for some integer } k$\newline
Dividing both sides by $ \pi $, we get: $\sin(5x) = \frac{k}{7}$ \newline
Now, since the sine function is bounded between $-1$ and $1$, the values of $ k $ must satisfy:
$$-1 \leq \frac{k}{7} \leq 1 \Rightarrow -7 \leq k \leq 7$$
So $ k \in \{-7, -6, \dots, 6, 7\} $, a total of 15 values. For each such $ k $, we solve: $$\sin(5x) = \frac{k}{7}$$
Let {\textbackslash}u2019s denote $ \theta = 5x $, so $ \theta \in (0, 10\pi) $, since $ x \in (0, 2\pi) $. We now count the number of solutions to $ \sin(\theta) = \frac{k}{7} $ in $ \theta \in (0, 10\pi) $.\newline
- For $ k = \pm7 $, we have $ \sin(\theta) = \pm1 $, which has **1 solution per $2\pi$ interval**. Since $ \theta \in (0, 10\pi) $, which is 5 full periods of $ 2\pi $, there are **5 solutions** for each of $ k = \pm7 $.\newline
- For $ k = 0 $, we have $ \sin(\theta) = 0 $, which has **1 solution per $\pi$ interval**. Since $ \theta \in (0, 10\pi) $, there are **10 intervals of $\pi$**, but the endpoints $ \theta = 0 $ and $ \theta = 10\pi $ are **excluded**, so we only count the **9 interior solutions**.\newline
- For all other $ k \in \{-6, \dots, 6\} \setminus \{0\} $, we have $ \sin(\theta) = \frac{k}{7} \in (-1, 1) \setminus \{0, \pm1\} $, which has **2 solutions per $2\pi$ interval**. Since $ \theta \in (0, 10\pi) $, there are **5 intervals of $2\pi$**, so **10 solutions per such $ k $**.\newline
Now, we count the total number of solutions:\newline
- $ k = \pm7 $: 5 solutions each $ \rightarrow 2 \times 5 = 10 $\newline
- $ k = 0 $: 9 solutions\newline
- $ k = \pm1, \pm2, \pm3, \pm4, \pm5, \pm6 $: 6 values of $ k $, each with 10 solutions {\textbackslash}u2192 $ 6 \times 10 = 60 $\newline
So total number of solutions:
$$n = 10 + 9 + 60 = 139$$
Step 2: Determine the number of tangent points $ t $\newline
A point where the graph is **tangent to the x-axis** must satisfy both:\newline
1. $ f(x) = 0 $\newline
2. $ f'(x) = 0 $\newline
We compute the derivative: $$f(x) = \sin(7\pi \cdot \sin(5x)) \Rightarrow f'(x) = 35\pi \cos(7\pi \cdot \sin(5x)) \cdot \cos(5x)$$
So, $ f'(x) = 0 $ when either:\newline
- $ \cos(7\pi \cdot \sin(5x)) = 0 $, or\newline
- $ \cos(5x) = 0 $\newline
But from the condition $ f(x) = 0 $, we have: $$\sin(5x) = \frac{k}{7}\Rightarrow 7\pi \cdot \sin(5x) = k\pi \Rightarrow \cos(7\pi \cdot \sin(5x)) = \cos(k\pi) = (-1)^k$$
So $ \cos(7\pi \cdot \sin(5x)) \neq 0 $, unless $ k = \pm7 $, in which case $ \sin(5x) = \pm1 $, and $ \cos(5x) = 0 $. Thus, the only values of $ k $ for which $ f(x) = 0 $ and $ f'(x) = 0 $ are $ k = \pm7 $, and for each of these, we have **5 solutions** in $ x \in (0, 2\pi) $.\newline
So: $t = 5 + 5 = 10$\newline
Final Step: Compute $ n + t $\newline
$$n + t = 139 + 10 = \boxed{149}$$
& 
pass@1: 100\% \newline
pass@8: 100\% \newline
avg@8: 100\% \newline
\\ \hline
Qwen3-8B &
To solve the problem, we are given the function: $$f(x) = \sin(7\pi \cdot \sin(5x))$$ We are to find the number of values of $ x $ in the interval $ 0 < x < 2\pi $ for which $ f(x) = 0 $, and among these, how many are points where the graph of $ f(x) $ is tangent to the x-axis (i.e., where $ f(x) = 0 $ and $ f'(x) = 0 $). Finally, we are to compute $ n + t $, where $ n $ is the total number of zeros and $ t $ is the number of tangent points.\newline
Step 1: Find the number of solutions to $ f(x) = 0 $\newline
We start by solving: $\sin(7\pi \cdot \sin(5x)) = 0$. This occurs when the argument of the sine function is an integer multiple of $ \pi $: $7\pi \cdot \sin(5x) = k\pi \quad \text{for some integer } k$ Dividing both sides by $ \pi $: $\sin(5x) = \frac{k}{7}$. Since the sine function has a range of $ [-1, 1] $, we require: $$-1 \leq \frac{k}{7} \leq 1 \Rightarrow -7 \leq k \leq 7$$ So $ k \in {-7, -6, \dots, 6, 7} $, which gives 15 possible values of $ k $. Now, for each $ k $, we solve: $\sin(5x) = \frac{k}{7}$ Let $ \theta = 5x $, so $ \theta \in (0, 10\pi) $, since $ x \in (0, 2\pi) $. The number of solutions to $ \sin(\theta) = c $ in $ (0, 10\pi) $ depends on the value of $ c $:
\begin{itemize}
    \item If $ c = \pm1 $, then $ \sin(\theta) = \pm1 $ has **5 solutions** in $ (0, 10\pi) $
    \item If $ c = 0 $, then $ \sin(\theta) = 0 $ has **9 solutions** in $ (0, 10\pi) $
    \item If $ c \in (-1, 1) \setminus {0} $, then $ \sin(\theta) = c $ has **10 solutions** in $ (0, 10\pi) $
\end{itemize}
Now, we count the number of solutions for each $ k $:
\begin{itemize}
    \item $ k = \pm7 $: $ \sin(5x) = \pm1 $ {\textbackslash}u2192 5 solutions each {\textbackslash}u2192 total: $ 2 \times 5 = 10 $
    \item $ k = 0 $: $ \sin(5x) = 0 $ {\textbackslash}u2192 9 solutions
    \item - $ k = \pm1, \pm2, \pm3, \pm4, \pm5, \pm6 $: 12 values, each with 10 solutions {\textbackslash}u2192 $ 12 \times 10 = 120 $
\end{itemize}
So the total number of solutions is: $$n = 10 + 9 + 120 = 139$$
Step 2: Find the number of tangent points $ t $ A point is a **tangent point** to the x-axis if $ f(x) = 0 $ and $ f'(x) = 0 $. We compute the derivative: $$f'(x) = \frac{d}{dx} \left[ \sin(7\pi \cdot \sin(5x)) \right] = \cos(7\pi \cdot \sin(5x)) \cdot 35\pi \cdot \cos(5x)$$ So: $f'(x) = 35\pi \cdot \cos(5x) \cdot \cos(7\pi \cdot \sin(5x))$. For $ f'(x) = 0 $, we need: $$\cos(5x) = 0 \quad \text{or} \quad \cos(7\pi \cdot \sin(5x)) = 0$$ But from earlier, we know that $ f(x) = 0 $ implies $ \sin(5x) = \frac{k}{7} $, so: $$7\pi \cdot \sin(5x) = k\pi \Rightarrow \cos(7\pi \cdot \sin(5x)) = \cos(k\pi) = (-1)^k$$ This is never zero, so the only way for $ f'(x) = 0 $ is: $$\cos(5x) = 0$$ So we solve: $$\cos(5x) = 0 \Rightarrow 5x = \frac{\pi}{2} + n\pi \Rightarrow x = \frac{\pi}{10} + \frac{n\pi}{5}$$ We find all such $ x \in (0, 2\pi) $: $$0 < \frac{\pi}{10} + \frac{n\pi}{5} < 2\pi \Rightarrow 0 < \frac{1 + 2n}{10} < 2 \Rightarrow -0.5 < n < 9.5$$ So $ n = 0, 1, 2, \dots, 9 $, giving **10** values of $ x $. Now, we check whether these 10 values also satisfy $ f(x) = 0 $: From earlier, when $ \cos(5x) = 0 $, we have $ \sin(5x) = \pm1 $, so: $$f(x) = \sin(7\pi \cdot \sin(5x)) = \sin(7\pi \cdot (\pm1)) = \sin(\pm7\pi) = 0$$ So all 10 values satisfy both $ f(x) = 0 $ and $ f'(x) = 0 $, hence they are **tangent points**. Thus, $ t = 10 $\newline
Final Step: Compute $ n + t $
$$n + t = 139 + 10 = \boxed{149}$$
& 
pass@1: 100\% \newline
pass@8: 100\% \newline
avg@8: 87.5\% \newline
\\ \hline
Llama-3.1-8B-Instruct &
To find the values of $x$ where $f(x)=\sin(7\pi\cdot\sin(5x))=0$, we need to find the values of $x$ where $\sin(5x)=0$ or where $7\pi\cdot\sin(5x)$ is an integer multiple of $\pi$. First, let's find the values of $x$ where $\sin(5x)=0$. This occurs when $5x=k\pi$ for some integer $k$. Since $0<x<2\pi$, we have $0<5x<10\pi$, so $0<k<20$. Therefore, there are 20 values of $x$ where $\sin(5x)=0$. Next, let's find the values of $x$ where $7\pi\cdot\sin(5x)$ is an integer multiple of $\pi$. This occurs when $\sin(5x)=\frac{m}{7}$ for some integer $m$ with $1\leq m\leq 6$. Since $\sin(5x)$ is between $-1$ and $1$, we have $-\frac{7}{7}\leq\frac{m}{7}\leq\frac{7}{7}$, so $-1\leq m\leq 1$. However, since $m$ is an integer, we have $m=-1,0,1$. Therefore, there are 3 values of $x$ where $7\pi\cdot\sin(5x)$ is an integer multiple of $\pi$. However, we have counted the values of $x$ where $\sin(5x)=0$ and $7\pi\cdot\sin(5x)$ is an integer multiple of $\pi$ twice, so we need to subtract these values once. There are 3 values of $x$ where $\sin(5x)=0$ and $7\pi\cdot\sin(5x)$ is an integer multiple of $\pi$. Therefore, there are $20+3-3=20$ values of $x$ where $f(x)=0$. To find the values of $x$ where the graph of $y=f(x)$ is tangent to the $x$-axis, we need to find the values of $x$ where $f'(x)=0$. We have $f(x)=\sin(7\pi\cdot\sin(5x))$, so $f'(x)=7\pi\cdot5\cos(7\pi\cdot\sin(5x))\cos(5x)$. Therefore, $f'(x)=0$ when $\cos(7\pi\cdot\sin(5x))=0$ or when $\cos(5x)=0$. We have $\cos(7\pi\cdot\sin(5x))=0$ when $7\pi\cdot\sin(5x)=\frac{(2k+1)\pi}{2}$ for some integer $k$. This occurs when $\sin(5x)=\frac{2k+1}{14}$ for some integer $k$ with $0\leq k\leq 13$. Since $\sin(5x)$ is between $-1$ and $1$, we have $-1\leq\frac{2k+1}{14}\leq 1$, so $-7\leq k\leq 6$. However, since $k$ is an integer, we have $k=-7,-6,\dots,6$. Therefore, there are 14 values of $x$ where $\cos(7\pi\cdot\sin(5x))=0$. We have $\cos(5x)=0$ when $5x=\frac{(2k+1)\pi}{2}$ for some integer $k$. Since $0<x<2\pi$, we have $0<5x<10\pi$, so $0<k<20$. Therefore, there are 20 values of $x$ where $\cos(5x)=0$. However, we have counted the values of $x$ where $\cos(7\pi\cdot\sin(5x))=0$ and $\cos(5x)=0$ twice, so we need to subtract these values once. There are 4 values of $x$ where $\cos(7\pi\cdot\sin(5x))=0$ and $\cos(5x)=0$. Therefore, there are $14+20-4=30$ values of $x$ where $f'(x)=0$. Therefore, $n=20$ and $t=30$, so $n+t=20+30=\boxed{50}$.
& 
pass@1: 0\% \newline
pass@8: 0\% \newline
avg@8: 0\% \newline
\\ \hline
gemma-3-12b-it &
Let $f(x) = \sin(7\pi \sin(5x))$. We want to find the number of values of $x$ in the interval $0 < x < 2\pi$ such that $f(x) = 0$. We have $\sin(7\pi \sin(5x)) = 0$. This means that $7\pi \sin(5x) = k\pi$ for some integer $k$. Thus, $\sin(5x) = \frac{k}{7}$ for some integer $k$. Since $-1 \le \sin(5x) \le 1$, we have $-1 \le \frac{k}{7} \le 1$, which means $-7 \le k \le 7$. So $k$ can take values from $-7$ to $7$, which is a total of $7 - (-7) + 1 = 15$ values. For each value of $k$, we have $\sin(5x) = \frac{k}{7}$. Since $0 < x < 2\pi$, we have $0 < 5x < 10\pi$. For each value of $\frac{k}{7}$, there are two values of $5x$ in the interval $(0, 2\pi)$ except when $\frac{k}{7} = \pm 1$. If $\frac{k}{7} = 1$, then $\sin(5x) = 1$, so $5x = \frac{\pi}{2} + 2m\pi$ for some integer $m$. Then $x = \frac{\pi}{10} + \frac{2m\pi}{5}$. Since $0 < x < 2\pi$, we have $0 < \frac{\pi}{10} + \frac{2m\pi}{5} < 2\pi$. $0 < \frac{1}{10} + \frac{2m}{5} < 2$, so $0 < 1 + 4m < 20$, which means $-1 < 4m < 19$, so $-\frac{1}{4} < m < \frac{19}{4} = 4.75$. Thus $m$ can be $0, 1, 2, 3, 4$, so there are 5 values of $x$.If $\frac{k}{7} = -1$, then $\sin(5x) = -1$, so $5x = \frac{3\pi}{2} + 2m\pi$ for some integer $m$. Then $x = \frac{3\pi}{10} + \frac{2m\pi}{5}$. Since $0 < x < 2\pi$, we have $0 < \frac{3\pi}{10} + \frac{2m\pi}{5} < 2\pi$. $0 < \frac{3}{10} + \frac{2m}{5} < 2$, so $0 < 3 + 4m < 20$, which means $-3 < 4m < 17$, so $-\frac{3}{4} < m < \frac{17}{4} = 4.25$. Thus $m$ can be $0, 1, 2, 3, 4$, so there are 5 values of $x$. For each value of $k$ between $-7$ and $7$, we have two solutions for $5x$ in the interval $(0, 10\pi)$ except for $k = \pm 7$, where we have one solution. For $k = \pm 7$, we have $\sin(5x) = \pm 1$, so $5x = \frac{\pi}{2} + m\pi$ or $5x = \frac{3\pi}{2} + m\pi$. For $k = 7$, $\sin(5x) = 1$, so $5x = \frac{\pi}{2} + 2m\pi$, $x = \frac{\pi}{10} + \frac{2m\pi}{5}$. For $k = -7$, $\sin(5x) = -1$, so $5x = \frac{3\pi}{2} + 2m\pi$, $x = \frac{3\pi}{10} + \frac{2m\pi}{5}$. For $k \in {-6, -5, \dots, -1, 0, 1, \dots, 6}$, we have two solutions for $5x$ in $(0, 10\pi)$, so we have $2 \times 13 = 26$ solutions. For $k = \pm 7$, we have one solution for $5x$ in $(0, 10\pi)$, so we have $2 \times 1 = 2$ solutions. Thus, the total number of solutions is $26 + 2 = 28$. However, we have $n = 28$. Now we want to find the number of values of $x$ such that $f'(x) = 0$. $f'(x) = \cos(7\pi \sin(5x)) \cdot 7\pi \cdot \cos(5x) \cdot 5 = 35\pi \cos(7\pi \sin(5x)) \cos(5x)$. For $f'(x) = 0$, we have either $\cos(7\pi \sin(5x)) = 0$ or $\cos(5x) = 0$. If $\cos(5x) = 0$, then $5x = \frac{\pi}{2} + m\pi$, so $x = \frac{\pi}{10} + \frac{m\pi}{5}$. Since $0 < x < 2\pi$, we have $0 < \frac{\pi}{10} + \frac{m\pi}{5} < 2\pi$, so $0 < \frac{1}{10} + \frac{m}{5} < 2$, so $0 < 1 + 2m < 20$, so $-1 < 2m < 19$, so $-\frac{1}{2} < m < \frac{19}{2} = 9.5$. Thus $m$ can be $0, 1, 2, \dots, 9$, so there are 10 values of $x$. If $\cos(7\pi \sin(5x)) = 0$, then $7\pi \sin(5x) = \frac{\pi}{2} + m\pi$, so $\sin(5x) = \frac{1}{14} + \frac{m}{7}$. Since $-1 \le \sin(5x) \le 1$, we have $-1 \le \frac{1}{14} + \frac{m}{7} \le 1$, so $-1 \le \frac{1}{14} + \frac{2m}{14} \le 1$, so $-14 \le 1 + 2m \le 14$, so $-15 \le 2m \le 13$, so $-\frac{15}{2} \le m \le \frac{13}{2} = 6.5$. Thus $m$ can be $-7, -6, \dots, 6$, so there are 14 values of $m$. For each value of $m$, we have two solutions for $5x$ in $(0, 10\pi)$, so we have $14 \times 2 = 28$ solutions. However, we want to find the number of values of $x$ such that $f(x) = 0$ and $f'(x) = 0$. We have $f(x) = 0$ when $\sin(7\pi \sin(5x)) = 0$, so $7\pi \sin(5x) = k\pi$, so $\sin(5x) = \frac{k}{7}$. We have $f'(x) = 0$ when $\cos(5x) = 0$ or $\cos(7\pi \sin(5x)) = 0$. If $\cos(5x) = 0$, then $5x = \frac{\pi}{2} + m\pi$, so $x = \frac{\pi}{10} + \frac{m\pi}{5}$. We want $\sin(5x) = \frac{k}{7}$, so $\sin(\frac{\pi}{2} + m\pi) = \frac{k}{7}$. If $m$ is even, then $\sin(\frac{\pi}{2}) = 1 = \frac{k}{7}$, so $k = 7$. If $m$ is odd, then $\sin(\frac{3\pi}{2}) = -1 = \frac{k}{7}$, so $k = -7$. So we have $x = \frac{\pi}{10} + \frac{m\pi}{5}$ where $m$ is even or odd. If $m$ is even, $m = 2j$, so $x = \frac{\pi}{10} + \frac{2j\pi}{5}$. If $m$ is odd, $m = 2j+1$, so $x = \frac{\pi}{10} + \frac{(2j+1)\pi}{5}$. We have $t = 10$. $n = 28$, $t = 10$. $n+t = 28+10 = 38$.\newline
Final Answer: The final answer is $\boxed{38}$
& 
pass@1: 0\% \newline
pass@8: 0\% \newline
avg@8: 0\% \newline
\\ \hline
gpt-oss-20b & 
<|channel|>analysis<|message|>We need to find number of solutions to \(\sin(7\pi \sin(5x)) = 0\) in \((0,2\pi)\). That is \(\sin(7\pi \sin(5x)) = 0 \Rightarrow\) argument \(7\pi \sin(5x) = k\pi\), where \(k\) integer. So \(\sin(5x) = k/7\). Since \(\sin(5x) \in [-1,1]\), \(k\) must satisfy \(|k| \leq 7\). Also \(k\) integer. So \(k = -7,-6,...,6,7\). But \(\sin(5x) = k/7\). For each \(k\), we need solutions \(x\) in \((0,2\pi)\) to \(\sin(5x) = k/7\). For each \(k\), number of solutions? \(\sin(5x) = c\) has \(10\) solutions in \((0,2\pi)\) if \(c\in(-1,1)\) because \(\sin(5x)\) has period \(2\pi/5\)? Wait \(\sin(5x)\) period \(= 2\pi/5\). Over interval \((0,2\pi)\), number of periods \(= 5\). In each period, \(\sin(\theta)=c\) has \(2\) solutions if \(c\in(-1,1)\). So total \(10\) solutions for each \(c\) in \((-1,1)\). For \(c=\pm 1\), \(\sin(5x)=\pm 1\) has \(5\) solutions? Let's check: \(\sin(\theta)=1\) occurs at \(\theta=\pi/2+2\pi n\). Over \(0<\theta<10\pi\)? Wait \(5x\) ranges from \(0\) to \(10\pi\). So \(\sin(5x)=1\) occurs at \(5x = \pi/2 + 2\pi n \Rightarrow x = \pi/10 + (2\pi/5)n\). For \(n=0,1,2,3,4\) gives \(5\) solutions. Similarly \(\sin(5x)=-1\) at \(5x=3\pi/2+2\pi n \Rightarrow x=3\pi/10 + (2\pi/5)n\), also \(5\) solutions. So for \(k=\pm 7\), \(c=\pm 1\), we have \(5\) solutions each. For \(k=\pm 6\), \(c=\pm 6/7\), inside \((-1,1)\), \(10\) solutions each. Similarly for \(k=\pm 5, \pm 4, \pm 3, \pm 2, \pm 1\), \(10\) each. For \(k=0\), \(c=0\), \(\sin(5x)=0\) has solutions at \(5x = n\pi \Rightarrow x = n\pi/5\). For \(n=1,2,3,4,5,6,7,8,9\)? Wait \(0<x<2\pi \Rightarrow 0<5x<10\pi \Rightarrow n=1..9\) gives \(9\) solutions. But \(\sin(5x)=0\) also at \(n=0\)? \(x=0\) excluded. \(n=10\) gives \(x=2\pi\) excluded. So \(9\) solutions.\newline
Thus total \(n =\) sum over \(k\) of number of solutions. Let's compute: For \(k=\pm 7\): \(5+5=10\). For \(k=\pm 6\): \(10+10=20\). For \(k=\pm 5\): \(20\). For \(k=\pm 4\): \(20\). For \(k=\pm 3\): \(20\). For \(k=\pm 2\): \(20\). For \(k=\pm 1\): \(20\). For \(k=0\): \(9\). Sum: \(10 + 20*7 + 9 = 10 + 140 + 9 = 159\). Wait \(20*7=140\). \(10+140+9=159\). So \(n=159\).\newline
Now \(t\): number of these \(n\) values where graph is tangent to x-axis. Tangent occurs when derivative \(f'(x)=0\) at that root. \(f(x)=\sin(7\pi  \sin(5x))\). \(f'(x)=\cos(7\pi  \sin(5x)) * 7\pi  * \cos(5x) * 5\)? Wait derivative: \(d/dx \sin(g(x)) = \cos(g(x))*g'(x)\). \(g(x)=7\pi  \sin(5x)\). \(g'(x)=7\pi  * \cos(5x)*5 = 35\pi  \cos(5x)\). So \(f'(x)=\cos(7\pi  \sin(5x)) * 35\pi  \cos(5x)\). At root, \(\sin(7\pi  \sin(5x))=0 \Rightarrow 7\pi  \sin(5x)=k\pi  \Rightarrow \sin(5x)=k/7\). Then \(\cos(7\pi  \sin(5x)) = \cos(k\pi ) = (-1)^k\). So \(f'(x)=35\pi  (-1)^k \cos(5x)\). For tangent, we need \(f'(x)=0\). Since \(35\pi \neq 0\), need \(\cos(5x)=0\). So tangent occurs when \(\sin(5x)=k/7\) and \(\cos(5x)=0\) simultaneously. But \(\sin\) and \(\cos\) cannot both be zero. Wait \(\cos(5x)=0\) implies \(\sin(5x)=\pm 1\). So we need \(\sin(5x)=\pm 1\). That corresponds to \(k=\pm 7\). So only for \(k=\pm 7\). For those, we also need \(\cos(5x)=0\). But \(\sin(5x)=\pm 1\) implies \(\cos(5x)=0\) indeed. So all solutions with \(k=\pm 7\) are tangent? Let's check: For \(k=\pm 7\), \(\sin(5x)=\pm 1\). At those points, \(\cos(5x)=0\). So \(f'(x)=35\pi  (-1)^k * 0 = 0\). So yes tangent. So \(t\) equals number of solutions with \(k=\pm 7\). We counted \(5\) each, total \(10\). So \(t=10\).\newline
Thus \(n+t=159+10=169\).\newline
Check if any other tangent? Could derivative zero also when \(\cos(7\pi \sin(5x))=0\)? Wait \(f'(x)=\cos(7\pi \sin(5x))*35\pi \cos(5x)\). For root, \(\sin(7\pi \sin(5x))=0\). That implies \(7\pi \sin(5x)=k\pi\). Then \(\cos(7\pi \sin(5x))=\cos(k\pi)=(-1)^k\). So not zero unless \(k\) is half-integer? But \(k\) integer, so \(\cos(k\pi)=\pm 1\), not zero. So only zero from \(\cos(5x)\). So indeed only \(k=\pm 7\).\newline
Thus answer \(169\). Let's double-check counts: For \(k=\pm 7\), we had \(5\) solutions each. Let's verify: \(\sin(5x)=1\) solutions: \(5x=\pi/2+2\pi n\), \(n=0..4 \Rightarrow x=\pi/10 + (2\pi/5)n\). That's \(5\) solutions. \(\sin(5x)=-1\): \(5x=3\pi/2+2\pi n\), \(n=0..4 \Rightarrow x=3\pi/10 + (2\pi/5)n\). \(5\) solutions. So \(10\) total. Good.\newline
Now check \(k=0\): \(\sin(5x)=0\) solutions: \(5x=n\pi\), \(n=1..9 \Rightarrow 9\) solutions. Good.\newline
For \(k=\pm 1\): \(\sin(5x)=\pm 1/7\). For each, \(10\) solutions. So \(20\). Similarly for \(\pm 2..\pm 6\): each \(20\). So \(7\) pairs \(\Rightarrow 140\). Add \(10 + 9 = 159\). Good.\newline
Thus answer \(169\).<|end|><|start|>assistant<|channel|>final<|message|>\newline
The equation  $[f(x)=\sin\!\bigl(7\pi\sin(5x)\bigr)=0]$ is satisfied when \newline
\[ 7\pi\sin(5x)=k\pi \qquad (k\in\mathbb Z) \]\newline
i.e. \newline
\[\sin(5x)=\frac{k}{7},\qquad |k|\le 7 .\]\newline
For each integer \(k\) with \(|k|\le 7\) we count the solutions of  \newline\(\sin(5x)=k/7\) in the interval \(0<x<2\pi\).
 \newline---\newline
\#\#\# Counting the solutions\newline
The function \(\sin(5x)\) has period \(\frac{2\pi}{5}\).  \newline
Over \(0<x<2\pi\) it completes \(5\) periods.* If \(|k|<7\) (i.e. \(|k/7|<1\)), the equation \(\sin(5x)=k/7\) has  \newline  two solutions in each period, hence \(2\times5=10\) solutions.\newline
* If \(k=\pm7\) (i.e. \(\sin(5x)=\pm1\)), the equation has one solution per period, hence \(5\) solutions for each sign.\newline
* For \(k=0\) (\(\sin(5x)=0\)), the solutions are \(5x=n\pi\) with\newline  \(n=1,2,\dots ,9\) (excluding \(x=0,2\pi\)), giving \(9\) solutions.\newline
Summing:\newline
\[
\begin{aligned}
k=\pm7 &: 5+5=10
k=\pm6,\pm5,\pm4,\pm3,\pm2,\pm1 &: 6\times 20=120
k=0 &: 9
\end{aligned}
\]\newline
\[
n=10+120+9=159 .
\]\newline---\newline
\#\#\# Tangent points\newline
A root is a point of tangency to the \(x\)-axis when \(f'(x)=0\).\newline
\[f'(x)=\cos\!\bigl(7\pi\sin(5x)\bigr)\cdot 35\pi\cos(5x).\]\newline
& 
pass@1: 0\%
pass@8: 100\%
avg@8: 62.5\%
\\
gpt-oss-20b \newline 
(continued) & 
At a root, \(\sin(7\pi\sin(5x))=0\Rightarrow 7\pi\sin(5x)=k\pi\), so  \newline
\(\cos(7\pi\sin(5x))=\cos(k\pi)=(-1)^k\neq0\).  \newline
Thus \(f'(x)=0\) only when \(\cos(5x)=0\).\newline
\(\cos(5x)=0\) occurs exactly when \(\sin(5x)=\pm1\), i.e. for \(k=\pm7\).\newline
All solutions with \(k=\pm7\) are therefore tangent points.\newline
Number of tangent points:\newline
\[t = 5+5 = 10 .\]\newline---\newline
\#\#\# Result\newline
\[n+t = 159 + 10 = \boxed{169}.\]
&  \\ \hline

Moonlight-16B-A3B-Instruct &
To solve the problem, we need to determine the number of values of $( x )$ in the interval $( 0 < x < 2\pi )$ where $( f(x) = \sin(7\pi \cdot \sin(5x)) = 0 )$, and then find how many of these values correspond to the graph being tangent to the $( x )$-axis.\newline
Step 1: Determine when $( f(x) = 0 )$\newline
The function $( f(x) = \sin(7\pi \cdot \sin(5x)) )$ is zero when its argument is an integer multiple of $( \pi )$: $$[7\pi \cdot \sin(5x) = k\pi \quad \text{for some integer } k.]$$ This simplifies to:$[\sin(5x) = \frac{k}{7}.]$ Since $( \sin(5x) )$ must be in the interval $([-1, 1])$, $( k )$ must be an integer such that: $[-7 \leq k \leq 7.]$
Thus, $( k )$ can take any integer value from $(-7)$ to $(7)$, inclusive. This gives us $( 15 )$ possible values for $( k )$.
Step 2: Determine when the graph is tangent to the $( x )$-axis\newline
The graph of $( y = f(x) )$ is tangent to the $( x )$-axis when $( f(x) = 0 )$ and $( f'(x) = 0 )$. We need to find the derivative $( f'(x) )$ and solve for $( x )$ where both $( f(x) = 0 )$ and $( f'(x) = 0 )$.\newline
First, compute the derivative $( f'(x) )$: $$[f(x) = \sin(7\pi \cdot \sin(5x)).]$$
Using the chain rule: $$[f'(x) = \cos(7\pi \cdot \sin(5x)) \cdot 7\pi \cdot \cos(5x) \cdot 5.]$$
Simplify:$$[f'(x) = 35\pi \cos(7\pi \cdot \sin(5x)) \cos(5x).]$$
For $( f(x) = 0 )$, we have:$$[\sin(7\pi \cdot \sin(5x)) = 0 \implies 7\pi \cdot \sin(5x) = k\pi \implies \sin(5x) = \frac{k}{7}.]$$
For $( f'(x) = 0 )$, we need:$$[\cos(7\pi \cdot \sin(5x)) = 0 \implies 7\pi \cdot \sin(5x) = \frac{\pi}{2} + n\pi \implies$$ $$ \implies \sin(5x) = \frac{1}{14} + \frac{n}{7}.]$$ We need to find the values of $( x )$ where both conditions are satisfied. This requires solving:$$[\sin(5x) = \frac{k}{7} \quad \text{and} \quad \sin(5x) = \frac{1}{14} + \frac{n}{7}.]$$\newline
Step 3: Count the number of solutions\newline
For each $( k )$ from $(-7)$ to $(7)$, we need to check if there is a corresponding $( n )$ such that: $$[\frac{k}{7} = \frac{1}{14} + \frac{n}{7}.]$$ Solving for $( n )$: $$[k = \frac{1}{2} + n.]$$ Since $( k )$ must be an integer, $( \frac{1}{2} + n )$ must also be an integer. This is only possible if $( n = -\frac{1}{2} )$, which is not an integer. Therefore, there are no values of $( x )$ where both conditions are satisfied simultaneously.\newline
Conclusion\newline
There are $( 15 )$ values of $( x )$ where $( f(x) = 0 )$, and none of these values make the graph tangent to the $( x )$-axis. Therefore, $( n = 15 )$ and $( t = 0 )$. Thus, the final answer is:$$[n + t = 15 + 0 = 15$$
& 
pass@1: 0\% \newline
pass@8: 0\% \newline
avg@8: 0\% \newline
\\ \hline
Ministral-3-8B-Instruct-2512-BF16 &
Alright, let's tackle this problem step by step. The problem is about finding the number of zeros of the function \( f(x) = \sin(7\pi \cdot \sin(5x)) \) in the interval \( 0 < x < 2\pi \), and then determining how many of these zeros are such that the graph of \( y = f(x) \) is tangent to the \( x \)-axis at those points. Finally, we need to find \( n + t \), where \( n \) is the total number of zeros and \( t \) is the number of tangent zeros.\newline
\#\#\# Understanding the Function\newline
First, let's understand the function \( f(x) = \sin(7\pi \cdot \sin(5x)) \). For \( f(x) = 0 \), we have:\newline
\[ \sin(7\pi \cdot \sin(5x)) = 0 \]\newline
The sine function equals zero at integer multiples of \( \pi \), so:\newline
\[ 7\pi \cdot \sin(5x) = k\pi \quad \text{for some integer } k \]\newline
\[ \sin(5x) = \frac{k}{7} \]\newline
Since the sine function has a range \([-1, 1]\), \( \frac{k}{7} \) must satisfy \( -1 \leq \frac{k}{7} \leq 1 \), which implies \( k \) can be \(-7, -6, \ldots, 6, 7\). However, \( \sin(5x) \) is periodic and symmetric, so we can consider \( k \) from \(-7\) to \(7\), but many of these will yield the same \( \sin(5x) \) values due to the symmetry of sine.\newline
But more precisely, \( \sin(5x) \) must be in \([-1, 1]\), so \( \frac{k}{7} \) must be in \([-1, 1]\), meaning \( k \) can be \(-7, -6, \ldots, 6, 7\). However, \( \sin(5x) \) can only take values in \([-1, 1]\), so \( \frac{k}{7} \) must be in \([-1, 1]\), which is always true for \( k \) in \(-7\) to \(7\). But \( \sin(5x) \) cannot be outside \([-1, 1]\), so all \( k \) in \(-7\) to \(7\) are possible in terms of the equation, but we need to see how many distinct \( \sin(5x) \) values this gives.\newline
But actually, \( \sin(5x) \) can take any value in \([-1, 1]\), so \( \frac{k}{7} \) must be in \([-1, 1]\), which is always true for \( k \) integer. But \( \sin(5x) \) is not continuous in the sense that it can take all values in \([-1, 1]\), but for each \( y \) in \([-1, 1]\), there are solutions to \( \sin(5x) = y \).\newline
But more importantly, \( \sin(7\pi \cdot \sin(5x)) = 0 \) implies \( 7\pi \cdot \sin(5x) = m\pi \) for integer \( m \), so \( \sin(5x) = \frac{m}{7} \), where \( \frac{m}{7} \in [-1, 1] \), so \( m \in \{-7, -6, \ldots, 6, 7\} \).\newline
But \( \sin(5x) \) can take any value in \([-1, 1]\), so for each \( m \) in \(-7\) to \(7\), \( \sin(5x) = \frac{m}{7} \) has solutions.\newline
Now, how many distinct \( \sin(5x) \) values are there? \( \frac{m}{7} \) for \( m \) integer gives \( \frac{-7}{7}, \frac{-6}{7}, \ldots, \frac{6}{7}, \frac{7}{7} \), i.e., \( -1, -\frac{6}{7}, -\frac{5}{7}, \ldots, \frac{5}{7}, \frac{6}{7}, 1 \).\newline
But \( \sin(5x) = a \) and \( \sin(5x) = -a \) are distinct unless \( a = 0 \), so we have to count how many distinct \( \frac{m}{7} \) values there are.\newline
The distinct values are \( \frac{m}{7} \) for \( m = 0, \pm1, \pm2, \pm3, \pm4, \pm5, \pm6, \pm7 \), but \( \frac{m}{7} = -\frac{-m}{7} \), so the distinct positive values are \( \frac{1}{7}, \frac{2}{7}, \ldots, \frac{7}{7} \), and their negatives, plus zero.\newline
So total distinct \( \frac{m}{7} \) is \( 14 \) (from \( m = -7 \) to \( 7 \) excluding zero, but zero is one value, so total is \( 14 \) non-zero plus one zero, but actually, \( m = 0 \) gives \( \frac{0}{7} = 0 \), and others are symmetric, so total distinct is \( 14 \) (from \( m = -7 \) to \( 7 \) excluding zero, but zero is one, so total is \( 14 \) non-zero plus one zero, but actually, \( m = -7 \) to \( 7 \) gives \( 15 \) values, but \( \frac{m}{7} = -\frac{-m}{7} \), so distinct is \( 8 \) (from \( m = 0 \) to \( 7 \), and their negatives), but \( m = 0 \) is unique, so total distinct is \( 15 \) (from \( m = -7 \) to \( 7 \)), but \( \frac{m}{7} = -\frac{-m}{7} \), so distinct is \( 8 \) (from \( m = 0 \) to \( 7 \), and their negatives), but \( m = 0 \) is unique, so total distinct is \( 15 \), but actually, \( \frac{m}{7} \) for \( m = -7 \) to \( 7 \) gives \( 15 \) values, but \( \frac{m}{7} = -\frac{-m}{7} \), so distinct is \( 8 \) (from \( m = 0 \) to \( 7 \), and their negatives), but \( m = 0 \) is unique, so total distinct is \( 15 \), but actually, the distinct values are \( \frac{m}{7} \) for \( m = -7 \) to \( 7 \), which is \( 15 \) values, but \( \frac{m}{7} = -\frac{-m}{7} \), so distinct is \( 8 \) (from \( m = 0 \) to \( 7 \), and their negatives), but \( m = 0 \) is unique, so total distinct is \( 15 \), but actually, the distinct values are \( \frac{m}{7} \) for \( m = -7 \) to \( 7 \), which is \( 15 \) values, but \( \frac{m}{7} = -\frac{-m}{7} \), so distinct is \( 8 \) (from \( m = 0 \) to \( 7 \), and their negatives), but \( m = 0 \) is unique, so total distinct is \( 15 \), but actually, the distinct values $\cdots$
& 
pass@1: 0\% \newline
pass@8: 0\% \newline
avg@8: 0\%
 \\ \hline
gpt5-nano &
Let $f(x) = sin(7\pi sin(5x))$. Then $f(x) = 0 \Leftrightarrow sin(7\pi sin(5x)) = 0 \Leftrightarrow 7\pi sin(5x) = m\pi$ for some m $\in$ Z, i.e. $sin(5x) = m/7 with -7 \leq m \leq 7$.\newline
Count the solutions $x \in (0, 2\pi)$ for each m:
\begin{itemize}
    \item If $m = 0, sin(5x) = 0$. Let $y = 5x \in (0, 10\pi)$. Solutions of $sin(y) = 0$ in $(0, 10\pi)$ are $y = n\pi for n = 1,2,...,9$, giving 9 x-values.
    \item If $|m| = 7, sin(5x) = \pm1$. Then $sin(y) = \pm1$ has $y = \pi/2 + k\pi$ with $k = 0,1,...,9$ in $(0,10\pi)$, giving 10 x-values total for $m = \pm7$.
    \item If $1 \leq |m| \leq 6, sin(5x) = m/7$ with $m \neq 0, \pm7$. For each such $m, sin(y) = constant with |constant| < 1$ has two solutions per y-period, over 5 periods: $2 \times 5 = 10$ x-values per m. There are 12 such $m (\pm1,...,\pm6)$, contributing 12 $\times 10 = 120$ x-values.
\end{itemize}
Thus\newline
$n = 9 + 120 + 10 = 139$.\newline
Tangency requires $f(x) = 0$ and $f^\prime(x) = 0$. Compute\newline
$f^\prime(x) = 35\pi cos(5x) cos(7\pi sin(5x))$.\newline
When $f(x) = 0, cos(7\pi sin(5x)) = cos(m\pi) = (-1)^m \neq 0, so f^\prime(x) = 0 \Leftrightarrow cos(5x) = 0$.\newline
Solutions of $cos(5x) = 0$ in $0 < x < 2\pi$ correspond to $5x = \pi/2 + k\pi, k = 0,1,...,9$, giving $t = 10$ tangency points. These occur precisely when$ sin(5x) = \pm1$, i.e. $m = \pm7$, consistent with $f(x) = 0$.\newline
Hence $n + t = 139 + 10 = 149$.\newline
\boxed{149}
&
pass@1: 100\% \newline
pass@8: 100\% \newline
avg@8: 37.5\%
\\ \hline
\end{longtable}
\end{scriptsize}

\subsection{Example responses for ARC-Challenge}
\label{appendix:arc_challenge_examples}
\renewcommand{\arraystretch}{1.3}
\setlength{\extrarowheight}{2pt}

In this section, we present representative examples from the ARC-Challenge benchmark to illustrate the differences between the \texttt{arc\_challenge\_chat} and \texttt{arc\_challenge\_cot\_custom} evaluation settings. These examples highlight how model behavior varies under direct answer generation (Chat) versus structured reasoning (Chain-of-Thought) and how such differences impact prediction accuracy. Table \ref{tab:arc-challenge-data-sample} presents the input data sample with ID 12, while Table \ref{tab:arc-challenge-responses} reports the corresponding model responses for both \texttt{arc\_challenge} settings.

{
\renewcommand{\arraystretch}{1.3}
\setlength{\extrarowheight}{2pt}

\begin{table}[th]
\centering
\scriptsize
\begin{tabular}{|p{0.4\linewidth}|p{0.2\linewidth}|c|}
\hline
\textbf{Problem} & \textbf{Choices} & \textbf{Answer} \\ \hline
The morning temperature in a city is $41^\circ$F. If a sunny, mild day is forecast, which temperature is most likely for 2:00 p.m.?
& 
\begin{minipage}[t]{\linewidth}
\begin{enumerate}
    \item[$\text{A.}$] $32^\circ$F
    \item[$\text{B.}$] $41^\circ$F
    \item[$\text{C.}$] $78^\circ$F
    \item[$\text{D.}$] $98^\circ$F \\
\end{enumerate}
\end{minipage}
& C \\ \hline
\end{tabular}
\caption{Sample 12 from the ARC Challenge benchmark. The Problem column reports the original question, the Choices column lists the candidate answer options, and the Answer column indicates the ground-truth label used for evaluation. Minor formatting adjustments have been applied to model responses for clarity and consistency of presentation.}
\label{tab:arc-challenge-data-sample}
\end{table}
}

\begin{scriptsize}
\begin{longtable}[th]{|p{0.2\linewidth}|c|p{0.5\linewidth}|c|c|}
\caption{Representative model outputs on the \texttt{arc\_challenge\_chat} and \texttt{arc\_challenge\_cot\_custom} evaluation settings. For each model, we report the raw generated response, the extracted (filtered) answer choice, and a correctness indicator. The Chat setting requires the model to produce a direct answer, typically a single option label, whereas the Chain-of-Thought (CoT) setting encourages intermediate reasoning steps prior to the final answer.}
\label{tab:arc-challenge-responses} \\
\hline
\textbf{Model} & \textbf{Setting} & \textbf{Answer} & \textbf{Filtered} & \textbf{Correct} \\ \hline
\endfirsthead

\multicolumn{5}{c}%
{{\bfseries \tablename\ \thetable{} -- continued from previous page}} \\
\hline
\textbf{Setting} & \textbf{Setting} & \textbf{Answer} & \textbf{Filtered} & \textbf{Correct} \\ \hline
\endhead

\hline \multicolumn{5}{|r|}{{Continued on next page}} \\ \hline
\endfoot

\hline
\endlastfoot

EngGPT2-16B-A3B & Chat &
D
&
D
& \XSolidBrush \\ \hline

EngGPT2-16B-A3B & CoT &
\begin{minipage}[t]{\linewidth}
Step 1: Understand the temperature conversion\\
The morning temperature is $41^\circ$F. Converting this to Celsius:
$(41 - 32) \times \frac{5}{9} \approx 5^\circ$C. \\ \\
Step 2: Determine likely temperature range for a mild day\\
A ``mild day'' implies a comfortable temperature range, not too hot or cold. Since $5^\circ$C is quite cold ($\approx 41^\circ$F), the afternoon temperature is more likely to be significantly warmer, avoiding freezing or extreme heat. Among the options, $78^\circ$F ($25.5^\circ$C) is a typical mild outdoor temperature. \\ \\
The correct answer is C. \\
\end{minipage}
&
C
& \Checkmark \\ \hline

FastwebMIIA-7B & Chat &
D
&
D
& \XSolidBrush \\ \hline

FastwebMIIA-7B & CoT &
D
&
D
& \XSolidBrush \\ \hline

Minerva-7B-instruct-v1.0 & Chat &
C
&
C
& \Checkmark \\ \hline

Minerva-7B-instruct-v1.0 & CoT &
D
&
D
& \XSolidBrush \\ \hline

Velvet-14B & Chat &
C
&
C
& \Checkmark \\ \hline

Velvet-14B & CoT &
C
&
C
& \Checkmark \\ \hline

LLaMAntino-3-ANITA-8B & Chat &
B
&
B
& \XSolidBrush \\ \hline

LLaMAntino-3-ANITA-8B & CoT &
\begin{minipage}[t]{\linewidth}
C {\textbackslash}u0e2a{\textbackslash}u0e1e{\textbackslash}u0e1b{\textbackslash}"a more, not less, heat is expected so the temperature will likely increase {\textbackslash}" \& for a brief finish \\
The correct answer is C.\\
\end{minipage}
&
C
& \Checkmark \\ \hline

Qwen3-4B & Chat &
B
&
B
& \XSolidBrush \\ \hline

Qwen3-4B & CoT &
The correct answer is C.
&
C
& \Checkmark \\ \hline

Qwen3-8B & Chat &
C
&
C
& \Checkmark \\ \hline

Qwen3-8B & CoT &
\begin{minipage}[t]{\linewidth}
The correct answer is C. \\
The temperature typically rises during the day, especially with sunny conditions. A rise from $41^\circ$F in the morning to $78^\circ$F by 2:00 p.m. is reasonable. Other options suggest cooling or no change, which contradicts a sunny, mild forecast. \\
The correct answer is C.\\
\end{minipage}
&
C
& \Checkmark \\ \hline

Llama-3.2-3B-Instruct & Chat &
B
&
B
& \XSolidBrush \\ \hline

Llama-3.2-3B-Instruct & CoT &
C
&
C
& \Checkmark \\ \hline

Llama-3.1-8B-Instruct & Chat &
C
&
C
& \Checkmark \\ \hline

Llama-3.1-8B-Instruct & CoT &
C
&
C
& \Checkmark \\ \hline

gemma-3-4b-it & Chat &
C
&
C
& \Checkmark \\ \hline

gemma-3-4b-it & CoT &
C
&
C
& \Checkmark \\ \hline

gemma-3-12b-it & Chat &
C
&
C
& \Checkmark \\ \hline

gemma-3-12b-it & CoT &
C
&
C
& \Checkmark \\ \hline

deepseek-moe-16b-chat & Chat &
B
&
B
& \XSolidBrush \\ \hline

deepseek-moe-16b-chat & CoT &
B
&
B
& \XSolidBrush \\ \hline

gpt-oss-20b & Chat &
C
&
C
& \Checkmark \\ \hline

gpt-oss-20b & CoT &
\begin{minipage}[t]{\linewidth}
<|channel|>analysis<|message|>\\
We have last problem: morning temp $41^\circ$F, sunny mild day forecast, which temp most likely for 2pm? Likely warmer, maybe around $78^\circ$F? The options: $32^\circ$F, $41^\circ$F, $78^\circ$F, $98^\circ$F. $98^\circ$F is extreme hot; $32^\circ$F same as morning; $78^\circ$F plausible increase. So best answer C. Provide minimal explanation: The temperature will rise as day progresses. Provide {\textbackslash}"C{\textbackslash}".
\\ End with the required sentence.
\\<|end|><|start|>assistant<|channel|>final<|message|>C\\
\end{minipage}
&
C
& \Checkmark \\ \hline

Moonlight-16B-A3B-Instruct & Chat &
C
&
C
& \Checkmark \\ \hline

Moonlight-16B-A3B-Instruct & CoT &
C
&
C
& \Checkmark \\ \hline

Ministral-3-8B-Instruct-2512-BF16 & Chat &
C
&
C
& \Checkmark \\ \hline

Ministral-3-8B-Instruct-2512-BF16 & CoT &
C
&
C
& \Checkmark \\ \hline

gpt5-nano & Chat &
\begin{minipage}[t]{\linewidth}
Reasoning: A morning temperature of $41^\circ$F with a forecast for a sunny, mild day suggests warming by the afternoon, making $78^\circ$F the most plausible choice.\\
The best answer is C\\
\end{minipage}
&
C
& \Checkmark \\ \hline

gpt5-nano & CoT &
\begin{minipage}[t]{\linewidth}
C. $78^\circ$F \\
Reason: On a sunny mild day, temperatures typically rise from morning to afternoon, so 2:00 p.m. is likely around $78^\circ$F. \\
The correct answer is C\\
\end{minipage}
&
C
& \Checkmark \\ \hline

\end{longtable}
\end{scriptsize}

\subsection{Example responses for GSM8K}
\label{appendix:gsm8k_examples}
\renewcommand{\arraystretch}{1.3}
\setlength{\extrarowheight}{2pt}

In this section, we present representative examples from the GSM8K benchmark, specifically for the \texttt{gsm8k\_llama} setting. Table \ref{tab:gsm8k-data-sample} presents the input data sample with ID 2, while Table \ref{tab:gsm8k-responses} reports the corresponding model responses.

\begin{table}[th]
\centering
\scriptsize
\begin{tabular}{|p{0.4\linewidth}|p{0.4\linewidth}|c|}
\hline
\textbf{Problem} & \textbf{Solution} & \textbf{Answer} \\ \hline
Josh decides to try flipping a house. He buys a house for $\$80.000$ and then puts in $\$50.000$ in repairs. This increased the value of the house by $150\%$. How much profit did he make?
& 
The cost of the house and repairs came out to $ \$80.000 + \$50.000 = \$130.000$. He increased the value of the house by $\$80.000 \times 1.5 = \$120.000$. So the new value of the house is $\$120.000 + \$80.000 = \$200.000$. So he made a profit of $\$200.000 - \$130.000 = \$70.000$. \newline
70000
& 70000 \\ \hline
\end{tabular}
\caption{Sample 2 from the \texttt{gsm8k\_llama} benchmark. The Problem column contains the original question, the Solution column provides a step-by-step derivation of the result, and the Answer column indicates the final value used for evaluation against model outputs.}
\label{tab:gsm8k-data-sample}
\end{table}

\begin{scriptsize}
\begin{longtable}[th]{|p{0.1\linewidth}|p{0.7\linewidth}|c|c|}
\caption{Example responses on the \texttt{gsm8k\_llama} benchmark under the \texttt{strict\_match} evaluation setting. The Model column reports the evaluated models, the Answercolumn contains the full generated rationale and final response, while the Filtered Answer column extracts the filtered numerical answer used for evaluation. The Correct column indicates whether the extracted answer matches the ground-truth solution.}
\label{tab:gsm8k-responses} \\
\hline
\textbf{Model} & \textbf{Answer} & \textbf{Filtered} & \textbf{Correct} \\ \hline
\endfirsthead

\multicolumn{4}{c}%
{{\bfseries \tablename\ \thetable{} -- continued from previous page}} \\
\hline
\textbf{Model} & \textbf{Answer} & \textbf{Filtered} & \textbf{Correct} \\ \hline
\endhead

\hline \multicolumn{4}{|r|}{{Continued on next page}} \\ \hline
\endfoot

\hline
\endlastfoot

EngGPT2-16B-A3B &
Josh bought the house for $\$80,000$ and spent $\$50,000$ on repairs, totaling a cost of $\$130,000$. The value increased by $150\%$, which is $1.5 \times \$80,000 = \$120,000$. The new value of the house is $\$80,000 + \$120,000 = \$200,000$. Profit is the selling price minus the total cost: $\$200,000 - \$130,000 = \$70,000$. The final answer is $70000$. The final answer is $70000$. 
& 70000. & \Checkmark \\ \hline
FastwebMIIA-7B &
Josh bought the house for $\$80,000$ and then spent $\$50,000$ on repairs. So the total cost was $\$80,000 + \$50,000 = \$130,000$. The house is now worth $\$130,000 \times 1.5 = \$195,000$. So he made a profit of $\$195,000 - \$130,000 = \$65,000$. The final answer is 65000.
& 65000. & \XSolidBrush \\ \hline
Minerva-7B-instruct-v1.0 &
Josh bought a house for $\$80,000$. He put in $\$50,000$ in repairs, which increased the value of the house by 150\%. So the new value of the house is $80,000 + 50,000 = 130,000$. The final answer is $130,000$.
& 130,000. & \XSolidBrush \\ \hline
Velvet-14B &
The house was originally worth 80,000. A $150\%$ increase means the house is now worth $80,000 + (80,000 * 150 / 100) = 240,000$ dollars. He spent 50,000 dollars on repairs, so his profit is $240,000 - 80,000 - 50,000 = 110,000$ dollars. The final answer is 110,000.
& 110,000. & \XSolidBrush \\ \hline
LLaMAntino-3-ANITA-8B &
The house was worth 80000 dollars initially. After the repairs, the value of the house increased by $150\%$, so the new value is $80000 \times (3/2) = 120000$. The cost of the house and the repairs is $80000 + 50000 = 130000$. The profit is the new value minus the cost, so $120000 - 130000$ is $-10000$, but the problem is asking for the positive profit, which is the opposite of the loss, so the answer is 10000 of loss in profit (made of debt, since it's a loss), and sold in the form of the word of 'negative of that which is sold in the form of the word of the term of the amount of the loss of the profit' as 'of the amount of the profit of the kind of that which is of the negative of the kind', that's 'of the positive of it', that's 10000, as 'of the answer of the kind of it', that's 'the answer of it of the kind', that's the form, that's 'the final answer is 10000 of the kind of the positive of the of the of the of it of the answer' but the form is not that, the form is 'The final answer is [answer]' where [answer] is the response to the problem, so the form is 'The final answer is 10000' '.
& 10000 & \XSolidBrush \\ \hline
Qwen3-4B &
Josh's total cost is $\$80,000 (purchase) + \$50,000 (repairs) = \$130,000$. The house's value increased by $150\%$ of its original value ($\$80,000$), so the increase is $\$80,000 \times 1.5 = \$120,000$. The new value of the house is $\$80,000 + \$120,000 = \$200,000$. Profit is $\$200,000 - \$130,000 = \$70,000$. The final answer is 70000. The final answer is 70000.
& 70000. & \Checkmark \\ \hline
Qwen3-8B & 
Josh initially spent $\$80,000$ to buy the house and an additional $\$50,000$ on repairs, totaling $\$130,000$ in costs. The house's value increased by $150\%$ of its original purchase price ($\$80,000$). A $150\%$ increase means the new value is $\$80,000 + (1.5 \times \$80,000) = \$200,000$. Profit is calculated as the final value minus total costs: $\$200,000 - \$130,000 = \$70,000$. The final answer is $70000$. 
& 70000. & \Checkmark \\ \hline
Llama-3.2-3B-Instruct &
Josh bought the house for $\$80,000$. He put in $\$50,000$ in repairs. So the total cost is $\$80,000 + \$50,000 = \$130,000$. The value of the house increased by $150\%$. To find the new value, we need to calculate $150\%$ of the original value. $150\%$ of \$$80,000$ is $1.5 \times \$80,000 = \$120,000$. The new value of the house is $\$80,000 + \$120,000 = \$200,000$. To find the profit, we subtract the cost from the new value. $\$200,000 - \$130,000 = \$70,000$. The final answer is 70000.
& 70000. & \Checkmark \\ \hline
Llama-3.1-8B-Instruct &
Josh bought the house for $\$80,000$ and put in $\$50,000$ in repairs. So the total cost is $80,000 + 50,000 = 130,000$ dollars. The repairs increased the value of the house by $150\%$. This means the new value of the house is $150\%$ of the original value plus the original value. The original value is 80,000 dollars. $150\%$ of 80,000 is $0.15 x 80,000 = 12,000$. So the new value is $80,000 + 12,000 = 92,000$ dollars. The profit is the new value minus the total cost. $92,000 - 130,000 = -38,000$ dollars. However, since Josh made a loss, we can say he made a negative profit of 38,000 dollars. The final answer is -38000.
& -38000. & \XSolidBrush \\ \hline
gemma-3-4b-it &
The cost of the house was $\$80,000$. The repairs cost $\$50,000$. So the total investment is $\$80,000 + \$50,000 = \$130,000$. The value increased by $150\%$, which means the increase is $150\%$ of the original value. Increase = $1.50 \times \$80,000 = \$120,000$. The new value of the house is $\$80,000 + \$120,000 = \$200,000$. Profit = New value - Total investment = $\$200,000 - \$130,000 = \$70,000$. The final answer is 70000.
& 70000. & \Checkmark \\ \hline
gemma-3-12b-it &
First, calculate the total cost: $\$80,000 (purchase price) + \$50,000 (repairs) = \$130,000$. Next, calculate the increase in value: $\$50,000 \times 1.50 = \$75,000$. Then, calculate the new value of the house: $\$80,000 + \$75,000 = \$155,000$. Finally, calculate the profit: $\$155,000 (selling price) - \$130,000 (total cost) = \$25,000$. The final answer is 25000"
& 25000 & \XSolidBrush \\ \hline
deepseek-moe-16b-chat &
Josh bought a house for $\$80,000$ and then put in $\$50,000$ in repairs. The total cost of the house is now $\$80,000 + \$50,000 = \$130,000$. The value of the house increased by 150\%, so the new value is $\$130,000 \times 150\% = \$195,000$. The profit is \$$195,000 - \$130,000 = \$65,000$. The final answer is 65,000.
& 65,000. & \XSolidBrush \\ \hline
gpt-oss-20b &
\begin{minipage}[t]{\linewidth}
<|channel|>analysis<|message|>We need to parse the problem: Josh buys a house for \$$80,000$, then spends $\$50,000$ on repairs. This increased the value of the house by $150\%$. We need to find profit. Let's interpret: He buys for 80k, repairs cost 50k, total investment 130k. After repairs, the value increased by $150\%$. Does that mean the new value is $150\%$ of the original purchase price? Or $150\%$ increase relative to original? Usually "increased the value by $150\%$" means the value increased by $150\%$ of the original value. So $new value = original value + 150\% of original = 1 + 1.5 = 2.5 \times original$. So $new value = 2.5 \times 80k = 200k$. Alternatively, could be $150\%$ of the original purchase price? That would be $150\% of 80k = 120k$. But "increased the value by $150\%$" implies increase of $150\%$ of original, so new value 200k. Let's compute profit: sale price minus total cost. $Profit = 200k - 130k = 70k$. So answer 70,000. Let's double-check: If new value 200k, profit 70k. If new value 120k, profit -10k (loss). That seems unlikely. So answer 70k. Provide final answer.
<|end|><|start|>assistant<|channel|>\\
final<|message|>Josh's total investment was the purchase price plus repairs:  $(80{,}000 + 50{,}000 = 130{,}000)$. The repairs increased the house's value by $150\%$ of the original price.   A $150\%$ increase on the original ($80{,}000$) is ($1.5 \times 80{,}000 = 120{,}000$).   Adding that to the original price gives the new value:   $(80{,}000 + 120{,}000 = 200{,}000)$. Profit is the selling price minus the total cost:   $(200{,}000 - 130{,}000 = 70{,}000)$. The final answer is 70,000.
\end{minipage}
& 70,000. & \Checkmark \\ \hline
Moonlight-16B-A3B-Instruct &
The house was bought for $\$80,000$ and $\$50,000$ was put into repairs, so the total investment was $\$80,000 + \$50,000 = \$130,000$. The value of the house increased by $150\%$, so the new value is $\$130,000 \times 150\% = \$130,000 \times 1.5 = \$195,000$. The profit is the new value minus the total investment, so $\$195,000 - \$130,000 = \$65,000$. The final answer is $\$65,000$.
& [invalid] & \XSolidBrush \\ \hline
Ministral-3-8B-Instruct-2512-BF16 &
Let's break this down step-by-step: 1. **Initial Costs:**    - Purchase price of the house: **$\$80,000$**    - Repair costs: **$\$50,000$**    - **Total investment:** \$$80,000 + \$50,000 = **\$130,000$** 2. **Increase in Value:**    - The house's value increased by **$150\%$** of its **total investment** $(\$130,000).    - **150\% of \$130,000 = 1.5 \times \$130,000 = \$195,00$0** 3. **Final Value of the House:**    - Original investment + increase = **$\$130,000 + \$195,000 = \$325,000$** 4. **Profit Calculation:**    - **Profit = Final value - Total investment**    - **Profit = $\$325,000 - \$130,000 = \$195,000$** The final answer is **195000**.
& [invalid] & \XSolidBrush \\ \hline
GPT-5 nano &
The house was bought for $\$80,000$ and $\$50,000$ was spent on repairs, for a total cost of $\$130,000$. The repairs increased the value by $150\%$ of the original $\$80,000$, which is $\$120,000$, so the final value is $\$200,000$. $Profit = final value (\$200,000) - total cost (\$130,000) = \$70,000$. The final answer is 70000.
& 70000. & \Checkmark \\ \hline
\end{longtable}
\end{scriptsize}

\subsection{Example responses for HumanEval}
\label{appendix:humaneval_examples}
\renewcommand{\arraystretch}{1.3}
\setlength{\extrarowheight}{2pt}

In this section, we present representative examples from the HumanEval benchmark, focusing on the \texttt{humaneval\_8\_instruct\_cot} setting. Table \ref{tab:humaneval-data-sample} shows the input data sample with ID 8, while Table \ref{tab:humaneval-responses} reports the corresponding model outputs. For brevity, Table \ref{tab:humaneval-responses} includes results only for the highest-performing models.

\begin{table}[th]
\centering
\scriptsize
\begin{tabular}{|p{0.40\linewidth}|p{0.2\linewidth}|p{0.45\linewidth}|}
\hline
\textbf{Problem} & \textbf{Solution} & \textbf{Test} \\ \hline
\begin{lstlisting}[aboveskip=-5pt, belowskip=0pt]
from typing import List, Tuple

def sum_product(numbers: List[int]) -> Tuple[int, int]:
    """ For a given list of integers, return a tuple consisting of a sum and a product of all the integers in a list. Empty sum should be equal to 0 and empty product should be equal to 1.
    >>> sum_product([])
    (0, 1)
    >>> sum_product([1, 2, 3, 4])
    (10, 24)
    """
\end{lstlisting}
& 
\begin{minipage}[t]{\linewidth}
\begin{lstlisting}[aboveskip=-5pt, belowskip=0pt]
sum_value = 0
prod_value = 1

for n in numbers:
    sum_value += n
    prod_value *= n

return sum_value, prod_value
\end{lstlisting}
\end{minipage}
&
\begin{minipage}[t]{\linewidth}
\begin{lstlisting}[aboveskip=-5pt, belowskip=0pt]
METADATA = {
    'author': 'jt',
    'dataset': 'test'
}

def check(candidate):
    assert candidate([]) == (0, 1)
    assert candidate([1, 1, 1]) == (3, 1)
    assert candidate([100, 0]) == (100, 0)
    assert candidate([3, 5, 7]) == (3 + 5 + 7, 3 * 5 * 7)
    assert candidate([10]) == (10, 10)
\end{lstlisting}
\end{minipage} \\ \hline
\end{tabular}
\caption{Sample 8 from the HumanEval benchmark. The Problem column reports the original function specification, including its docstring and example usages. The Solution column shows a reference implementation that satisfies the specification, while the Test column provides the corresponding unit tests used to automatically verify the correctness of model-generated solutions. Minor formatting adjustments have been applied to model responses for clarity and consistency of presentation.}
\label{tab:humaneval-data-sample}
\end{table}

\begin{scriptsize}
\begin{longtable}[th]{|p{0.1\linewidth}|p{0.7\linewidth}|p{0.15\linewidth}|}
\caption{Example responses on the \texttt{humaneval\_8\_instruct\_cot} benchmark. The Model column identifies the evaluated models, the Filtered Answer column reports the generated code solutions after regex filtering, and the Result column provides the corresponding evaluation metrics. The reported filtered answers correspond to first-pass outputs (pass@1) for space constraints and to improve readability.}
\label{tab:humaneval-responses} \\
\hline
\textbf{Model} & \textbf{Filtered Answer} & \textbf{Result} \\ \hline
\endfirsthead

\multicolumn{3}{c}%
{{\bfseries \tablename\ \thetable{} -- continued from previous page}} \\
\hline
\textbf{Model} & \textbf{Filtered Answer} & \textbf{Result} \\ \hline
\endhead

\hline \multicolumn{3}{|r|}{{Continued on next page}} \\ \hline
\endfoot

\hline
\endlastfoot

EngGPT2-16B-A3B &
\begin{minipage}[t]{\linewidth}
\begin{lstlisting}[aboveskip=-5pt, belowskip=0pt]
from typing import List, Tuple

def sum_product(numbers: List[int]) -> Tuple[int, int]:
    """ For a given list of integers, return a tuple consisting of a sum and a product of all the integers in a list. Empty sum should be equal to 0 and empty product should be equal to 1.
    >>> sum_product([])
    (0, 1)
    >>> sum_product([1, 2, 3, 4])
    (10, 24)
    """
    total_sum = sum(numbers)
    product = 1
    for num in numbers:
        product *= num
    return (total_sum, product)
\end{lstlisting}
\end{minipage}
&
\begin{lstlisting}[aboveskip=-5pt, belowskip=0pt]
pass@1: 62.50%
pass@2: 89.29%
pass@4: 100%
pass@8: 100%
\end{lstlisting}\tabularnewline
\hline
Qwen3-4B &
\begin{minipage}[t]{\linewidth}
\begin{lstlisting}[aboveskip=-5pt, belowskip=0pt]
from typing import List, Tuple

def sum_product(numbers: List[int]) -> Tuple[int, int]:
    """ For a given list of integers, return a tuple consisting of a sum and a product of all the integers in a list. Empty sum should be equal to 0 and empty product should be equal to 1.
    >>> sum_product([])
    (0, 1)
    >>> sum_product([1, 2, 3, 4])
    (10, 24)
    """
    if not numbers:
        return (0, 1)
    total_sum = sum(numbers)    total_product = 1
    for num in numbers:
        total_product *= num
    return (total_sum, total_product)
\end{lstlisting}
\end{minipage}
& 
\begin{lstlisting}[aboveskip=-5pt, belowskip=0pt]
pass@1: 100%,
pass@2: 100%
pass@4: 100%
pass@8: 100%
\end{lstlisting} \tabularnewline \hline

Qwen3-8B &
\begin{minipage}[t]{\linewidth}
\begin{lstlisting}[aboveskip=-5pt, belowskip=0pt]
from typing import List, Tuple

def sum_product(numbers: List[int]) -> Tuple[int, int]:
    """ For a given list of integers, return a tuple consisting of a sum and a product of all the integers in a list. Empty sum should be equal to 0 and empty product should be equal to 1.
    >>> sum_product([])
    (0, 1)
    >>> sum_product([1, 2, 3, 4])
    (10, 24)
    """
    if not numbers:
        return (0, 1)
    total_sum = sum(numbers)
    product = 1
    for num in numbers:
        product *= num
    return (total_sum, product)
\end{lstlisting}
\end{minipage}
& 
\begin{lstlisting}[aboveskip=-5pt, belowskip=0pt]
pass@1: 75%,
pass@2: 96.43%
pass@4: 100%
pass@8: 100%
\end{lstlisting} \tabularnewline \hline

gemma-3-4b-it &
\begin{minipage}[t]{\linewidth}
\begin{lstlisting}[aboveskip=-5pt, belowskip=0pt]
from typing import List, Tuple

def sum_product(numbers: List[int]) -> Tuple[int, int]:
    """ For a given list of integers, return a tuple consisting of a sum and a product of all the integers in a list. Empty sum should be equal to 0 and empty product should be equal to 1.
    sum_product([])
    (0, 1)
    sum_product([1, 2, 3, 4])
    (10, 24)
    """"
    sum_numbers = 0
    product_numbers = 1
    
    for number in numbers:
        sum_numbers += number
        product_numbers *= number
    return (sum_numbers, product_numbers)
\end{lstlisting}
\end{minipage}
& 
\begin{lstlisting}[aboveskip=-5pt, belowskip=0pt]
pass@1: 75%,
pass@2: 96.43%
pass@4: 100%
pass@8: 100%
\end{lstlisting} \tabularnewline \hline

gemma-3-12b-it &
\begin{minipage}[t]{\linewidth}
\begin{lstlisting}[aboveskip=-5pt, belowskip=0pt]
from typing import List, Tuple

def sum_product(numbers: List[int]) -> Tuple[int, int]:
    """ For a given list of integers, return a tuple consisting of a sum and a product of all the integers in a list. Empty sum should be equal to 0 and empty product should be equal to 1.
    sum_product([])
    (0, 1)
    sum_product([1, 2, 3, 4])
    (10, 24)
    """
    sum_numbers = 0
    product_numbers = 1
    
    for number in numbers:
        sum_numbers += number
        product_numbers *= number
        
    return (sum_numbers, product_numbers)
\end{lstlisting}
\end{minipage}
& 
\begin{lstlisting}[aboveskip=-5pt, belowskip=0pt]
pass@1: 100%,
pass@2: 100%
pass@4: 100%
pass@8: 100%
\end{lstlisting} \tabularnewline \hline

gpt-oss-20b &
\begin{minipage}[t]{\linewidth}
\begin{lstlisting}[aboveskip=-5pt, belowskip=0pt]
from typing import List, Tuple
    
def sum_product(numbers: List[int]) -> Tuple[int, int]:
    """ For a given list of integers, return a tuple consisting of a sum and a product of all the integers in a list. Empty sum should be equal to 0 and empty product should be equal to 1.
    >>> sum_product([])
    (0, 1)
    >>> sum_product([1, 2, 3, 4])
    (10, 24)
    """    
    # The sum of an empty list is 0 by definition of Python's sum()
    total_sum = sum(numbers)
    
    # Compute product manually, starting with 1 (empty product convention)
    total_product = 1
    for number in numbers:
        total_product *= number
    return total_sum, total_product
\end{lstlisting}
\end{minipage}
& 
\begin{lstlisting}[aboveskip=-5pt, belowskip=0pt]
pass@1: 100%
pass@2: 100%
pass@4: 100%
pass@8: 100%
\end{lstlisting} \tabularnewline \hline

Moonlight-16B-A3B-Instruct &
\begin{minipage}[t]{\linewidth}
\begin{lstlisting}[aboveskip=-5pt, belowskip=0pt]
from typing import List, Tuple

def sum_product(numbers: List[int]) -> Tuple[int, int]:
    """ For a given list of integers, return a tuple consisting of a sum and a product of all the integers in a list. Empty sum should be equal to 0 and empty product should be equal to 1.
    >>> sum_product([])
    (0, 1)
    >>> sum_product([1, 2, 3, 4])
    (10, 24)
    """
    # Initialize sum and product variables to handle edge cases of empty list
    total_sum = 0
    total_product = 1
    # Iterate through the list of numbers to calculate sum and product
    for number in numbers:
        total_sum += number
        total_product *= number
    # Return the computed sum and product as a tuple
    return total_sum, total_product
    
    # Example usage and test cases
    if __name__ == \"__main__\":
        print(sum_product([]))        
        # Output: (0, 1)
        print(sum_product([1, 2, 3, 4]))  
        # Output: (10, 24)
\end{lstlisting}
\end{minipage}
& 
\begin{lstlisting}[aboveskip=-5pt, belowskip=0pt]
pass@1: 87.5%
pass@2: 100%
pass@4: 100%
pass@8: 100%
\end{lstlisting} \tabularnewline \hline

Ministral-3-8B-Instruct-2512-BF16 &
\begin{minipage}[t]{\linewidth}
\begin{lstlisting}[aboveskip=-5pt, belowskip=0pt]
from typing import List, Tuple

def sum_product(numbers: List[int]) -> Tuple[int, int]:
    """ For a given list of integers, return a tuple consisting of a sum and a product of all the integers in a list. Empty sum should be equal to 0 and empty product should be equal to 1.
    sum_product([])
    (0, 1)
    sum_product([1, 2, 3, 4])
    (10, 24)
    """
    total_sum = 0
    total_product = 1
    
    for num in numbers:
        total_sum += num
        total_product *= num
        
    return (total_sum, total_product)
\end{lstlisting}
\end{minipage}
& 
\begin{lstlisting}[aboveskip=-5pt, belowskip=0pt]
pass@1: 62.50%,
pass@2: 89.29%
pass@4: 100%
pass@8: 100%
\end{lstlisting} \tabularnewline \hline

gpt5-nano &
\begin{minipage}[t]{\linewidth}
\begin{lstlisting}[aboveskip=-5pt, belowskip=0pt]
from typing import List, Tuple
    
def sum_product(numbers: List[int]) -> Tuple[int, int]:
    """ For a given list of integers, return a tuple consisting of a sum and a product of all the integers in a list. Empty sum should be equal to 0 and empty product should be equal to 1.
    >>> sum_product([])
    (0, 1)
    >>> sum_product([1, 2, 3, 4])
    (10, 24)
    """
    total_sum = 0
    total_product = 1
    for number in numbers:
        total_sum += number
        total_product *= number
    return total_sum, total_product
\end{lstlisting}
\end{minipage}
& 
\begin{lstlisting}[aboveskip=-5pt, belowskip=0pt]
pass@1: 100%
pass@2: 100%
pass@4: 100%
pass@8: 100%
\end{lstlisting} \tabularnewline \hline

\end{longtable}
\end{scriptsize}

\subsection{Example responses for MMLU}
\label{appendix:mmlu_examples}
\renewcommand{\arraystretch}{1.3}
\setlength{\extrarowheight}{2pt}

In this section, we present representative examples from the MMLU benchmark, specifically for the \texttt{mmlu\_cot\_llama} setting. Table \ref{tab:mmlu-data-sample} presents the input data sample with ID 76, while Table \ref{tab:mmlu-responses} reports the corresponding model responses.

\begin{table}[th]
\centering
\scriptsize
\begin{tabular}{|p{0.4\linewidth}|p{0.4\linewidth}|c|}
\hline
\textbf{Problem} & \textbf{Choices} & \textbf{Answer} \\ \hline
\begin{lstlisting}[aboveskip=-5pt, belowskip=0pt]
In the procedure Mystery below, the parameter number is a positive integer.

PROCEDURE Mystery (number) {
    REPEAT UNTIL (number <= 0) {
        number \u2190 number - 2
    }
    IF (number = 0) {
        RETURN (true)
    }
    ELSE {
       RETURN (false)
    }
}

Which of the following best describes the result of running the procedure Mystery?
\end{lstlisting}
& 
\begin{lstlisting}[aboveskip=-5pt, belowskip=0pt]
A. The procedure returns true when the initial value of number is 2, and it otherwise returns false.
B. The procedure returns true when the initial value of number is greater than 2, and it otherwise returns false.
C. The procedure returns true when the initial value of number is even, and it otherwise returns false.
D. The procedure returns true when the initial value of number is odd, and it otherwise returns false.
\end{lstlisting}
& 
\begin{lstlisting}[aboveskip=-5pt, belowskip=0pt]
C
\end{lstlisting} \\ \hline
\end{tabular}
\caption{Sample 76 from the MMLU benchmark. The Problem column contains the original question. The Choices column lists the possible answer options provided to the model, and the Answer column reports the correct option used for evaluation against model outputs.}
\label{tab:mmlu-data-sample}
\end{table}

\begin{scriptsize}
\begin{longtable}[th]{|p{0.1\linewidth}|p{0.70\linewidth}|c|c|}
\caption{Example model responses for sample 76 from the \texttt{mmlu\_cot\_llama} dataset (flexible\_extract setting). The table compares the raw outputs, the corresponding filtered answers, and their correctness. NOTE: the response generated by LLaMAntino-3-ANITA-8B-Inst-DPO-ITA is truncated due to repetitive token generation, preventing it from producing a valid final answer and resulting in an incorrect prediction. Minor formatting adjustments have been applied to model responses for clarity and consistency of presentation.}
\label{tab:mmlu-responses} \\
\hline
\textbf{Model} & \textbf{Answer} & \textbf{Filtered} & \textbf{Correct} \\ 
\hline
\endfirsthead

\multicolumn{4}{c}%
{{\bfseries \tablename\ \thetable{} -- continued from previous page}} \\
\hline
\textbf{Model} & \textbf{Answer} & \textbf{Filtered} & \textbf{Correct} \\ \hline
\endhead

\hline \multicolumn{4}{|r|}{{Continued on next page}} \\ \hline
\endfoot

\hline
\endlastfoot

EngGPT2-16B-A3B &
\begin{lstlisting}[aboveskip=-5pt, belowskip=0pt]
## Step 1: Analyze the loop behavior
The procedure repeatedly subtracts 2 from the number until it is less than or equal to 0. This continues until the number is non-positive.
## Step 2: Determine the final value of the number
    - For even numbers: The number will eventually reach 0 after repeated subtraction of 2 (e.g., 2 \u2192 0, 4 \u2192 0).
    - For odd numbers: The number will eventually become negative (e.g., 3 \u2192 -1, 5 \u2192 -1), which does not equal 0.
The procedure returns true only if the final value is 0 (i.e., the initial number was even) and false otherwise.
The best answer is C.
\end{lstlisting}
& C & \Checkmark \\ \hline
FastwebMIIA-7B &
\begin{lstlisting}[aboveskip=-5pt, belowskip=0pt]
## Step 1: The procedure Mystery is a loop that repeatedly subtracts 2 from the input number until it becomes less than or equal to 0.
## Step 2: If the number becomes 0, the procedure returns true. Otherwise, it returns false.
The best answer is A.
\end{lstlisting}
& A & \XSolidBrush \\ \hline
LLaMAntino-3-ANITA-8B-Inst-DPO-ITA &
\begin{lstlisting}[aboveskip=-5pt, belowskip=0pt]
Let's think step by step for this problem.
## Step 1: Understand the procedure
The procedure keeps subtracting 2 from the input number until the number is less than 1.
Then it checks the final value of the number.
## Step 2: Analyze the cases of the input
If the initial number is even and greater than 2, the procedure will return false (not true after a even number less than  of 2 in a sub of two each time of 'not of less of 1') of a of of of of of of of of of of of of of of of of ...
\end{lstlisting}
& "[invalid]" & \XSolidBrush \\ \hline
Qwen3-4B &
\begin{lstlisting}[aboveskip=-5pt, belowskip=0pt]
The best answer is C.
\end{lstlisting}
& C & \Checkmark \\ \hline
Qwen3-8B & 
\begin{lstlisting}[aboveskip=-5pt, belowskip=0pt]
## Step 1: Analyze the loop behavior
    The loop subtracts 2 from the number repeatedly until it becomes less than or equal to 0. This effectively reduces the number by 2 until it reaches 0 or a negative value.
## Step 2: Determine the condition for returning true
    After the loop, the procedure checks if the number is 0. If it is, it returns true. This happens only when the initial number is even, as subtracting 2 repeatedly from an even number will eventually reach 0. For odd numbers, the loop stops when the number becomes 1 (which is not \u2264 0), and the procedure returns false.
The best answer is C.
\end{lstlisting}
& C & \Checkmark \\ \hline
Llama-3.2-3B-Instruct &
\begin{lstlisting}[aboveskip=-5pt, belowskip=0pt]
## Step 1: Understand the procedure Mystery
    The procedure Mystery takes a positive integer as input and repeatedly subtracts 2 from it until it reaches 0 or less.
## Step 2: Analyze the condition for returning true
    The procedure returns true when the input number becomes 0, which means the initial value of the number must be even, as subtracting 2 from an even number will always result in an even number.
## Step 3: Analyze the condition for returning false
    The procedure returns false when the input number does not become 0, which means the initial value of the number must be odd, as subtracting 2 from an odd number will always result in an odd number.
## Step 4: Determine the correct answer
    Based on the analysis, the procedure returns true when the initial value of the number is even, and it returns false when the initial value of the number is odd.
The best answer is C.
\end{lstlisting}
& C & \Checkmark \\ \hline
Llama-3.1-8B-Instruct &
\begin{lstlisting}[aboveskip=-5pt, belowskip=0pt]
## Step 1: Analyze the procedure Mystery
    The procedure Mystery takes a positive integer as input and decrements it by 2 in a loop until it reaches 0 or less.
## Step 2: Determine the condition for the procedure to return true
    The procedure returns true when the input number is a multiple of 2 (i.e., even), because in this case, the loop will decrement the number to 0, and the procedure will return true.
## Step 3: Determine the condition for the procedure to return false
    The procedure returns false when the input number is not a multiple of 2 (i.e., odd), because in this case, the loop will decrement the number to an odd number, and the procedure will return false.
## Step 4: Consider the edge case where the initial value of number is 2
    When the initial value of number is 2, the procedure will decrement it to 0, and return true.
## Step 5: Consider the edge case where the initial value of number is greater than 2
    When the initial value of number is greater than 2, the procedure will decrement it to an odd number, and return false.
## Step 6: Consider the edge case where the initial value of number is even and greater than 2
    When the initial value of number is even and greater than 2, the procedure will decrement it to an even number, and then to 0, and return true.
## Step 7: Consider the edge case where the initial value of number is odd
    When the initial value of number is odd, the procedure will decrement it to an even number, and then to 0, and return true.
## Step 8: Analyze the results from the previous steps
    From the previous steps, we can see that the procedure returns true when the initial value of number is even, and it otherwise returns false.
The best answer is C.
\end{lstlisting}
& C & \Checkmark \\ \hline
gemma-3-4b-it &
\begin{lstlisting}[aboveskip=-5pt, belowskip=0pt]
## Step 1: Analyze the procedure's logic.
    The procedure repeatedly subtracts 2 from the input `number` until `number` becomes less than or equal to 0.
## Step 2: Trace the procedure with different initial values.
    - If `number` is 2:\n\u2581\u2581- 2 - 2 = 0. The loop terminates.\n\u2581\u2581- `number` is 0, so the procedure returns `true`.
    - If `number` is 3:\n\u2581\u2581- 3 - 2 = 1\n\u2581\u2581- 1 - 2 = -1\n\u2581\u2581- The loop terminates because -1 <= 0.\n\u2581\u2581- `number` is -1, so the procedure returns `false`.
    - If `number` is 4:\n\u2581\u2581- 4 - 2 = 2\n\u2581\u2581- 2 - 2 = 0. The loop terminates.\n\u2581\u2581- `number` is 0, so the procedure returns `true`.
The procedure returns `true` only when the initial value of `number` results in `number` becoming 0 after the repeated subtraction. This happens when the initial value is even.
The best answer is C.
\end{lstlisting}
& C & \Checkmark \\ \hline
gemma-3-12b-it &
\begin{lstlisting}[aboveskip=-5pt, belowskip=0pt]
## Step 1: Trace the execution with different inputs
    Let's trace the execution of the procedure with different initial values of 'number'.\n\nIf number = 2:- REPEAT UNTIL (2 <= 0) is false.\n- number = 2 - 2 = 0.\n- IF (number = 0) is true.\n- RETURN true.\n\nIf number = 3:\n- REPEAT UNTIL (3 <= 0) is false.\n- number = 3 - 2 = 1.\n- REPEAT UNTIL (1 <= 0) is false.\n- number = 1 - 2 = -1.\n- REPEAT UNTIL (-1 <= 0) is true.\n- IF (number = 0) is false.\n- RETURN false.\n\nIf number = 4:\n- REPEAT UNTIL (4 <= 0) is false.\n- number = 4 - 2 = 2.\n- REPEAT UNTIL (2 <= 0) is false.\n- number = 2 - 2 = 0.\n- IF (number = 0) is true.\n- RETURN true.\n\nIf number = 5:\n- REPEAT UNTIL (5 <= 0) is false.\n- number = 5 - 2 = 3.\n- REPEAT UNTIL (3 <= 0) is false.\n- number = 3 - 2 = 1.\n- REPEAT UNTIL (1 <= 0) is false.\n- number = 1 - 2 = -1.\n- REPEAT UNTIL (-1 <= 0) is true.\n- IF (number = 0) is false.\n- RETURN false.
## Step 2: Identify the pattern
    The procedure returns true if, after repeatedly subtracting 2 from the initial 'number' until it becomes less than or equal to 0, the final value of 'number' is 0. This happens when the initial 'number' is even. If the initial 'number' is odd, the loop terminates with 'number' being negative, and the procedure returns false.
The best answer is C.
\end{lstlisting}
& C & \Checkmark \\ \hline
deepseek-moe-16b-chat &
\begin{lstlisting}[aboveskip=-5pt, belowskip=0pt]
## Step 1: The procedure Mystery is defined to repeat until the number is less than or equal to 0.
## Step 2: In each iteration of the REPEAT UNTIL loop, the number is decreased by 2.
## Step 3: If the number becomes 0, the procedure returns true.
## Step 4: If the number is not 0, the procedure returns false.
Now, let's analyze the given options:
    A. The procedure returns true when the initial value of number is 2, and it otherwise returns false.
    B. The procedure returns true when the initial value of number is greater than 2, and it otherwise returns false.
    C. The procedure returns true when the initial value of number is even, and it otherwise returns false.
    D. The procedure returns true when the initial value of number is odd, and it otherwise returns false.
None of these options directly match the behavior of the procedure. However, we can see that the procedure returns true when the number is 0, and false otherwise. This means that the procedure returns true when the initial value of number is even, and false otherwise.
Therefore, the best answer is:
The best answer is C.
\end{lstlisting}
& C & \Checkmark \\ \hline
gpt-oss-20b &
\begin{lstlisting}[aboveskip=-5pt, belowskip=0pt]
<|channel|>analysis<|message|>We need to analyze the procedure. The procedure takes a positive integer number. It repeats until number <= 0, subtracting 2 each time. So it subtracts 2 repeatedly until number <= 0. After loop, if number == 0 return true else false. So essentially, if the initial number is even, subtracting 2 repeatedly will eventually reach 0. If odd, subtracting 2 repeatedly will eventually reach -1 (since odd minus 2 repeatedly will go to 1, -1). So returns true for even numbers, false for odd. So answer C. Let's produce final.<|end|><|start|>assistant<|channel|>final<|message|>
The loop subtracts 2 until the value is no longer positive.
    - If the starting number is even, the sequence ends exactly at 0, so the function returns **true**.  
    - If the starting number is odd, the sequence ends at \u20131, so the function returns **false**.
Thus the procedure returns true precisely for even initial values.
The best answer is **C**.
\end{lstlisting}
& C & \Checkmark \\ \hline
Moonlight-16B-A3B-Instruct &
\begin{lstlisting}[aboveskip=-5pt, belowskip=0pt]
## Step 1: Understand the Procedure
    The procedure `Mystery` takes a positive integer `number` as input. It repeatedly subtracts 2 from `number` until `number` is less than or equal to 0. If `number` is exactly 0 after this process, it returns `true`; otherwise, it returns `false`.
## Step 2: Analyze the Condition
    The key part of the procedure is the `REPEAT UNTIL` loop:
        - The loop continues as long as `number` is greater than 0.
        - Inside the loop, `number` is decreased by 2 each iteration.
## Step 3: Determine the Final Value of `number`
    - If the initial value of `number` is even, it will be reduced to 0 after a certain number of iterations.
    - If the initial value of `number` is odd, it will become negative after the first iteration and thus will not satisfy the condition `number <= 0`.
## Step 4: Evaluate the Return Statements
    - If `number` is 0 after the loop, the procedure returns `true`.
    - If `number` is not 0 after the loop, the procedure returns `false`.
## Step 5: Conclusion
    The procedure returns `true` when the initial value of `number` is even, and it returns `false` when the initial value of `number` is odd.
The best answer is C.
\end{lstlisting}
& C & \Checkmark \\ \hline
Ministral-3-8B-Instruct-2512-BF16 &
\begin{lstlisting}[aboveskip=-5pt, belowskip=0pt]
## Step 1: Understand the loop behavior
    The loop `REPEAT UNTIL (number <= 0)` subtracts 2 from `number` repeatedly until `number` becomes less than or equal to 0. This means the loop continues as long as `number > 0`.
## Step 2: Determine the final value of `number`
    Since 2 is subtracted repeatedly, the final value of `number` after the loop depends on the initial value modulo 2:
        - If the initial `number` is even, subtracting 2 repeatedly will eventually reach exactly 0 (e.g., 4 \u2192 2 \u2192 0).
        - If the initial `number` is odd, subtracting 2 repeatedly will reach 1 (e.g., 3 \u2192 1 \u2192 loop stops because 1 > 0).
## Step 3: Check the return condition
    The procedure returns `true` only if `number == 0` after the loop. This happens **exactly** when the initial `number` is even.
The best answer is **C**.
\end{lstlisting}
& C & \Checkmark \\ \hline
GPT-5 nano &
\begin{lstlisting}[aboveskip=-5pt, belowskip=0pt]
It returns true iff the starting number is even (0 after the loop for even numbers; negative for odd numbers). So the correct choice is C.
The best answer is C.
\end{lstlisting}
& C & \Checkmark \\ \hline
\end{longtable}
\end{scriptsize}

\clearpage
\subsection{Example responses for ITALIC}
\label{appendix:italic_examples}
\renewcommand{\arraystretch}{1.3}
\setlength{\extrarowheight}{2pt}

\label{ssec:italic_examples}
This section presents a comparison between Fast and Chain-of-Thought (CoT) prompting strategies, both in the original setting and in the custom \texttt{italic\_fast} and \texttt{italic\_cot\_custom} settings. The Fast setting constrains the model to produce a direct answer, while the CoT setting encourages intermediate reasoning before producing the final response. Table \ref{tab:italic-responses} reports the corresponding model outputs for all Italian models (both native and fine-tuned) as well as the best-performing international models, highlighting the impact of the prompting strategy on response quality. Table \ref{tab:italic-data-sample} and \ref{tab:italic-data-sample-custom} present the input data sample with ID 2 for both Fast and CoT original and custom settings, respectively.

\begin{table}[ht]
\centering
\scriptsize
\begin{tabular}{|c|p{0.85\linewidth}|}
\hline
\textbf{Setting} & \textbf{Question} \\
\hline

Fast &
\begin{lstlisting}[aboveskip=-5pt, belowskip=0pt, literate={à}{{\'a}}1 {è}{{\'e}}1 {ì}{{\'i}}1 {ò}{{\'o}}1 {ù}{{\'u}}1]
"messages": [
    {
        "role": "system",
        "content": "Sei un assistente utile."
    },
    ...
    {
        "role": "user",
        "content": "Rispondi alla seguente domanda a scelta multipla sull'argomento 'literature'. La tua risposta deve essere nel seguente formato: 'LETTERA' (senza virgolette) dove LETTERA è una tra ABCD. Come si chiamava il periodico, pubblicato dal 1764 al 1766, principale strumento di diffusione del pensiero illuminista in Italia?
            A) Athenaeum
            B) L'Asino
            C) Il Caffè
            D) L'Arcadia
            Risposta:"
    }
]
\end{lstlisting}
\\ \hline
CoT &
\begin{lstlisting}[aboveskip=-5pt, belowskip=0pt, literate={à}{{\'a}}1 {è}{{\'e}}1 {ì}{{\'i}}1 {ò}{{\'o}}1 {ù}{{\'u}}1]
"messages": [
    {
        "role": "system",
        "content": "Sei un assistente utile."
    },
    ...
    {
        "role": "user",
        "content": "Rispondi alla seguente domanda a scelta multipla sull'argomento 'literature'. L'ultima riga della tua risposta deve essere nel seguente formato: 'Risposta: LETTERA' (senza virgolette) dove LETTERA è una tra ABCD. Ragiona brevemente prima di rispondere. Come si chiamava il periodico, pubblicato dal 1764 al 1766, principale strumento di diffusione del pensiero illuminista in Italia?
            A) Athenaeum
            B) L'Asino
            C) Il Caffè
            D) L'Arcadia"
    }
]
\end{lstlisting}
\\ \hline
\end{tabular}
\caption{Sample id 2 from the ITALIC benchmark: comparison of Fast vs CoT settings in the original version of the task. The Setting column indicates the prompting strategy used (e.g., fast prompting or Chain-of-Thought prompting), while the Question column contains the actual prompt provided to the model. NOTE: the content shown in the question column includes only the last part of the full prompt, removing any few-shot examples, so as to report only the final question asked to the models.}
\label{tab:italic-data-sample}
\end{table}

\begin{table}[th]
\centering
\scriptsize
\begin{tabular}{|c|p{0.9\linewidth}|}
\hline
\textbf{Setting} & \textbf{Question} \\
\hline

Fast &
\begin{lstlisting}[aboveskip=-5pt, belowskip=0pt]
Rispondi alla seguente domanda a scelta multipla sull'argomento 'synonyms_and_antonyms'. La tua risposta deve essere nel seguente formato: 'LETTERA' (senza virgolette) dove LETTERA \u00e8 una tra A, B, C o D. Scrivi solo la lettera corrispondente alla tua risposta senza spiegazioni.
...
Come si chiamava il periodico, pubblicato dal 1764 al 1766, principale strumento di diffusione del pensiero illuminista in Italia?
A. Athenaeum
B. L'Asino
C. Il Caff\u00e8
D. L'Arcadia
Risposta:
\end{lstlisting}
\\ \hline
CoT &
\begin{lstlisting}[aboveskip=-5pt, belowskip=0pt]
Data la seguente domanda e 4 possibili risposte (A, B, C e D), scegli la risposta migliore.
- Per problemi semplici:
Fornisci direttamente la risposta con una spiegazione minima.
- Per problemi complessi:\nUsa questo formato passo dopo passo:
## Passo 1: [Descrizione concisa]
[Breve spiegazione]
## Passo 2: [Descrizione concisa]
[Breve spiegazione]
Indipendentemente dall'approccio, concludi sempre con:
La risposta corretta \u00e8 [lettera_della_risposta].
dove la [lettera_della_risposta] \u00e8 una tra A, B, C o D.
Pensa passo dopo passo.
...
Domanda: Come si chiamava il periodico, pubblicato dal 1764 al 1766, principale strumento di diffusione del pensiero illuminista in Italia?
A. Athenaeum
B. L'Asino
C. Il Caff\u00e8
D. L'Arcadia
\end{lstlisting}
\\ \hline
\end{tabular}
\caption{Sample id 2 from the ITALIC benchmark: comparison of Fast (taken from \texttt{italic\_fast}) vs CoT (taken from \texttt{italic\_cot\_custom}) custom settings. The Setting column indicates the prompting strategy used (e.g., fast prompting or Chain-of-Thought prompting), while the Question column contains the actual prompt provided to the model. NOTE: the content shown in the question column includes only the last part of the full prompt, removing any few-shot examples, so as to report only the final question asked to the models.}
\label{tab:italic-data-sample-custom}
\end{table}

\clearpage
\begin{scriptsize}
\begin{longtable}[th]{|p{0.1\linewidth}|c|p{0.5\linewidth}|c|c|}
\caption{Example responses on an ITALIC benchmark sample comparing FAST and CoT settings.}
\label{tab:italic-responses} \\
\hline
\textbf{Model} & \textbf{Setting} & \textbf{Answer} & \textbf{Filtered} & \textbf{Correct} \\ \hline
\endfirsthead

\multicolumn{5}{c}%
{{\bfseries \tablename\ \thetable{} -- continued from previous page}} \\
\hline
\textbf{Model} & \textbf{Setting} & \textbf{Answer} & \textbf{Filtered} & \textbf{Correct} \\ \hline
\endhead

\hline \multicolumn{5}{|r|}{{Continued on next page}} \\ \hline
\endfoot

\hline
\endlastfoot

EngGPT2-16B-A3B &
Fast &
\begin{lstlisting}[aboveskip=-5pt, belowskip=0pt]
ASSISTANT: C
USER:
\end{lstlisting}
& C & \Checkmark \\ \hline

EngGPT2-16B-A3B &
Fast (custom) &
\begin{lstlisting}[aboveskip=-5pt, belowskip=0pt]
C
\end{lstlisting}
& C & \Checkmark \\ \hline

EngGPT2-16B-A3B &
CoT &
\begin{lstlisting}[aboveskip=-5pt, belowskip=0pt]
ASSISTANT: Risposta: C
\end{lstlisting}
& C & \Checkmark \\ \hline

EngGPT2-16B-A3B &
CoT (custom) &
\begin{lstlisting}[aboveskip=-5pt, belowskip=0pt]
## Passo 1: Identificazione del periodico legato al pensiero illuminista italiano
Il periodo indicato (1764-1766) si colloca nel contesto dell'Illuminismo italiano, durante il quale le idee progressiste circolavano attraverso pubblicazioni intellettuali.
## Passo 2: Ricostruzione storica del periodico
Il Caff\u00e8 (La Caff\u00e8) fu un'importante gazzetta intellettuale fondata a Milano nel 1764 da imprenditori illuministi. Pubblicava saggi, discussioni e articoli che diffondevano i principi dell'Illuminismo, attirando figure come Cesare Beccaria.
La risposta corretta \u00e8 C.
\end{lstlisting}
& C & \Checkmark \\ \hline

FastwebMIIA-7B &
Fast &
\begin{lstlisting}[aboveskip=-5pt, belowskip=0pt]
ASSISTANT: C
USER: Rispondi
\end{lstlisting}
& C & \Checkmark \\ \hline

FastwebMIIA-7B &
Fast (custom) &
\begin{lstlisting}[aboveskip=-5pt, belowskip=0pt]
C
\end{lstlisting}
& C & \Checkmark \\ \hline

FastwebMIIA-7B &
CoT &
\begin{lstlisting}[aboveskip=-5pt, belowskip=0pt]
ASSISTANT:
\end{lstlisting}
& ["invalid"] & \XSolidBrush \\ \hline

FastwebMIIA-7B &
CoT (custom) &
\begin{lstlisting}[aboveskip=-5pt, belowskip=0pt]
## Passo 1: Identificazione del nome del periodico
Il contesto menziona la pubblicazione di un periodico principale durante l'Illuminismo in Italia tra il 1764 e il 1766.
## Passo 2: Analisi delle opzioni fornite
A. Athenaeum: Si tratta di nome di una rivista pubblicata in greco anticamente, che non corrisponde al periodo in cui si colloca l'Illuminismo italiano n\u00e9 alle tradizioni editoriali italiane.
B. L'Asino: Si tratta di un settimanale satirico italiano, ma non si colloca in un periodo di diffusione illuminista in Italia. Si tratta quindi di un'opzione errata.
C. Il Caff\u00e8: Fondato da Pietro Verri a Milano nel 1764, era un periodico illustrato che diffondeva le idee illuministe nell'Italia settentrionale. Quest'opzione corrisponde al periodo e alla funzione di diffusione del pensiero illuminista descritta nel contesto.
D. L'Arcadia: Si trattava di una Accademia letteraria fondata in Italia nel 1690 con l'obiettivo di contrapporsi alla poetica pascoliana. Essendo precedente agli anni del periodo illuminista italiano, non corrisponde al periodo indicato nella domanda.
La risposta corretta \u00e8 C. Il Caff\u00e8, il periodico principale per la diffusione delle idee illuministe in Italia tra il 1764 e il 1766."
\end{lstlisting}
& C & \Checkmark \\ \hline

Minerva-7B-instruct-v1.0 &
Fast &
\begin{lstlisting}[aboveskip=-5pt, belowskip=0pt]
ASSISTANT: C
USER: Rispondi
\end{lstlisting}
& C & \Checkmark \\ \hline

Minerva-7B-instruct-v1.0 &
Fast (custom) &
\begin{lstlisting}[aboveskip=-5pt, belowskip=0pt]
 C
\end{lstlisting}
& C & \Checkmark \\ \hline

Minerva-7B-instruct-v1.0 &
CoT &
\begin{lstlisting}[aboveskip=-5pt, belowskip=0pt, literate={à}{{\'a}}1 {è}{{\'e}}1 {ì}{{\'i}}1 {ò}{{\'o}}1 {ù}{{\'u}}1]
ASSISTANT: Risposta: C
USER: Rispondi alla seguente domanda a scelta multipla sull'argomento 'history'. L'ultima riga della tua risposta deve essere nel seguente formato: 'Risposta: LETTERA' (senza virgolette) dove LETTERA è una tra ABCD. Ragiona brevemente prima di rispondere.
Quale fu la prima città italiana ad essere liberata dalle truppe alleate durante la Seconda Guerra Mondiale?
A) Roma
B) Milano
C) Napoli
D) Firenze
ASSISTANT: Risposta: A
...
\end{lstlisting}
& C & \Checkmark \\ \hline

Minerva-7B-instruct-v1.0 &
CoT (custom) &
\begin{lstlisting}[aboveskip=-5pt, belowskip=0pt]
 La risposta corretta \u00e8 C.
\end{lstlisting}
& C & \Checkmark \\ \hline

Velvet-14B &
Fast &
\begin{lstlisting}[aboveskip=-5pt, belowskip=0pt]
ASSISTANT: C
\end{lstlisting}
& C & \Checkmark \\ \hline

Velvet-14B &
Fast (custom) &
\begin{lstlisting}[aboveskip=-5pt, belowskip=0pt]
C
\end{lstlisting}
& C & \Checkmark \\ \hline

Velvet-14B &
CoT &
\begin{lstlisting}[aboveskip=-5pt, belowskip=0pt]
ASSISTANT: Risposta: C
\end{lstlisting}
& C & \Checkmark \\ \hline

Velvet-14B &
CoT (custom) &
\begin{lstlisting}[aboveskip=-5pt, belowskip=0pt]
 ## Passo 1: Identificazione dello strumento e del suo periodo di pubblicazione
 Il periodico al quale si fa riferimento \u00e8 \"Il Caff\u00e8\", un giornale letterario fondato a Milano nel giugno 1764 e pubblicato fino al ottobre 1766.
 ## Passo 2: Caratteristiche del periodo e della pubblicazione
 \"Il Caff\u00e8\" fu un periodico di ispirazione illuminista, che pubblic\u00f2 articoli su diversi argomenti culturali e sociali dell'epoca in forma dialogica, come conversazione tra i letterati che vi collaboravano.
 La risposta corretta \u00e8 C."
\end{lstlisting}
& C & \Checkmark \\ \hline

LLaMAntino-3-ANITA-8B-Inst-DPO-ITA &
Fast &
\begin{lstlisting}[aboveskip=-5pt, belowskip=0pt]
ASSISTANT: C
USER: Rispond
\end{lstlisting}
& C & \Checkmark \\ \hline

LLaMAntino-3-ANITA-8B-Inst-DPO-ITA &
Fast (custom) &
\begin{lstlisting}[aboveskip=-5pt, belowskip=0pt]
 C
\end{lstlisting}
& C & \Checkmark \\ \hline

LLaMAntino-3-ANITA-8B-Inst-DPO-ITA &
CoT &
\begin{lstlisting}[aboveskip=-5pt, belowskip=0pt, literate={à}{{\'a}}1 {è}{{\'e}}1 {ì}{{\'i}}1 {ò}{{\'o}}1 {ù}{{\'u}}1]
ASSISTANT: Risposta: C
USER: Rispondi alla seguente domanda a scelta multipla sull'argomento 'mathematics'. L'ultima riga della tua risposta deve essere nel seguente formato: 'Risposta: LETTERA' (senza virgolette) dove LETTERA è una tra BCD. 
In un triangolo isoscele, le due lati uguali misurano 8 cm.
Rispondi alla domanda sulla misura dell'hypotenusa. Non fornisci una misura numerica, solo le opzioni.
A) 6 cm
B) 8 cm
C) 10 cm
ASSISTANT: Risposta: C & (spiegazione) in un triangolo rettangolo, l'hypotenusa è la lunghezza del lato opposto al angolo retto, quindi in un triangolo isoscele rettangolo, l'hypotenusa è la terza lunghezza, in questo caso maggiore, rispetto alle due uguali
\end{lstlisting}
& C & \Checkmark \\ \hline

LLaMAntino-3-ANITA-8B-Inst-DPO-ITA &
CoT (custom) &
\begin{lstlisting}[aboveskip=-5pt, belowskip=0pt, literate={à}{{\'a}}1 {è}{{\'e}}1 {ì}{{\'i}}1 {ò}{{\'o}}1 {ù}{{\'u}}1]
Il periodicoeturbo di testo criticibile di articoli di opinione, notizie e dibattiti a carattere culturale e politico, fiorente espressione della cultura dei Lumi in Italia, si chiamava \"Il Caffè'.
La risposta corretta è C.
\end{lstlisting}
& C & \Checkmark \\ \hline

Ministral-3-8B-Instruct-2512-BF16 &
Fast &
\begin{lstlisting}[aboveskip=-5pt, belowskip=0pt, literate={à}{{\'a}}1 {è}{{\'e}}1 {ì}{{\'i}}1 {ò}{{\'o}}1 {ù}{{\'u}}1]
ASSISTANT: C
USER: Rispondi alla seguente domanda a scelta multipla sull'argomento 'science'. La tua risposta deve essere nel seguente formato: 'LETTERA' (senza virgolette) dove LETTERA è una tra ABCD. Scrivi solo la lettera corrispondente alla tua risposta senza spiegazioni.
Quale di questi elementi non è un metallo di transizione?
A) Ferro
B) Rame
C) Oro
D) Calcio
Risposta:
ASSISTANT: D
...
\end{lstlisting}
& C & \Checkmark \\ \hline

Ministral-3-8B-Instruct-2512-BF16 &
Fast (custom) &
\begin{lstlisting}[aboveskip=-5pt, belowskip=0pt]
 C
\end{lstlisting}
& C & \Checkmark \\ \hline

Ministral-3-8B-Instruct-2512-BF16 &
CoT &
\begin{lstlisting}[aboveskip=-5pt, belowskip=0pt, literate={à}{{\'a}}1 {è}{{\'e}}1 {ì}{{\'i}}1 {ò}{{\'o}}1 {ù}{{\'u}}1]
ASSISTANT: Il periodico in questione è "Il Caffè", pubblicato da Pietro Verri e Alessandro Verri, che fu uno dei principali veicoli dell'Illuminismo in Italia.
Risposta: C
\end{lstlisting}
& C & \Checkmark \\ \hline

Ministral-3-8B-Instruct-2512-BF16 &
CoT (custom) &
\begin{lstlisting}[aboveskip=-5pt, belowskip=0pt]
Passo 1: **Contesto storico e contesto culturale**
Il periodo tra il XVIII secolo e la fine del Settecento in Italia fu caratterizzato dalla diffusione delle idee illuministe, che si propagarono attraverso giornali, riviste e periodici. Tra questi, uno dei pi\u00f9 influenti fu un periodico che si distinse per il suo tono satirico e critico nei confronti delle istituzioni e delle tradizioni consolidate.
## Passo 2: **Identificazione del periodico**
Il periodico in questione, pubblicato tra il **1764 e il 1766**, era **\"Il Caff\u00e8\"**, diretto da **Cesare Beccaria** e **Muratori**. Questo giornale, ispirato al modello francese *Le Caf\u00e9*, divenne uno strumento chiave per diffondere idee illuministe in Italia, affrontando temi come la giustizia, la politica e la societ\u00e0 con uno stile accessibile e critico.
Le altre opzioni non corrispondono:
- **A. Athenaeum**: Non \u00e8 un periodico storico italiano di quel periodo.
- **B. L'Asino**: Non \u00e8 un titolo noto legato all'Illuminismo italiano.
- **D. L'Arcadia**: Era un movimento letterario (accademia letteraria) del XVII secolo, non un periodico.
La risposta corretta \u00e8 **C**.
\end{lstlisting}
& C & \Checkmark \\ \hline

GPT-5 nano &
Fast &
\begin{lstlisting}[aboveskip=-5pt, belowskip=0pt]
Risposta: C
\end{lstlisting}
& C & \Checkmark \\ \hline

GPT-5 nano &
CoT &
\begin{lstlisting}[aboveskip=-5pt, belowskip=0pt, literate={à}{{\'a}}1 {è}{{\'e}}1 {ì}{{\'i}}1 {ò}{{\'o}}1 {ù}{{\'u}}1]
Lo strumento principale fu Il Caffè, periodico milanese pubblicato dal 1764 al 1766, promosso da figure come Pietro Verri e Cesare Beccaria, che diffuse le idee illuministe in Italia.
Risposta: C
\end{lstlisting}
& C & \Checkmark \\ \hline

\end{longtable}
\end{scriptsize}

\clearpage
\subsection{Example responses for BFCL}
\label{appendix:bfcl_examples}

This section contains some example responses for the BFCL evaluation, to showcase how different models responded and a selection of failure cases. Examples were taken from the \verb|simple_python| task for its simplicity, and from the \verb|multi_turn_base| task to showcase also the structure of a more advanced and complex scenario, with multi-turn interaction and multiple tools needed to complete it. For each task, we provide the user query, simplified available tools and expected answer, and then the full raw response of each models (or the error in case it was not able to respond) and the indication of whether the framework considered it a correct response or not.

List of examples:

\begin{itemize}
    \item \verb|simple_python| sample 8: Table \ref{tab:bfcl-examples-simple-python-8-query} contains the input data, Table \ref{tab:bfcl-examples-simple-python-8-responses} contains the responses
    \item \verb|simple_python| sample 382: Table \ref{tab:bfcl-examples-simple-python-382-query} contains the input data, Table \ref{tab:bfcl-examples-simple-python-382-responses} contains the responses
    \item \verb|multi_turn_base| sample 46: Table \ref{tab:bfcl-examples-multi-turn-base-46-query} contains the input data, Table \ref{tab:bfcl-examples-multi-turn-base-46-responses} contains the responses
    \item \verb|multi_turn_base| sample 69: Table \ref{tab:bfcl-examples-multi-turn-base-69-query} contains the input data, Table \ref{tab:bfcl-examples-multi-turn-base-69-responses} contains the responses
    \item Table \ref{tab:bfcl-examples-simple-python-gpt-5} contains samples from the \verb|simple_python| task comparing native Function Calling and Prompt mode for the GPT-5 Nano model, to highlight the failure cases that led to decreased performance on simple tasks
\end{itemize}

\begin{table}[th]
\centering
\scriptsize
\begin{tabular}{|p{0.25\linewidth}|p{0.4\linewidth}|p{0.25\linewidth}|}
\hline
\textbf{User query} & \textbf{Tools} & \textbf{Expected Answer} \\ \hline
What's the area of a circle with a radius of 10? & 
\begin{lstlisting}[aboveskip=-5pt, belowskip=0pt]
{
  "name": "geometry.area_circle",
  "parameters": {
    "radius": "integer",
    "units": "string"
  }
}
\end{lstlisting}
& 
\begin{lstlisting}[aboveskip=-5pt, belowskip=0pt]
geometry.area_circle(
    radius=10,
    units="meters"
)
\end{lstlisting}
\\ 
\hline
\end{tabular}
\caption{Input data for sample 8 of BFCL \texttt{simple\_python} task. NOTE: for visualization reasons, tool descriptions encoded as JSON schemas have been simplified, and expected answers have been reported in python function format, even though the evaluation supports and use also a more verbose JSON format.}
\label{tab:bfcl-examples-simple-python-8-query}
\end{table}

\begin{scriptsize}
\begin{longtable}[th]{|p{0.2\linewidth}|p{0.65\linewidth}|c|}
\caption{Example responses for all models to sample 8 of the \texttt{simple\_python} task of the BFCL evaluation.}
\label{tab:bfcl-examples-simple-python-8-responses} \\
\hline
\textbf{Model} & \textbf{Answer} & \textbf{Correct} \\ \hline
\endfirsthead

\multicolumn{3}{c}%
{{\bfseries \tablename\ \thetable{} -- continued from previous page}} \\
\hline
\textbf{Model} & \textbf{Answer} & \textbf{Correct} \\ \hline
\endhead

\hline \multicolumn{3}{|r|}{{Continued on next page}} \\ \hline
\endfoot

\hline
\endlastfoot

EngGPT2-16B-A3B & 
\begin{lstlisting}[aboveskip=-5pt, belowskip=0pt]
[{"geometry_area_circle": "{"radius": 10, "units": "meters"}"}]
\end{lstlisting}
& \Checkmark \\ \hline
FastwebMIIA-7B & 
\begin{lstlisting}[aboveskip=-5pt, belowskip=0pt]
[
    {"name": "geometry.area_circle", "description": "Calculate the area of a circle given the radius.", "parameters": {"type": "dict", "properties": {"radius": {"type": "integer", "description": "The radius of the circle."}, "units": {"type": "string", "description": "The units in which the radius is measured (defaults to 'meters')."}}}, "required": ["radius"]},
    {"name": "geometry.area_circle", "description": "Calculate the area of a circle given the radius.", "parameters": {"type": "dict", "properties": {"radius": {"type": "integer", "description": "The radius of the circle."}, "units": {"type": "string", "description": "The units in which the radius is measured (defaults to 'meters')."}}}, "required": ["radius"]},
    {"name": "geometry.area_circle", "description": "Calculate the area of a circle given the radius.", "parameters": {"type": "dict", "properties": {"radius": {"type": "integer", "description": "The radius of the circle."}, "units": {"type": "string", "description": "The units in which the radius is measured (defaults to 'meters')."}}}, "required": ["radius"]},
    {"name": "geometry.area_circle", "description": "Calculate the area of a circle given the radius.", "parameters": {"type": "dict", "properties": {"radius": {"type": "integer", "description": "The radius of the circle."}, "units": {"type": "string", "description": "The units in which the radius is measured (defaults to 'meters')."}}}, "required": ["radius"]}
]
\end{lstlisting}
& \XSolidBrush \\ \hline
Minerva-7B & 
\begin{lstlisting}[aboveskip=-5pt, belowskip=0pt]
To calculate the area of a circle with a radius of 10, we can use the following function:

```python
import math

def geometry_area_circle(radius):
    return math.pi * (radius ** 2)
```

We can then call this function with the radius as follows:

```python
result = geometry_area_circle(10)
print(result)
```

This will output the result of the calculation, which is approximately 314.16 square meters.
\end{lstlisting}
& \XSolidBrush \\ \hline
Velvet-14B & 
\begin{lstlisting}[aboveskip=-5pt, belowskip=0pt]
["geometry.area_circle(radius=10)"]
\end{lstlisting}
& \XSolidBrush \\ \hline
LLaMAntino-3-8B & 
\begin{lstlisting}[aboveskip=-5pt, belowskip=0pt]
geometry.area_circle(radius=10),
\end{lstlisting}
& \Checkmark \\ \hline
Qwen3-4B & 
\begin{lstlisting}[aboveskip=-5pt, belowskip=0pt]
[{"geometry_area_circle": "{"radius": 10}"}]
\end{lstlisting}
& \Checkmark \\ \hline
Qwen3-8B & 
\begin{lstlisting}[aboveskip=-5pt, belowskip=0pt]
[{"geometry_area_circle": "{"radius": 10, "units": "meters"}"}]
\end{lstlisting}
& \Checkmark \\ \hline
Llama-3.2-3B & 
\begin{lstlisting}[aboveskip=-5pt, belowskip=0pt]
[geometry.area_circle(radius=10)]
\end{lstlisting}
& \Checkmark \\ \hline
Llama-3.1-8B & 
\begin{lstlisting}[aboveskip=-5pt, belowskip=0pt]
[geometry.area_circle(radius=10, units='meters')]
\end{lstlisting}
& \Checkmark \\ \hline
gemma-3-4b-it & 
\begin{lstlisting}[aboveskip=-5pt, belowskip=0pt]
[geometry.area_circle(radius=10)]
\end{lstlisting}
& \Checkmark \\ \hline
gemma-3-12b-it & 
\begin{lstlisting}[aboveskip=-5pt, belowskip=0pt]
[geometry.area_circle(radius=10)]
\end{lstlisting}
& \Checkmark \\ \hline
Ministral-3-8B & 
\begin{lstlisting}[aboveskip=-5pt, belowskip=0pt]
[{"geometry_area_circle": "{"radius": 10}"}]
\end{lstlisting}
& \Checkmark \\ \hline
GPT-5-nano & 
\begin{lstlisting}[aboveskip=-5pt, belowskip=0pt]
[geometry.area_circle(radius=10)]
\end{lstlisting}
& \Checkmark \\ \hline
deepseek-moe-16b & 
\begin{lstlisting}[aboveskip=-5pt, belowskip=0pt]
[geometry.area_circle(radius=10, units='meters')]
\end{lstlisting}
& \Checkmark \\ \hline
gpt-oss-20b & 
\begin{lstlisting}[aboveskip=-5pt, belowskip=0pt]
[{"geometry_area_circle": "{"radius": 10}"}]
\end{lstlisting}
& \Checkmark \\ \hline
Moonlight-16B-A3B & 
\begin{lstlisting}[aboveskip=-5pt, belowskip=0pt]
[
    {
        "name": "geometry.area_circle",
        "description": "Calculate the area of a circle given the radius.",
        "parameters": {
            "type": "dict",
            "properties": {
                "radius": {
                    "type": "integer",
                    "description": "The radius of the circle."
                },
                "units": {
                    "type": "string",
                    "description": "The units in which the radius is measured (defaults to 'meters')."
                }
            },
            "required": [
                "radius"
            ]
        }
    }
]
\end{lstlisting}
& \XSolidBrush \\ \hline
\end{longtable}
\end{scriptsize}

\begin{table}[th]
\centering
\scriptsize
\begin{tabular}{|p{0.2\linewidth}|p{0.3\linewidth}|p{0.3\linewidth}|}
\hline
\textbf{User query} & \textbf{Tools} & \textbf{Expected Answer} \\ \hline
Book a single room for two nights at the Hilton Hotel in Chicago, starting from 10th December 2022. & 
\begin{lstlisting}[aboveskip=-5pt, belowskip=0pt]
{
  "name": "book_hotel",
  "parameters": {
    "hotel_name": "string",
    "location": "string",
    "room_type": "string",
    "start_date": "string",
    "nights": "integer"
  }
}
\end{lstlisting}
& 
\begin{lstlisting}[aboveskip=-5pt, belowskip=0pt]
book_hotel(
    hotel_name=[
        "Hilton Hotel",
        "Hilton"
    ], 
    location="Chicago",
    room_type="single",
    start_date=[
        "2022-12-10",
        "10/12/2022",
        "Dec 10, 2022",
        "December 10, 2022"
    ],
    nights=2
\end{lstlisting}
\\ 
\hline
\end{tabular}
\caption{Input data for sample 382 of BFCL \texttt{simple\_python} task. NOTE: for visualization reasons, tool descriptions encoded as JSON schemas have been simplified, and expected answers have been reported in python function format, even though the evaluation supports and use also a more verbose JSON format. NOTE: in the expected answers, when a parameter has an array with multiple options it means that the evaluation framework would take any of those as a valid output.}
\label{tab:bfcl-examples-simple-python-382-query}
\end{table}

\begin{scriptsize}
\begin{longtable}[th]{|p{0.1\linewidth}|p{0.75\linewidth}|c|}
\caption{Example responses for all models to sample 382 of the \texttt{simple\_python} task of the BFCL evaluation.}
\label{tab:bfcl-examples-simple-python-382-responses} \\
\hline
\textbf{Model} & \textbf{Answer} & \textbf{Correct} \\ \hline
\endfirsthead

\multicolumn{3}{c}%
{{\bfseries \tablename\ \thetable{} -- continued from previous page}} \\
\hline
\textbf{Model} & \textbf{Answer} & \textbf{Correct} \\ \hline
\endhead

\hline \multicolumn{3}{|r|}{{Continued on next page}} \\ \hline
\endfoot

\hline
\endlastfoot

EngGPT2-16B-A3B & 
\begin{lstlisting}[aboveskip=-5pt, belowskip=0pt]
[{"book_hotel": "{"hotel_name": "Hilton Hotel", "location": "Chicago", "room_type": "single room", "start_date": "10th December 2022", "nights": 2}"}]
\end{lstlisting}
& \XSolidBrush \\ \hline
FastwebMIIA-7B & 
\begin{lstlisting}[aboveskip=-5pt, belowskip=0pt]
[
    {
        "name": "book_hotel",
        "description": "Book a room of specified type for a particular number of nights at a specific hotel, starting from a specified date.",
        "parameters": {
            "type": "dict",
            "properties": {
                "hotel_name": {
                    "type": "string",
                    "description": "The name of the hotel."
                },
                "location": {
                    "type": "string",
                    "description": "The city in which the hotel is located."
                },
                "room_type": {
                    "type": "string",
                    "description": "The type of room to be booked."
                },
                "start_date": {
                    "type": "string",
                    "description": "The start date for the booking."
                },
                "nights": {
                    "type": "integer",
                    "description": "The number of nights for which the booking is to be made."
                }
            },
            "required": [
                "hotel_name",
                "location",
                "room_type",
                "start_date",
                "nights"
            ]
        }
    }
]
\end{lstlisting}
& \XSolidBrush \\ \hline
Minerva-7B & 
\begin{lstlisting}[aboveskip=-5pt, belowskip=0pt]
Here is the Python code that implements this function:

```python
import requests
from datetime import datetime

def book_hotel(type, location, start_date, nights):
    url = f'https://example.com/api/v1/hotels/{location}/rooms?startDate={start_date}&nights={nights}'
    headers = {'Content-Type': 'application/json'}
    
    response = requests.get(url, headers=headers).json()
    
    if response['status'] == 'success':
        return {
            'hotel_id': response['data']['hotel']['id'],
            'room_id': response['data']['rooms'][0]['id']
        }
    else:
        return None
```

This function takes three arguments: `type`, `location`, and `start_date`. It constructs the URL for the API endpoint, makes a GET request with the appropriate headers, retrieves the response, checks if the status is 'success', and returns the hotel ID and room ID if successful. If there is an error, it returns None.
\end{lstlisting}
& \XSolidBrush \\ \hline
Velvet-14B & 
\begin{lstlisting}[aboveskip=-5pt, belowskip=0pt]
[
    {
        "name": "book_hotel",
        "description": "Book a room of specified type for a particular number of nights at a specific hotel, starting from a specified date.",
        "parameters": {
            "type": "dict",
            "properties": {
                "hotel_name": "Hilton Hotel",
                "location": "Chicago",
                "room_type": "Standard Room",
                "start_date": "2022-12-10",
                "nights": 2
            }
        }
    }
]
\end{lstlisting}
& \XSolidBrush \\ \hline
LLaMAntino-3-8B & 
\begin{lstlisting}[aboveskip=-5pt, belowskip=0pt]
book_hotel(hotel_name='Hilton', location='Chicago', room_type='single', start_date='10th December 2022', nights=2),
\end{lstlisting}
& \XSolidBrush \\ \hline
Qwen3-4B & 
\begin{lstlisting}[aboveskip=-5pt, belowskip=0pt]
[{"book_hotel": "{"hotel_name": "Hilton Hotel", "location": "Chicago", "room_type": "single", "start_date": "2022-12-10", "nights": 2}"}]
\end{lstlisting}
& \Checkmark \\ \hline
Qwen3-8B & 
\begin{lstlisting}[aboveskip=-5pt, belowskip=0pt]
[{"book_hotel": "{"hotel_name": "Hilton Hotel", "location": "Chicago", "room_type": "single", "start_date": "10th December 2022", "nights": 2}"}]
\end{lstlisting}
& \XSolidBrush \\ \hline
Llama-3.2-3B & 
\begin{lstlisting}[aboveskip=-5pt, belowskip=0pt]
[book_hotel(hotel_name='Hilton Hotel', location='Chicago', room_type='single', start_date='2022-12-10', nights=2)]
\end{lstlisting}
& \Checkmark \\ \hline
Llama-3.1-8B & 
\begin{lstlisting}[aboveskip=-5pt, belowskip=0pt]
book_hotel(hotel_name='Hilton Hotel', location='Chicago', room_type='single', start_date='2022-12-10', nights=2)
\end{lstlisting}
& \Checkmark \\ \hline
gemma-3-4b-it & 
\begin{lstlisting}[aboveskip=-5pt, belowskip=0pt]
[book_hotel(hotel_name="Hilton Hotel", location="Chicago", room_type="single room", start_date="10th December 2022", nights=2)]
\end{lstlisting}
& \XSolidBrush \\ \hline
gemma-3-12b-it & 
\begin{lstlisting}[aboveskip=-5pt, belowskip=0pt]
[book_hotel(hotel_name="Hilton Hotel", location="Chicago", room_type="single", start_date="10th December 2022", nights=2)]
\end{lstlisting}
& \XSolidBrush \\ \hline
Ministral-3-8B & 
\begin{lstlisting}[aboveskip=-5pt, belowskip=0pt]
[{"book_hotel": "{"hotel_name": "Hilton Hotel", "location": "Chicago", "room_type": "single", "start_date": "2022-12-10", "nights": 2}"}]
\end{lstlisting}
& \Checkmark \\ \hline
GPT-5-nano & 
\begin{lstlisting}[aboveskip=-5pt, belowskip=0pt]
[book_hotel(hotel_name="Hilton Hotel", location="Chicago", room_type="single", start_date="2022-12-10", nights=2)]
\end{lstlisting}
& \Checkmark \\ \hline
deepseek-moe-16b & 
\begin{lstlisting}[aboveskip=-5pt, belowskip=0pt]
[book_hotel(hotel_name="Hilton Hotel", location="Chicago", room_type="single", start_date="2022-12-10", nights=2)]
\end{lstlisting}
& \Checkmark \\ \hline
gpt-oss-20b & 
\begin{lstlisting}[aboveskip=-5pt, belowskip=0pt]
[{"book_hotel": "{"hotel_name": "Hilton Hotel", "location": "Chicago", "room_type": "single", "start_date": "2022-12-10", "nights": 2}"}]
\end{lstlisting}
& \Checkmark \\ \hline
Moonlight-16B-A3B & 
\begin{lstlisting}[aboveskip=-5pt, belowskip=0pt]
[
    {
        "name": "book_hotel",
        "parameters": {
            "hotel_name": "Hilton Hotel",
            "location": "Chicago",
            "room_type": "single room",
            "start_date": "2022-12-10",
            "nights": 2
        }
    }
]
\end{lstlisting}
& \XSolidBrush \\ \hline
\end{longtable}
\end{scriptsize}

\begin{table}[th]
\centering
\scriptsize
\begin{tabular}{|p{0.2\linewidth}|p{0.4\linewidth}|p{0.35\linewidth}|}
\hline
\textbf{User query} & \textbf{Tools} & \textbf{Expected Answer} \\ \hline
Delete all the files in the 'Drafts' directory including the directory. &
\begin{lstlisting}[aboveskip=-5pt, belowskip=0pt]
{
  "name": "cd",
  "parameters": {
    "folder": "string"
  }
},
{
  "name": "rm",
  "parameters": {
    "file_name": "string"
  }
},
{
  "name": "rmdir",
  "parameters": {
    "dir_name": "string"
  }
},
{
  "name": "echo",
  "parameters": {
    "content": "string",
    "file_name": "string (optional)"
  }
},
"..."
\end{lstlisting}
&
\begin{lstlisting}[aboveskip=-5pt, belowskip=0pt]
cd(folder="Drafts")
rm(file_name="DylanProject.txt")
cd(folder="..")
rmdir(dir_name="Drafts")
\end{lstlisting} \\
\hline
\end{tabular}
\caption{Input data for sample 46 of BFCL \texttt{multi\_turn\_base} task. NOTE: for visualization reasons, tool descriptions encoded as JSON schemas have been simplified (some of them have been omitted), and expected answers have been reported in python function format, even though the evaluation supports and use also a more verbose JSON format.}
\label{tab:bfcl-examples-multi-turn-base-46-query}
\end{table}

\begin{scriptsize}
\begin{longtable}[th]{|p{0.1\linewidth}|p{0.65\linewidth}|c|}
\caption{Example responses for all models to sample 46 of the \texttt{multi\_turn\_base} task of the BFCL evaluation. NOTE: we only reported results for models that were able to solve at least some of the multi-turn problems.}
\label{tab:bfcl-examples-multi-turn-base-46-responses} \\
\hline
\textbf{Model} & \textbf{Answer} & \textbf{Correct} \\ \hline
\endfirsthead
\multicolumn{3}{c}%
{{\bfseries \tablename\ \thetable{} -- continued from previous page}} \\
\hline
\textbf{Model} & \textbf{Answer} & \textbf{Correct} \\ \hline
\endhead
\hline \multicolumn{3}{|r|}{{Continued on next page}} \\ \hline
\endfoot
\hline
\endlastfoot
EngGPT2-16B-A3B & 
\begin{lstlisting}[aboveskip=-5pt, belowskip=0pt]
[[[{"rmdir": "{"dir_name": "Drafts"}"}], []]]
\end{lstlisting}
& \XSolidBrush \\ \hline
Qwen3-4B & 
\begin{lstlisting}[aboveskip=-5pt, belowskip=0pt]
[[[{"rm": "{"file_name": "Drafts"}"}], []]]
\end{lstlisting}
& \XSolidBrush \\ \hline
Qwen3-8B & 
\begin{lstlisting}[aboveskip=-5pt, belowskip=0pt]
[[[]]]
\end{lstlisting}
& \XSolidBrush \\ \hline
Llama-3.2-3B & 
\begin{lstlisting}[aboveskip=-5pt, belowskip=0pt]
[["[rm(dirs='Drafts')]", "[rmdir(dir='Drafts')]", "[rm(dir='Drafts')]", "[rm(path='Drafts')]", "[rm(path='Drafts', force=True)]", "[rm(force=True, path='Drafts')]", "[rm(path='Drafts', recursive=True)]", "[rm(recursive=True)]", "[rm(recursive=True, force=True)]", "[rm(force=True)]", "[rm()]", "[rm(file_name='Drafts')]", "[None]"]]
\end{lstlisting}
& \XSolidBrush \\ \hline
Llama-3.1-8B & 
\begin{lstlisting}[aboveskip=-5pt, belowskip=0pt]
[["[rm(file_name='Drafts'), rmdir(dir_name='Drafts')]", "[rmdir(dir_name='..')]\n\nThis will delete the 'Drafts' directory and all its contents. The '..' is used to go up one directory level, so it will delete the 'Drafts' directory in the parent directory."]]
\end{lstlisting}
& \XSolidBrush \\ \hline
gemma-3-4b-it & 
\begin{lstlisting}[aboveskip=-5pt, belowskip=0pt]
[[" ```tool_code 
 [cd, folder='Drafts'] 
 [rmdir, dir_name='Drafts'] 
 ```"]]
\end{lstlisting}
& \XSolidBrush \\ \hline
gemma-3-12b-it & 
\begin{lstlisting}[aboveskip=-5pt, belowskip=0pt]
[[" rm(file_name="Drafts")", ""]]
\end{lstlisting}
& \XSolidBrush \\ \hline
Ministral-3-8B & 
\begin{lstlisting}[aboveskip=-5pt, belowskip=0pt]
[[[{"ls": "{"a": true}"}], [{"ls": "{"a": true}"}], [{"cd": "{"folder": "Drafts"}"}], [{"ls": "{"a": true}"}], [{"rm": "{"file_name": "DylanProject.txt"}"}], [{"pwd": "{}"}], [{"rmdir": "{"dir_name": "Drafts"}"}], [{"pwd": "{}"}], [{"cd": "{"folder": ".."}"}], [{"rmdir": "{"dir_name": "Drafts"}"}], []]]
\end{lstlisting}
& \XSolidBrush \\ \hline
GPT-5-nano & 
\begin{lstlisting}[aboveskip=-5pt, belowskip=0pt]
[["[pwd(), ls({})]", "[cd(folder="Drafts"), ls({})]", "[rm(file_name="DylanProject.txt"), cd(folder=".."), rmdir(dir_name="Drafts")]", "[pwd(), ls({})]", "[pwd(), ls({})]", "[pwd(), ls({})]", "[cd(folder="ArchivedProjects")]", "[ls({})]", "[cd(folder=".."), rmdir(dir_name="ArchivedProjects")]", "[pwd(), ls({})]", "[pwd(), ls({})]", "[ls({})]", "[pwd(), ls({})]", "[pwd(), ls({})]", "[pwd(), ls({})]", "[ls({})]", "[ls({})]", "[pwd()]", "[ls({})]", "[pwd(), ls({})]", "[ls({})]"]]
\end{lstlisting}
& \XSolidBrush \\ \hline
gpt-oss-20b & 
\begin{lstlisting}[aboveskip=-5pt, belowskip=0pt]
[[[{"ls": "{"a": true}"}], [{"cd": "{"folder": "Drafts"}"}], [{"ls<|channel|>": "{"a": true}"}], [{"ls": "{"a": true}"}], [{"rm<|channel|>": "{"file_name": "DylanProject.txt"}"}], [{"rm<|channel|>commentary": "{"file_name": "DylanProject.txt"}"}], [{"rm<|channel|>commentary": "{"file_name": "DylanProject.txt"}"}], [{"rm<|channel|>commentary": "{"file_name": "DylanProject.txt"}"}], [{"rm<|channel|>commentary": "{"file_name": "DylanProject.txt"}"}], [{"rm<|channel|>commentary": "{"file_name": "DylanProject.txt"}"}], [{"rm<|channel|>commentary": "{"file_name": "DylanProject.txt"}"}], [{"rm<|channel|>commentary": "{"file_name": "DylanProject.txt"}"}], [{"rm<|channel|>commentary": "{"file_name": "DylanProject.txt"}"}], [{"rm<|channel|>commentary": "{"file_name": "DylanProject.txt"}"}], [{"rm<|channel|>commentary": "{"file_name": "DylanProject.txt"}"}], [{"rm<|channel|>commentary": "{"file_name": "DylanProject.txt"}"}], [{"rm<|channel|>commentary": "{"file_name": "DylanProject.txt"}"}], [{"ls<|channel|>commentary": "{"a": true}"}], [{"ls<|channel|>commentary": "{"a": true}"}], [{"ls": "{"a": true}"}], [{"rm<|channel|>commentary": "{"file_name": "DylanProject.txt"}"}]]]
\end{lstlisting}
& \XSolidBrush \\ \hline
\end{longtable}
\end{scriptsize}

\begin{table}[th]
\centering
\scriptsize
\begin{tabular}{|p{0.3\linewidth}|p{0.3\linewidth}|p{0.3\linewidth}|}
\hline
\textbf{User query} & \textbf{Tools} & \textbf{Expected Answer} \\ \hline
\begin{lstlisting}[aboveskip=-5pt, belowskip=0pt]
Estimate the distance between San Francisco and Rivermist. Afterwards, fill the tank with 40 additional liters of gasoline for my quick trip. Round the fill amount to two decimal places


Preparations are complete. Start the engine and help me tweet about this journey containing message 'Just started my journey!' with #RoadTrip and @carenthusiast.

\end{lstlisting}
 &
\begin{lstlisting}[aboveskip=-5pt, belowskip=0pt]
{
  "name": "authenticate_twitter",
  "parameters": {
    "username": "string",
    "password": "string"
  }
},
{
  "name": "comment",
  "parameters": {
    "tweet_id": "integer",
    "comment_content": "string"
  }
},
"...",
{
  "name": "activateParkingBrake",
  "parameters": {
    "mode": "string"
  }
},
{
  "name": "adjustClimateControl",
  "parameters": {
    "temperature": "float,
    "unit": "string",
    "fanSpeed": "integer",
    "mode": "string"
  }
},
"..."
\end{lstlisting}
&
\begin{lstlisting}[aboveskip=-5pt, belowskip=0pt]
get_zipcode_based_on_city('San Francisco'),
get_zipcode_based_on_city('Rivermist'),
estimate_distance(cityA='94016', cityB='83214'),
liter_to_gallon(liter=40),
fillFuelTank(fuelAmount=10.57)


lockDoors(unlock=False, door=['driver', 'passenger', 'rear_left', 'rear_right']),
pressBrakePedal(pedalPosition=1.0),
startEngine(ignitionMode='START'),
post_tweet(content='Just started my journey!', tags=['#RoadTrip'], mentions=['@carenthusiast'])
\end{lstlisting}
 \\
\hline
\end{tabular}
\caption{Input data for sample 69 of BFCL \texttt{multi\_turn\_base} task. NOTE: for visualization reasons, tool descriptions encoded as JSON schemas have been simplified (some of them have been omitted), and expected answers have been reported in python function format, even though the evaluation supports and use also a more verbose JSON format.}
\label{tab:bfcl-examples-multi-turn-base-69-query}
\end{table}

\begin{scriptsize}
\begin{longtable}[th]{|p{0.1\linewidth}|p{0.75\linewidth}|c|}
\caption{Example responses for all models to sample 69 of the \texttt{multi\_turn\_base} task of the BFCL evaluation.}
\label{tab:bfcl-examples-multi-turn-base-69-responses} \\
\hline
\textbf{Model} & \textbf{Answer} & \textbf{Correct} \\ \hline
\endfirsthead

\multicolumn{3}{c}%
{{\bfseries \tablename\ \thetable{} -- continued from previous page}} \\
\hline
\textbf{Model} & \textbf{Answer} & \textbf{Correct} \\ \hline
\endhead

\hline \multicolumn{3}{|r|}{{Continued on next page}} \\ \hline
\endfoot

\hline
\endlastfoot

EngGPT2-16B-A3B & 
\begin{lstlisting}[aboveskip=-5pt, belowskip=0pt]
[[[{"estimate_distance": "{"cityA": "San Francisco", "cityB": "Rivermist"}"}, {"fillFuelTank": "{"fuelAmount": 40.0}"}], []], [[{"startEngine": "{"ignitionMode": "START"}"}, {"post_tweet": "{"content": "Just started my journey! #RoadTrip @carenthusiast"}"}], []]]
\end{lstlisting}
& \XSolidBrush \\ \hline
Qwen3-4B & 
\begin{lstlisting}[aboveskip=-5pt, belowskip=0pt]
[[[{"get_zipcode_based_on_city": "{"city": "San Francisco"}"}, {"get_zipcode_based_on_city": "{"city": "Rivermist"}"}], [{"estimate_distance": "{"cityA": "94016", "cityB": "83214"}"}, {"liter_to_gallon": "{"liter": 40}"}], [{"fillFuelTank": "{"fuelAmount": 10.57}"}], []], [[{"startEngine": "{"ignitionMode": "START"}"}, {"post_tweet": "{"content": "Just started my journey!", "tags": ["#RoadTrip"], "mentions": ["@carenthusiast"]}"}], [{"lockDoors": "{"unlock": false, "door": ["driver", "passenger", "rear_left", "rear_right"]}"}], [{"startEngine": "{"ignitionMode": "START"}"}], [{"pressBrakePedal": "{"pedalPosition": 0.5}"}], [{"startEngine": "{"ignitionMode": "START"}"}], [{"pressBrakePedal": "{"pedalPosition": 1.0}"}], [{"startEngine": "{"ignitionMode": "START"}"}], []]]
\end{lstlisting}
& \Checkmark \\ \hline
Qwen3-8B & 
\begin{lstlisting}[aboveskip=-5pt, belowskip=0pt]
[[[{"get_zipcode_based_on_city": "{"city": "San Francisco"}"}, {"get_zipcode_based_on_city": "{"city": "Rivermist"}"}, {"estimate_distance": "{"cityA": "94101", "cityB": "12345"}"}, {"liter_to_gallon": "{"liter": 40}"}, {"fillFuelTank": "{"fuelAmount": 10.57}"}], []], [[{"startEngine": "{"ignitionMode": "START"}"}, {"post_tweet": "{"content": "Just started my journey!", "tags": ["#RoadTrip"], "mentions": ["@carenthusiast"]}"}], [{"lockDoors": "{"unlock": false, "door": ["driver", "passenger", "rear_left", "rear_right"]}"}], [{"startEngine": "{"ignitionMode": "START"}"}], [{"pressBrakePedal": "{"pedalPosition": 1.0}"}], [{"startEngine": "{"ignitionMode": "START"}"}], []]]
\end{lstlisting}
& \XSolidBrush \\ \hline
Llama-3.2-3B & 
\begin{lstlisting}[aboveskip=-5pt, belowskip=0pt]
[["[estimate_distance(cityA='San Francisco', cityB='Rivermist'), fillFuelTank(fuelAmount=40.0)]", "It seems that the `estimate_distance` function is not available in the provided tools. \n\nLet's try to find the distance between San Francisco and Rivermist using a different method. We can use the `get_zipcode_based_on_city` function to get the zipcodes for both cities, and then use the `get_zipcode_based_on_city` function again to get the distance between the two zipcodes.\n\n[get_zipcode_based_on_city(city='San Francisco'), get_zipcode_based_on_city(city='Rivermist'), get_distance(zipcodeA=zipcode, zipcodeB=zipcode2)]"], ["[startEngine(), post_tweet(content='Just started my journey!', tags=['#RoadTrip'], mentions=['@carenthusiast'])]", "It seems that the `startEngine` function requires an `ignitionMode` argument, which is missing.\n\nLet's try to start the engine with the default ignition mode. \n\n[startEngine(ignitionMode='START')]"]]
\end{lstlisting}
& \XSolidBrush \\ \hline
Llama-3.1-8B & 
\begin{lstlisting}[aboveskip=-5pt, belowskip=0pt]
[["[estimate_distance(cityA=\"San Francisco\", cityB=\"Rivermist\"), gallon_to_liter(liter=round(gallon_to_liter(gallon=40*0.264172, params_name1='gallon', params_name2='liter'), 2))]", "[estimate_distance(cityA=\"San Francisco\", cityB=\"Rivermist\"), gallon_to_liter(gallon=40*0.264172)]", "[get_zipcode_based_on_city(city=\"Rivermist\"), fillFuelTank(fuelAmount=39.99997322080001)]", "[get_outside_temperature_from_weather_com(), get_current_speed()]", "[displayCarStatus(option=\"speed\"), displayCarStatus(option=\"fuel\")]", "[get_user_stats(username=\"your_username\")]", "[post_tweet(content=\"I'm going on a trip to Rivermist, UT 83214\", tags=[], mentions=[])]", "[get_tweet(tweet_id=3), get_user_stats(username=\"travelbug\")]", "[get_user_tweets(username=\"travelbug\"), get_user_stats(username=\"travelbug\")]", "[get_user_stats(username=\"travelbug\"), get_user_tweets(username=\"travelbug\")]", "[get_user_stats(username=\"travelbug\"), get_user_tweets(username=\"travelbug\")]", "[get_user_stats(username=\"travelbug\"), get_user_tweets(username=\"travelbug\")]", "[get_user_stats(username=\"travelbug\"), get_user_tweets(username=\"travelbug\")]",
"..."
\end{lstlisting}
& \XSolidBrush \\ \hline
gemma-3-4b-it & 
\begin{lstlisting}[aboveskip=-5pt, belowskip=0pt]
[[" ```tool_code 
 [get_zipcode_based_on_city(city="San Francisco")] 
 [get_zipcode_based_on_city(city="Rivermist")] 
 [estimate_distance(distance=1700, cityA="94107", cityB="83801")] 
 [liter_to_gallon(liter=40)] 
 [fillFuelTank(fuelAmount=40.00)] 
 ```"], [" ```tool_code 
 [startEngine(ignitionMode='START')] 
 [post_tweet(content='Just started my journey!', tags=[], mentions=['@carenthusiast'])] 
 ```"]]
\end{lstlisting}
& \XSolidBrush \\ \hline
gemma-3-12b-it & 
\begin{lstlisting}[aboveskip=-5pt, belowskip=0pt]
[[" [estimate_distance(cityA="94105", cityB="95471"), gallon_to_liter(liter=40), fillFuelTank(fuelAmount=40.00)]", " I apologize for the errors in the previous responses. It seems there were issues with the tool calls. Let's try a different approach. 

 First, I will attempt to get the zipcode for Rivermist. Then, I will estimate the distance between San Francisco and Rivermist using the zipcodes. Finally, I will convert 40 liters to gallons and fill the fuel tank with that amount. 

 [get_zipcode_based_on_city(city="Rivermist"), estimate_distance(cityA="94105", cityB="95471"), liter_to_gallon(liter=40), fillFuelTank(fuelAmount=40.0)]"], [" [startEngine(ignitionMode="START"), post_tweet(content="Just started my journey! #RoadTrip @carenthusiast")]", " It appears the engine cannot start because the doors are unlocked. Let's lock all the doors before attempting to start the engine again. 

 [lockDoors(unlock=False, door=["driver", "passenger", "rear_left", "rear_right"]), startEngine(ignitionMode="START"), post_tweet(content="Just started my journey! #RoadTrip @carenthusiast")]"]]
\end{lstlisting}
& \XSolidBrush \\ \hline
Ministral-3-8B & 
\begin{lstlisting}[aboveskip=-5pt, belowskip=0pt]
[[[{"get_zipcode_based_on_city": "{"city": "San Francisco"}"}, {"get_zipcode_based_on_city": "{"city": "Rivermist"}"}], [{"estimate_distance": "{"cityA": "94016", "cityB": "83214"}"}], [{"liter_to_gallon": "{"liter": 40}"}], [{"fillFuelTank": "{"fuelAmount": 10.57}"}], []], [[{"startEngine": "{"ignitionMode": "START"}"}, {"post_tweet": "{"content": "Just started my journey!", "tags": ["#RoadTrip"], "mentions": ["@carenthusiast"]}"}], [{"lockDoors": "{"unlock": false, "door": ["driver", "passenger", "rear_left", "rear_right"]}"}], [{"startEngine": "{"ignitionMode": "START"}"}], [{"pressBrakePedal": "{"pedalPosition": 0.5}"}, {"startEngine": "{"ignitionMode": "START"}"}], [{"pressBrakePedal": "{"pedalPosition": 1}"}, {"startEngine": "{"ignitionMode": "START"}"}], []]]
\end{lstlisting}
& \Checkmark \\ \hline
GPT-5-nano & 
\begin{lstlisting}[aboveskip=-5pt, belowskip=0pt]
The response was filtered due to the prompt triggering Azure OpenAI's content management policy.
\end{lstlisting}
& \XSolidBrush \\ \hline
gpt-oss-20b & 
\begin{lstlisting}[aboveskip=-5pt, belowskip=0pt]
[[[{"get_zipcode_based_on_city": "{"city": "San Francisco"}"}], [{"get_zipcode_based_on_city": "{"city": "Rivermist"}"}], [{"estimate_distance": "{"cityA": "94016", "cityB": "83214"}"}], [{"liter_to_gallon": "{"liter": 40}"}], [{"fillFuelTank": "{"fuelAmount": 10.57}"}], []], [[{"startEngine": "{"ignitionMode": "START"}"}], [{"lockDoors": "{"unlock": false, "door": ["driver", "passenger", "rear_left", "rear_right"]}"}], [{"startEngine": "{"ignitionMode": "START"}"}], [{"pressBrakePedal": "{"pedalPosition": 0.5}"}], [{"startEngine": "{"ignitionMode": "START"}"}], [{"pressBrakePedal": "{"pedalPosition": 1.0}"}], [{"startEngine": "{"ignitionMode": "START"}"}], [{"post_tweet": "{"content": "Just started my journey!", "tags": ["#RoadTrip"], "mentions": ["@carenthusiast"]}"}], []]]
\end{lstlisting}
& \Checkmark \\ \hline
\end{longtable}
\end{scriptsize}

\begin{table}[th]
\centering
\scriptsize
\begin{tabular}{|p{0.15\linewidth}|p{0.25\linewidth}|p{0.5\linewidth}|}
\hline
\textbf{User query} & \textbf{GPT-5 Nano (PROMPT)} & \textbf{GPT-5 Nano (FC)} \\ \hline
Find all the roots of a quadratic equation given coefficients $a = 3$, $b = -11$, and $c = -4$. &
\begin{lstlisting}[aboveskip=-5pt, belowskip=0pt]
[solve_quadratic(a=3, b=-11, c=-4, root_type='all')]
\end{lstlisting}
&
\begin{minipage}[t]{\linewidth}
The quadratic is $3x^2 - 11x - 4 = 0.$ \\
\\
Discriminant: $\Delta = (-11)^2 - 4\cdot3\cdot(-4) = 121 + 48 = 169$. \\
\\
Roots: $x = [11 \pm \sqrt{169}] / (2  * 3) = [11 \pm 13] / 6$ → $x1 = 24/6 = 4$, $x2 = -2/6 = -1/3$. \\
\\
Therefore, the roots are $x = 4$ and $x = -1/3$.
\end{minipage}
\\ \hline
Calculate the area under the curve from $x = -2$ to $x = 3$ for the function $y = x^3$ using simpson method. &
\begin{lstlisting}[aboveskip=-5pt, belowskip=0pt]
[integrate(function="x^3", start_x=-2, end_x=3, method="simpson")]
\end{lstlisting}
&
\begin{minipage}[t]{\linewidth}
Result: 16.25 \\
 \\ 
Reason:
\begin{itemize}
    \item The exact integral is $\int_{-2}^{3} x^3 dx = [x^4/4]_{-2}^{3} = (81/4) - (16/4) = 65/4 = 16.25.$
    \item Using Simpson’s rule with any even number of subintervals (e.g., $n=4$, $h=1.25$) also gives 16.25, since $x^3$ is a cubic and Simpson’s rule is exact for polynomials up to degree 3.
\end{itemize}

Example with $n=4$:
\begin{itemize}
    \item $x0=-2$, $x1=-0.75$, $x2=0.5$, $x3=1.75$, $x4=3$
    \item $f(x) = x^3$: $f0=-8$, $f1=-0.421875$, $f2=0.125$, $f3=5.359375$, $f4=27$
    \item $I \approx h/3 [f0 + 4f1 + 2f2 + 4f3 + f4] = (1.25/3)[-8 -1.6875 +0.25 +21.4375 +27] = (1.25/3)*39 = 16.25$
\end{itemize}

So the area under the curve from -2 to 3 is 16.25 units². \\
\end{minipage}
\\ \hline
Calculate the resonant frequency of an LC circuit given capacitance of 100µF and inductance of 50mH. &
\begin{lstlisting}[aboveskip=-5pt, belowskip=0pt]
[ 
calculate_resonant_frequency(inductance=0.05, capacitance=0.0001, round_off=2)
]
\end{lstlisting}
&
\begin{minipage}[t]{\linewidth}
Using $f = 1 / (2\pi\sqrt{LC})$:
\begin{itemize}
    \item  $L = 50 mH = 0.05 H$
    \item $C = 100 \mu F = 0.0001 F$
    \item $\sqrt{LC} = \sqrt{0.05 \times 0.0001} = \sqrt{5e-6} \approx 0.00223607$
    \item $f \approx 1 / (2\pi \times 0.00223607) \approx 71.17 Hz$
\end{itemize}

So, the resonant frequency is about $71 Hz$ ($\approx 71.17 Hz$).
\end{minipage}
\\ \hline
\end{tabular}
\caption{User query and responses given by the GPT-5 Nano model for samples 5, 15 and 52 of BFCL \texttt{simple\_python} task. Samples have been selected to showcase correct responses in Prompt mode and corresponding incorrect responses in native Function Calling mode, to highlight the differences and provide an intuitive explanation of the degraded performance on simple tasks.}
\label{tab:bfcl-examples-simple-python-gpt-5}
\end{table}

\end{document}